\def\mode{1}

\if 0\mode
	\documentclass{article}
	\usepackage[utf8]{inputenc}
	\usepackage[margin=1in]{geometry}

	\usepackage{acronym}
\usepackage{graphicx}
\usepackage{mathtools, amscd, mathrsfs}
\usepackage{multirow}
\usepackage{url}
\usepackage{tikz}
\usepackage{tikz-cd}

\usepackage{tabu}

\usetikzlibrary{shapes.geometric}
\usepackage{amsmath}
\usepackage{amsfonts}
\usepackage{amsthm}
\usepackage{multirow}
\usepackage{colortbl}
\usepackage{bm}
\usepackage{array}

\def\vxadv{{\bm{x}'}}

\def\eqref#1{equation~\ref{#1}}

\def\1{\bm{1}}

\def\eps{{\epsilon}}

\def\vtheta{{\bm{\theta}}}

\def\vb{{\bm{b}}}

\def\vg{{\bm{g}}}

\def\vw{{\bm{w}}}
\def\vx{{\bm{x}}}
\def\vy{{\bm{y}}}
\def\vz{{\bm{z}}}

\def\mW{{\bm{W}}}

\DeclareMathAlphabet{\mathsfit}{\encodingdefault}{\sfdefault}{m}{sl}
\SetMathAlphabet{\mathsfit}{bold}{\encodingdefault}{\sfdefault}{bx}{n}

\def\gD{{\mathcal{D}}}

\def\gF{{\mathcal{F}}}

\def\gP{{\mathcal{P}}}

\def\gS{{\mathcal{S}}}

\def\gU{{\mathcal{U}}}

\def\gX{{\mathcal{X}}}
\def\gY{{\mathcal{Y}}}
\def\gZ{{\mathcal{Z}}}

\def\sB{{\mathbb{B}}}

\def\sR{{\mathbb{R}}}
\def\sS{{\mathbb{S}}}

\def\sZ{{\mathbb{Z}}}

\newcommand{\E}{\mathbb{E}}
\newcommand{\Ls}{\mathcal{L}}
\newcommand{\R}{\mathbb{R}}

\DeclareMathOperator*{\argmax}{arg\,max}
\DeclareMathOperator*{\argmin}{arg\,min}

\DeclareMathOperator{\sign}{sign}

\newcommand{\specialcell}[2][c]{%
  \begin{tabular}[#1]{@{}c@{}}#2\end{tabular}}

\newcommand{\etal}{et al.}
\newcommand{\etc}{etc.}
\newcommand{\ie}{i.e.}
\newcommand{\eg}{e.g.}
\newcommand{\wrt}{w.r.t.}
\newcommand{\eqq}{eq.}
\newcommand{\stt}{s.t.}

\if 0\mode

\else

\fi

\newcommand{\newpara}
    {
    \vskip 0.2in
    }

\usepackage{changes}[2019/01/26 v3.1.2]

\allowdisplaybreaks

\if 0\mode
    \usepackage{newclude}
    \usepackage{dirtree}
    \usepackage{booktabs}
    \usetikzlibrary{backgrounds}

    \usepackage{fixltx2e}
    \usepackage{caption}
    \usepackage{subfig}
    \usepackage{lscape}
    \usepackage{longtable}

\else
    \usepackage{caption}
    \usepackage{subcaption}
    \usepackage{dirtree}
    \usepackage{fixltx2e}
    \usepackage{lscape}
    \usepackage{afterpage}

\fi

\begin{document}
	\title{Adversarial Examples~--~A Complete Characterisation of the Phenomenon}

	\author{Alex Serban\thanks{a.serban@cs.ru.nl}\\ Radboud University \\ Software Improvement Group \\ The Netherlands \and Erik Poll \\ Radboud University \\ The Netherlands \and Joost Visser\\ Radboud University \\ Software Improvement Group \\ The Netherlands}

    \date{}
    \maketitle

    \begin{abstract}
    \if 0\mode
We provide a complete characterisation of the phenomenon of adversarial examples~--~inputs intentionally crafted to fool machine learning models.
We aim to cover all the important concerns in this field of study:
(1) the conjectures on the existence of adversarial examples, (2) the security, safety and robustness implications, (3) the methods used to generate and (4) protect against adversarial examples and (5) the ability of adversarial examples to transfer between different machine learning models.
We provide ample background information in an effort to make this document self-contained.
Therefore, this document can be used as survey, tutorial or as a catalog of attacks and defences using adversarial examples.

\else

Deep neural networks are at the forefront of machine learning research.
However, despite achieving impressive performance on complex tasks, they
can be very sensitive:
small perturbations of inputs can be sufficient to induce incorrect
behavior.
Such perturbations, called adversarial examples, are intentionally designed to test the network's sensitivity to distribution drifts.
Given their surprisingly small size, a wide body of literature conjectures on their existence and how this phenomenon can be mitigated.
In this paper we discuss the impact of adversarial examples on security, safety and robustness of neural networks.
We start by introducing the hypotheses behind their existence, the methods used to construct or protect against them and the capacity to transfer adversarial examples between different machine learning models.
Altogether, the goal is to provide a comprehensive and self-contained survey of this growing field of research.

\fi

    \end{abstract}

    \acrodef{ML}[ML]{Machine Learning}
\acrodef{DL}[DL]{Deep Learning}
\acrodef{DNN}[DNN]{Deep Neural Network}
\acrodef{SVM}[SVM]{Support Vector Machine}
\acrodef{RL}[RL]{Reinforcement learning}
\acrodef{PAC}[PAC]{Probably Approximately Correct}

\acrodef{MILP}[MILP]{mixed integer linear programming}

\acrodef{wrt}[\emph{w.r.t}]{with respect to}
\acrodef{st}[\emph{s.t.}]{such that}

\acrodef{fgsm}[FGS]{Fast Gradient Sign}
\acrodef{bim}[BI]{Basic Iterative}
\acrodef{illcm}[ILC]{Iterative Least-likely Class}
\acrodef{jsma}[JSMA]{Jacobian-based Saliency Map Attack}
\acrodef{uap}[UAP]{Universal Adversarial Perturbations}
\acrodef{opa}[OPA]{One Pixel Attack}
\acrodef{pgd}[PGD]{projected gradient descent}
\acrodef{rssa}[RSSA]{Randomised Single Step Attack}
\acrodef{eat}[EAT]{Ensemble Adversarial Training}
\acrodef{gaa}[GAA]{Generative Adversarial Attacks}
\acrodef{gan}[GAN]{Generative Adversarial Networks}
\acrodef{nae}[NAE]{Natural Adversarial Examples}
\acrodef{atn}[ATN]{Adversarial Transformation Networks}
\acrodef{vae}[VAE]{Variational Auto-Encoders}
\acrodef{cfoa}[CFOA]{Complete First Order Adversary}
\acrodef{iid}[i.i.d]{independent and identically distributed}
\acrodef{bpda}[BPDA]{Backward Pass Differentiable Approximation}
\acrodef{alp}[ALP]{Adversarial Logit Pairing}
\acrodef{fbgan}[FB-GAN]{Featurized Bidirectional Generative Adversarial Networks}
\acrodef{sap}[SAP]{Stochastic Activation Pruning}
\acrodef{mat}[MAT]{Multi-strength Adversarial Training}
\acrodef{dam}[DAM]{Dense Associative Memory}
\acrodef{zoo}[ZOO]{Zeroth Order optimisation}
\acrodef{sa}[STA]{Strong Adversary}
\acrodef{lm}[LM]{Linear Models}
\acrodef{dt}[DT]{Decision Trees}
\acrodef{knn}[KNN]{K-nearest Neighbour}

    \section{Introduction}
\label{sec:intro}

There is no doubt \ac{ML} and, in particular, \acp{DNN} achieve impressive results on tasks where it is not possible to specify procedural rule-sets.
Some examples are object recognition~\cite{he2016deep}, machine
translation~\cite{sutskever2014sequence} or speech recognition~\cite{vaswani2017attention}.
Fueled by the increasing size of  available  data and a decrease in computing cost, \ac{ML} algorithms are explored in a variety of new tasks and commercial applications, many of which are safety- and mission-critical.

Facing commercial deployment and the possibility of use in safety-critical systems, new properties of \ac{ML} algorithms become important:
in particular, their ability to maintain performance whenever faced with data coming from slightly different distributions than trained with or cope with uncertainties in the operational environment.
These properties are defined as the algorithm's power to generalize outside the training data and, respectively, the algorithm's robustness.

In optimization, a robust solution has the ability to perform well under a certain level of uncertainty~\cite{ben2009robust}.
Recent publications~\cite{szegedy2013intriguing, papernot2016transferability} showed \ac{ML} algorithms exhibit low robustness and triggered an impressive wave of publications.
Notably, \acp{DNN} are highly sensitive to small, \emph{intentional}, distribution drifts~--~inputs which substantially decrease their performance, while being in close resemblance to training data.
The term \emph{adversarial examples} was first used to describe such inputs by \citeauthor{szegedy2013intriguing}~\cite{szegedy2013intriguing}.

Since an \emph{intention} is required, many publications claim security consequences, \eg~\cite{nguyen2015deep, kurakin2016aadversarial, guo2017countering, papernot2017practical, su2017one, hein2017formal, sinha2018certifying, fawzi2018adversarial}, and
hypothesize that commercial deployment is hindered by low robustness.
In contrast, other publications show these claims are sometimes exaggerated and demand  that clear security requirements are formulated before security consequences are claimed~\cite{gilmer2018motivating, oneval}.
In between, many publications investigate the existence of adversarial examples from a theoretical perspective and shed light on this particular behavior of \ac{ML} algorithms.
Overall, there are two emergent reasons to study adversarial examples: (1) because attackers might use them to exploit \ac{ML} algorithms and (2) because they show \ac{ML} algorithms are not robust, which may stop them from being adopted in some domains.

Another phenomenon presented in this paper is the potential to transfer adversarial examples between different \ac{ML} algorithms.
This means an input designed to fool \acp{DNN} can trigger the same behavior for kernel methods. 
From a security standpoint, this phenomenon suggests an attacker does not need precise information about the algorithm she plans to attack.
Moreover, from a learning theory standpoint it suggests that (1) algorithms extrapolate similar decision boundaries despite using different \ac{ML} constructs and (2) sensitivity to similar distribution drifts is an universal phenomenon, independent of the \ac{ML} algorithm.

\if 0\mode
The goal of this report is to provide a comprehensive and complete overview of this research field.
\else
The goal of this paper is to provide a comprehensive survey of this research field.
\fi
We characterize the phenomenon of adversarial examples from its inception by discussing its causes, position it in the context of security with relevant threat models, introduce methods to construct and defend against adversarial examples and explore the capacity to transfer them between different \ac{ML} algorithms.
We strive for completeness, but given the high activity on this topic, with new papers constantly coming out, there will be further improvements in attacks and defences which are not covered here.
Nonetheless, the most representative attacks and defenses, which reveal how broad this research field is and how distinct the proposed solutions are, can be found in this paper.
The taxonomies used and the different perspectives on security, safety and robustness discussed in this paper equip the reader with a broad framework in which new developments will fit.
Altogether, the goal is to provide enough information so this document becomes a self-contained survey of the field, able to capture its different nuances and inspire new research directions.

\if 0\mode
Therefore, this report can be used in tutorials or as a catalog of attacks and defenses against adversarial examples.
\else
\fi

Although adversarial examples can be found for a variety of tasks, we restrict our presentation to object recognition because (1) most publications target this task and (2) examples from this field are easier to illustrate.
Nevertheless, adversarial example are constantly explored in other tasks.
Of particular interest is malware detection~\cite{grosse2016adversarial, hu2017generating, laskov2014practical, xu2016automatically, kreuk2018adversarial} because it implies direct consequences on security.
Other tasks such as reinforcement learning~\cite{behzadan2017vulnerability, huang2017adversarial, lin2017tactics},
speech recognition~\cite{carlini2016hidden, carlini2018audio}, facial recognition~\cite{sharif2016accessorize}, semantic segmentation~\cite{xie2017adversarial} or video processing~\cite{wei2018transferable, li2018adversarial, thys2019fooling} are also explored.
Moreover, some practical experiments are not covered in detail, \eg~deploying adversarial examples in the physical world by printing corrupted images~\cite{evtimov2017robust, kurakin2016adversarial}, altering the image acquisition device (\eg~a phone camera)~\cite{moosavi2017universal} or playing adversarial examples through speakers~\cite{yakura2018robust}.

A general remark about the terminology used in the paper: we make a distinction between \ac{ML} algorithms and \ac{ML} systems, in which the latter is any type of system which uses \ac{ML} algorithms and other components.
Whenever we talk about security, we consider the entire system under attack.
Moreover, the terms \ac{ML} algorithms and \ac{ML} models are used with the same meaning.
The rest of the paper is organized as follows.
Section~\ref{sec:background} provides \if 0\mode sufficient background information to comprehend the phenomenon\else some background information on machine learning, a formal definition of adversarial examples and positions the phenomenon in its historical context\fi.
Section~\ref{sec:taxonomy} presents taxonomies for attacks and defenses and uses these to classify existing approaches from the literature.
Section~\ref{sec:robustness_eval} discusses the property of robustness and Section~\ref{sec:causes} the hypotheses concerning the existence of adversarial examples.
Sections~\ref{sec:attacks} and~\ref{sec:defences} introduce the methods used to generate adversarial examples or protect against them, followed by the phenomenon of transferability  in Section~\ref{sec:transferability}.
\if 0\mode We conclude with a body of distilled knowledge in Section~\ref{sec:distilled_knowledge}.
\else
We conclude with a discussion in Section~\ref{sec:discussion} and lay down directions for future research in Section~\ref{sec:conclusions}.
\fi

    \section{Background and Related Work}
\label{sec:background}

\if 0\mode
In this section we cover several topics.
At first, we introduce related work presenting surveys or tutorials on adversarial examples.
Secondly, we formally define machine learning, deep learning and the task of object recognition.
Thirdly, we give more information about the underlying assumptions of machine learning models and some properties needed to understand adversarial examples.
Lastly, we formally introduce adversarial examples.

\else 
\paragraph{Prerequisites}
A computer is said to learn from experience 
\wrt~a task 
and a performance measure 
if its measured performance on the task
increases with experience
~\cite{mitchell1997machine}.
In this paper, we focus on the task of object recognition: given a set of images defined on the input space $\gX$ with their labels from the output space $\gY$, sampled from a fixed, but unknown probability distribution $\gD$ over the space $\gZ = \gX \times \gY$, a \ac{ML} algorithm attempts to find a mapping $f:\gX \rightarrow \gY$ which minimizes the number of misclassified images.
We assume that $\gX$ is a metric space and we can define distance functions between two points of the space.
The error made by a prediction $f(\vx_i) = \hat{y_i}$ when the true label is $y_i$ is measured by a loss function $l : \gY \times \gY \rightarrow \sR $.
Through learning, we select a function $f^{*}$ from a hypotheses space $\gF$ such that the expected loss $r(f) = \E_{(\vx, y \sim \gD)}[l(f(\vx), y)]$ is minimal: $f^{*} =  \argmin_{f \in \gF} r(f)$.
In practice, $\gD$ is not known and only a set of samples $\gS$ (defined as a set of pairs $\{(\vx_i, y_i)\}_{i=1}^{n}$) is available for training.
Thus, a \ac{ML} algorithm uses the empirical loss to approximate the expected loss:
\begin{equation}
	\hat{f} = \argmin_{f \in \gF} \quad \E_{(\vx, y) \sim \gS}[l(f(\vx), y)].
 \label{eq:emprirical}
\end{equation}
The hypotheses space $\gF$ can be any mapping from $\gX$ to $\gY$ such as a linear function or a \ac{DNN}.
Choosing $\gF$ for a task adds an inductive bias from the algorithm designer and involves a tradeoff between expressivity and generalization: if $\gF$ is not expressive enough, the algorithm will not be able to learn complex hypotheses.
On the opposite, if $\gF$ is too expressive, the algorithm will overfit on the training data. 
The loss function is generally chosen to be zero when $f(\vx_i) = y_i$ and positive otherwise.
A common loss function for object recognition is the cross-entropy loss.

The \ac{PAC}~\cite{valiant1984theory} theoretical model for statistical learning guarantees that given enough samples for a desired accuracy $\epsilon$ and for the probability of getting non-representative samples from the training distribution $\delta$ ($0 < \epsilon, \delta < 1$), the empirical risk will have an error less than or equal to $\epsilon$ with probability $1 - \delta$: $P(|r(\hat{f}) - r(f^*)| \leq \epsilon) \geq 1 - \delta$. 
In this framework, given the choice for $\epsilon$ and $\delta$, we can derive the sample complexity for learning a hypothesis with minimal risk.
An important assumption of this model is that training, test and inference data are drawn from the same probability distribution $\gD$.
Moreover, all data are sampled independently from distribution $\gD$ (also called \ac{iid}).

\paragraph{Adversarial Examples.}
Adversarial examples are inputs intentionally designed to be in close resemblance with samples from the distribution $\gD$, but cause a misclassification.
Formally, given a classification function $f$ and a clean sample $\vx$, which gets correctly classified by $f$ with label $y$, an adversarial example $\vxadv$ is constructed by applying the minimal perturbation $\eta$ to input $\vx$ such that $\vxadv$ gets classified with a different label $\hat{y}$: $\argmin_{\eta} f(\vx + \eta) = \hat{y}$.
Similarly, in the initial paper on adversarial examples,~\citeauthor{szegedy2013intriguing}~\cite{szegedy2013intriguing} search for the perturbation solving the following optimization problem:
\begin{equation}
	\begin{aligned}
	\label{eq:adversarial_generic}
		&\min_{\vxadv}. & & \|\vxadv - \vx\|_p, \\
		&s.t. & & f(\vxadv) = \hat{y}, \\
	\end{aligned}
\end{equation}
where $ || \cdot \|_p$ is a distance function defined on the metric space $\gX$.
Searching for the minimal perturbation is often a complex task because the search space is non-linear and non-convex~\cite{papernot2017practical, larochelle2009exploring}.
However, many approximation solutions have been proposed.
Finding solutions to \eqq~(\ref{eq:adversarial_generic}) is  illustrated in Figure~\ref{fig:adversarial_expl}.
Some examples of perturbations are also illustrated in~Figure~\ref{fig:adv_exam}.

\begin{figure}[h]
	\begin{minipage}[c]{0.6\textwidth}
		\includegraphics[height=6cm, keepaspectratio]{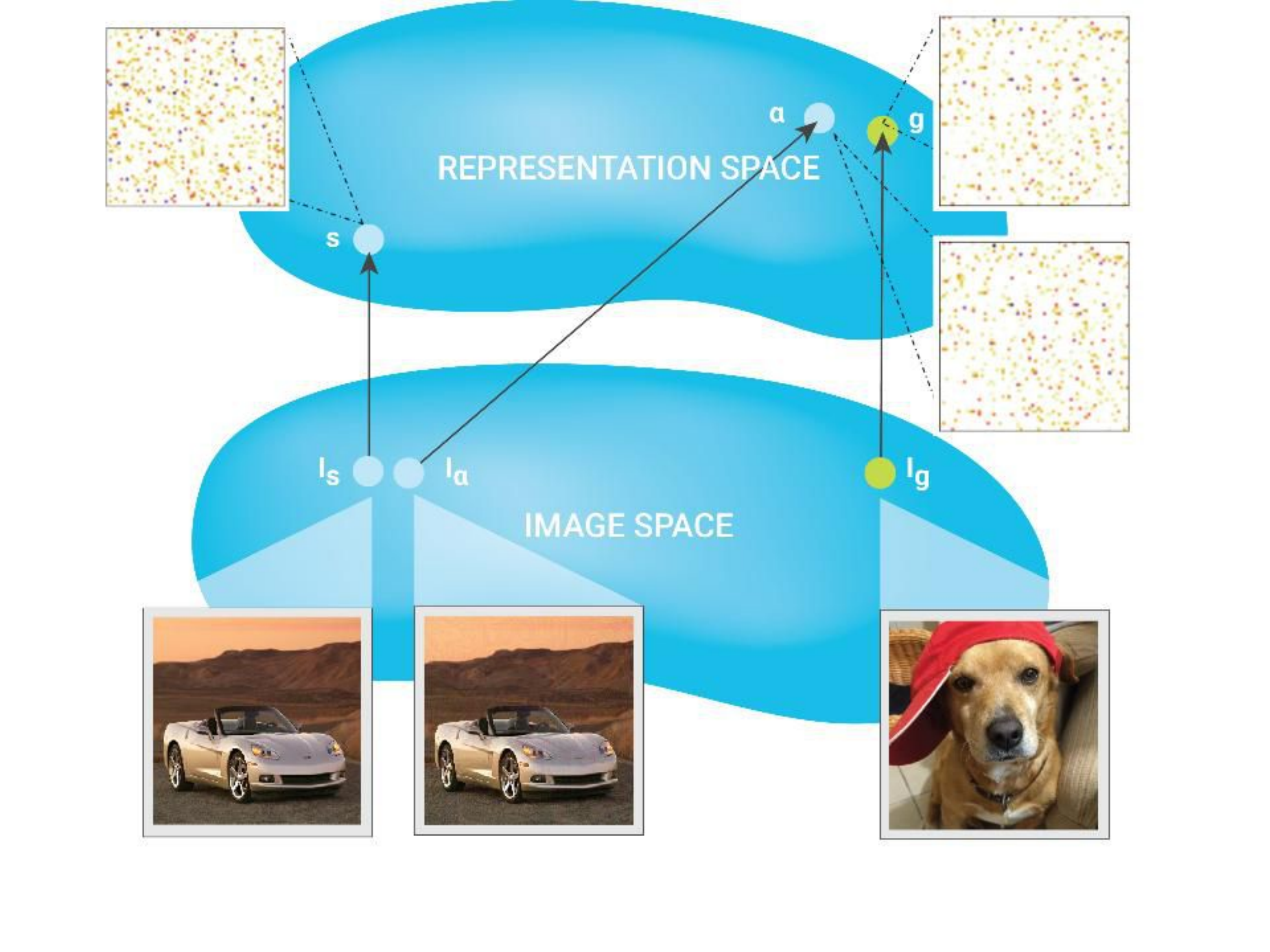}
	\end{minipage}\hfill
	\begin{minipage}[c]{0.4\textwidth}
		\caption{
			Adversarial example in input and representation space~\cite{sabour2015adversarial}.
			While the two pictures of cars are similar in the image space, the activation patterns of the second car are close to the activation patterns of the dog. Therefore, the second car gets classified as a dog.
			Moving the activation patterns from cars to dogs while keeping the representation in the image space similar is equivalent to searching for a solution to~\eqq~(\ref{eq:adversarial_generic}) and generating an adversarial example.}
		\label{fig:adversarial_expl}
	\end{minipage}
\end{figure}

The distance function most commonly used for adversarial examples in the object recognition domain is the p-norm:
\begin{equation}
	\| \vx \|_p = \left(\sum\limits_{i=1}^n \lvert \vx_i \rvert ^p\right)^{1\over p},
\end{equation}
where $p \in \{0, 2, \infty\}$. 
The choice for $p$ influences the coordinates changed in the initial sample as follows:
\begin{itemize}
	\item when $p=0$ the distance measures the number of different coordinates between the normal input and the adversarial; corresponding to the number of pixels altered in the original image.
	\item when $p=2$ the distance measures the Euclidian distance between the original and the adversarial image. This metric remains small when there are many small changes to many pixels and increases when there is a big change in one or multiple pixels.
	\item when $p=\infty$ the distance measures the maximum change in any of the coordinates and is equivalent to the maximum bound for changing each pixel in an image, without restricting the number of changed pixels.
\end{itemize}  
Although most publications use the p-norm distance, there is an increased interest to move away from it and explore new metrics. 
One proposed alternative is the Wasserstein distance, which represents the cost of moving pixel mass from the original image to the adversarial example~\cite{wong2019wasserstein}.
Selecting the right metric is still an open question and will be discussed in Section~\ref{sec:discussion}.
\fi

\if 0\mode
\subsection{Related Work}
\label{subsec:related_work}
\else
\paragraph{Historical considerations.}
\fi

Even though the term adversarial examples was first coined around 2014 in research by \citeauthor{szegedy2013intriguing} into \acp{DNN}~\cite{szegedy2013intriguing}, adversarial machine learning was established long before.
Unfortunately, as other authors have also observed~\cite{Biggio2018, gilmer2018adversarial}, recent publications concerning \acp{DNN} seem unaware of the earlier research on adversarial machine learning and loose important perspective in this field.
In particular, the importance of thread modeling to security is overlooked.

The first publication regarding adversarial \ac{ML} was published in 2004, when \citeauthor{dalvi2004adversarial}~\cite{dalvi2004adversarial}, followed by \citeauthor{lowd2005adversarial}~\cite{lowd2005adversarial}, managed to fool linear classifiers for spam detection by making changes to spam e-mails \cite{Biggio2018}.
\citeauthor{barreno2006can}~\cite{barreno2006can} first introduced a taxonomy for attacks and defenses in adversarial settings, and later refined it in~\cite{barreno2010security}.
This early taxonomy defines \ac{ML} threat models and is comprehensive enough to include adversarial examples.
However, the notion of minimal perturbation was not yet adopted.

Thereafter, a large body of publications discussed adversarial attacks against \ac{ML} models at both \emph{training} time~\cite{rubinstein2009antidote, biggio2012poisoning} and \emph{test} or \emph{inference} time~\cite{lowd2005adversarial, globerson2006nightmare} or defense mechanisms against such attacks~\cite{kolcz2009feature, bruckner2012static}.
Adversarial attacks at training time  modify or \emph{poison} the training data set (before training), while attacks at test time only modify the samples used for inference (after training).
In parallel to developing attacks and defenses, several publications proposed methods to evaluate the security of \ac{ML} models against adversarial attacks~\cite{biggio2014security, barreno2010security}.
\citeauthor{Biggio2018}~\cite{Biggio2018} trace an interesting parallel between the evolution of adversarial \ac{ML} and the rise of \acp{DNN}.

Adversarial examples represent attacks against machine learning models at inference time.
Moreover, they have a special trait: the perturbations used to fool classifiers are desired to be minimal, or as small as possible.
In practice, such perturbations are very small and barely noticeable to human observers.
In this \if 0\mode report \else paper \fi we are concerned with recent literature, triggered by~\citeauthor{szegedy2013intriguing}~\cite{szegedy2013intriguing} and the widely adopted definition of adversarial examples presented in \eqq~(\ref{eq:adversarial_generic}).
This body of work focuses on \acp{DNN} and was triggered by the surprisingly small perturbations needed to fool such algorithms.

From a security standpoint, we can make another distinction between
publications before and after~\citeauthor{szegedy2013intriguing}~\cite{szegedy2013intriguing}:
in general, publications before~\citeauthor{szegedy2013intriguing}
look at attacks on systems providing \emph{security functionality} (\eg~spam or virus detection), in contrast to more recent papers~\cite{Biggio2018, Papernot2018} which look at \emph{secure functionality} of \emph{any} application of \ac{ML} algorithms, \ie~if any application of \ac{ML} algorithms is secure.
This distinction will be further developed in Section~\ref{sec:discussion}.

\if 0\mode \bigskip \else \paragraph{Related work.}   \fi

Two previous publications surveyed the field of adversarial examples.
Firstly, \citeauthor{liu2018survey}~\cite{liu2018survey} investigated security threats at both training and test time.
Their work, together with~\cite{Biggio2018} represents a bridge between the two positions mentioned earlier: before and after \citeauthor{szegedy2013intriguing}~\cite{szegedy2013intriguing} coined the term adversarial examples.
The paper successfully maps the phenomenon of adversarial examples to the initial taxonomy of adversarial attacks~\cite{barreno2010security} and positions the field in the general context of \ac{ML} security.
Secondly, \citeauthor{akhtar2018threat}~\cite{akhtar2018threat} present an overview of attacks and defenses against adversarial examples for object recognition, focused on technical details.

We build on previous work by relating the threats posed by adversarial examples to security, safety and robustness of \acp{DNN}.
Moreover, we discuss the hypotheses on the existence of adversarial examples and their property of being transferable between different \ac{ML} models.


\if 0\mode

    \subsection{Formal Definition of Machine and Deep Learning}
\label{subsec:machine_learning}

\ac{ML} algorithms are computer programs conceived to learn from data.
Instead of manually specifying decision rules about a scenario, the algorithms automatically learn them by examining descriptive data.
Mitchell defines a computer program to learn from \emph{experience} $E$ \wrt~some class of \emph{tasks} $T$ and \emph{performance measure} $P$, if its performance at tasks in $T$ as measured by $P$ \emph{improves} with experience $E$~\cite{mitchell1997machine, goodfellow2016deep}.
Following this definition, the three ingredients of \ac{ML} algorithms are: (1) a task, (2) data about the task and (3) a measure of performance or improvement on the task.

As mentioned in the introduction, we will focus on the task of object detection (or classification) because (1) the majority of papers targeting adversarial examples focus on this task and (2) it is easier to provide visual examples.
Formally, a classification algorithm identifies a mapping from an input space (experiences) to an output space that represents a class (label) the input belongs to:

\begin{equation}
	 f : \sR^{n} \rightarrow \{1,2, \dots, k \}.
\end{equation}
\noindent
For object recognition, the input is a vector containing the image pixels -~$\vx$~- and the output is a number corresponding to a class (label)~--~$\hat{y}$.
If we group the parameters of the classification function in a vector~--~$\vtheta$~--~we can define a machine learning model as $\hat{y} = f(\vtheta, \vx)$.
Instead of manually specifying the parameters~--~$\vtheta$~--~the goal of \ac{ML} is to automatically learn them by looking at multiple examples.

In order to evaluate the performance of a \ac{ML} model we need to define a \emph{cost function} which computes the \emph{error rate} \ie~the proportion of examples for which the model produces incorrect labels.
If we know the true label of an input before designing the model, the task is called \emph{supervised} learning.
In this case we can define a cost function that takes the correct label and the one returned by our model and outputs an error score:
\begin{equation}
	J(\vtheta) = g(f(\vx, \vtheta), y).
\end{equation}
\noindent
Some examples of functions $g$ are mean square error or cross entropy.

We say that, in order to build a machine learning algorithm, one needs to design an algorithm that will automatically adjust its parameters when seeing new data, in order to increase its performance for a task.
The range of possible parameters describe a hypothesis space; from which a \ac{ML} algorithm selects the best hypothesis for the given task.
The \ac{ML} approach to this problem is to minimise the cost function, $J(\vtheta)$, \wrt~parameters $\vtheta$, on each sample in the data set, through a process called \emph{training}.
In order to evaluate the performance of an algorithm in a neutral manner, it is common to separate the available data into a \emph{training data set} - used only during the training phase - and \emph{testing data set} - a smaller portion of the data that is only used for testing.

The choice for a cost function may seem straightforward and objective, but in practice it is difficult to choose a function that corresponds well to the desired behaviour of a system.
In some cases, this is a consequence of the fact that it is difficult to decide what should be measured~\cite{goodfellow2016deep}.
Moreover, the dependence of $J$ on $f(\vx, \vtheta)$ impacts both the algorithm design and the optimisation strategy.
It is well known that non-convex functions can have multiple local optimum points and deciding if a local optimum is a global one or wether the problem has any solutions is computationally expensive.

%
%

\bigskip
In complex settings, such as object recognition, the function $f(\vx, \vtheta)$ is highly non-linear and hard to engineer.
Moreover, features relevant to the task are hard to extract due to high dimensionality, computational costs or lack of expert knowledge.
A solution to this problem is to compose $f(\vx, \vtheta)$ from several functions and use raw, unprocessed data as input.

\ac{DNN} are \ac{ML} models loosely inspired by neuroscience, in which a model is built through composition of linear and non-linear functions.
The model is analogous to a directed acyclic graph describing how the functions are composed~\cite{goodfellow2016deep}.
For example, a model can be composed of three functions $f^{(1)}, f^{(2)}, f^{(3)}$, connected in a chain to form $f(\vx) = f^{(3)}(f^{(2)}(f^{(1)}(\vx)))$.
In this case $f^{(1)}$ is called the \emph{first layer}, $f^{(2)}$ is called the \emph{second layer}, \etc~
Because an algorithm must learn the parameters of these functions, they are called \emph{hidden layers}.
The final layer of a neural network is also called the \emph{output layer}.
Deep Learning is a term attributed to models with many hidden layers (Figure~\ref{fig:fwdprop}).

In practice, each hidden layer applies a linear and a non-linear (activation) transformation to its input.
The goal of deep learning is to find the parameters of each function in the hidden layers that obtain maximum performance on the training data set.
The activation function is typically chosen to be a function applied element-wise (\eg~ReLU~\cite{nair2010rectified}).
Formally, we can define the output of one hidden layer as:

\begin{equation}
	h_i = g(\mW_{:,i}^{T} \vx  + \vb_i),
\end{equation}
\noindent
where $g$ is a the non-linear, element-wise, function, $\mW$ is the weight vector for the linear transformation and $\vb_i$ is a vector of bias terms.

The parameters are learned by minimising a cost function (taking a step in the opposite direction of its gradient, at each iteration through the data set).
Because we have a composition of functions, the gradient is computed through the hidden layers using a chain-rule based algorithm, called \emph{back-propagation}~\cite{chauvin2013backpropagation}.
One of the necessary conditions for back-propagations is to have differentiable activation functions.

\begin{figure}[h]
	\begin{equation*}
	  \begin{CD}
    	\vx @>>> f^{(1)} @>>> \dots @>>> f^{(n)} @>>> \hat{y} @>>> J \\
		@. @AAA @. @AAA \\
    	@. \vw_{1} @. \dots @. \vw_{n}
	  \end{CD}
	\end{equation*}
	\caption{Forward propagation of error (increases the cost).}
	\label{fig:fwdprop}

\end{figure}

\begin{figure}[h]
	\begin{equation*}
	  \begin{CD}
    	\frac{\partial J}{\partial \vx} @<\frac{\partial f^{(1)}}{\partial \vx}<< \frac{\partial J}{\partial f^{(1)}} @<\frac{\partial f^{(2)}}{\partial f^{(1)}}<< \dots @<\frac{\partial f^{(n)}}{\partial f^{(n-1)}}<< \frac{\partial J}{\partial f^{(n)}} @<\frac{\partial \hat{y}}{\partial f^{(n)}}<< \frac{\partial J}{\partial \hat{y}} \\
		@. @V\frac{\partial f^{(1)}}{\partial \vw_{1}}VV @. @V\frac{\partial f^{(n)}}{\partial \vw_{n}}VV  \\
		@. \frac{\partial J}{\partial \vw_{1}} @.\dots @. \frac{\partial J}{\partial \vw_{n}}
	  \end{CD}
	\end{equation*}
	\caption{Backward propagation of error (decreases the cost).}
	\label{fig:backprop}
\end{figure}

One may think of a neural network as increasing the error function from left to right through forward-propagation, as showcased in Figure~\ref{fig:fwdprop} and decreasing it from right to left through back-propagation, as showcased in Figure~\ref{fig:backprop}.
By learning the right parameters for the hidden layers, \ac{DNN} allow to trace discriminants between classes in high dimensional spaces (hyper-spaces), where the discriminants are often non-linear.

    \subsection{The Manifold Assumption}
\label{subsec:manifold}

An important concept underlying many ideas in \ac{ML} is that of a manifold~\cite{goodfellow2016deep}.
Intuitively, a manifold is an n-dimensional surface.
Accurately, a manifold is a set of connected points associated with their neighbourhoods and transformations that allow transitions from one point to another.
In \ac{ML}, the term manifold is used to describe a set of connected points that can be well approximated considering only a small numbers of dimensions.
Moreover, in \ac{ML}, the dimensionality of the manifold can vary from one point to another~--~when the manifold intersects itself.

Because approximations across all of $\sR^n$ are impossible, \ac{ML} algorithms assume that most of $\sR^n$ consists of invalid inputs and that interesting inputs lie in a collection of manifolds.
Moreover, it is assumed that variations of the learned function occur only in the directions that lie on the manifold or when moving from one manifold to another.
For example, one can imagine in $\sR^n$ several manifolds describing 10 classes, corresponding to digits from $0-9$.
By moving on a manifold which describes one class, we can identify variations of the same input (such as rotations, translations, \etc).
When moving across manifolds, however, the input can denote a change of class.
This behaviour is called \emph{the manifold assumption}.
Learning the structure of the manifold where the data lies is usually easier because the manifold is described by less dimensions than $\sR^n$.

Although the concept of manifold is mostly used in un-supervised and semi-supervised learning, it is of importance when talking about adversarial examples.

    \subsection{Independent and Identically Distributed Random Variables}
\label{subsec:iid}

\ac{ML} algorithms assume training and test data sets are drawn from the same probability distribution.
Moreover, the examples in each data set are assumed to be \emph{independent}, \ie~they convey no information about each other and, as a consequence, knowing any information about one does not change the probability distribution of the others.
These assumptions are called the \ac{iid} assumptions and allow the data generation process to described with a probability distribution over a single sample.
The same distribution can later be used to generate all training and testing examples.

The \ac{iid} assumptions are a fundamental tool to study the relationship between training and testing errors.
A key requirement of \ac{ML} algorithms, that distinguishes them from other optimisation solutions, is the capacity to obtain low error on both training and testing data sets \ie~\emph{generalise} to unseen data, such as the testing set.
Because data in the training set is limited, all \ac{ML} algorithms can only approximate the true data generating distribution, \ie~they can only \emph{estimate} it.
Whenever this estimate or the true distribution shifts,  the accuracy of \ac{ML} models drops.

A popular example of distribution shift is \emph{covariate shift} - a change in the distribution of the independent variables that should not impact the output of a \ac{ML} model.
For example, a change in the brightness of all images in the test set should not cause any misclassifications.
Covariate shift is a consequence of a change in the state of latent variables from the distribution.
Adversarial examples have been interpreted in some contexts as an instantiation of covariate shift~\cite{song2017pixeldefend}.
However, common techniques to alleviate the impact of covariate shift in \ac{DNN}~\cite{ioffe2015batch} do not help with adversarial examples.

    \subsection{Norms and Norm Ball}
\label{subsec:norm_balls}

In order to measure the size of a vector or the distance between two vectors, we use a function called a \emph{norm}.
Formally, the $L_p$ norm, $\| \vx \|_p$, is defined as:

\begin{equation}
	\| \vx \|_p = \left(\sum\limits_{i=1}^n \lvert \vx_i \rvert ^p\right)^{1\over p}.
\end{equation}
\noindent
Norms are functions that map vectors to non-negative scalars.
In order to measure the distance between two vectors, we can take the norm of their difference: $\| x' - x \|_p$, which will return a positive scalar.

Three $p$ values are used in the context of adversarial examples, which lead to the following distance measures:

\begin{enumerate}
	\item $L_0$ (based on the absolute norm) - which measures the number of coordinates $i$ such that $\vx_i \ne \vxadv_i$. It corresponds to the number of pixels that have been altered in an image.\footnote{In RGB images, there are
			three channels that each can change. We count the number of \emph{pixels} that
			are different, where two pixels are considered different if \emph{any} of the three colours are different.}

	\item $L_2$ (based on the Euclidian norm) - which measures the Euclidean distance between $\vx$ and $\vxadv$. The $L_2$ distance can remain small when there are many small changes to many pixels. This distance metric was used in the initial adversarial example work~\cite{szegedy2013intriguing}.

	\item $L_{\infty}$ (based on the maximum norm) - which measures the maximum change to any coordinate:
		\begin{equation*}
			\|\vxadv-\vx\|_\infty = \max (|\vxadv_1-\vx_1|,\dots,|\vxadv_n-\vx_n|).
		\end{equation*}
		For images, one can imagine a maximum limit which bounds the change in each pixel, without restricting the number of pixels that are modified.
\end{enumerate}

\bigskip
In some circumstances, we would like to measure the set of points that lie at a certain norm distance from a chosen point.
This set is called a \emph{norm ball} and is formally defined as:
\begin{equation}
	\sB(\vx_c, r) = \{\vx \mid \| \vx - \vx_c \|_p \leq r\},
\end{equation}
where $\vx_c$ is the point chosen as the centre of the ball and $r$ is the maximum distance from the centre aka radius.

    \subsection{Lipschitz Continuity}
\label{subsec:lipschitz}

In order to limit how fast a function changes, given a change in inputs, one can restrict to functions that are either Lipschitz continuous or have Lipschitz continuous derivatives.
A Lipschitz continuous function is a function whose rate if change is bounded by a Lipschitz constant $\Ls$:
\begin{equation}
	| f(\vx) - f(\vy) | \leq \Ls \| \vx - \vy \|_2.
\end{equation}
\noindent
This property allows us to quantify the assumption that a small change in the input will have a small impact in the output of a function.
Enforcing this constraint is fairly easy, and often used to prove a bound to robustness of \ac{DNN}. 
Szegedy \etal~\cite{szegedy2013intriguing} showed that deep  neural  networks with  half-rectified  layers  (\ie~convolutional  or  fully  connected layers with ReLU activation functions), max pooling and  contrast-normalisation  layers  are  Lipschitz  continuous.
Later it has been proved that the softmax output layer and the sigmoid and hyperbolic tangent activation functions also satisfy Lipschitz continuity~\cite{ruan2018reachability}.

    \subsection{Formal Definition of Adversarial Examples}
\label{subsec:formal_adversarial}

We \if 0\mode are now ready to \else\fi provide a formal definition of adversarial examples, as first stated in the paper that coined the term~\cite{szegedy2013intriguing}.
Given a classification function $f$ and a clean sample $\vx$, which gets correctly classified by $f$ with the label $l$, we construct an adversarial example $\vxadv$ by applying the minimal \emph{perturbation} $\eta$ to input $\vx$ such that it gets classified by the model with a different label, $l'$:
\begin{equation}
	\begin{aligned}
	\label{eq:adversarial_generic}
		&\min_{\vxadv} & & \|\vxadv - \vx\|_p, \\
		&s.t. & & f(\vxadv) = l', \\
		& & & f(\vx) = l, \\
		& & & l \neq l', \\
		& & & \vxadv \in [0,1]^m,
	\end{aligned}
\end{equation}
\noindent
where $ || \cdot \|_p$ is the distance between the samples (usually the p-norm) and $\vxadv - \vx = \eta$ is called a \emph{perturbation}.
Searching the minimal perturbation is "not trivial"~\cite{papernot2017practical} because different properties of \ac{DNN} make the search non-linear and non-convex~\cite{larochelle2009exploring}.
However, many approximation methods have been proposed (Section~\ref{sec:attacks}).
\if 0\mode
The condition that a perturbation is \emph{minimal} does not always hold because, in contrast to humans, a \ac{ML} model can not distinguish between large or small perturbations.
However, it is enforced in most methods used to generate adversarial examples.
\else \fi

We illustrate the adversarial generation process and give some examples of adversarial inputs in Figure~\ref{fig:adv_examples}.
Determining the minimal perturbation is equivalent to moving the data-point corresponding to an input from the region of a correct class to another, as illustrated in Figure~\ref{adv_space}.
If the perturbation is small, the result can not be perceived by humans as malicious, as illustrated in Figure~\ref{adv_exam}.

\if 0\mode
\begin{figure}[h]%
    \centering
    \subfloat[The adversarial generation process~\cite{tramer2017space}.]{\label{adv_space}{\includegraphics[height=4cm, keepaspectratio]{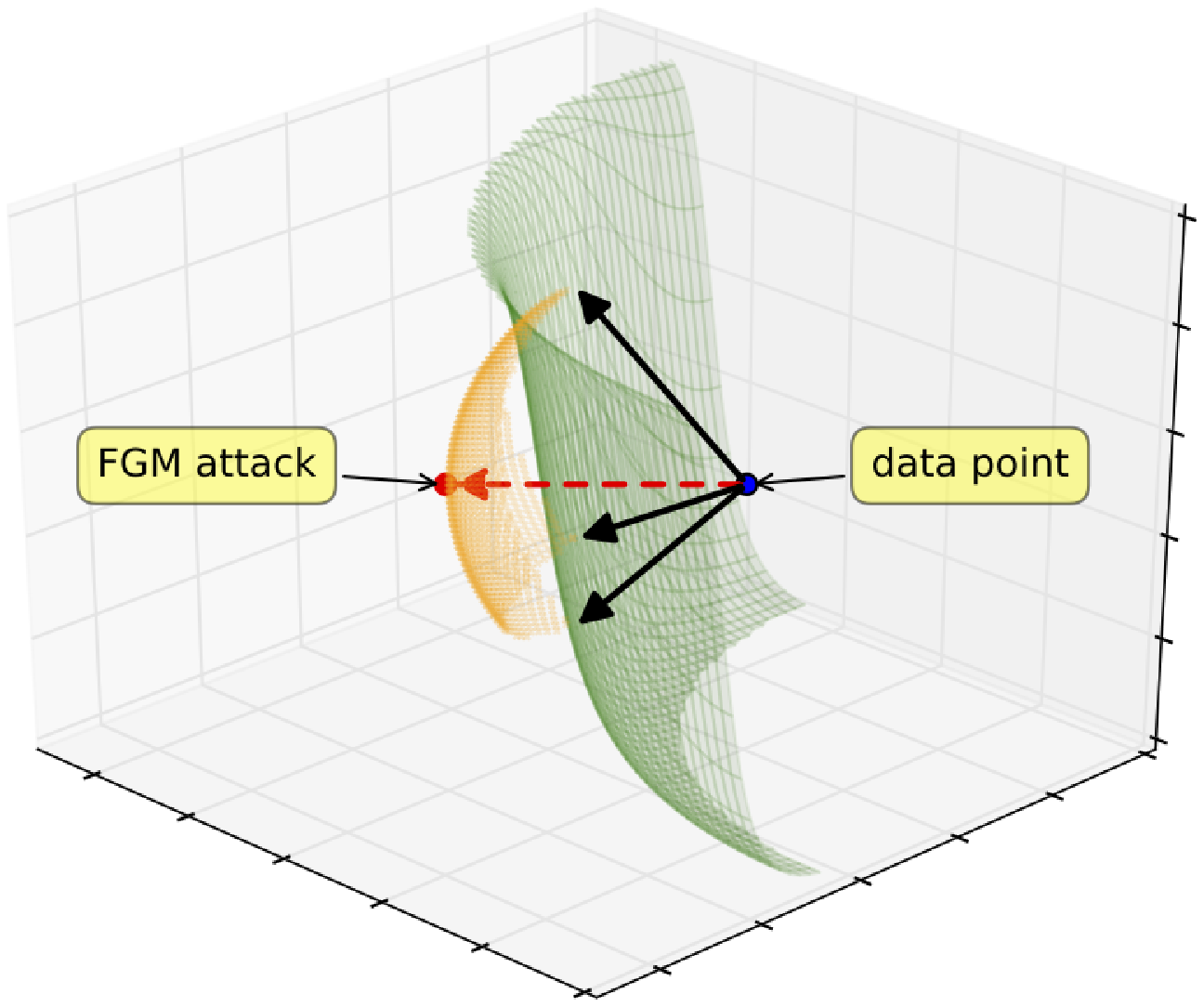} }}%
    \qquad \qquad
    \subfloat[Adversarial inputs~\cite{szegedy2013intriguing}.]{\label{adv_exam}{\includegraphics[height=4cm, width=5cm, keepaspectratio]{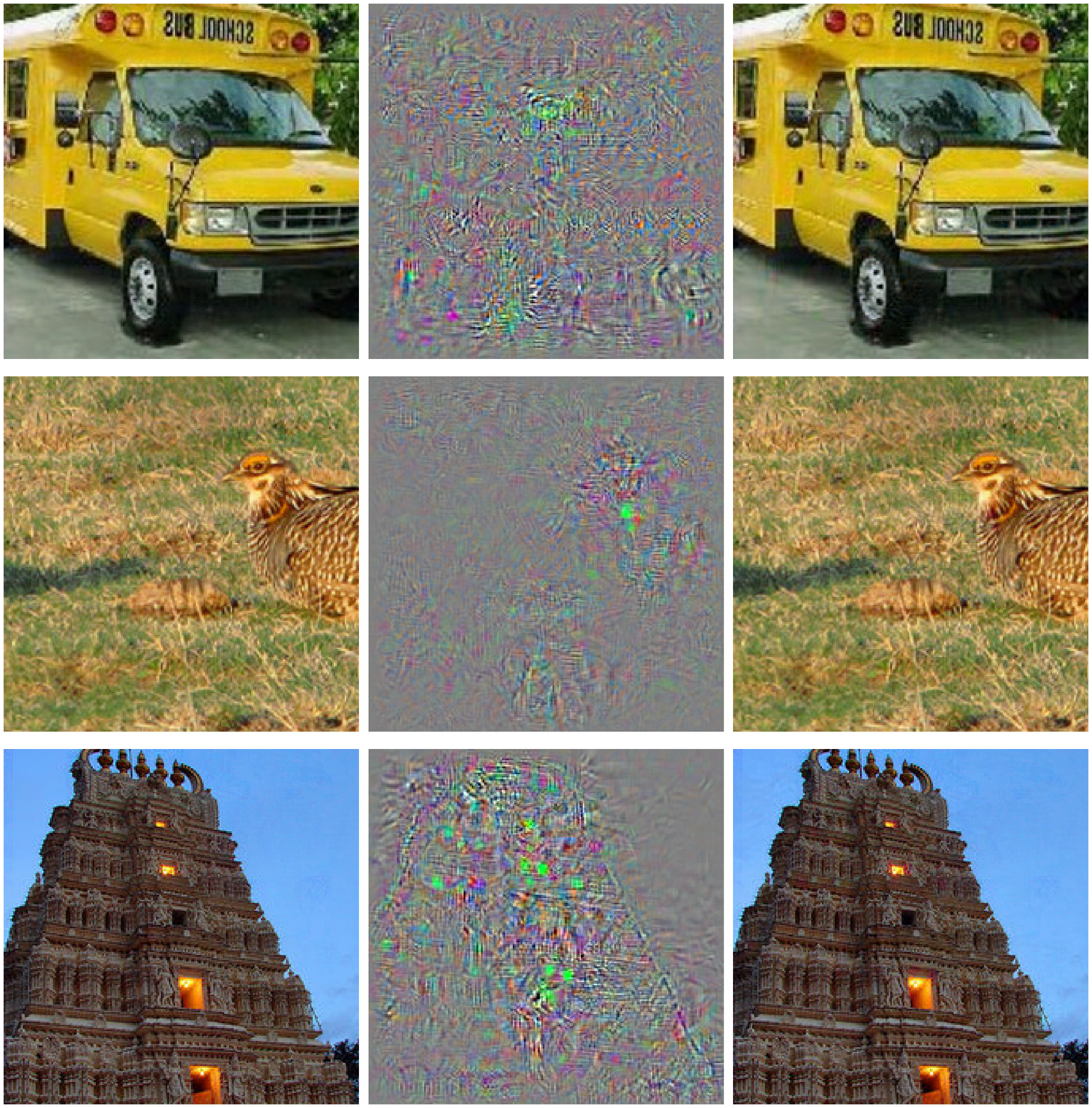} }}%
    \caption{An illustration of crafting an adversarial examples (a)~--~equivalent to moving the data-point corresponding to an input from the region of a correct class to an incorrect class and (b) an illustration of adversarial examples with very low perturbations~--~the pictures in the first column are correctly classified inputs, while the pictures in the last column are adversarial examples, crafted to cause a misclassification. In between, the perturbation used to cause the misclassification is illustrated.}%
    \label{fig:adv_examples}%
\end{figure}
\else

 \begin{figure}[h]
 		\centering
		\begin{subfigure}[]{0.49\textwidth}
			\centering
			\includegraphics[height=4cm, keepaspectratio]{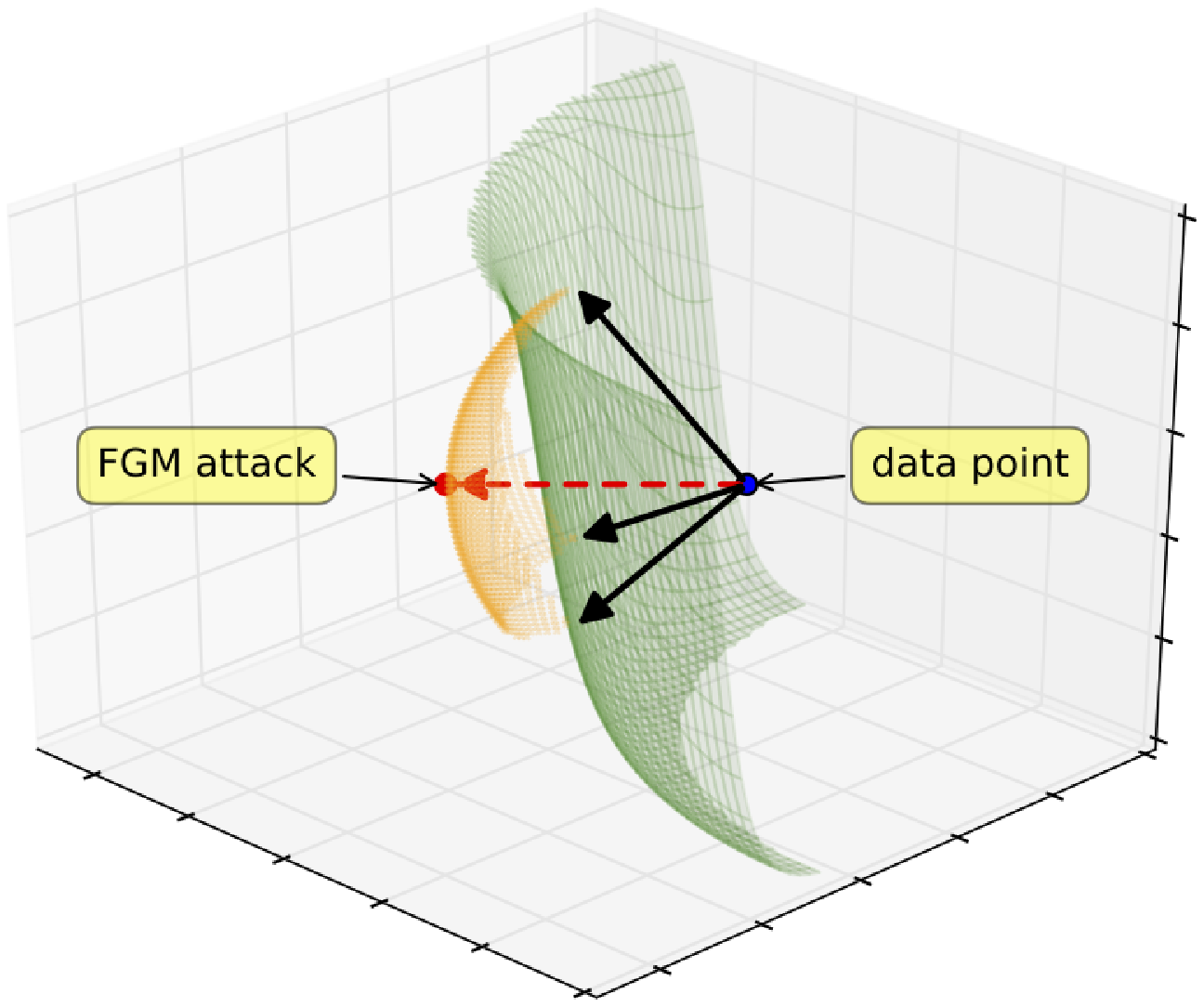}
            \subcaption{The adversarial generation process~\cite{tramer2017space}.}
			\label{adv_space}
		\end{subfigure}
		\begin{subfigure}{0.49\textwidth}
        	\centering
            \includegraphics[height=4cm, width=5cm, keepaspectratio]{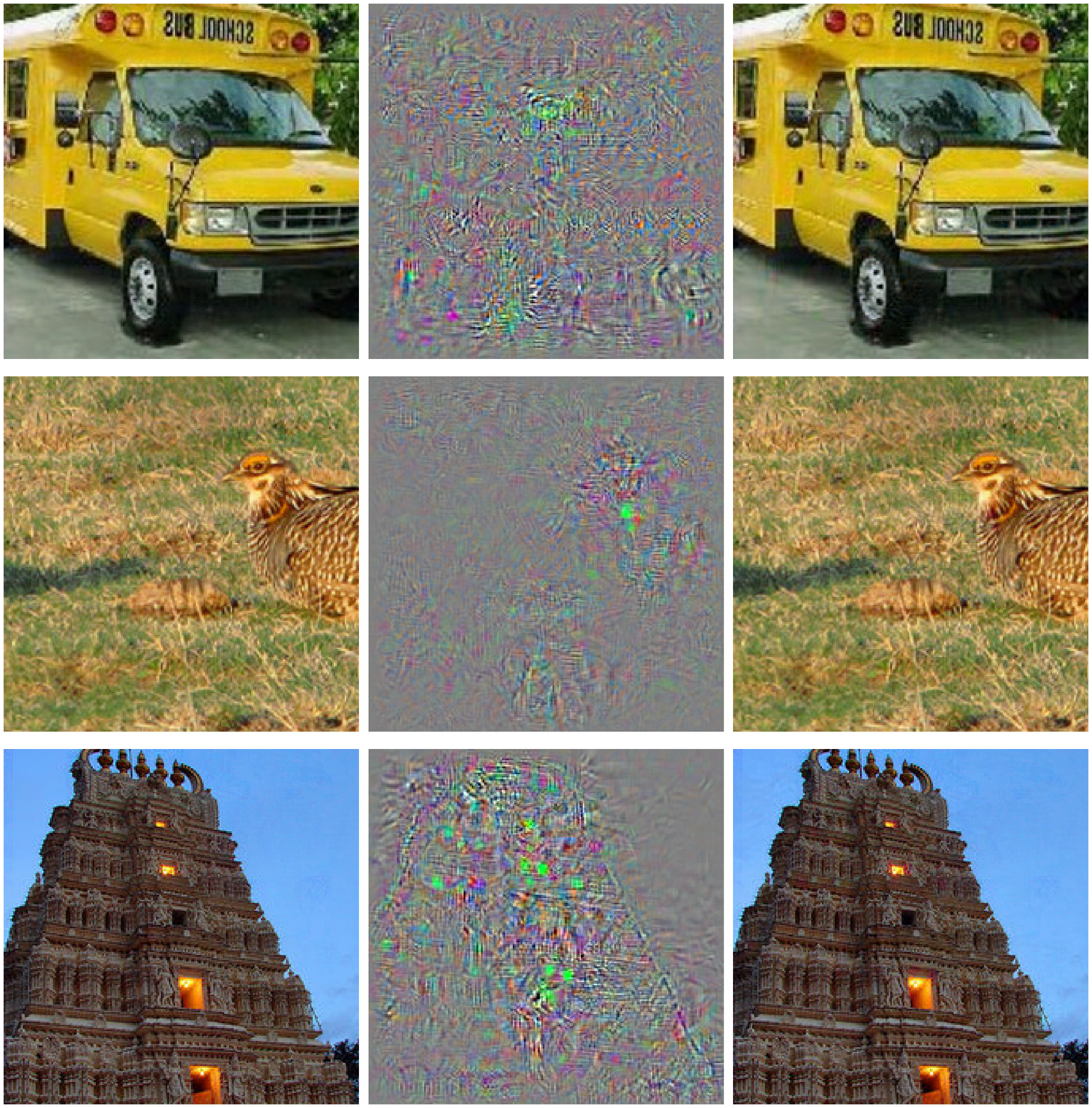}
            \subcaption{Adversarial examples~\cite{szegedy2013intriguing}.}
            \label{adv_exam}
		\end{subfigure}
	\caption{An illustration of crafting an adversarial examples (a)~--~equivalent to moving the data-point corresponding to an input from the region of a correct class to an incorrect class and (b) an illustration of adversarial examples with very low perturbations~--~the pictures in the first column are correctly classified inputs, while the pictures in the last column are adversarial examples, crafted to cause a misclassification. In between, the perturbation used to cause the misclassification is illustrated.}%
    \label{fig:adv_examples}%
\end{figure}
\fi

Because the perturbation is not perceivable by human beings, many publications claimed protecting against adversarial examples is important from a security perspective.
Therefore, before presenting the algorithms used to craft adversarial examples, we first discuss the threat models and position the field of adversarial examples in a security context.

\else 

\fi
    \if 0\mode
\section{Threat Models}
\label{sec:attack_models}

While most publications neglect the threat model when discussing attacks using adversarial examples, we chose to introduce the attacker's goals and capabilities before discussing the attacks because this information \if 0\mode can shape \else helps shaping \fi an initial taxonomy.
Moreover, on the same basis, we introduce the defender's \if 0\mode goals and \else \fi capabilities.
Towards the end of this section, we discuss the need for clear threat models whenever security consequences of adversarial examples are claimed.


\subsection*{Attacker Goals}

Threats are defined in regard to an objective that must be defended.
In our case, "the integrity of the classification is of paramount importance"~\cite{papernot2016limitations}, but not the only property to be defended~\cite{Papernot2018}.
Some attackers target the availability or the confidentiality of a model.
In the case of adversarial examples, an attacker targets the integrity of a classifier at \emph{inference} time, supplying an input that causes an incorrect output.
In practice, an attacker can cause:
\begin{enumerate}
	\item Confidence reduction. In this case an attacker can reduce the output confidence score of a classifier, thus introducing class ambiguity,
	\item Random misclassification. In this case an attacker modifies an input in order to output \emph{any class} different than the correct one.
	\item Targeted misclassification. In this case an attacker modifies an input in order to output a \emph{specific, target, class}.
\end{enumerate}
\noindent
Although confidence reduction is useful in active learning contexts, it is not explored in the adversarial example literature.

Adversarial examples consist in \emph{modifying} - or crafting - an input before sending it to a classifier.
When the input is known, an attacker strives to modify it as little as possible.
Thus, the input has common elements with a real object.
Other methods from literature generate examples that are not recognisable by human observers and can still lead to a misclassification~\cite{nguyen2015deep}.
The focus of this \if 0\mode report \else paper \fi is on adversarial examples that resemble real life objects because they are the main driver of this research field.
However, we do not neglect the case of un-recognisable images.

\subsection*{Attacker Capabilities}
As mentioned in the past section, an attacker can craft an adversarial example by modifying a normal input or by generating it from noise.
Although crafting an adversarial example by applying a very small perturbation to a normal input will make it not recognisable by human observers, in most cases no humans will supervise a \ac{ML} algorithm.

Attackers can be characterised using the information they can use and the actions they can take.
In the case of \ac{DNN}, the information available to an attacker is related to the training data or the neural network's architecture and parameters.
Since the development of \ac{DNN} models is tightly coupled with powerful hardware, the capability of an attacker can be limited by her inability to train a large model.
However, we exclude this restriction and only consider attackers that are not bounded by hardware constraints.
Therefore, the information available for an attacker, that limits her capabilities is:
\begin{itemize}
	\item Training data, \ac{DNN} architecture and hyper-parameters (\emph{white-box} scenario). In this case an attacker knows everything about the target \ac{DNN}~(architecture, hyper-parameters, weights, \etc), has access to the training data set and knows about any defence mechanisms employed (\eg~adversarial detection systems). Therefore, an attacker has the ability to \emph{completely} replicate the model under attack.
	\item Knowledge of the training data, the \ac{DNN} architecture, or \emph{some} knowledge about the defences employed (\emph{grey-box} scenario). In this case the attacker can collect some information about the network's architecture (\eg~she knows a certain model uses an open-source architecture), she knows the model under attack was trained using a certain data set (\eg~the ImageNet~\cite{krizhevsky2012imagenet} data set is very common for object recognition) or she has information about some defence mechanisms. In any of these cases, the information is neither complete or certain and provides the attacker an ability to \emph{partially} simulate the model under attack.
	\item No knowledge at all (\emph{black-box} scenario). In this case an attacker does not know anything about the model under attack, however, she has the ability to use the model as an \emph{oracle}. Therefore, an attacker can supply \emph{limited} or \emph{unlimited} inputs and collect output information.
\end{itemize}
\noindent
An attacker might also have access to pairs of inputs and outputs collected from a classifier, but no ability to modify the inputs or test them with an oracle.
This scenario can help to extract some information about the model, by reverse-engineering the data, however it is not explored in the field of adversarial examples.

\subsection*{Defender Capabilities}
\label{subsec:defender_goals}
While some publications take into consideration the attacker's capabilities, \eg~\cite{huang2017adversarial, kurakin2016adversarial, papernot2016limitations},  almost none discuss the defender's.
An incipient taxonomy is presented in~\cite{Papernot2018}, however, it is far from complete.
We consider equally important to define the capabilities of a defender, whenever designing a defence.
As in the case of attackers, defenders aim to protect the model at \emph{inference} time.
However, a defender can choose different processing stages to work with:
\begin{itemize}
	\item Input pre-processing. In this case, the defender aims to apply some techniques before the inputs reach the \ac{ML} model. Given that a defender knows the model in details, she can train a classifier to either spot adversarial examples early in the pipeline or apply different pre-processing techniques (\eg~geometrical transformations) that can alleviate the impact of adversarial examples.
	\item Hide (obscure) relevant information for attackers. In some cases the defender aims to mitigate weak points of the model that can be exploited by attackers. A common method to generate adversarial examples is to exploit sensitive features, estimated using gradients taken \wrt~to an input. It follows that a natural class of defences seeks to reduce these sensitivities and minimise the gradients during the learning phase. These defences only minimise the impact of adversarial examples and can be defeated by adaptive attacks (as discussed in Section~\ref{subsec:overall_defences}).
	\item Model hardening against small perturbations. A large body of publications focuses on defences against small, unrecognisable, perturbations. Although impossible to detect by human observers, such perturbations can easily fool \ac{DNN}. In this case, the defender can alter the model in any way, aiming to improve its robustness against small perturbations.
	\item Native defence against all types of adversarial examples. This category is, at the moment, hypothetical. It is not yet clear what the existence of adversarial examples tells us about state-of-the-art \ac{DNN} and either it can help us to develop robust \ac{ML} models (or models that will better resemble our perception system). This category leaves room for models that not only aim to defend against small perturbations, but against any kind of perturbations.

\end{itemize}


\subsection*{The security of adversarial examples in real world scenarios}
Although it was shown that adversarial examples persist during the image acquisition process~\cite{kurakin2016adversarial} and can  be deployed in real-life scenarios~\cite{eykholt2018robust}, the economics of using adversarial examples are often mistaken.
In particular, it is not clear if an attacker would prefer an attack through adversarial examples over other methods (some with no machine learning components)~\cite{gilmer2018motivating}.
For example, in~\cite{eykholt2018robust, lu2017standard} the attacker makes small changes to a stop sign in order to fool autonomous vehicles.
While generating adversarial examples is one way of performing this attack, obscuring the stop sign or going as far as removing it are somehow easier choices.

Moreover, the motivation for minimal or very small perturbation is often over-emphasised from a security \if 0\mode perspective \else standpoint\fi.
Indeed, human observers can spot big perturbations and act accordingly, however, if machines can not, why enforce this requirement?
Until motivated by real-world scenarios, this constraint is confusing and adds clutter to the field.
In some cases, the adversary will benefit from a less distinguishable attack - for example the attacker could use the same attack for a longer time, before being detected - however this adds as a requirement for the flexibility of an attack~\cite{gilmer2018motivating} and is not always a strong requirement.

Nevertheless, minimal perturbations become important when the property of \emph{robustness} is discussed.
There are real-life scenarios such as sensor wear or small changes in distributions (\eg~sales in some region), which should not drastically impact the behaviour of a model.
We argue these use cases belong to the field of \emph{robustness}, sometimes mistakenly called \emph{safety}, and not security.
Therefore, in the next section we present the safety perspective of adversarial examples and introduce the property of robustness, as discussed in literature.

For a complete taxonomy and overview of real-life scenarios where adversarial examples can be used, see~\cite{gilmer2018motivating}.
In order to claim security consequences for adversarial examples, it is important to specify a complete description of the scenarios and threat models involved.
Moreover, it is important to discuss the economics of using adversarial examples, compared to other methods of achieving the same objective.
The goal of this paper is to characterise the complete research field; therefore, we do not omit defence mechanisms that are poorly motivated or, sometimes, impossible to find in real life scenarios.
However, we signal these drawbacks whenever possible.

\else

\section{Threat Model and Taxonomies of Attacks and Defenses}
\label{sec:taxonomy}

For any meaningful discussion of security it is crucial to have a
clear description of a \emph{threat model}, aka an \emph{attacker
model}, which describes the goals
of attacker~--~\ie~what does the attacker want to achieve~--~along
with a description of the attacker's capabilities and knowledge~--~\ie~what means does the attacker have to make that happen.

Adversarial examples are only one possible attack vector on \ac{ML}
systems. Instead of targeting a system at \emph{inference} time by
feeding it an adversarial example, an attacker could also try to
compromise the system in the \emph{training} phase. This obviously
requires different capabilities of the attacker, namely the ability to
influence or compromise the training set.  Attackers may have
other objectives than attacking the correct functioning of a system.
For example, an attacker might be interested in obtaining information
about the model and reverse engineering some of its parameters~--~which
is an attack on confidentiality rather than integrity.

This paper only consider attacks at inference time, \ie~attacks with
adversarial examples. A more general threat model, which also considers
attacks on the learning phase, was proposed in~\cite{barreno2010security}.
A more recent
and comprehensive threat model is given by \citeauthor{Papernot2018}
\cite{Papernot2018}.
We introduce a new way to classify attacks and defenses
not considered in these earlier publications:
in sections~\ref{sec:taxonomy_attacks} and ~\ref{sec:taxonomy_defense}
we introduce a notion of \emph{strategy} to classify different techniques
to generate adversarial examples or protect against them,
providing a taxonomy we use to classify existing research.

Whether or not a threat model is a good threat model for a specific
system~--~\ie~whether it is realistic, relevant, and complete~--~is a
separate issue: a threat model defines a \emph{hypothetical} attacker
which may not have any bearing on attackers out there in the real
world.  Threat modeling depends on the specific application and its
context: these have to be known to do a good risk analysis, which
should also consider impacts, efforts, and possibilities to recover
from attacks.  This topic will be further discussed in
Section~\ref{sec:discussion}.

\subsection{Taxonomy of adversarial attacks.}
\label{sec:taxonomy_attacks}

The basic threat model outlined above can be refined in different
ways, depending on the attacker's goal or the attacker's
knowledge, as discussed below.

\paragraph*{Attacker Goal}
The basic goal of causing misclassifications can be further refined in:
\begin{itemize}
 \item \emph{Untargeted attacks}, where it is the attacker's goal
      to produce an input that will be misclassified as \emph{any}
      incorrect class, and
 \item \emph{Targeted attacks}, where the input is incorrectly
      classified as a \emph{specific} incorrect class.
\end{itemize}
This distinction is also called \emph{error specificity} in~\cite{Biggio2018}.
Targeted attacks are also called source-targeted attacks in~\cite{papernot2016limitations}.

The distinction above considers the outputs of the \ac{ML} algorithm
that the attacker is interested in. An orthogonal distinction can be
made by considering inputs the attacker is interested in: the
attacker's goal may be to simply misclassify any input, but it may
also be to misclassify a specific input, or an input from a specific
set (for example, those inputs that should be classified as some
specific class).  This distinction is also called \emph{attack
specificity} \cite{Biggio2018}.
In most cases, adversarial examples search for perturbations specific to one
input drawn from the data generation distribution.
Therefore, we do not consider the attack specificity in this paper.

\paragraph*{Attacker Knowledge}
When it comes to the attacker's knowledge, a common distinction is
between a \emph{white box} scenario, where the attacker has
complete knowledge of the model, its parameters, and can completely
replicate the model under attack, and a \emph{black box}
scenario, where the attacker has no knowledge of the model and only has
access to query the system \cite{papernot2016limitations, Papernot2018}.  Note that the system here also
encompasses any preprocessing of raw inputs before these are fed to
the \ac{ML} algorithm.
For the black box scenario, one can then still make different
assumptions about the attacker's ability to query the model
indefinitely or only a limited numer of, to access the output
probability distribution or the final class, \etc

\citeauthor{Biggio2018}~\cite{Biggio2018} also consider the
gray-box scenario, in which an attacker has only partial knowledge
about the model.  However, this scenario is not common in the
adversarial examples literature and it is often seen as a special case
of the black box scenario, in which the attacker has some restrictions (as suggested earlier)~\cite{oneval}.

\paragraph*{Attack Strategies}
We use the notion of attack strategy to classify ways to construct
adversarial examples. This involves two aspects: (1) what types of
perturbations can an attacker use? and (2) which classes of algorithms
are then used to find interesting perturbations?  Regarding the first question, we distinguish
between perturbations based on \emph{noise} and perturbations based on
\emph{geometric transformations}.  Methods in the first class involve
adding white noise to specific areas of an image,
while methods from the second class use natural geometric
transformations~--~\eg~rotations or translations~--~to induce
misclassifications.  These two perturbation types have been used until
now, but others may well exist.
Searching for new perturbation types is an active research area and will be later discussed in Section~\ref{sec:discussion} and Section~\ref{sec:conclusions}.

Regarding the second question, we distinguish between three classes of
algorithms:
\begin{itemize}
	\item \emph{Optimization.} Here attackers use optimization algorithms to search for solutions to \eqq~(\ref{eq:adversarial_generic}), alternative forms or constraints.
	\item \emph{Sensitivity Analysis.} Here attackers use sensitivity analysis~--~a class of algorithms used to determine the contribution of each input feature to the output~--~ in order to find sensitive features and perturb them.
	\item \emph{Generative.} Here the probability distribution of adversarial perturbations is learned using generative models and used to sample new adversarial examples.
\end{itemize}

\begin{table}[t]
	\centering
	\scalebox{0.52}{
    \begin{tabu}{|c|c|c|c|c|c|c|c|c|c|c|c|c|} 
    \hline
    \toprule%
    \rowfont{\bfseries\large} \multirow{4}{*}{Attack} & \multicolumn{2}{c|}{Attacker Goal} & \multicolumn{2}{c|}{Attacker Knowledge} & \multicolumn{4}{c|}{Attack Strategy} & \multicolumn{4}{c|}{\specialcell{Attack Performance}} \\ \cline{2-13}

	 & \multirow{2}{*}{Untargeted} & \multirow{2}{*}{Targeted} & \multirow{2}{*}{White Box} & \multirow{2}{*}{Black Box} & \multicolumn{3}{c|}{Noise} & \multicolumn{1}{c|}{Geometric} & \multirow{2}{*}{Strength}& \multirow{2}{*}{Complexity} & \multirow{2}{*}{ \specialcell{Experimental \\ Setup}} & \multirow{2}{*}{ \specialcell{Research \\ Impact}} \\ \cline{6-9}

    & & & & &    Optimization & \specialcell{Sensitivity \\ Analysis} & Generative  & Optimization & & &  & \\ \hline
    
    L-BFGS~\cite{szegedy2013intriguing} & - & x & x &  - & x & - & - & - & *** & *** & *** & *** \\ \hline 
    Deep Fool~\cite{moosavi2016deepfool} & x & - & x &  - & x & - & - & - & * & *** & *** & *** \\ \hline 
    UAP~\cite{moosavi2017universal} & x & - & x &  - & x & - & - & - & *** & ** & *** & **  \\ \hline 
    Carlini~\cite{carlini2017towards} & x & x & x &  - & x & - & - & - & *** & *** & *** & ***   \\ \hline 

    FGS~\cite{goodfellow2014explaining} & x & - & x &  - & - & x & - & - & * & * & ** & ***  \\ \hline 
    JSMA~\cite{papernot2016limitations} & x & x & x & - & - & x & - & - & * & *** & ** & ***   \\ \hline 
    STA~\cite{huang2015learning} & - & x & x & - & - & x & - & - & ** & *** & * & *  \\ \hline   
    SV-UAP~\cite{khrulkov2017art} & - & x & x & - & - & x & - & - & * & *** & ** & *   \\ \hline 
   	RSSA~\cite{tramer2017ensemble} & x & - & x & - & - & x & - & - & * & * &  ***& ***  \\ \hline 
    BPDA~\cite{athalye2018obfuscated} & - & x & x & - & - & x & - & - & *** & *** & *** & ***  \\ \hline 
    Elastic-Net~\cite{chen2017ead} & - & x & x & - & - & x & - & - & *** & ***  & ** & ** \\ \hline 
    BI~\cite{kurakin2016adversarial} & x & - & x & - & - & x & - & - & ** & ** & *** & ***  \\ \hline 
    ILC~\cite{kurakin2016adversarial} & - & x & x & - & - & x & - & - & ** & ** & *** & ***   \\ \hline 
    Madry~\cite{madry2017towards} & x & - & x &  - & - & x & - & - & *** & ** & *** & ***  \\ \hline  
    Momentum~\cite{dong2017boosting} & x & - & x & - & - & x & - & - & ** & ** & *** & **  \\ \hline 
    
    ATN~\cite{baluja2017adversarial} & x & x & x & - & - & - & x & - & ** & *** & ** & * \\ \hline 
    NAE~\cite{zhao2017adversarial} & - & x & x & - & - & - & x & - & ** & *** & ** & *  \\ \hline 
	Univ. GM~\cite{poursaeed2018generative} & - & x & x & - & - & - & x & - & ** & *** & *** & *  \\ \hline 
	Unrestr. GM~\cite{song2018constructing} & - & x & x & - & - & - & x & - & *** & *** & *** & *  \\ \hline 

    ManiFool~\cite{kanbak2017geometric} & x & x & x & - & - & - & - & x & ** & *** & ** & *  \\ \hline 
    \specialcell{Spatial Tr.}~\cite{xiao2018spatially} & - & x & x & - & - & - & - & x & *** & ** & ** & * \\ \hline 
    \specialcell{Expectation}~\cite{athalye2017synthesizing} & - & x & x & - & - & - & - & x & ** & ** & ** & *** \\ \hline 
    \specialcell{Formal Tr.}~\cite{pei2017towards} & - & x & x & - & - & - & - & x &*** & ** & *** & * \\ \hline 
    \specialcell{Rotation Tr.}~\cite{engstrom2017rotation} & x & - & x & x & - & - & - & x & ** & * & ** & * \\ \hline 
     
    Grad. Est.~\cite{bhagoji2017exploring} & x & x &  - & x & x & - & - & - & *** & ** & ** & *   \\ \hline 
    ZOO~\cite{chen2017zoo} & x & x &  - & x & x & - & - & - & *** & *** & ** & *   \\ \hline 
   	IS~\cite{narodytska2017simple} & x & x &  - & x & x & - & - & - & * & * & ** & *  \\ \hline 
	Substitute~\cite{papernot2017practical} & - & x & - & x & - & x & - & - & ** & *** & *** & ***  \\ \hline 
    \end{tabu}}
	\caption{Catalog of adversarial attacks following the taxonomy introduced in Section~\ref{sec:taxonomy_attacks} and the quality attributes defined in Section~\ref{sec:attacks}.}
	\label{tbl:attacks}
\end{table}

\paragraph*{Classification of Attacks}

Table~\ref{tbl:attacks} classifies representative attacks using the
taxonomy outlined above and some quality attributes that will be discussed
later, in Section~\ref{sec:attacks}.
Note that untargeted and targeted attacks are approximately equally distributed, suggesting both goals have been explored in depth.
There is clearly more research on white box attacks than
black box attacks, and more attacks use noise perturbations
than geometric transformations.
Maybe it is not so surprising that research has concentrated on white
box attacks: here there is more information, and hence more opportunities
to explore how to use this. But note that in many attack scenarios it
is more realistic that the attacker does not have full knowledge of
the system under attack; for these, research into black box attacks is
much more relevant.

Zooming in to noise based perturbations, we observe that most attacks make use of sensitivity analysis.
There may be several reasons for it. Firstly, these attacks are, in general, faster than optimization based attacks.
Therefore, they are better suited to be incorporated in the training process of \ac{ML} models and used to improve their robustness.
They are also simpler than optimization based attacks, which rely on different constructs~--~\eg~L-BFGS~--~than commonly used in training or analyzing neural networks~--~\eg~gradient descent or the Jacobian matrix.
Secondly, generative attacks have generally received less attention in literature.

Although attacks based on sensitivity analysis are more common, these require full knowledge of the system under attack.
Even the black box approach of \citeauthor{papernot2017practical} \cite{papernot2017practical}, called `substitute' in Table~\ref{tbl:attacks}, an attacker trains a white box model and uses it to create adversarial examples, which are then transferred to a black box model.
In contrast to sensitivity-based methods, optimization-based attacks are used more in black box scenarios.
This result is not unexpected: without white box access to an algorithm it is hard to perform sensitivity analysis.
However, an optimizer can still minimize an objective by sending queries to a black box algorithm and use various constraints to reflect the operational environment~--~\eg~limited number of queries.

We observe that no attacks based on generative methods are used in black box settings.
Recall that generative models involve learning the probability distribution of adversarial perturbations.
Without access to any data, learning the underlying probability distribution is difficult.
As in the substitute approach of \citeauthor{papernot2017practical}, one can train a substitute  generative model and try to transfer the examples to other algorithms.
However, this scenario has not yet been explored.

Regarding attacks based on geometric transformations, we observe that these only use optimization methods and are usually applied in white box scenarios.
The reason for making extensive use of optimization methods is the constraint on the perturbation size: these attacks search for a very small perturbation which should not alter the overall geometry of the scene.
We can imagine that rotating an image of the digit six by 180 degrees will generate a misclassification.
Maybe similar scenarios can be found for images with objects, however, the goal now is to find a very small perturbation which does not change the scene.
This goal can be more easily formulated as an optimization problem and solved by an optimizer.

\subsection{Taxonomy of Defenses}
\label{sec:taxonomy_defense}

A defender can be \emph{reactive} and improve the system \emph{in response} to
new attacks as these are discovered or \emph{proactive} and try to \emph{anticipate} attacks and design the system with security in mind.
A disadvantage of reactive security is that it can only protect
against known attacks.
This distinction has been used for \ac{ML} security~\cite{biggio2012poisoning, liu2018survey}. However, the field of adversarial examples mainly focused on defenses against perturbations in the p-norm ball around one input.
Since this threat is already known, one can argue that most literature  focuses on reactive defenses. A scalable and flexible solution to this threat has not yet been found, so this line of research is still ongoing.
Therefore, we classify defenses only in terms of the defense strategy,
a notion introduced below,
which is similar to notion of the attack strategy from Section~\ref{sec:taxonomy_attacks}.
This classification is meant to give an overview of the large volume and highly varied work that has been done on this topic.
It is worth mentioning that protecting against perturbations in the p-norm ball is not the only requirement to guarantee security, as argued in a series of publications~\cite{gilmer2018motivating, oneval, papernot2018marauder}.
This issue is discussed in Section~\ref{sec:discussion}.
Moreover, most defenses proposed lead to a false sense of security because they assume an attacker does not know a defense is employed. This threat is discussed in Section~\ref{subsec:gradient_ob}.

\paragraph*{Defense Strategies}
\label{subsec:defense}
Similar to the attack strategies, defense strategies describe the types of algorithms used to defend against adversarial examples.
We start by classifying defenses based on their place in the processing pipeline.
Some defenses act early in the pipeline, before an input reaches the model, while others strengthen the model directly (and are part of it).
We call the first class of defenses \emph{guards} because they do not interact with the  under attack and only build precautions around it.
The second class of defenses acts directly on the model, by modifying its architecture, the training data or the loss function.
Therefore, we call them \emph{defenses by design}.

We decompose these two classes further based on the types of algorithms used:
\begin{itemize}
	\item \emph{Guards:}
		\begin{itemize}
			\item \emph{Detection.} These methods assume that adversarial examples have special characteristics or are sampled from different data distributions than normal inputs. Therefore, we can train a separate detector to identify and discard them.
			\item \emph{Input Transformation.} Defenses in this class use pre-processing techniques such as compression or bit-depth reduction in order to remove the effect of adversarial perturbations and diminish their impact on the system under attack.
		\end{itemize}
	\item \emph{Defense by Design:}
	\begin{itemize}
		\item \emph{Adversarial Training.} Given that learning is a data driven process, a normal defense strategy is to include adversarial examples in the training process.
		\item \emph{Architectural Defenses.} Another strategy is to design new architectures and models with constraints related to adversarial examples, such as custom regularization techniques.
		\item \emph{Certified Defenses.} An interesting approach to defend against adversarial examples is to use formal verification to certify that within some bounds no adversarial examples exist.
	\end{itemize}
\end{itemize}

\begin{table}[t]
	\centering
	\scalebox{0.605}{
	\begin{tabu}{|c|c|c|c|c|c|}
		\toprule%
   \rowfont{\bfseries\large}  \multirow{2}{*}{Defense} & \multirow{2}{*}{Defense Strategy} & \multicolumn{4}{c|}{Defense Performance} \\ \cline{3-6}

	&  & \multirow{1}{*}{Defense Strength} & \multirow{1}{*}{Defense Complexity} & \multirow{1}{*}{Experimental Setup} & \multirow{1}{*}{Research Impact} \\


		\midrule
		Statistical Detection~\cite{grosse2017statistical} & \emph{Guard} - Adversarial Detector & * & ** & ** & ** \\ \hline 
		Binary Classification~\cite{gong2017adversarial} & \emph{Guard} - Adversarial Detector & * & ** & * & * \\ \hline 
		In-Layer Detection~\cite{metzen2017detecting} & \emph{Guard} - Adversarial Detector & * & **  & ***  & **  \\ \hline 
		Detecting from Artifacts~\cite{feinman2017detecting} & \emph{Guard} - Adversarial Detector & * & ** & ** & ** \\ \hline 
		SafetyNet~\cite{lu2017safetynet} & \emph{Guard} - Adversarial Detector & * & ** & ** & * \\ \hline 
		Convolutional Statistics Detector~\cite{li2017adversarial} & \emph{Guard} - Adversarial Detector & * & ** & ** & * \\ \hline 
		Saliency Data Detector~\cite{zhang2018detecting} & \emph{Guard} - Adversarial Detector & * & ** & * & * \\ \hline 
		Ensemble Detectors~\cite{abbasi2017robustness} &  \emph{Guard} - Adversarial Detector & * & ** & * & * \\ \hline 

		MagNet~\cite{meng2017magnet} & \emph{Guard} - Adversarial Detector & * & ** & *** & ** \\ \hline 
		Generative  Detector~\cite{lee2017generative} & \emph{Guard} - Adversarial Detector & * & ** & * & *  \\ \hline 
		PixelDefend~\cite{song2017pixeldefend}  & \emph{Guard} - Adversarial Detector & * & ** & *** & *  \\ \hline 
		VAE Detector~\cite{ghosh2018resisting} & \emph{Guard} - Adversarial Detector & * & *** & ** & * \\ \hline 
		Bit-Depth~\cite{guo2017countering} & \emph{Guard} - Input Transformation & * & * & ** & ** \\ \hline 
		Basis Transformations~\cite{shaham2018defending} & \emph{Guard} - Input Transformation & * & *  & ** & * \\ \hline 
		Randomized Transformations~\cite{xie2017mitigating} & \emph{Guard} - Input Transformation & * & * &  *** & * \\ \hline 
		Thermometer Encoding~\cite{buckman2018thermometer} & \emph{Guard} - Input Transformation & * & * & *** & * \\ \hline 
		Blind Pre-Processing~\cite{rakin2018blind} & \emph{Guard} - Input Transformation & * & * & * & * \\ \hline 
		Data Discretization~\cite{chen2018improving} & \emph{Guard} - Input Transformation & * & * & ** & * \\ \hline 
		Adaptive Noise~\cite{liang2017detecting} &  \emph{Guard} - Input Transformation & * & * & * & * \\ \hline 
		FGSM Training~\cite{goodfellow2014explaining} & \emph{Design}~--~Adversarial Training &  * & * & ** & *** \\ \hline 
		Gradient Training~\cite{sinha2018gradient} & \emph{Design}~--~Adversarial Training & * & * & * & * \\ \hline 
		Gradient Regularization~\cite{lyu2015unified} & \emph{Design}~--~Adversarial Training  & * & * & * & * \\ \hline 
		Structured Regularization~\cite{roth2018adversarially} & \emph{Design}~--~Adversarial Training & * & * & ** & * \\ \hline 
		Robust Training~\cite{shaham2015understanding} & \emph{Design}~--~Adversarial Training & ** & * & ** &  ** \\ \hline 
		Strong Adversary Training~\cite{huang2015learning} & \emph{Design}~--~Adversarial Training  & * & ** & * & ** \\ \hline
		Madry~\cite{madry2017towards} & \emph{Design}~--~Adversarial Training  & *** & ** & *** & *** \\ \hline 
		Ensemble Training~\cite{tramer2017ensemble} & \emph{Design}~--~Adversarial Training & ** & ** & ** & *** \\ \hline 
		Stochastic Pruning~\cite{dhillon2018stochastic} & \emph{Design}~--~Adversarial Training & ** & ** & ** & ** \\ \hline 
		Distillation~\cite{papernot2016distillation} &  \emph{Design}~--~Architecture & * & ** & ** & *** \\ \hline 
		Parseval Networks~\cite{cisse2017parseval} & \emph{Design}~--~Architecture & * & ** & ** & ** \\ \hline 
		Deep Contractive Networks~\cite{gu2014towards} & \emph{Design}~--~Architecture & * & *** & ** & ** \\ \hline 
		Biological Networks~\cite{nayebi2017biologically} & \emph{Design}~--~Architecture & * & * & *** & * \\ \hline 
		DeepCloak~\cite{gao2017deepcloak} & \emph{Design}~--~Architecture & * & * & * & *  \\ \hline 
		Fortified Networks~\cite{lamb2018fortified} & \emph{Design}~--~Architecture & ** & ** & ** & * \\ \hline 
		Rotation-Equivariant Networks~\cite{dumont2018robustness} & \emph{Design}~--~Architecture & * & * & * & * \\ \hline 
		HyperNetworks~\cite{sun2017hypernetworks} & \emph{Design}~--~Architecture & * & *** & * & * \\ \hline 
		Bidirectional Networks~\cite{pontes2018bidirectional} & \emph{Design}~--~Architecture &  * & ** & * & * \\ \hline 
		DAM~\cite{krotov2016dense} & \emph{Design}~--~Architecture & ** & ** & ** & * \\ \hline 
		Safety Verification~\cite{huang2017safety} & \emph{Design}~--~Certified & *** & *** & *** & ** \\ \hline 
		Reluplex~\cite{katz2017reluplex} & \emph{Design}~--~Certified & *** & *** & *** & *** \\ \hline 
		Planet~\cite{ehlers2017formal} & \emph{Design}~--~Certified & ** & *** & * & *  \\ \hline 
		Convex polytope~\cite{wong2017provable} & \emph{Design}~--~Certified & ** & *** & ** & * \\ \hline 
		Dual~\cite{dvijotham2018dual} & \emph{Design}~--~Certified & *** & *** & *** & *  \\ \hline 
		Abstract Interpretation~\cite{mirman2018differentiable}  & \emph{Design}~--~Certified & *** & *** & *** & * \\ \hline 
		Interval Bound~\cite{gowal2018effectiveness} & \emph{Design}~--~Certified & *** & ** & *** & * \\ 

	\bottomrule
	\end{tabu}
	}

	\caption{Catalog of defenses against adversarial examples following the taxonomy introduced in Section~\ref{subsec:defense} and the quality attributes from Section~\ref{sec:defences}.}
	\label{tbl:defences_acm}
\end{table}

\paragraph*{Classification of Defenses}

Table~\ref{tbl:defences_acm} classifies representative defenses using
the taxonomy outlined above and the quality attributes
discussed later, in Section~\ref{sec:defences}.

Note that all strategies are well represented, which means they all showed some potential to defend against adversarial examples and are worth looking into.
Guards are of interest because they do not impose any restrictions on the \ac{ML} algorithm we want to defend.
In particular, adversarial detectors exploit perturbation specific characteristics in an attempt to detect adversarial examples and discard them.
These techniques are suited for scenarios in which we can discard or choose not to classify an input.
However, the perturbations produced by different attacks are sometimes different and require retraining the detector.
Moreover, adversarial detectors which rely on \ac{ML} constructs can also suffer from low robustness and can be attacked with adversarial examples.
Input transformations aim to reduce the space where adversarial perturbations lie and diminish their impact.
These are lightweight techniques,~--~\eg~image compression~--~easy to apply and require low computational resources.
Such properties are  important for a defense because they make it easy to implement and adopt.
Unfortunately, as we will discuss in Section~\ref{subsec:reactive}, guards are not very effective.

Defenses by design require retraining the models adding custom changes to the training data or its architecture.
Therefore, they require more resources than guards.
Since \ac{ML} is a data driven process, a normal reaction to adversarial examples is to include them in the training set.
This method, called adversarial training, is a regularization technique used for robustness and shows good results when the space of the perturbations can be well approximated.
Moreover, adversarial training provides benefits for the model, such as more interpretable gradients~\cite{tsipras2018robustness}.

In an analogous manner, architectural defenses rely on regularization penalties designed to offer robustness against adversarial examples.
This time, however, the constraints are applied layer-wise, to the input data or to the final layer.
These constraints go beyond enhancing the training data set with adversarial examples and require new architectural designs or constraints.

The strategies presented above can only give approximate (often empirical) guarantees about their efficacy against adversarial examples.
In contrast, certified defenses borrow methods from formal verification to certify that adversarial examples can not be found within some bounds.
These defenses have great potential for improving \ac{ML} models and finding spots where they fail.
However, they are not yet scalable to deep models and often require more computational resources than other defenses.
More details about each defense strategy follow in Section~\ref{sec:defences}.

\fi
    \section{Robustness}
\label{sec:robustness_eval}

Most publications use the property of robustness as a proxy to safety or security.
Whether this is a relevant aspect for safety or security is left for the discussion in Section~\ref{sec:discussion}.
For now, we introduce robustness and discuss several methods used to measure it.

Two general definitions of robustness are valid for adversarial examples: (1) \emph{distributional robustness}, defined as insensitivity to \emph{slight deviations} of the underlying distribution from the assumed model~\cite{huber2011robust} and (2) \emph{optimization robustness}, defined as an algorithm's ability to perform well under a certain level of uncertainty in the input space~\cite{ben2009robust}.
This means the uncertainty margins are defined beforehand.
In real world applications of optimization, small uncertainty in the data can heavily affect the quality of the output, therefore, instead of deploying uncertain solutions it is recommended to deploy the associated \emph{robust counterpart}~\cite{ben2009robust}.

Formally, given a class of distributions $\gP$ around the data generation distribution $\gD \sim \ \gP_i$, distributional robustness is defined as:
\begin{equation*}
	\E_{(\vx, y)\sim \gP_i}[l(f(\vx), y)] \simeq  \E_{(\vx, y)\sim \gD}[l(f(\vx), y)].
\end{equation*}
\noindent
The choice of $\gP$ can influence the robustness guarantee and the ability to compute it.
For example, one can choose a family of distributions defined on a convex metric space around the empirical distribution, measured using a metric on this space (\eg~relative entropy).
The robust counterpart of this problem can be formulated as $\hat{f} = \argmin_{f \in \gF} \E_{(\vx, y) \sim \gP} [ \max l(f(\vx), y)]$, which is similar to \eqq~(\ref{eq:emprirical}), but the minimization is performed on the maximum loss given training data sampled from the class of distributions we want to provide robustness for.
In the case of adversarial examples, the samples from $\gP_i$ account for input data in close resembles with data sampled from $\gD$, but perturbed with perturbations equivalent to solving \eqq~(\ref{eq:adversarial_generic}).

Optimization robustness aims to protect against a strict set of perturbations around an input $\vx$, defined using a distance function $d(\cdot)$ on the input space $\gX$:
\begin{equation}
	\label{eq:uset}
	\gU = \{\vxadv | d(\vx, \vxadv) \leq \epsilon\},
\end{equation}
\noindent
where $\epsilon$ controls the set size.
Similar to distributional robustness, the robust counterpart is defined as:
\begin{equation}
	\label{eq:minmax}
	\hat{f} = \argmin_{f \in \gF} \E_{(\vx, y) \sim \gS} [ \max_{\vxadv \in \gU_{}} l(f(\vxadv, y)],
\end{equation}
\noindent
where $\vxadv$ is a realization of $\vx$ in the uncertainty set described by $d(\cdot)$.
The  metric $d(\cdot)$ and the size of $\epsilon$ from~\eqq~(\ref{eq:uset}) control the size and the direction of the perturbation, and should account for perturbation equivalent to solving~\eqq~(\ref{eq:adversarial_generic}).
Minimizing on solutions (or approximations) of the inner maximization problem in~\eqq~(\ref{eq:minmax}) increases the robustness of models against perturbations from $\gU$.
In fact, solutions to eq.~(\ref{eq:minmax}) result in state-of-the-art defenses, as will be discussed later, in Section~\ref{sec:defences}.
A discussion about $d(\cdot)$ was already provided in Section~\ref{sec:background}.

\if 0\mode
At this point we can draw similarities between robustness and factors of safety - described as the capability of a system to perform well beyond maximum loads.
For example, we would like an autonomous vehicle to perform well in a temperature interval of $\pm 40\deg$.
However, if the temperature exceeds the upper bound by $1\deg$, the system should perform reasonably well and not instantly crash.
Safety factors, however, assume some bounds are known and the behavior within these bounds is proven to be safe.
Therefore, some methods to guarantee safety are still needed.
We advise caution when using the concept of safety as a verifiable property of an algorithm.
\else\fi

Judging adversarial examples through these lenses, we observe that (1) according to some publications (guards-detectors) adversarial examples violate the \ac{iid} assumption and belong to a class of distributions $\gP$ dissimilar to $\gD$, for which  \acp{DNN} do not provide distributional robustness in standard training settings and (2) in order to build robust models, the uncertainty bounds have to be defined up-front \stt~the training procedure is adjusted for robustness.
In this context, it is important to decide if one wants to guarantee performance for inputs drawn from a different distribution or only within some known bounds.
The problem is context dependent and a scalable, universal, solution is missing for the moment.

\if 0\mode \bigskip \else  \newpara  \fi

Practical robust counterparts for linear models or \acp{SVM}~\cite{sra2012optimization} rely on adding penalty terms to the loss function and have been used to protect against adversarial examples~\cite{demontis2016security, russu2016secure}.
However, for tasks where complex and highly non-linear models are used~--~such as object recognition~--~finding a robust counterpart is often complex or intractable~\cite{huang2015learning}.
The natural question that rises is how to measure, quantify and test robustness of these models.
The solutions presented in this paper rely on finding lower or upper bounds to it, as follows:
\begin{itemize}
	\item \emph{Lower bound.} The minimum space around an input (defined by a distance function) where no adversarial examples can be found: $\gU$~s.t.~$\forall \vxadv \in \gU, f(\vxadv) = y $.
	\item \emph{Upper bound.} The maximum size of a perturbation for which no adversarial examples can be constructed: $\epsilon$~s.t.~$\|\vx - \vxadv\| \leq \epsilon, f(\vxadv) = y$.
\end{itemize}
\noindent
\citeauthor{Biggio2018}~\cite{Biggio2018} proposed to also measure the model's accuracy while increasing the attack strength.
This method is recommended to evaluate the security of an algorithm and shows when it starts to misbehave or 'break'.
In this paper we focus on evaluating robustness using the two bounds presented above because all publications discussed use any of them.
Nonetheless, the method proposed by~\citeauthor{Biggio2018}~\cite{Biggio2018} is better suited for evaluating security and can be used to approximate an upper bound to robustness.

Several definitions and methods to measure robustness have been proposed in the literature and are discussed below.
In the adversarial examples inception paper, \citeauthor{szegedy2013intriguing}~\cite{szegedy2013intriguing} measure robustness using spectral analysis of each layer.
Under the assumption that all layers are Lipschitz continuous, one can inspect the upper Lipschitz constant for each layer.
It follows that a lower bound stability measure can be derived for a \ac{DNN} by multiplying the Lipschitz upper bounds of each layer.
However, this global Lipschitz constant often gives a very loose bound \cite{weng2018evaluating}.

\citeauthor{fawzi2018analysis}~\cite{fawzi2018analysis} propose to average over the minimal perturbations required to cause a misclassification, for each example in the data set:
\if 0\mode
\begin{equation}
	\rho(f, \vx) = \mathbb{E}_{f(\vx) \sim p_{data}} [\| \eta \|_{p}],
\end{equation}
\noindent \else $	\rho(f, \vx) = \mathbb{E}_{f(\vx) \sim \gS} [\| \eta \|_{p}],$
and provide a theoretical upper bound guarantee for linear and quadratic classifiers.
However, this approximate boundary can also be affected by distribution drifts.\fi

\citeauthor{bastani2016measuring}~\cite{bastani2016measuring} provide a formalism for lower bound robustness to adversarial examples, independent of the Lipschitz constant.
The authors abstract from robustness of a point, defined locally as $\rho(f, \vx) = \inf \{ \alpha \geq 0 \mid \|\vxadv-\vx\|_{p} \leq \alpha \text{,} f(\vxadv) \neq f(\vx) \}$, the notion of adversarial frequency: $\phi(f, \epsilon) = P_{\vx \sim \gD} [\rho(f, \vx) \le \epsilon]$  \ie~the probability mass function of a point not being robust.
%
%
\if 0\mode
%
\else
\fi
%
%
The authors also define a metric called adversarial severity, as the average minimal space where $f$ fails to be robust, conditioned by the upper bound $\epsilon$:
\if 0\mode
\begin{equation}
	\mu(f, \epsilon) = \mathbb{E}_{f(\vx) \sim \gD} [\rho(f, \vx) \mid \rho(f,\vx) \le \epsilon].
\end{equation}
\noindent
\else $\mu(f, \epsilon) = \mathbb{E}_{f(\vx) \sim p_{data}} [\rho(f, \vx) \mid \rho(f,\vx) \le \epsilon].$ \fi
\if 0\mode
\emph{Smaller} values for $\mu(f,\epsilon)$ correspond to \emph{worse} adversarial severity, because $f$ is more susceptible to misclassifications when the average distance to the nearest adversarial example is small.
Frequency and severity capture different robustness behaviors.
A neural network may have high adversarial frequency, but low adversarial severity; indicating that most adversarial examples are about $\epsilon$ distance away from the original point $\vx$.
Analogously, a neural network may have low adversarial frequency but high adversarial severity, indicating that it is typically robust, but occasionally severely fails to be robust.
Frequency is, therefore, more important, because a neural network with low adversarial frequency is robust most of the time.
\else\fi
However, the generalization of point-wise robustness still involves an upper bound on the perturbation.
\citeauthor{weng2018evaluating}~\cite{weng2018evaluating} developed a lower bound metric for robustness based on Lipschitz continuity.
CLEVER~\cite{weng2018evaluating} generalizes a metric introduced by \citeauthor{hein2017formal}~\cite{hein2017formal} for kernel methods and neural networks with only one layer.
Consider $f(\vx)$  with continuously differentiable components $f_i$ and define the class which $f$ predicts for an input $\vx_0$ as $y = \argmax_{1\leq i \leq K} f_i(\bm{x_0})$, then the lower bound robustness of $f$ is defined as:
\if 0\mode
\begin{equation}
	\label{eq:our_delta_bnd}
	\beta_L=\min_{y' \neq y} \frac{f_y(\vx_0)-f_y'(\vx_0)}{L_q^{y'}},
\end{equation}
\noindent
\else $	\beta_L=\min_{y' \neq y} \frac{f_y(\vx_0)-f_{y'}(\vx_0)}{L_q^{y'}},$ \fi
where $L_q^{y'}$ is the Lipschitz constant for the function $f_y(\vx)-f_{y'}(\vx)$ in p-norm.
\citeauthor{weng2018evaluating}~propose to use extreme value theory in order to approximate $\beta_L$.
However, \citeauthor{goodfellow2018gradient}~\cite{goodfellow2018gradient} showed that CLEVER fails to correctly estimate lower bound robustness, even in theoretical settings.
Moreover, \citeauthor{huster2018limitations}~\cite{huster2018limitations} showed that the existing approaches to compute the Lipschitz constant of \acp{DNN} have representational learning limitations, which may limit the robustness guarantees we can obtain using it.

The question of accurately measuring robustness remains open.
Some publications, presented in \if 0\mode Section~\ref{subsubsec:provable}\else Section~\ref{subsec:proactive}\fi,~exhaustively search for the space constrained by a lower bound or provide convex relaxations in order to accurately approximate it~\cite{salman2019convex}.
In practice, however, a large body of literature uses the expected accuracy of a model tested with upper bounded adversarial examples and ignore the dichotomy between lower and upper bounds.
In these cases, the upper bounds are chosen arbitrary and often lead to incorrect evaluations.
\citeauthor{salman2019convex}~\cite{salman2019convex} investigated the gap between upper and  lower bounds computed using exact solvers and showed it can grow up to 5 orders of magnitude.


    \section{Hypotheses on the Existence of Adversarial Examples}
\label{sec:causes}

Since the discovery of adversarial examples, there is no universally accepted hypothesis on their existence.
Many conjectures have been proposed and are discussed in this section.
The presentation follows a chronological order, but new developments or evidence for a conjecture are added in line with the initial publication which advanced it.

\paragraph{Initial hypothesis~--~low-probability spaces.} At first, adversarial examples were thought to lie in low-probability spaces from the data manifold, which are hard to reach by randomly sampling the space around an input~\cite{szegedy2013intriguing}.
Searching for solutions to \eqq~(\ref{eq:adversarial_generic}), however, spans the input space in search for adversarial examples and enables the solver to find perturbations.
While state-of-the-art \acp{DNN} models are already trained with data augmentation techniques in order to increase their robustness, the transformed inputs are highly correlated and drawn from the same distribution.
Adversarial examples were thought to be neither correlated or identically distributed, thus leading to the theory that they lie in 'pockets' of the data manifold~\cite{szegedy2013intriguing}.

\citeauthor{gu2014towards}~\cite{gu2014towards} investigated the size of these pockets and discovered they are relatively large in volume and locally continuous.
The authors hypothesized that sensitivity to adversarial examples relates to choosing a wrong objective function or to deficiencies of the training method~--~instead of being a consequence of the model's topology.
Therefore, coming up with a training procedure that can efficiently output regions where the data variance around a training input is low should solve this issue.
The authors made an attempt to design a defense which minimizes the network's output variance, with some success for small perturbations and small data sets.
However, this was not enough to train robust models for larger data sets or any perturbation.

\paragraph{The linearity hypothesis.} \citeauthor{goodfellow2014explaining}~\cite{goodfellow2014explaining} refuted the hypothesis that adversarial examples lie in small regions of the data manifold and advanced the conjecture that adversarial examples span large and high-dimensional regions.
The authors argued adversarial examples exist because \acp{DNN} have, in fact, very linear behavior, despite non-linear transformations within hidden layers.
The choice for activation functions that are easy to optimize (\eg~ReLU) drive \acp{DNN} to behave more linearly.
Therefore, summing small perturbations in all dimensions of a high dimensional input forces the entire sum in a direction that will likely cause a misclassification.
This hypothesis lead to the discovery of more efficient methods to generate adversarial examples, as discussed in Section~\ref{sec:noise}.
Empirical evidence for the linearity hypothesis was also provided by~\cite{tabacof2015exploring, tramer2017space, krotov2016dense}.
\citeauthor{luo2015foveation}~\cite{luo2015foveation} proposed a variant of this conjecture in which \acp{DNN} operate linearly in certain regions of the input manifold, but non-linear in others.

\paragraph{Vanishing gradients.} \citeauthor{rozsa2016towards}~\cite{rozsa2016towards} believe that the gradients of correctly classified inputs diminish during training and fail to create flat regions around the training data.
Therefore, most training data lie close to a decision boundary and small perturbations are able to push inputs over the boundary.
The authors hypothesise that coming up with a training algorithm that will avoid this phenomenon will mitigate the threat to adversarial examples.
However, as we will discuss in Section~\ref{subsec:gradient_ob}, imposing constraints on the gradients is not an efficient defense.

\paragraph{The boundary tilting hypothesis.} \citeauthor{tanay2016boundary}~challenge the linear hypothesis as not "convincing"~\cite{tanay2016boundary}.
At first, because small perturbations are taken relatively to the activations, which increase linearly to the problem.
Therefore, the ratio between inputs and perturbations remains constant.
Secondly, the authors argue that linear behavior is \emph{not sufficient} to explain the adversarial examples phenomenon and demonstrate the possibility to build linear models that are not sensitive to adversarial examples.
In contrast, the authors propose the \emph{boundary tilting perspective}, based on the assumption that a learned class boundary lies close to the data manifold, but the boundary is tilted with respect to it.
Adversarial examples can then be found by perturbing points from the data manifold towards the classification boundary until the perturbed input crosses the boundary.
If the boundary is only slightly tilted, the distance required by the perturbation to cross the decision-boundary is very small, leading to strong adversarial examples that are visually almost imperceptibly close to the data.
The authors argue that adversarial examples are likely to occur along directions of low variance in the data and thus speculate that adversarial examples can be considered an effect of an overfitting phenomenon, which can be alleviated through regularization.
\citeauthor{izmailov2018enablers}~\cite{izmailov2018enablers} investigated this claim by removing low-variability features from inputs during classification and found out that removing them barely improves robustness.
However, removing the features with low mutual information has a significant impact on robustness.
On a similar note, \citeauthor{ilyas2019adversarial}~\cite{ilyas2019adversarial} showed that adversarial examples are a consequence of non-robust features, which are derived from patterns in the data that can be easy to predict by computers, but not understood by humans.

\paragraph{Relation to decision boundaries.} \citeauthor{moosavi2017analysis}~\cite{moosavi2017analysis} showed that it is possible to generate universal perturbations~--~which can be applied to any input.
While investigating the phenomenon, the authors hypothesized that adversarial examples exploit geometric correlations in the space between decision boundaries.
Precisely, the authors suggest the existence of a low dimensional sub-space which contains the vectors normal to the decision boundaries around an input.
\citeauthor{fawzi2016robustness}~\cite{fawzi2016robustness} examined the sensitivity to adversarial examples in relation to the curvature of decision boundaries.
Their results show that a small curvature in the decision boundary increases the classifier's robustness to adversarial examples.
Thus, it is assumed that limiting the curvature of decision boundaries can increase sensitivity to adversarial examples
A similar hypothesis was proposed in~\cite{tramer2017ensemble} and more theoretical analyses are presented in~\cite{moosavi-dezfooli2018robustness}.

\paragraph{Not i.i.d hypothesis.} A different hypothesis assumes that adversarial examples lie off the data manifold, and are sampled from a different distribution~\cite{song2017pixeldefend, meng2017magnet, ghosh2018resisting, lee2017generative}.
This hypothesis lead to the proposal of adversarial detection methods \if 0\mode (Section~\ref{subsubsec:detection}) \else (Section~\ref{subsec:reactive}) \fi and the attempt to learn this new distribution with generative models.
While interesting in nature, because the proof of this hypothesis means adversarial examples break the \ac{iid} assumption, more empirical data is needed.
\citeauthor{carlini2017adversarial}~\cite{carlini2017adversarial} also questioned this hypothesis by developing attacks that can easily bypass adversarial detectors.

\paragraph{The manifold geometry hypothesis.} \citeauthor{gilmer2018adversarial}~\cite{gilmer2018adversarial} hypothesize that adversarial examples are a result of the high-dimensional (and possibly intricate) geometry of the data manifold.
The authors used a synthetic data set which is easier to explore and found that whenever the classifier has the slightest test error, most data points in the input distribution which get correctly classified lie in the neighborhood of a misclassified input.
Therefore, whenever training is performed on an approximation of the real distribution, the model is sensitive to adversarial examples.
This result raises the question if the sensitivity to adversarial examples could ever be removed.
Moreover, the authors refute the hypothesis that adversarial examples lie off the data manifold.




\paragraph{Relation to training resources.}
Following the PAC-learning model briefly discussed in Section~\ref{sec:background}, \citeauthor{schmidt2018adversarially}~\cite{schmidt2018adversarially} showed  the sample complexity for training robust models learning can be significantly higher than for training non-robust models.
In particular, for achieving robustness for the $l_\infty$~norm 	requires an increase of sample complexity polynomial in the input dimension.
On a similar note, \citeauthor{bubeck2018adversarial}~\cite{bubeck2018adversarial} suggest that robust learning in the statistical query model increases the number of queries exponentially.
Somehow dissimilar, \citeauthor{cullina2018pac}~\cite{cullina2018pac}, show that the sample complexity does not increase in the presence of adversaries bounded by convex constraint sets.
This result suggests that robustness can be achieved under some constraints, yet the practicality of such sets was not evaluated.
\citeauthor{tsipras2018robustness}~\cite{tsipras2018robustness} show that gaining robustness involves loosing accuracy and that this tradeoff prevails independent of the \ac{ML} model.

\if 0\mode \bigskip \else\fi
\smallskip
In summary, although \citeauthor{tanay2016boundary}~\cite{tanay2016boundary} claim to refute the linearity hypothesis, there is still not enough empirical evidence to completely reject it.
One can argue that linear transformations in high dimensional spaces can be sufficient to move a sample in the direction of a tilting boundary, thus causing a misclassification.
However, the authors succeed to show that adversarial examples are \emph{not only} due to linear behavior of \acp{DNN}.
The complicated geometry of the manifold can be a root cause of adversarial examples, as suggested by \citeauthor{gilmer2018adversarial}~\cite{gilmer2018adversarial}.
However, there is still no study to investigate if the probability of finding adversarial examples is constrained by the geometry of the manifold.
On a similar note, there is no study to connect non-robust features to the geometry of the manifold.
Will removing non-robust features, as suggested by~\cite{ilyas2019adversarial}, lead to a smooth manifold on which the data is better represented?
Further on, there is no evidence to show that adversarial examples lie off the data manifold.
The publications using this claim tried to develop adversarial detectors with some degree of success, however, their efficacy is still low.

It is not yet settled if adversarial examples lie in small or large spaces in the decision space.
According to \citeauthor{gilmer2018adversarial}~\cite{gilmer2018adversarial} they are proportional with the testing error and the capacity of a model to correctly approximate the input distribution.
However, more evidence is needed to support this conjecture for models with high capacity.

Some publications suggest there are limits to adversarial robustness~\cite{gilmer2018adversarial, fawzi2015fundamental, bubeck2018adversarial} and even that sensitivity to adversarial examples can not be removed~\cite{gilmer2018adversarial}.
Such a consequence sparks several questions regarding future research in this field, some of which are discussed in Section~\ref{sec:conclusions}.
We argue that more fundamental research, similar to~\cite{gilmer2018adversarial, schmidt2018adversarially, cullina2018pac, ilyas2019adversarial}, is needed in order to explain both the causes and the effects of this particular behavior of \acp{DNN} and develop the topic in Section~\ref{sec:discussion}.

    \section{Attacks}
\label{sec:attacks}

\if 0\mode

In this section we aim to fulfil three objectives: (1) provide an overview of the methods used to generate adversarial examples,  (2) classify them according to common characteristics and (3) describe all important attacks in detail, such that we can later use this section as a catalog of attacks.

Because most characteristics of adversarial examples are orthogonal, some classes will overlap (and might lead to confusion).
At a high level, there are two classes of algorithms used to craft adversarial examples: (M) attacks which \emph{modify} a real input - \eg~an image - and (G) attacks which \emph{generate} adversarial inputs from noise.
Following the hypothesis presented in Section~\ref{sec:causes}, of the first class we distinguish between (OP) attacks that efficiently search the input space for adversarial examples using \emph{optimisation} techniques (Section~\ref{subsec:optimisation_attacks}) and (SA) attacks that exploit \emph{sensitive features} (Section~\ref{subsec:sensitivity}).
In some cases, (GT), adversarial examples can be generated only using \emph{geometric transformations} (Section~\ref{subsec:geometric_attacks}).
\emph{Generative models} (GM) aim to learn the  adversarial or input data generation distribution and use it in order to convert inputs to adversarial examples or generate adversarial examples from noise (Section~\ref{subsec:gan_attacks}).

One level deeper, we distinguish between attacks that use (IT) \emph{iterative} or (SS) \emph{single-shot} methods~--~thus illustrating a tradeoff between precision and speed.
Some algorithms approximate the minimal perturbation, at the cost of speed, while others use very fast methods, trading precision.
Fast algorithms are used to enhance the training set with adversarial examples - a procedure called adversarial training and presented in Section~\ref{subsec:adv_training} - but are weaker and easier to protect against.
Iterative algorithms require more processing time, but are powerful and hard to protect against.

A dominant theme in the publications about adversarial examples is to reduce or minimise the size of the perturbation.
As discussed in Section~\ref{sec:attack_models}, this requirement needs better motivation and clear threat models.
Without information about the context and the economics of an attack, searching for minimal perturbations is futile.
In Section~\ref{subsubsec:unrecognisable} we present an attack that uses evolutionary algorithms to generate inputs that are indistinguishable for human observers, but still fool \ac{DNN}.

Further on, when classifying adversarial attacks we have to consider the adversarial goals and capabilities introduced in Section~\ref{sec:attack_models}.
At first, given the adversarial goals, one could aim to generate a \emph{non-targeted} (NTG), random, misclassification or a \emph{targeted} (TG) misclassification.
This goal often shapes the algorithms used to generate adversarial examples because simpler methods (such as single-shot sensitivity attacks) are fitted for non-targeted attacks, but harder to adapt to targeted attacks.
Similarly, iterative attacks are sometimes harder to implement and require more processing time, therefore they are more qualified for targeted attacks.
Once again, the economics of carrying an attack matters.
Without clear security requirements, it is impossible to draw a line between the efficiency of any of the methods.

Secondly, given the knowledge available to an attacker, we distinguish between \emph{white-box} (WB), \emph{grey-box} (GB) and \emph{black-box} attacks (BB).
Most attacks assume full-knowledge of the model under attack and, therefore, fall in the white-box category.
Therefore, we will present grey and black-box attacks in a separate section (Section~\ref{subsec:black_box}), although they sometimes use optimisation methods.
The rather small number of black-box attacks, in contrast with white-box attacks, is another way of judging the security implications of adversarial examples.

We note again that most attack traits are orthogonal and the existence of one does not imply the absence of others.
For example, one may generate an iterative, targeted, white-box attack as well as an iterative, not-targeted, grey-box attack.
Selecting a method is, therefore, context dependent and requires a clear threat model.
Unfortunately, most publications do not present any threat model, therefore, we have to accept some overlap in the classification.

Lastly, we distinguish between \emph{specific} attacks (SP)~--~that aim to modify or generate an input against able to fool \emph{one} model~--~and \emph{universal} attacks (UN)~--~that aim to modify or generate an input that can work against \emph{all} models.

Before delving into the details of each attack, we present an overview in Table~\ref{tbl:attacks} where we enumerate all attacks in relation to their features.
We observe that most attacks are based on modifying a known input.
We also note that the most powerful attacks~--~Carlini and PGD~--~are based on optimisation methods.

In the following sections we introduce, in details, the most important attacks.
We try to extensively cover all attacks in the current literature, however, new attacks might be rolled out after writing this paper.
The classification follows the criteria introduced earlier.

\if 0\mode
\begin{landscape}
\begin{table}
	\centering
	\begin{tabular}{|l|l|l|l|l|l|l|}
		\toprule%
		Attack & \specialcell{Modify (M) or \\ Generate (G) \\ Input} &%
		  		 \specialcell{Optimisation (OP), \\ Sensitivity (SA), \\ Geometric \\ Transformations (GT) \\ Generative Models (GM)} &%
		  		 \specialcell{Targeted (TG), \\ Non-Targeted (NTG)} &%
		  		 \specialcell{Single-Shot (SS), \\ Iterative (IT)} &%
  				 \specialcell{White-box (WB),  \\ Grey-box (GB), \\ Black-box (BB)}&%
	 			 \specialcell{Specific (SP), \\ Universal (UN)} \\
		\midrule
	L-BFGS~\cite{szegedy2013intriguing} & M & OP & TG & IT & WB & SP  \\ \hline
	Deep Fool~\cite{moosavi2016deepfool} & M & OP & NTG & IT & WB & SP \\ \hline
	UAP~\cite{moosavi2017universal} & M & OP & NTG & IT & WB & UN \\ \hline
	Carlini~\cite{carlini2017towards} & M & OP & TG / NTG & IT & WB & SP  \\ \hline 
	CFOA (Madry / PG)~\cite{madry2017towards} & M & OP & TG / NTG & IT & WB & SP   \\ \hline 
	STA~\cite{huang2015learning} & M & OP & TG / NTG & IT & WB & SP  \\ \hline
	ZOO~\cite{chen2017zoo} & M & OP & TG / NTG & IT & BB & SP  \\ \hline
	IS~\cite{narodytska2017simple} & M & OP &  TG / NTG & IT & BB & SP   \\ 
\specialrule{1pt}{1pt}{1pt}
	FGS~\cite{goodfellow2014explaining} & M & SA & NTG & SS & WB & SP  \\ \hline
	JSMA~\cite{papernot2016limitations} & M & SA & TG & IT & WB & SP  \\ \hline
	RSSA~\cite{tramer2017ensemble} & M & SA & NTG & SS / IT & WB & SP  \\ \hline
	BPDA~\cite{athalye2018obfuscated} & M & SA & TG & IT & WB & SP  \\ \hline
	Elastic-Net~\cite{chen2017ead} & M & SA & TG & IT & WB & SP  \\ \hline 
	BI~\cite{kurakin2016adversarial} & M & SA & NTG & IT & WB & SP  \\ \hline
	ILC~\cite{kurakin2016adversarial} & M & SA & TG & IT & WB & SP  \\ \hline
	Momentum~\cite{dong2017boosting} & M & SA & NTG & IT & WB & SP  \\  	\hline	
	Substitute~\cite{papernot2017practical} & M & SA & TG & SS / IT & BB & SP \\ 
	\specialrule{1pt}{1pt}{1pt}
	\specialcell{Rotation Tr.}~\cite{engstrom2017rotation} & M & GT & NTG & SS / IT & WB / GB & SP  \\ \hline
	ManiFool~\cite{kanbak2017geometric} & M & GT & TG / NTG & IT & WB & SP  \\ \hline
	\specialcell{Spatial Tr.}~\cite{xiao2018spatially} & M & GT  & TG & IT & WB & SP  \\  
	\specialrule{1pt}{1pt}{1pt}
	ATN~\cite{baluja2017adversarial} & G & GM & TG / NTG & IT & WB & SP  \\ \hline
	NAE~\cite{zhao2017adversarial} & G & GM & TG & IT & WB & SP \\ 
	\bottomrule
	\end{tabular}	
	\caption{Catalog of Adversarial Attacks.}	
	\label{tbl:attacks}
\end{table}
\end{landscape}

\else 

\begin{landscape}

\begin{table}
	\centering
	\begin{tabular}{|l|l|l|l|l|l|l|}
		\toprule%
		Attack & \specialcell{Modify (M) or \\ Generate (G) \\ Input} &%
		  		 \specialcell{Optimisation (OP), \\ Sensitivity (SA), \\ Geometric \\ Transformations (GT) \\ Generative Models (GM)} &%
		  		 \specialcell{Targeted (TG), \\ Non-Targeted (NTG)} &%
		  		 \specialcell{Single-Shot (SS), \\ Iterative (IT)} &%
  				 \specialcell{White-box (WB),  \\ Grey-box (GB), \\ Black-box (BB)} \\
		\midrule
	L-BFGS~\cite{szegedy2013intriguing} & M & OP & TG & IT & WB   \\ \hline
	Deep Fool~\cite{moosavi2016deepfool} & M & OP & NTG & IT & WB  \\ \hline
	UAP~\cite{moosavi2017universal} & M & OP & NTG & IT & WB \\ \hline
	Carlini~\cite{carlini2017towards} & M & OP & TG / NTG & IT & WB   \\ \hline 
	CFOA (Madry / PG)~\cite{madry2017towards} & M & OP & TG / NTG & IT & WB    \\ \hline 
	STA~\cite{huang2015learning} & M & OP & TG / NTG & IT & WB   \\ \hline
	ZOO~\cite{chen2017zoo} & M & OP & TG / NTG & IT & BB \\ \hline
	IS~\cite{narodytska2017simple} & M & OP &  TG / NTG & IT & BB    \\ 
\specialrule{1pt}{1pt}{1pt}
	FGS~\cite{goodfellow2014explaining} & M & SA & NTG & SS & WB   \\ \hline
	JSMA~\cite{papernot2016limitations} & M & SA & TG & IT & WB   \\ \hline
	SV-UAP~\cite{khrulkov2017art} & M & SA & NTG & IT & WB \\ \hline
	RSSA~\cite{tramer2017ensemble} & M & SA & NTG & SS / IT & WB   \\ \hline
	BPDA~\cite{athalye2018obfuscated} & M & SA & TG & IT & WB   \\ \hline
	Elastic-Net~\cite{chen2017ead} & M & SA & TG & IT & WB   \\ \hline 
	BI~\cite{kurakin2016adversarial} & M & SA & NTG & IT & WB   \\ \hline
	ILC~\cite{kurakin2016adversarial} & M & SA & TG & IT & WB   \\ \hline
	Momentum~\cite{dong2017boosting} & M & SA & NTG & IT & WB   \\  	\hline	
	Substitute~\cite{papernot2017practical} & M & SA & TG & SS / IT & BB  \\ 
	\specialrule{1pt}{1pt}{1pt}
	\specialcell{Rotation Tr.}~\cite{engstrom2017rotation} & M & GT & NTG & SS / IT & WB / GB   \\ \hline
	ManiFool~\cite{kanbak2017geometric} & M & GT & TG / NTG & IT & WB   \\ \hline
	\specialcell{Spatial Tr.}~\cite{xiao2018spatially} & M & GT  & TG & IT & WB   \\  
	\specialrule{1pt}{1pt}{1pt}
	ATN~\cite{baluja2017adversarial} & G & GM & TG / NTG & IT & WB   \\ \hline
	NAE~\cite{zhao2017adversarial} & G & GM & TG & IT & WB \\ 
	\bottomrule
	\end{tabular}	
	\caption{Catalog of Adversarial Attacks.}	
	\label{tbl:attacks}
\end{table}

\end{landscape}
\fi


\subsection{Attacks based on Optimisation Methods (OP)}
\label{subsec:optimisation_attacks}

In this section we review attacks which use optimisation methods in order to search for adversarial examples.
We start with the initial paper, which revealed the sensitivity of \ac{DNN} to adversarial examples.
Later, we introduce a series of attacks that use optimisation techniques similar to sensitivity analysis - \eg~gradient descent - but with different objectives.

\subsubsection{The L-BFGS Attack}
\label{subsubsec:lbfgs}

Szegedy \etal~\cite{szegedy2013intriguing} were the firsts to discover the phenomenon and coin the term \emph{adversarial examples}.
The authors propose an equivalent formalism for \eqq~(\ref{eq:adversarial_generic}) and use limited memory box constrained optimisation (L-BFGS) in order to approximate the minimal perturbation needed to change a label.
They propose a targeted attack, formally defined as follows. Given $c > 0$ and a target label $l'$:
\begin{equation}
	\begin{aligned}
	\label{eq:lbfgs}
		&\min_{\eta} & & c \| \eta \|_2 + J(\vtheta, \vx + \eta, y), \\
		&s.t. & & \eta \in [0,1]^{m}, \\
		& & & f(\vx + \eta) = l'. \\
	\end{aligned}
\end{equation}
\noindent
The final adversarial example is defined as $x' = x + \eta$.
Experimental results show the minimal average distortion drops as low as $0.062$, which means adversarial examples are almost identical to normal inputs.
The authors also propose to augment the training set with adversarial examples in order to increase the robustness to adversarial examples.
However, L-BFGS is a slow procedure, making adversarial training almost impossible for big data sets, where one would have to find adversarial examples for all inputs.
\subsubsection{The Deep Fool Attack}
\label{subsubsec:deepfool}

The Deep Fool~\cite{moosavi2016deepfool} attack is equivalent to moving a sample towards a hyperplane that separates two classes. 
In this case, the shortest distance from a sample $\vx_0$ to the separation hyperplane is equivalent to the algorithm's robustness for sample $\vx_0$ (similar to point-wise robustness in Section \ref{sec:robustness_eval}).
If we assume that a classifier behaves in a linear fashion, this distance is equivalent to the orthogonal projection of a point on a plane (or separation line).
For a binary, linear, classifier we can define the minimal perturbation as:
\begin{equation}
	\begin{aligned}
		\eta_{\vx_0} = \frac{f(\vx_0)}{\| \vw \|_2^2}\vw,
	\end{aligned}
	\label{eq:deep_fool_linear_binary}
\end{equation}
\noindent
where $\vw$ is equivalent to the \emph{geometrical normal}. 

Further on, for a general differentiable function, we assume an implicit equation\footnote{An implicit equation is a function of form f(x\textsubscript{1}, x\textsubscript{2}, ..., x\textsubscript{n}) = 0} defining the surface that separates two classes such that the geometrical normal is equal to its gradient vector.
The Deep Fool algorithm then takes an iterative approach assuming that, at every iteration $i$, the classification function is linearised around the input and computes the minimal perturbation as:
\begin{equation}
	\begin{aligned}
		\eta_{\vx_i} = \frac{f(\vx_i)}{\| \nabla f(x_i) \|_2^2}\nabla f(x_i).
	\end{aligned}	
	\label{eq:deep_fool_binary}
\end{equation}
\noindent
The final perturbation is computed as the sum of all intermediate perturbations $\eta_i$.

The generalisation to linear multi-class classifiers follows from the example above.
If instead of a single separation surface there are many, one can imagine that the initial sample $\vx_0$ is mapped to a region equivalent to a polyhedron, $P$, whose faces are discriminants from other classes.
Following the same rationale, by computing the distance from $\vx_0$ to one of the faces of $P$ one can measure the classifier's robustness for input $\vx_0$.
In practice, the attack selects the closest hyperplane to the boundary of $P$ and projects $\vx_0$ on its surface.
Formally, the minimal perturbation for a linear multi-class classifier is:
\begin{equation}
	\eta_{\vx_0} = \frac{\left|f_{\hat{y}(\vx_0)}(\vx_0)-f_{y(\vx_0)}(\vx_0)\right|}{\| \vw_{\hat{y}(\vx_0)}-\vw_{y(\vx_0)}\|_2^2}(\vw_{\hat{y}(\vx_0)}-\vw_{y(\vx_0)}),
	\label{eq:deep_fool}
\end{equation}
\noindent	
where $\hat{y}(\vx_0)$ is the closest face of $P$ to $\vx_0$.
Further more, the generalisation to a differentiable multi-class classifier can be obtained using the same rationale as for \eqq~(\ref{eq:deep_fool_binary}) in \eqq~(\ref{eq:deep_fool}).
An important observation is that Deep Fool can converge on the discriminant hyperplane. 
In order to "push" the classifier over the border, the final perturbation is multiplied by a small constant.
Another important observation is that Deep Fool assumes that classifiers behave in a linear fashion, according to the linearity conjecture from Section~\ref{sec:causes}.
\subsubsection{Attacks using \ac{uap}}
\label{subsubsec:universal}

Moosavi-Dezfooli \etal~\cite{moosavi2017universal} demonstrate the existence of image and \ac{DNN} agnostic perturbations that cause misclassification with high probability.
The algorithm iteratively applies the Deep Fool attack (Section \ref{subsubsec:deepfool}) for all images in the training data set, until one perturbation can fool a large part of the training set.
Formally, given the input distribution $\hat{p}_{data}$, the algorithms searches for an upper-bounded universal perturbation $\eta$ such that:
\begin{equation}
	\begin{aligned}
	\label{eq:universal}
		& & \| \eta \| &  \le \epsilon  \text{    and}, & \\
		& & \underset{\vx \sim \hat{p}_{data}}{\mathbb{P}} ( f(\vx+\eta) &  \neq f(\vx) ) \geq 1 - \delta, &
	\end{aligned}
\end{equation}
\noindent
where $\delta$ quantifies the desired fooling rate for all images sampled from the distribution $\hat{p}_{data}$.
At each iteration step, the perturbation is found using the Deep Fool attack:
\begin{equation}
	\begin{aligned}
		\eta' & \gets \arg\min_{\eta_{\vx_i}} \| \eta_{\vx_i} \|_{2} \text{ s.t. } f(\vx_i + \eta_{\vx_{i-1}} + \eta_{\vx_i}) \neq  f(\vx_i),
	\end{aligned}
\end{equation}
\noindent
where $\eta_{\vx_i}$ is the output of Deep Fool.
In order to ensure the upper bound constraint, the perturbation is projected on the $l_p$ ball of radius $\epsilon$ and centred at 0, leading to the following update rule:
\begin{equation}
	\begin{aligned}
		\eta & = \arg\min_{\eta'} \| \eta_{\vx_{i-1}} - \eta' \| _2\text{ subject to } \| \eta' \|_p \leq \epsilon.
	\end{aligned}
\end{equation}
\noindent
Several passes over the training data set are performed in order to improve the quality of an universal perturbation.
The algorithm is able to find multiple universal perturbations that are different in nature and can fool \ac{DNN} with high accuracy.

\subsubsection{The Carlini $L_{p}$ Attacks}
\label{subsubsec:carlini}

Carlini and Wagner~\cite{carlini2017towards} propose three new attacks based on the $L_{p}$ distance metrics discussed in Section~\ref{subsec:norm_balls}.
All three attacks use upper-bound robustness and are a variance of \eqq~(\ref{eq:adversarial_generic}), where the classification function is replaced in the constraints set.
Because the classification constraint is highly non-linear, the authors propose to replace it with another function such that $f(\vx + \eta) = l'$ if and only if $f(\vx + \eta) \le 0$.
Thus, the optimisation objective becomes:
\begin{equation}
	\begin{aligned}
	\label{eq:adversarial_carlini}
		&\min_{\eta} & & \|\eta\|_p + c \cdot f(\vx+\eta),\\
		&s.t. & & \vx + \eta \in [0,1]^n,
	\end{aligned}
\end{equation}
\noindent
where $c$ is a constant determined empirically through binary search.
In order to ensure the optimisation result yields a valid image, $\eta$ is constrained as follows: $0 \le x_i+\eta_i \le 1$ for all $i$ (box-constraints).
However, in order to use optimisation algorithms that do not support box constraints, the authors optimise $\eta$ through a change of variable, $w$, such that:
\begin{equation}
	\label{eq:distorsion_carlini}
	\eta= \frac12 (\tanh(w) + 1) - \vx.
\end{equation}
\noindent
Since $-1 \le \tanh(w) \le 1$, it follows that $0 \le \vx+ \eta \le 1$, so the solution will automatically be valid.

The authors then substitute \eqq~(\ref{eq:distorsion_carlini}) in \eqq~(\ref{eq:adversarial_carlini}) and optimise for $L_{0}, L_{2} \text{ and } L{\infty}$ norms.
They are able to craft performant adversarial examples that remain state-of-the-art today.
\subsubsection{\ac{cfoa}}
\label{subsubsec:madry}

Madry \etal~\cite{madry2017towards} study the adversarial robustness of \ac{DNN} through the lens of robust optimisation.
The authors propose an attack and a defence mechanism formulated in a min-max fashion.
Because the attack can be independently used, we discuss it here and outline the defence in Section~\ref{subsubsec:madry_learning}.

Assuming for each data point in the training set $\vx_i$ a set of allowed perturbations $ \mathcal{U} \subseteq \R^d $ is known, the attack searches for perturbations that maximise the loss function across the training data set. 
For the image classification task, $\mathcal{U}$ is chosen to be the $L_\infty$-ball around the normal input $\vx$, limited by a constant $\epsilon$.
The attack is formalised for all training set as:
\begin{equation}
\label{eq:minmax}
	\rho(\theta) = \mathbb{E}_{(\vx,y)\sim p_{data} }\left[\max_{\eta \in \mathcal{U}} J(\vtheta, x + \eta, y) \right].
\end{equation}
\noindent
The main contribution of this attack is the use of \ac{pgd} for the maximisation problem, which suggests the problem is tractable.
In order to explore a large part of the loss landscape, \ac{pgd} is restarted from many points in the $L_{\infty}$ ball around an input.
Surprisingly, although there are many local maxima spread widely apart within $\vx_i + \mathcal{U}$, they tend to have very well concentrated loss values.
This suggests that an adversarial example found by this method is representative for \emph{all} adversarial examples generated with first order methods.
Because of this, the attack was named \emph{complete first order adversary}.

The general aspect of \ac{cfoa} suggests that training against complete first order examples can yield robustness against all possible examples generated with first order methods.
The training results are discussed in Section~\ref{subsubsec:madry_learning}.

%
%


\subsubsection{\ac{sa}}
\label{subsubsec:strong_adversary}

Huang \etal~\cite{huang2015learning} develop two methods to generate adversarial examples and use them to train robust \ac{DNN}.
In this section we present the adversarial generation process and postpone the training procedure for Section~\ref{subsubsec:learning_strong_adversary}.

The search for a minimal perturbation is based on the \emph{linear approximation}\footnote{The linear approximation follows from Taylor's theorem when n=1.} of the \ac{DNN} output. 
If we denote the output of the softmax layer as $y = (\alpha_1, \alpha_2, \dots, \alpha_l)$, the linear approximation for a perturbation $\eta$ can be written as $ \hat{y}(\vx+\eta) = y(\vx) + \mathbf{H} \eta $; where $\mathbf{H}$ is the Jacobian matrix \wrt~$\vx$.
It follows that for a target $l' \neq l$ the necessary condition for $\eta$ is $	\mathbf{H}_{l'} \eta - \mathbf{H}_{l} \eta \leq \alpha_{l} - \alpha_{l'} $. 
Thus the norm for the optimal perturbation can be formally defined as:
\begin{equation}
	\label{eq:strong_adversary}
	\min_{\eta} \| \eta \|_p \quad \text{s.t.} \quad \mathbf{H}_{l'} \eta - \mathbf{H}_{l} \eta \leq \alpha_{l} - \alpha_{l'}.
\end{equation}
\noindent
The final perturbation is computed as $\eta = \epsilon \eta_{i} / \| \eta_{i}\|_p$ where $\eta_{i}$ is the output of \eqq~(\ref{eq:strong_adversary}) for a chosen $p$ norm.

The authors propose a second method to generate a perturbation that follows from using the linear approximation, introduced earlier, when maximising the loss function.
The objective is to find a perturbation smaller than $\epsilon$ that can maximise the loss function: $ \eta = \arg\max_{\|\eta_{i}\|\leq \epsilon} J\left( \vtheta, \vx + \eta_{i}, y_i \right) $.
Using the linear approximation detailed earlier, a perturbation can be formally defined as:
\begin{equation}
	\eta = \{\eta :\, \|\eta\|_{p} \le \epsilon;\, \langle H_{y_i},\,\eta_{i}\rangle = \epsilon \|H_{y_i}\|_p\}.
\end{equation}
\noindent
Depending of the chosen norm, the optimal solution takes different forms.
For example, if we use $L_\infty$, the attacks is identical to \ac{fgsm}.

The solely performance of the attacks is not discussed in the paper.
However, the attacks prove efficient for training a robust neural network (Section~\ref{subsubsec:learning_strong_adversary}).

\subsection{Attacks based on Sensitive Features (SA)}
\label{subsec:sensitivity}
Until now we reviewed attacks that use optimisation methods in order to search for adversarial perturbations in the input space, project an input on a discriminant surface or optimise alternatives of \eqq~(\ref{eq:adversarial_generic}).
In this section we introduce attacks that search for sensitive features of one input and modify them in order to generate adversarial examples.
Although these attacks also use optimisation methods, their objective is to determine sensitive features or directions of perturbations, and later use them to build an adversarial example.

\subsubsection{The \ac{fgsm} Attack}
\label{subsubsec:fgsm}

The \ac{fgsm}~\cite{goodfellow2014explaining} attack is as a consequence of the linearity conjecture, presented in Section~\ref{sec:causes}.
It assumes that summing small perturbations in the direction of the gradient taken \wrt~one input can lead to adversarial examples.
In order to overcome the speed limitations of L-BFGS, Goodfellow \etal~\cite{goodfellow2014explaining} propose a fast, upper-bounded, method to generate an example, formally defined as:
\begin{equation}
	\label{eq:fgsm}
	\eta = \eps \sign \left( \nabla_\vx J(\vtheta, \vx, y) \right),
\end{equation}
\noindent
where $\epsilon$ is a hyper-parameter that controls the size of a perturbation.
The final adversarial example results from $\vxadv = \vx + \eta$.

The gradient is a vector of partial derivatives, where each partial derivative gives the local rate of change of the output \wrt~the corresponding input, holding the other inputs fixed.
Analysing the gradient of the loss function \wrt~the input is often referred to as \emph{sensitivity} or \emph{saliency} analysis~\cite{yeung2010sensitivity} and can reveal the importance of one feature in the decision process.
Moreover, the gradient points the direction of the maximum of a function.
By taking a small step in this direction, the \ac{fgsm} attack takes a small step in increasing the loss function \wrt~each input feature.

The value of $\epsilon$ controls the size of a perturbation and impacts the sensitivity of a human observer.
It represents an upper bound for the robustness of a model.
In practice, values as low as $.1$ generate error rate of over $85\%$ on small convolutional max-out networks.
\subsubsection{The \ac{jsma}}
\label{subsubsec:jsma}

Papernot \etal~\cite{papernot2016limitations} introduce an attack based on saliency analysis \cite{yeung2010sensitivity}.
In order to discover the importance of each pixel in the decision process, a \emph{saliency map}
\footnote{In computer vision a saliency map is an image which shows each pixel's unique quality.}
is generated by computing the forward derivative (Jacobian) of the function learned by a \ac{DNN}, $f(\cdot)$.
This method contrasts with gradient based methods, which take the backward gradient of the loss function \wrt~network parameters or the input vector (as in the case of \ac{fgsm}).
The forward derivative allows to find input features that lead to significant changes in the \ac{DNN} output.

For image classification, the input features are image pixels.
In order to construct the saliency map, one can analyse the forward derivative \wrt~each pixel and see how it influences the newly selected target.
A formal definition for saliency map is introduced~\cite{papernot2016limitations} as:
\begin{equation}
	\label{eq:saliency-map-increasing-features}
	 S(\vx, l')[i] = \left\lbrace
\begin{array}{c}
0  \mbox{ if }   \frac{\partial f_{l'}(\vx)}{\partial \vx_i}<0  \mbox{ or } \sum_{j\neq l'} \frac{\partial f_{j}(\vx)}{\partial \vx_i}>0,\\
\left(  \frac{\partial f_{l'}(\vx)}{\partial \vx_i}\right)  \left| \sum_{j\neq l'} \frac{\partial f_{j}(\vx)}{\partial \vx_i}\right| \mbox{ otherwise },
\end{array}\right.
\end{equation}
\noindent
where $i$ is an input feature and $l'$ is the desired label.

The first condition rejects input features with a negative target derivative or an overall positive derivative for other classes.
Similarly, $ \sum_{j\neq l'} \frac{\partial f_{j}(\vx)}{\partial \vx_i}$ needs to be negative to decrease or stay constant when feature $\vx_i$ is increased.
The product on the second line makes it possible to consider all other forward derivative such that we can compare all $S(\vx,l')[i]$.

In summary, high values of $S(\vx,l')[i]$ correspond to input features that will either increase the target class, decrease other classes significantly, or both. By increasing these features, the adversary eventually misclassifies the sample into the target class.
The algorithm selects, at each iteration, relevant input features using the saliency map and increases or decreases their intensity.
An upper bound parameter limits the overall distortion that can be applied.

An important limitation of the \ac{jsma} attack, as noted in~\cite{Papernot2018}, is its inherent computational cost for big images (\eg~ImageNet~\cite{deng2009imagenet}).

\subsubsection{\ac{rssa}}
\label{subsubsec:radomised_attack}

Sharp curvatures near a data point can mask the true direction of steepest ascent and burden the discovery of adversarial examples with single-shot gradient methods.
In order to escape this phenomenon, Tramer \etal~\cite{tramer2017ensemble} introduce an attack that precedes single-shot attacks with a randomisation step.
\noindent
Formally, for hyper-parameters $\epsilon$ and $\alpha$ (where $\alpha < \epsilon$) the attack can be described as follows:
\begin{equation}
	\begin{aligned}
		\vxadv & = \vx_{rand} + (\epsilon - \alpha) \cdot \sign \nabla_{\vx_{rand}} J(\vtheta, \vx_{rand}, y), \\
		\text{where} \quad \vx_{rand} & = \vx + \alpha \cdot \sign(\mathcal{N}(\mathbf{0}^d, \mathbf{I}^d)),
	\end{aligned}
\end{equation}
\noindent
and $ \mathcal{N}(\mathbf{0}^d, \mathbf{I}^d) $ is a normal distribution with mean 0 and variance 1. 
The proposed attack can be seen as a single-step alternative to the complete first order adversary, computed with \ac{pgd} (Section~\ref{subsubsec:madry}).

\ac{rssa} searches for an adversarial example starting every time from a random vicinity of the input data point, thus avoiding gradient masking.
The extra random step yields a stronger attack for all models under evaluation, even those not trained on adversarial examples and suggests that a model's loss function is generally less smooth near the data points.

\subsubsection{\ac{bpda}}
\label{subsubsec:backward_pass}

In order to avoid cases when the gradient can not be well approximated - a phenomenon called gradient obfuscation and presented in Section~\ref{sec:defences} - the authors of \cite{athalye2018obfuscated} introduced a new type of attack, called \ac{bpda}.


To approximate the gradient of a non-differentiable layer of a neural network, $f^{i}(\cdot)$, one can search for a differentiable approximation s.t. $g(x) \approx f^{i}(x)$.
Then, the gradient of $f(x)$ can be approximating by performing the forward pass through the whole network, but on the backward pass $f^{i}(x)$ is replaced with $g(x)$.
As long as the two functions are similar, the slightly inaccurate gradients still prove useful in constructing an adversarial example and avoid gradient masking and obfuscation.

\subsubsection{Elastic-Net Regularisation}
\label{subsubsec:elastic_attack}

Chen \etal~\cite{chen2017ead} explore the use of $L_1$ norm in creating adversarial examples.
They propose an attack based on elastic-net regularisation - a mixture of penalty functions used for high-dimensional feature selection~\cite{zou2005regularization}.
Elastic-net regularisation uses $L_1$ and $L_2$ norms as penalty functions. 
Formally, this regularisation technique is defined as:

\begin{equation}
    \min_{\vz \in \sZ} f(\vz) + \lambda_1 \| \vz \|_1 + \lambda_2 \| \vz \|_2^2,
\end{equation}
\noindent
where $\vz$ is a vector of optimisation variables, $f(\vz)$ is a loss function and $\lambda_1$, $\lambda_2$ are regularisation parameters.
\noindent
In practice, the authors extrapolate from the Carlini attack \eqq~(\ref{eq:adversarial_carlini}) using the elastic-net regularisation:

\begin{equation}
    \begin{aligned}
        \label{eq:ead-attack}
            &\min_{\vxadv} & & c f(\vxadv, t) + \beta \| \vxadv - \vx \|_1 + \| \vxadv - \vx \|_2^2 
            &s.t. & & \vxadv \in [0,1]^n,
        \end{aligned}
\end{equation}
\noindent
where $t$ is the target class.

Experimental results show $L_1$-based adversarial examples crafted with this method can be as succesful as $L_2$ and $L_\infty$ attacks.

\subsubsection{The \ac{bim} Attack}
\label{subsubsec:bim}

Kurakin \etal~\cite{kurakin2016adversarial} are one of the first groups to test adversarial examples in the physical world.
Towards this end, they introduce two new methods to generate adversarial examples based on the \ac{fgsm} attack (Section~\ref{subsubsec:fgsm}).
The first one, \ac{bim}, uses the \ac{fgsm} attack iteratively in order to generate adversarial examples, but limits the amount of perturbation on a pixel basis.
The limit allows a more versatile attack, where the perturbation is smoothened among pixels.

Similar to gradient \emph{clipping}~\cite{pascanu2013difficulty}, the authors define a clipping function which limits the value of a pixel by an upper bound ($255$) and forces the adversarial example to stay in the $L_p$-neighbourhood of the original image $\vx$. 
The clipping function is formally defined as follows. 
Given an adversarial example generated through \ac{fgsm}, $\vxadv = \vx + \eps \sign \left( \nabla_\vx J(\vtheta, \vx, y) \right) $, the limit is:
\begin{equation}
	Clip_{\vx, \epsilon} \left\{ \vxadv \right\} (x, y, z) = \min \Bigl\{ 255, \vx(x,y,z) + \epsilon, \max \bigl\{ 0, \vx(x,y,z) - \epsilon, \vxadv(x,y,z) \bigr\} \Bigr\},
\end{equation}
\noindent
where $\vx(x,y,z)$ is the value of channel $z$ from the image $\vx$, at coordinates $(x, y)$ and $\epsilon$ is an upper bound for the model's robustness.

The \ac{bim} attack extends \ac{fgsm} by applying it iteratively, with a small step size, and clipping the pixel values of the intermediate results on each step.
Formally, the \ac{bim} attack is defined as:
\begin{equation}
	\label{eq:bim}
	\vxadv_{0} = \vx, \quad
	\vxadv_{N+1} = Clip_{\vx, \epsilon}\Bigl\{ \vxadv_{N} + \alpha \sign \bigl( \nabla_\vx J(\vxadv_{N}, y_{true})  \bigr) \Bigr\},
\end{equation}
\noindent
where $\alpha$ represents the step size. 
In practice, the authors use $\alpha = 1$ and select the number of iterations using the formula $min(\epsilon+4, 1,25\epsilon)$, determined heuristically.


\subsubsection{The \ac{illcm} Attack}
\label{subsubsec:illcm}

By using the \ac{fgsm} or the \ac{bim} attack one can generate an un-targeted misclassification.
However, for classification problems with a large number of classes, but little difference between them, an un-targeted attack might have insignificant impact.
For example, mistaking one breed of dog for another can have a low impact for the classification task, while mistaking a person for a bird can take an interesting turn.
In order to create more interesting attacks, Kurakin \etal~\cite{kurakin2016adversarial} propose he \ac{illcm} attack.
This iterative method uses the \ac{bim} attack (Section~\ref{subsubsec:bim}) in a targeted fashion.
However, instead of manually choosing the target class, \ac{illcm} prefers the \emph{least likely} class according to the model's prediction for an input:
\begin{equation}
	y_{LL} = \argmin_{y} \bigl\{ p( y | \vx ) \bigr\}.	
\end{equation}
\noindent
Thus \eqq~(\ref{eq:bim}) can be re-written, for the \ac{illcm} attack, as follows:
\begin{equation}
	\vxadv_{0} = \vx, \quad
	\vxadv_{N+1} = Clip_{\vx, \epsilon}\Bigl\{ \vxadv_{N} - \epsilon \sign \bigl( \nabla_\vx J(\vxadv_{N}, y_{LL})  \bigr) \Bigr\},
\end{equation}
\noindent
because we want to maximise the likelihood for class $y_{LL}$, which is equivalent to minimising the loss function for this class - take a step in the opposite direction of its gradient.
\subsubsection{Momentum Based Gradient Attacks}
\label{subsubsec:attack_momentum}

Dong \etal~\cite{dong2017boosting} propose a class of momentum-based, iterative, adversarial attacks.
As in the case of gradient descent, momentum can stabilise the update directions and help escape poor local minimum/maxima by accumulating a velocity vector in the gradient direction.
Formally, the iterative \ac{fgsm} with momentum is defined as:

\begin{equation}
    \begin{aligned}
        \vg_{t+1} = \mu \vg_t + \frac{\nabla_\vxadv J(\vtheta, \vx, y) }{\| \nabla_\vxadv J(\vtheta, \vx, y) \|_{1}} \\
        \vxadv_{t+1} = \vxadv_{t} + \epsilon \text{sign}(\vg_{t+1})
    \end{aligned}
\end{equation}
\noindent
where $ \mu $ is a decay factor.
When the decay factor is equal to $0$, the method is equivalent to \ac{fgsm} (Section~\ref{subsubsec:fgsm}).


\subsection{Attacks which Exploit Geometric Transformations (GT)}
\label{subsec:geometric_attacks}

The attacks introduced by now are based on special crafted perturbations.
It was not yet explored if such perturbations are common in real-life scenarios \eg~due to scratches on cameras or sensor wear.
In this section, however, we review attacks that explore natural and common perturbations.

At first, we present an attack based on very simple geometric transformations of images, followed by an algorithm that measures the robustness of \ac{DNN} to geometric transformations.
Towards the end we introduce an approach that changes the geometry of a scene, without altering the pixels.

\subsubsection{Adversarial Examples Using Rotations and Translations}
\label{subsubsec:rotation_translation}

Engstrom \etal~\cite{engstrom2017rotation} show that only simple transformations - rotations and translations - are sufficient to fool \ac{DNN}.
These transformations are easy to craft and realistic in real-life scenarios.
The authors introduce three methods to search for the best transformations: at first, they propose a gradient-based method, similar to the attacks in Section~\ref{subsec:sensitivity}, which optimises pixel positions in order to increase the cost function.
Secondly, they propose a grid search method that can be ran in black-box mode and, lastly, they propose to randomly sample $k$ different transformations and keep the one on which the model performs worse.

The drop in accuracy ranges from $34-90\%$, depending on the model and data set used, even though the models were trained with data augmentation techniques (which include affine transformations).
This result leads us to think that adversarial robustness is a concern which surpasses strict adversarial settings.

\subsubsection{ManiFool}
\label{subsubsec:geometric_robustness}

ManiFool~\cite{kanbak2017geometric} is an algorithm for finding small worst-case geometrical transformations of images.
Moreover, the authors define a quantitative measure of robustness to geometrical transformations based on the geodesic distance between two transformations.

The idea behind ManiFool is simply to iteratively move from an image sample towards the decision boundary of the classifier where the classification decision changes, while staying on the transformation manifold.
Each iteration is composed of two steps: choosing the movement direction and mapping this movement onto the manifold. 
The iterations continue until the algorithm reaches the decision boundary and finds a fooling transformation example~\cite{kanbak2017geometric}.


\subsubsection{Spatially Transformed Adversarial Examples}
\label{subsubsec:spatially_transformed}

Xiao \etal~\cite{xiao2018spatially} propose to change the geometry of the scene, while keeping the original appearance.
Instead of imposing norm constraints on the pixel space, the authors introduce a new regularisation loss on the local geometric distortion, thus producing higher perceptual quality for adversarial examples.
Surprisingly, spatially transformed adversarial examples prove harder to defend with defences such as adversarial training.

\subsection{Attacks based on Generative Models (GM)}
\label{subsec:gan_attacks}

In this section we cover adversarial attacks based on generative models - a class of machine learning algorithms that learn to estimate a probability distribution by looking at samples drawn from it.
The  model is used to produce artificial examples belonging to the same distribution.
For example, one could start with a database with raw pictures of plants and generate more examples in order to design a forest.
The goal is to generate examples that are alike the training samples, but not exactly the same.

In particular, we study two generative models: (1) \ac{vae}~\cite{kingma2013variational, danilo2014stochastich} and (2) \ac{gan}~\cite{goodfellow2014generative}.

\ac{vae} search for an optimal function that returns entries very close to those in the input distribution $P(\vx)$, given a vector of latent variables $\vz$.
Formally, we wish to optimise $\vtheta$ s.t. we can sample $\vz$ from $P(\vz)$ and get a high probability that $f(\vtheta, \vz)$ is close to a sample in the input data set.
This translates as maximising the probability of each sample $\vx$ in the training data set, under the entire generative process:
\begin{equation}
	\label{eq:vae_1}
	P(\vx) = \int P(\vx|\vz; \vtheta)P(\vz)d\vz,
\end{equation}
\noindent
where $P(\vx |\vz; \vtheta)$ replaces $f(\vtheta, \vz)$.
This allows the problem to be formulated in a maximum likelihood fashion.
In order to solve \eqq~(\ref{eq:vae_1}), generative models choose the latent variables $\vz$ s.t. task specific information is captured.
In practice, \ac{vae} assume that samples of $\vz$ can be drawn from a simple distribution, \ie~$P(\vz) = \mathcal{N}(\vz | 0, I)$ and use powerful approximators (\eg~\ac{DNN}) to automatically learn a mapping between the independent, normally-distributed $\vz$ values and the required latent variables~\cite{doersch2016tutorial}.
Afterwards, the latent variables are mapped to $\vx$.


\bigskip
\ac{gan} is a generative technique that simultaneously trains two models: a generator - which captures the data distribution - and a discriminator - which estimates the probability that a sample comes from the training data, rather than from the generator.
In order to train \ac{gan}, both models play a min-max game in the following setting: (1) the discriminator, $D$, trains in order to maximise the probability to assign a correct label to training data or samples coming from the generator and (2) the generator, $G$, trains in order to minimise the output of $D$.

Formally, in order to learn the generator's distribution $P_g$ we define a prior on input noise variables $P_\vz(\vz)$, then represent a mapping to data space as $G(\vz, \vtheta_g)$.
$G$ is a differentiable function represented by a multi-layer perceptron with parameters $\vtheta_g$.
We also define a second multilayer perceptron $D(\vx, \vtheta_d)$ that outputs a single scalar representing the probability that $\vx$ came from the data rather than $p_g$~\cite{goodfellow2014generative}.
The min-max game with value function $V(G,D)$ can be described as follows:
\begin{equation}
\label{eq:gan}
	\min_G \max_D V(D,G) = V(D, G) = \mathbb{E}_{\vx \sim p_{\text{data}}(\vx)}[\log D(\vx)] + \mathbb{E}_{\vz \sim p_{\vz}(\vz)}[\log (1 - D(G(\vz)))].
\end{equation}
\noindent
Informally, the generative model can be thought of as analogous to a team of counterfeiters, trying to produce fake currency and use it without detection, while the discriminative model is analogous to the police, trying to detect the counterfeit currency~\cite{goodfellow2014generative}.
The same setting can be applied to generate adversarial examples - the generative model aims to generate examples that cause misclassifications and appear normal to human observers and the discriminant enforces such constraints.

\subsubsection{\ac{atn}}
\label{subsubsec:atn}

Baluja and Fischer~\cite{baluja2017adversarial} train a \ac{DNN} that transforms an input into an adversarial example.
The transformation network is trained to fool a target network or to generate examples transferable to a range of networks.
Although \ac{atn} can be potentially used in black-box scenarios, the authors demonstrate its use in targeted, white-box settings.
Formally, \ac{atn} can be defined as:
\begin{equation}
	g_{f, \vtheta_g}(\vx) = \vx \in X \rightarrow \vxadv,
\end{equation}
\noindent
where $f$ is the target network.
In order to train $g_{f, \vtheta_g}$, the following optimisation problem is solved:
\begin{equation}
	\arg \min_{\vtheta_g} \sum_{\vx_i \in X} \beta L_{X} (g_{f, \vtheta_g}(\vx_i), \vx_i) + L_y (f(g_{f, \vtheta_g}(\vx_i)), f(\vx_i)),
\end{equation}
\noindent
where $L_{X}$ is a loss function in the input space ($L_p$ norm for images), $L_{y}$ is a custom loss function and $\beta$ is a weight that balances the loss.
The $L_{y}$ function attempts to change just one probability in the softmax layer, corresponding to the target class, while keeping all others the same.

The authors use two approaches to generate adversarial examples with an \ac{atn}: (1) use a residual network~\cite{he2016deep} to generate a perturbation and (2) use auto-encoders with $L_{y}$ regulariser.
In practice, using auto-encoders yields the best results and successfully scales to large data sets.
\ac{atn} are efficient to train, fast to execute, and produces diverse  adversarial examples.
Because the network is pre-trained, the generation of adversarial examples only takes one step, suggesting the use of \ac{atn} in adversarial training.
However, this approach was not investigated.

\subsubsection{\ac{nae}}
\label{subsubsec:gan_natural}

The intuition behind generating natural adversarial examples~\cite{zhao2017adversarial} is to perform the search for adversarial examples in a \emph{deep representation} of the input data, instead of searching in the input data space directly.

For this, a generator $G$ is trained to map random noise vectors $\vz$ to samples $\vx$ from the input distribution $P(X)$ (from noise to input domain).
A second generative model called \emph{inverter}, $I$, is trained to map data instances to corresponding dense representations (from input domain to $G$).
This is equivalent to finding an adversary $\vz'$ in an underlying vector space which defines the distribution $P(\vx)$ and then map it back to $\vxadv$ with the help of a generative model.

Both the generator and the inverter are trained with Wasserstein \ac{gan}~\cite{arjovsky2017wgan} - a \ac{gan} model that uses the Wasserstein distance in the objective function.
Using the learned functions, a natural adversarial example can be described as follows:
\begin{equation}
	\vxadv = G(\vtheta_g, \vz') \quad\text{where}\quad \vz' = \arg \min_{\tilde{\vz}} \| \tilde{\vz} - I(\vtheta_i, \vx) \| \quad\text{s.t.}\quad f(G(\vtheta_g, \tilde{\vz})) \neq f(x).
\end{equation}
\noindent
Instead of $\vx$, its dense representation $\vz' = I(\vtheta_i, \vx)$ is perturbed.
The generator is used to test whether a perturbation $\tilde{\vz}$ fools the classifier by querying $f$ with $\tilde{\vx}=G(\vtheta, \tilde{\vz})$.
An intermediary step minimises the reconstruction error of $\vx$ and the divergence between sample $\vz$ and the inversion $I(\vtheta_i, G(\vtheta_g, \vz))$ in order to encourage the latent space to be normally distributed.

The authors propose two approaches to search for an adversary: (1) by incrementally increasing the search space in the vicinity of $\vz'$ until an adversary $\vxadv$ is found and (2) searches for adversaries in a wide search range, then recursively tighten the upper bound of the search range with denser sampling in bisections.
The examples are tested on both computer vision and natural language processing tasks and prove that natural adversaries can help evaluating the accuracy of black-box classifiers even in absence of labeled training data.

The fact that adversarial examples generated for a model can transfer to others, regardless of architecture, was first observed by \citeauthor{szegedy2013intriguing}~\cite{szegedy2013intriguing}.
However, this property was only later explored by \citeauthor{papernot2016transferability}~\cite{papernot2016transferability} in an attempt to evaluate black-box attacks.
The authors examine how adversarial examples crafted on one model can transfer between several \ac{ML} techniques such as linear regression, \acp{SVM} or \acp{DNN}.
In such cases, an attacker has partial or full knowledge of the training data and can train a substitute model with it.

Later,~\citeauthor{papernot2017practical}~\cite{papernot2017practical} developed and evaluated practical black-box attacks against \acp{DNN}.
In order to construct adversarial examples for a target model without any information available, the authors train a substitute model with data generated by the adversary and labeled by querying the target model.
Afterwards, adversarial examples are crafted using attacks based on sensitivity analysis on the substitute model and transferred to the target model.
In order to properly evaluate their technique, the authors attack various \acp{DNN} models hosted online by large companies and are able to generate misclassifications in most cases.

\citeauthor{liu2016delving}~\cite{liu2016delving} generate adversarial examples for an ensemble of methods, hypothesizing that examples that transfer across several substitute models are more likely to transfer across a large array of black-box models.
Their work is also the first attempt to scale black-box attacks on large data sets.
Experimental results show that, in a targeted fashion, the precision of back-box attacks is quite low \ie~the adversarial examples do not maintain their intended class.
However, if adversarial examples are used for untargeted black-box attacks, the success rate increases significantly.

\citeauthor{bhagoji2017exploring}~\cite{bhagoji2017exploring} present an attack that removes the need to train substitute models.
The authors approximate the gradient of a black-box model \wrt~one input using the finite difference method and show  its success by attacking pre-trained models or cloud deployed ones.
Similarly, \citeauthor{chen2017zoo}~\cite{chen2017zoo} present an attack called \ac{zoo}, that uses a derivative free optimization model.
This method can estimate the gradient across the perturbation's direction taking into consideration the value of the objective function at two neighboring points (corresponding to adding or subtracting a small perturbation).
The experimental results show \ac{zoo} outperforms black box attacks that artificially train a substitute model~\cite{papernot2017practical, papernot2016transferability}.

Another method that avoids training a substitute models was proposed by \citeauthor{narodytska2017simple}~\cite{narodytska2017simple}.
The authors use an iterative search procedure to explore the local neighborhood of a datapoint and refine the adversarial perturbation.
This information provides an approximation of the gradient of the loss function \wrt~to the input and, thus, provides valuable information about the sensitive pixels.
Experimental results show good accuracy even for very deep networks.
Other publications use genetic algorithms to synthesize adversarial examples~\cite{alzantot2018genattack} or to estimate the gradients~\cite{ilyas2018black} in order to generate adversarial examples in limited domains (such as limited queries or information).
Outside image recognition, black-box adversarial examples have been used against malware detection systems~\cite{rosenberg2017generic}.

Black-box attacks rely on estimating the gradient's direction and using it to generate perturbations.
The applicability of substitute models is further discussed in Section~\ref{sec:transferability}, where the ability to transfer adversarial examples between different models is presented.
Besides, other estimation techniques show good performance and have been tested with success in real life scenarios such as cloud \ac{ML} service providers.

\subsection{Other Attacks}

In this section we explore peculiar attacks - which do not fit any of the classes presented before.
In particular, we present two attacks that use evolutionary algorithms: one which evolves a population of unrecognisable images which still fool \ac{DNN} and one which only modifies one pixel of an image.
The last section (Section~\ref{subsec:real_world}) is reserved to adversarial examples used for different \ac{ML} tasks than object recognition.

\subsubsection{Unrecognisable Images}
\label{subsubsec:unrecognisable}

Although most publications consider adversarial examples in close resemblance to inputs drawn from the training set, the authors of~\cite{nguyen2015deep} show that it is easy to produce images that are completely unrecognisable by human which are able to fool \ac{DNN} with over $90\%$ accuracy.
The authors use an evolutionary algorithm called multi-dimensional archive of phenotypic elites~\cite{cullyrobots} which allows to simultaneously evolve a population that contains individuals which score well on many classes.
Fitness is determined by sending the image to a \ac{DNN}. 
If the image generates a higher prediction score for any class than has been seen before, the newly generated individual becomes the champion for that class.

The algorithm finds images that are completely unrecognisable, but get classified with very high confidence by \ac{DNN}.
It is believed that synthetic images are far from the decision boundary and deep into a classification region, therefore the images generate high confidence scores even though they are far from the natural images in the class.
\subsubsection{The \ac{opa}}
\label{subsec:one_pixel}

The One Pixel attack~\cite{su2017one} is based on the $L_0$ norm, described in Section~\ref{subsec:norm_balls}. 
However, instead of using gradient based optimisation techniques - which require access to the underlying model - the authors use differential evolution~\cite{storn1997differential} - an evolutionary optimisation method that ensures high population diversity.
This technique only requires access to the softmax layer output.

Formally, the perturbation is encoded into an array optimised by differential evolution. 
One candidate solution contains a fixed number of perturbations and each perturbation is a tuple holding five elements: x-y coordinates and RGB values of one pixel. 
At each iteration candidate solutions are produced using the following formula:
\begin{equation}
	\begin{aligned}
		\eta_i(g + 1)   &=  \eta_{r1}(g) + F(\eta_{r2}(g) + \eta_{r3}(g)), \\
		r1 &\neq  r2 \neq r3, 	
	\end{aligned}
\end{equation}
\noindent
where $\eta_i$ is an element of the candidate solution, $r1, r2, r3$ are random numbers, $F$ is the scale parameter set to be 0.5 and $g$ is the current index of generation.

In order to define a fitness function, only the output of the softmax function is required.
For targeted attacks, the fitness function will aim to increase the probability of a target class; while for un-targeted attacks it will aim to decrease the probability of the true class.

While it requires less information about the model under attack, the \ac{opa} attack performs poorly when compared to gradient-based methods.

\subsubsection{Adversarial Examples for Other \ac{ML} Tasks}
\label{subsec:real_world}

The threat to adversarial example was also explored for applications other than object recognition.
Of particular interest is the task of malware detection~\cite{grosse2016adversarial, hu2017generating, laskov2014practical, xu2016automatically, kreuk2018adversarial}.
Other tasks explored come from the field of reinforcement learning~\cite{behzadan2017vulnerability, huang2017adversarial, lin2017tactics},
speech recognition~\cite{carlini2016hidden, carlini2018audio} or facial recognition~\cite{sharif2016accessorize}.
However, it is not common to present a accurate threat model or consider the economics of carrying an attack using adversarial examples as opposed to other methods.
Therefore, it is not clear if these threats are really important.

Other publications explore the impact of adversarial examples in the physical world~--~by printing corrupted images~\cite{evtimov2017robust, kurakin2016adversarial} or altering the image acquisition device (\eg~phone camera, digital camera, \etc)~\cite{moosavi2017universal}.

\else

Besides the attacker model introduced in Section~\ref{sec:taxonomy}, we  characterize the performance of every attack in a qualitative and a quantitative manner.
Qualitatively, we evaluate the attack's performance against different \ac{ML} models and defenses, while quantitatively, we evaluate the attack's impact in the literature.
Each dimension for both the qualitative and the quantitative assessment is measured on a categorical scale, ranging from low(*), medium(**) and high(***).
The final score is computed by averaging over the attributes described below.

We select three dimensions for the qualitative evaluation:
\begin{itemize}
	\item \emph{Attack strength.} The attack strength evaluates how powerful an attack is against various \ac{ML} models and defenses. It is based on the lower and upper bounds discussed in Section~\ref{sec:robustness_eval} (*, ** or ***, depending on the bound size and the fooling rate). We also consider untargeted attacks (*), less powerful than targeted ones (***).  Moreover, some attacks can be used to find universal perturbations (***)~--~which can be used with any testing sample~--~while others can only discover  specific perturbations specific to one sample (* or **, depending on the perturbation size).
	Further on, some attacks have been successfully tested against a large array of defenses (** or ***, depending on the defense types), while others not (*).  The attack strength hides an inherent trade-off with the attack's complexity.
	\item \emph{Attack complexity.} The complexity of an attack evaluates the resources needed to mount it, but also the underlying assumptions of the technique used. Some attacks use simpler, single-shot, methods to generate the perturbation (*), while others use more complex, iterative methods (** or ***, depending on the number of steps needed and the method used).
	\item \emph{Experimental setup.} The experimental setup evaluates how thorough the attack was tested, on which data sets (* for MNIST or similar, ** for CIFAR-100 or similar and *** for ImageNet or similar) and which models were used during evaluation (* for simple, feed forward models, ** for deep feed forward or convolutional models and *** for deep convolutional models).
		Moreover, the experimental setup evaluates if the attack has been tested in practice (** or ***, depending on the use case presented) or not (*).
\end{itemize}
\noindent
The quantitative evaluation is based on bibliometrics indicators, namely the ratio between the number of citations and the number of months since publication as indicated by Google Scholar.
Although bibliometrics are not a direct indicator of quality, in this paper they are used to explore the areas where research concentrates most and which papers have potential for a novice reader.

The results are presented along the threat modeling introduced in Section~\ref{sec:taxonomy_attacks} in Table~\ref{tbl:attacks}.
Note that except attacks based on geometric transformations, which have lower complexity for high strength, most strong attacks are also complex.
In particular, optimization based attacks are, on average, both strong and complex.
The distribution is different for sensitivity based attacks, where we can find complex attacks with minimum strength and strong attacks with medium complexity (\eg~Madry).
Training generative models is also more complex because, as we will discuss in the next section, they require to train a generator and a discriminator, but also to perform extra operations.
Nonetheless, generative models have medium to high strength.
More details follow in the next section where the attacks are presented based on the attacker knowledge and the attack strategies introduced earlier.

\subsection{White box attacks}
\label{sec:white_box}

In the white box scenario, an attacker has complete knowledge about the model under attack, its parameters and the data used for training or testing.
Therefore, an attacker can completely replicate the model or learn the data generation distribution, such that it can generate new samples.

\subsubsection{\textbf{Noise based attacks}}
\label{sec:noise}
We begin with attacks which craft perturbations from white noise because they are more common. The presentation follows the attack strategies discussed in Section~\ref{sec:taxonomy_attacks}.
An illustration of noise perturbations is shown in Figure~\ref{fig:adv_examples}.

\begin{figure}[h]
	\centering
	\begin{subfigure}{0.49\textwidth}
		\centering
		\includegraphics[height=4cm, width=5cm, keepaspectratio]{figs/attack-ex-1}
		\subcaption{Adversarial examples~\cite{szegedy2013intriguing}.}
		\label{fig:adv_exam}
	\end{subfigure}
	\begin{subfigure}[]{0.49\textwidth}
		\centering
		\includegraphics[height=4cm, keepaspectratio]{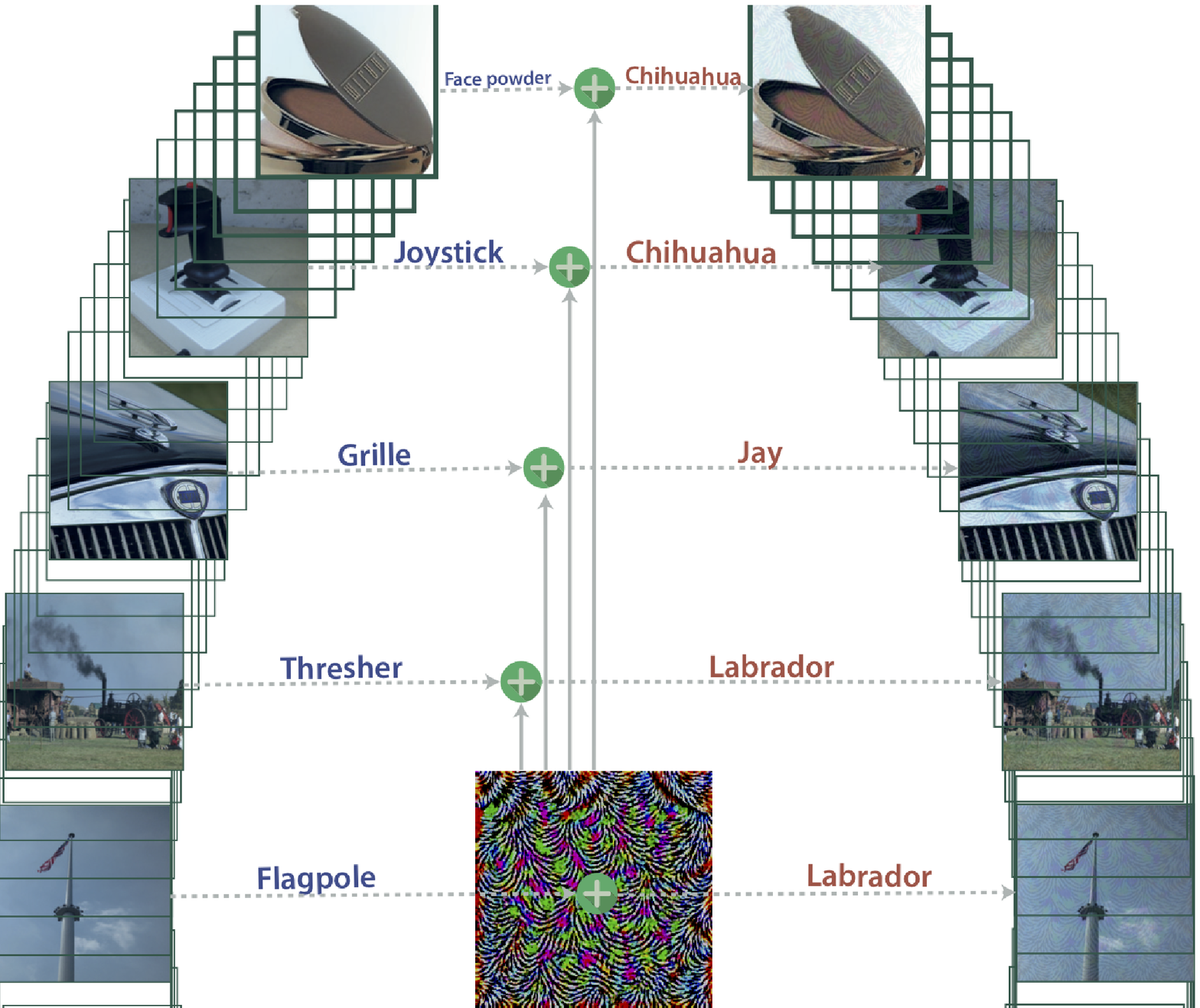}
		\subcaption{Universal Adversarial Perturbation~\cite{moosavi2017universal}.}
		\label{fig:universal}
	\end{subfigure}
	\caption{An illustration of adversarial examples. (a) Specific perturbations for each new input. The images in the first column are inputs correctly classified, the ones in the middle are the perturbations used to induce a misclassification and the images on the last column (right) are the resulting adversarial examples. (b) Universal perturbations~--~only one perturbation can be applied to any picture on the left to generate adversarial examples on the right.}%
	\label{fig:adv_examples}%
\end{figure}

\paragraph{Optimization Attacks.}
All attacks rely on optimization constructs.
For example, in order to discover sensitive features, one can use the gradient taken \wrt~one input.
However, in this section we review the attacks that use optimization methods in order to search for solutions to \eqq~(\ref{eq:adversarial_generic}) or alternative forms and constraints.

\citeauthor{szegedy2013intriguing}~\cite{szegedy2013intriguing} were the firsts to discover the \acp{DNN} sensitivity to adversarial examples and coin the term adversarial examples.
The authors used limited memory box constrained optimization (L-BFGS) in order to approximate the minimum perturbation needed to change the label of an input, as in \eqq~(\ref{eq:adversarial_generic}).
The initial experimental results showed that the minimum average distortion is very low, and can not be distinguished by human observers.
This publication was the first to discover this behavior of \acp{DNN} and triggered a large array of publications.

\citeauthor{carlini2017towards}~\cite{carlini2017towards} proposed an alternative form to \eqq~(\ref{eq:adversarial_generic}) in which the classification function is replaced due to its non-linear character.
The alternative function is chosen \stt~$f(\vx + \eta) = \hat{y}$ if and only if $f(\vx + \eta) \le 0$ (this is a linear constraint).
This formulation allows the use of powerful optimization methods in order to search for very small perturbations and remains, at the time of writing this paper, one of the state-of-the-art attacks.
Given its popularity, we will refer to this attack as the Carlini and Wagner attack.

The DeepFool attack~\cite{moosavi2016deepfool} assumes \acp{DNN} behave linear around an input and projects the point to a separation plane between two classes.
In the case of multi-class classifiers, the separation plane represents the face of a polyhedron whose faces are discriminants to other classes.
The attack assumes the surface that separates two classes is defined by an implicit equation \stt~the geometrical normal is equal to its gradient vector.

\citeauthor{moosavi2017universal}\cite{moosavi2017universal} use the DeepFool attack iteratively for all images in the training data set in order to find an universal perturbation~--~one that can be applied to any sample from the training set.
An illustration of universal perturbation can be seen in Figure~\ref{fig:universal}.
Several passes over the training data set are required in order to improve the quality of the perturbation.
At the end of the procedure, however, the method succeeds in finding \emph{image-agnostic} perturbations which cause misclassifications with high confidence.

Although most publications consider adversarial examples in close resemblance to inputs drawn from the training set, \citeauthor{nguyen2015deep}~\cite{nguyen2015deep}
	craft special images which can not be understood by human observers, but can generate a targeted classification with high accuracy.
The authors leverage evolutionary algorithms in order to evolve candidate solutions that can fool \acp{DNN}.
The fitness function evaluates candidates by sending the image to a target \ac{DNN}.

Similarly, \citeauthor{su2017one}~\cite{su2017one} use differential evolution~--~an evolutionary optimization method that ensures high population diversity~--~to generate adversarial perturbations.
The output of the final softmax layer of a \ac{DNN} is used as a fitness function for the algorithm; for targeted attacks the fitness function aims to increase the probability of a certain class, while for untargeted attacks it aims to decrease the probability of the true class.
While it requires less information about the model under attack, this attack performs poorly when compared to gradient-based methods.

The attacks using the optimization strategy are generally precise in finding minimum perturbations or very good approximations of it.
Although they are more complex than other types of attacks (because they require more iterations or use different optimization methods), most optimization based attacks are still state-of-the-art.
From a security standpoint, an interesting approach is also to generate images that are not intelligible to humans, but can cause misclassifactions.
This phenomenon links to several points in Section~\ref{sec:discussion}.
\paragraph{Sensitivity Analysis Attacks.}
In order to overcome the speed drawbacks of the L-BFGS attack and following the linearity hypothesis introduced in the previous section, \citeauthor{goodfellow2014explaining}~\cite{goodfellow2014explaining} proposed to sum small perturbations in the direction of the loss gradient taken \wrt~one input.
Moving a small step in the direction of the gradient results in taking a step towards maximizing the loss function.
Analyzing the gradient of the loss function \wrt~the input is often referred to as \emph{sensitivity} or \emph{saliency} analysis~\cite{yeung2010sensitivity} and reveals the importance of one feature in the decision process.
Formally, the perturbation resulting from this simple attack (called \ac{fgsm}) is defined as: $\eta = \eps \sign \left( \nabla_\vx l(\vtheta, \vx, y) \right)$.
The value of $\epsilon$ controls the size of a perturbation and impacts the sensitivity of a human observer.
Because this attack only requires the computation of the gradient vector, it is fast to apply and can be used to quickly generate new training data.
However, this method trades precision for speed.

In order to increase the precision of \ac{fgsm}, \citeauthor{kurakin2016aadversarial}~\cite{kurakin2016aadversarial} propose to apply this method iteratively and, similar to the gradient clipping procedure, limit the value of a pixel by an upper bound.
Moreover, the authors propose to use this method in a targeted fashion, by maximizing the likelihood for a chosen class.
\citeauthor{madry2017towards}~\cite{madry2017towards} extend the iterative attack proposed in~\cite{kurakin2016aadversarial} by iteratively applying \ac{pgd}~\cite{boyd2004convex} in order to search for a perturbation that can approximate the p-norm ball around an input.
The use of \ac{pgd} suggests the approximation is tractable and a large part of the loss landscape can be explored through it.
If such a perturbation can be found and can approximate the entire p-norm ball, then protecting against it means one can protect against any perturbation in the norm ball.

Sharp curvatures near a data point can mask the true direction of steepest ascent and burden the discovery of adversarial examples with single-shot gradient methods~\cite{kurakin2016adversarial}.
In order to escape this phenomenon, \citeauthor{tramer2017ensemble}\cite{tramer2017ensemble} introduce an attack that precedes single-shot attacks with a randomization step.
\ac{rssa}~\cite{tramer2017ensemble} searches for an adversarial example starting every time from a random vicinity of the input data point, thus avoiding gradient masking (a phenomenon discussed in Section~\ref{subsec:gradient_ob}).

Dong \etal~\cite{dong2017boosting} proposed to boost the iterative version of \ac{fgsm} using gradient momentum.
As in the case of gradient descent, momentum can stabilize the update directions and help escape poor local minimum/maxima by accumulating a velocity vector in the gradient direction.
Setting the velocity vector to $0$ is equivalent to the normal \ac{fgsm} attack.

\citeauthor{papernot2016limitations}~\cite{papernot2016limitations} introduced an attack based solely on saliency analysis~\cite{yeung2010sensitivity}.
In order to discover the importance of each pixel in the decision process, a saliency map is generated by computing the forward derivative (forward Jacobian) of the function learned by a \ac{DNN}.
This method contrasts with early methods introduced in this section, which use the backward gradient of the loss function.
The forward derivative allows to find better input feature, which ultimately lead to significant changes in the \ac{DNN} output.
However, its inherent computational costs for big images limits the impact of this method.

Similarly, \citeauthor{khrulkov2017art}~\cite{khrulkov2017art} used the  Jacobian matrix to construct universal adversarial perturbations exploiting the singular value vectors of the feature maps while
\citeauthor{huang2015learning}~\cite{huang2015learning} use the Jacobian matrix to compute the linear approximation of a \ac{DNN} output.
The approximation for the normal and the perturbed case gives the minimal perturbation.

For cases when the gradient can not be well approximated~--~a phenomena called gradient masking and presented in Section~\ref{subsec:gradient_ob}~--~\citeauthor{athalye2018obfuscated}~\cite{athalye2018obfuscated} introduced an attack which replaces the gradient of a non-differentiable layer with a differentiable approximation.
Thus, the gradient of a \ac{DNN} can be approximated by performing the forward pass through the whole network, but on the backward pass each layer is replaced by its approximation.
As long as the two functions are similar, the slightly inaccurate gradients prove useful in constructing adversarial examples.

\citeauthor{chen2017ead}~\cite{chen2017ead} extrapolate the Carlini and Wagner attack~\cite{carlini2017towards} from elastic-net regularization~--~a mixture of penalty functions used for high-dimensional feature selection.
This algorithm is a bridge between optimization and sensitivity analysis methods because it uses different minimization techniques to discover sensitive features and, thus, adversarial examples.

The attacks based on sensitivity analysis are, generally, faster than the attacks based on the optimization strategy and can be more easily applied in training models with adversarial examples.
In fact, training with an approximation of the p-norm ball around an input generated with \ac{pgd} is one of the state-of-the-art defenses.
However, these attacks are often less precise than the attacks based on the optimization strategy.
Evaluating defenses against weak attacks might lead to a false sense of security, a topic we touch upon in Section~\ref{subsec:gradient_ob}.
\paragraph{Generative attacks.}
Until now we introduced attacks that modify a sample from the data generating distribution by adding an adversarial perturbation.
However, data samples might not always be available.
In this section, we cover adversarial attacks based on generative models - a class of machine learning algorithms that learn to estimate a probability distribution by looking at samples drawn from it.
The  model is later used to produce artificial examples belonging to the same distribution.
The goal is to generate examples that are similar to training samples, but not exactly the same.
In particular, two generative models are used: (1) \ac{vae}~\cite{kingma2013variational, danilo2014stochastich} and (2) \ac{gan}~\cite{goodfellow2014generative}.

\citeauthor{baluja2017adversarial}~\cite{baluja2017adversarial} trained a \ac{DNN} that transforms an input into an adversarial example.
The transformation network is trained to fool a target network or to generate examples transferable to a larger range of networks.
The authors use two approaches to generate adversarial examples: (1) using a residual network to generate a perturbation and (2) using auto-encoders.
In practice, using auto-encoders yields better results and successfully scales to large data sets.
The model is efficient to train, fast to execute, and produces diverse adversarial examples.
Moreover, once the network is trained, the generation of adversarial examples only takes one step, suggesting its efficiency in adversarial training.
However, this approach was not investigated.

\citeauthor{zhao2017adversarial}~\cite{zhao2017adversarial} search for adversarial examples in the deep representation of the input data (instead of searching directly in the input space).
For this, a generator is trained to map random noise vectors  to samples  from the input distribution (from noise to input domain).
A second generative model, called inverter, is trained to map data instances to corresponding dense representations (from input domain to noise).
This is equivalent to finding an adversary in an underlying vector space which defines the data generation distribution and then map it back to the input space with the help of a generative model.
However, the perturbations obtained through this method are far from the original inputs and can easily be spotted by human observers.

\citeauthor{poursaeed2018generative}~\cite{poursaeed2018generative} trained a generative model to generate image dependent and independent perturbations; leading to specific or universal adversarial examples.
However, in the universal case, the generator's loss function is a linear combination of the loss functions of the target models, which makes it heavily dependent on the model under attack.
\citeauthor{song2018constructing}~\cite{song2018constructing} used a generative model, similar to~\cite{baluja2017adversarial} against strong, certified defenses.
Their results show that generative models can easily break defenses focusing on the p-norm ball, even though the defense guarantees no adversarial perturbation can be found in this region.

Generative models learn to approximate a distribution, from which we can sample new data.
In the case of adversarial examples, generative models learn the distribution of perturbations, assuming that all perturbations are identically distributed.
Although this assumption can be restrictive, even in this setting they are able to find powerful perturbations.
However, training generative models often requires more resources and are rarely used in practical applications of adversarial examples.

\subsubsection{\textbf{Geometric Attacks}}
\label{subsec:geometric_attacks}

By now we have introduced algorithms that craft perturbations from noise.
In this section, we review some attacks that use more natural, geometric transformations such as rotation or translation to construct adversarial examples.

\citeauthor{engstrom2017rotation}~\cite{engstrom2017rotation} showed that only simple transformations - rotations and translations - are sufficient to create adversarial examples.
These transformations are easy to craft and realistic in various operational scenarios.
The authors propose several methods ranging from randomly sampling different transformations to grid search or gradient approaches.
Depending on the chosen method, the drop in accuracy ranges from $34-90\%$ on models trained with data augmentation techniques (already including affine transformations).

ManiFool~\cite{kanbak2017geometric} searches for the smallest, worst-case, geometrical transformation that can fool \acp{DNN}.
Similar to the perturbations generated by optimization or sensitivity based methods, these perturbations are imperceptible to human observers.
The main idea behind ManiFool is simply to iteratively move from an image sample towards the decision boundary where the classification decision changes, while staying on the geometrical transformation manifold.
In particular, the authors use a combination of rotation, translation and scaling transformations.

\citeauthor{xiao2018spatially}~\cite{xiao2018spatially} proposed to change the geometry of the scene, while keeping the original appearance.
Instead of imposing norm constraints on the pixel space, the authors introduce a new regularization loss on the local geometric distortion.
The perceptual quality of the adversarial examples remains high, while most defenses fail against this attack.
In a similar fashion, \citeauthor{zhang2019limitations}~\cite{zhang2019limitations} change the geometry of the scene before applying the Carlini and Wagner attack. 
In this case, small transformations find inputs that are far away from the training data and lie in 'blind spots' which are not covered by defenses.  
Once perturbed, these inputs lead to powerful attacks able to fool certified defenses.
Although in this case the transformation is not the attack, it is an important preprocessing step which enables  it.

Instead of using one transformation, \citeauthor{athalye2017synthesizing}~\cite{athalye2017synthesizing} search for perturbations over a distribution of transformations. 
In the 2D case, the authors use a random distribution of affine transformations, while in the 3D case they consider textures and shape.
This technique is able to synthesize adversarial examples robust to defenses which use input transformations.

\citeauthor{pei2017towards}~\cite{pei2017towards} proposed a framework for verifying the robustness of computer vision algorithms against natural perturbation such as rotation, translation or reflection (but also contrast, brightness or erosion). 
These adversarial examples are proved to be strong against state-of-the art classifiers, making the authors argue they are even stronger than gradient-based attacks.
However, more evidence is needed to test this claim.

Geometrical transformations are thought to occur more frequently in the data acquisition process.
However, most methods introduced in this section require carefully tuned transformations, which might limit this claim.
Nonetheless, sensitivity to geometric transformations for models trained with data augmentation techniques shows that algorithms do not learn to abstract general transformations, but only to fit the training data.


\subsection{Black box Attacks}
\label{subsec:black_box}

\fi
    \section{Defences}
\label{sec:defences}

\if 0\mode

	\paragraph{Taxonomy of defences.} Following the defender's capabilities, presented in Section~\ref{subsec:defender_goals}, we identify three major classes of defences:
\begin{enumerate}
	\item \emph{Reactive} defences. In this case defenders leverages pre-processing techniques in order to alleviate the impact of adversarial examples, or employ detection mechanisms in order to discard them completely.
	\item \emph{Obfuscation} defences. In this case a defender aims to hide or obfuscate sensitive traits of a model (\eg~gradients) in order to alleviate the impact of adversarial examples.
	\item \emph{Proactive} defences. In this case defenders aim to build models that are natively robust to small adversarial perturbations.
\end{enumerate}

\if -\mode \paragraph{Overview of reactive defences (Section~\ref{subsec:reactive_defences}).} \else \fi%
In the following sections we will review two classes of \emph{reactive defences}, namely, \emph{detection of adversarial examples} \if 0\mode (Section~\ref{subsubsec:detection} \else\fi and \emph{input transformations} \if 0\mode (Section~\ref{subsec:input_transform}) \else\fi.
Detection techniques assume that adversarial examples have special characteristics or come from a different data generation distribution than normal inputs.
Therefore, it was hypothesised that a detector can learn to distinguish between the two~\cite{song2017pixeldefend, meng2017magnet, ghosh2018resisting, lee2017generative}.
In some cases, the learning procedure is embedded in the main model, by defining a special class corresponding to adversarial inputs.
In others, a detector is trained separately, using specially crafted features.
\if 0\mode
For example, using the output of hidden layers~\cite{metzen2017detecting}, using statistics on the output of convolutional layers~\cite{li2017adversarial} or measuring the distance between a point and the data manifold, at the output layer of \ac{DNN}~\cite{feinman2017detecting}.
An interesting approach, besides detection of adversarial examples, is the attempt to recover the original samples and correctly classify them.
In spite of the fact that promising results are presented in all papers, most detectors can be bypassed with iterative attacks (Section~\ref{subsec:overall_defences}).
\else \fi

Defences belonging to the \emph{input-transformation} class use pre-processing techniques such as compression or bit-depth reduction~\cite{guo2017countering} in order to remove the effect of adversarial perturbations.
The field sometimes overlaps with adversarial detection methods, where input transformations are used in order to build adversarial detectors.
However, in the former case, the purpose is not to discard an input, but to pre-process it in such a way that adversarial perturbations become useless.
A promising feature of input transformation methods is their speed of adoption and adaptation.
In most cases, they can be applied at test time and do not require any training time.
However, input transformations are domain specific and hard to transfer between different machine learning tasks.
For example, one has to develop different pre-processing techniques for speech and image recognition.

\if 0\mode \bigskip \else \fi
\if 0\mode \paragraph{Overview of defences by obfuscation.} \else\fi%
A natural answer to attacks that use sensitive features or special traits of models, is to diminish or \emph{obfuscate} such features, in order to restrict the attacker's capabilities.
This phenomenon was first described as \emph{gradient masking} - the use of models that do not have useful gradients~\cite{papernot2017practical}.
However, the range of defences in this category extends from the use of models that do not have useful gradients to modifying the models in order to lower the gradient amplitudes or to remove features that do not contribute to classification.

A peculiar instantiation of the gradient masking phenomenon, that affects the evaluation of many defences, was named \emph{obfuscated gradients}~\cite{athalye2018obfuscated}.
This corresponds to instances where (1) the gradients are non-existent or incorrect, caused intentionally through non-differentiable operations or un-intentionally through numerical instability, (2) the gradients depend on test-time randomness or (3) the model is subject to vanishing or exploding gradients~\cite{pascanu2013difficulty}.
Obfuscated gradients is a phenomenon prevalent across \ac{ML} models and leads to an amplified effect of defences.
The authors of~\cite{athalye2018obfuscated} develop an offensive technique able to overcome these limitations and exploit the gradient informations even in such scenario.
This result shows that defence by obscurity is not a strong method to employ against adversarial examples.
Defence obfuscation can be achieved through different techniques that heavily overlap with adversarial training or new architectures~--~which also serve other purposes.
Because of this overlap, we choose to not discuss these techniques in a separate section and only signal the gradient obfuscation effect whenever the case.

\if 0\mode \bigskip \else \fi
\if 0\mode \paragraph{Overview of proactive defences (Section~\ref{subsec:proactive_defences}).} \else\fi%
Defences that fall in the last class~--~\emph{proactive}~--~aim to build models that are natively robust to small perturbations.
An initial attempt to actively increase the \ac{DNN} robustness to adversarial examples, suggested in the initial paper~\cite{szegedy2013intriguing}, is to include adversarial examples in the training set.
This procedure was scalable with the advent of fast methods to generate adversarial examples, which increased the training speed and lead to scalable \emph{adversarial training}\if 0\mode~(Section~\ref{subsec:adv_training})\else\fi.
Within this category we can distinguish \if 0\mode and separate \else between \fi a set of defences that formulate the learning problem as a \emph{min-max} (or saddle) problem and employ optimisation techniques\if 0\mode(Section~\ref{subsec:minmax})\else\fi.
\if 0\mode
The reason behind this choice is that min-max learning uses only perturbed samples during the training process; as opposed to adversarial training where both perturbed and normal inputs are used.
\else
The distinction can be made because min-max learning uses only perturbed samples, as opposed to adversarial training where both perturbed and normal inputs are used.
\fi

As discussed in Section~\ref{sec:attack_models}, most defences act against small perturbations.
Some techniques try to completely remove the sensitivity to adversarial examples by designing new models and \ac{DNN} architectures.
\if 0\mode
We gather all defences that alter the learning process in any way - \eg~modify any of the layers or imposing other restrictions than gradient obfuscation~--~into one category called \emph{architectural defences} \if 0\mode (Section~\ref{subsec:defences_arch}) \else\fi.
\else
We can, thus, separate the defences that alter the learning process by modifying any of the layers or imposing other restrictions than gradient obfuscation under the tag \emph{architectural defences}.
\fi

A normal follow up of the linear hypothesis, introduced in Section~\ref{sec:causes}, is to increase the non-linearity of \ac{DNN}.
Other authors have suggested to increase the capacity of a neural network in order to increase robustness.
Both methods affect training time.
Although, at the beginning, these approaches seem similar to architectural defences, they only change the \emph{hyper-parameters} of a network (number of layers or activation function).
\if 0\mode Therefore, they are reviewed in a different section~(Section~\ref{subsec:hyperparam}).\else \fi

An interesting approach to solving the adversarial examples issue is to search for the smallest space at $L_p$ distance from an input where the classifier is robust.
Defences falling in this category aim to certify that a classifier can work under some levels of uncertainty \ie~is robust within some bounds.
\if 0\mode
Therefore, they are reviewed separately, as \emph{provable adversarial defences}, in Section~\ref{subsubsec:provable}.
Towards the end we review a set of defences that use \emph{generative models} (Section~\ref{subsec:defences_gans}).
\else

\fi
Before delving into each individual class, we provide a complete list of the defences in Table~\ref{tbl:defences}.

	\begin{table}
	\centering
	\begin{tabular}{|l|l|l|l|l|}
		\toprule%
		Defence & \specialcell{Type} &%
		  		 \specialcell{Method} &%
		  		 \specialcell{Threat Model} &%
		  		 \specialcell{Defeated} \\
		\midrule
		Statistical Detection~\cite{grosse2017statistical} & Reactive & Detection & Yes  & Yes \\ \hline
		Binary Classification~\cite{gong2017adversarial} & Reactive & Detection & No &  Yes  \\ \hline
		In-Layer Detection~\cite{metzen2017detecting} & Reactive & Detection & No & Yes  \\ \hline
		Detecting from Artifacts~\cite{feinman2017detecting} & Reactive & Detection & No & Yes \\ \hline
		SafetyNet~\cite{lu2017safetynet} & Reactive & Detection & No & Yes  \\ \hline
		Saliency Data Detector~\cite{zhang2018detecting} & Reactive & Detection & No & Yes  \\ \hline
		Linear Transformations Detector~\cite{bhagoji2018enhancing} & Reactive & Detection & No & Yes \\ \hline
		Key-based Networks~\cite{zhao2018detecting} & Reactive & Detection & No & Yes  \\ \hline
		Ensemble Detectors~\cite{abbasi2017robustness} & Reactive & Detection & No & Yes  \\ \hline
		Generative  Detector~\cite{lee2017generative} & Reactive & Detection & No & -  \\ \hline
		Convolutional Statistics Detector~\cite{li2017adversarial} & Reactive & Detection & No & Yes  \\ \hline
		Feature Squeezing~\cite{xu2017feature} & Reactive & Detection & Yes & Yes  \\ \hline
		PixelDefend~\cite{song2017pixeldefend}  & Reactive & Detection & No & Yes \\ \hline
		MagNet~\cite{meng2017magnet} & Reactive & Detection & Yes & Yes  \\ \hline
		VAE Detector~\cite{ghosh2018resisting}  & Reactive & Detection & No & Yes  \\ \hline
		Bit-Depth~\cite{guo2017countering} & Reactive & Input Transformation & Yes & Yes  \\ \hline
		Basis Transformations~\cite{shaham2018defending} & Reactive & Input Transformation & Yes &  Yes \\ \hline
		Randomised Transformations~\cite{xie2017mitigating} & Reactive & Input Transformation & Yes & No \\ \hline
		Thermometer Encoding~\cite{buckman2018thermometer} & Reactive & Input Transformation & No & Yes \\ \hline
		Blind Pre-Processing~\cite{rakin2018blind} & Reactive & Input Transformation & No & Yes   \\ \hline
		Data Discretisation~\cite{chen2018improving} & Reactive & Input Transformation & No & Yes  \\ \hline
		Adaptive Noise~\cite{liang2017detecting} & Reactive & Input Transformation & Yes & Yes  \\ \hline
		FGSM Training~\cite{goodfellow2014explaining} & Proactive & Training & No & Yes \\ \hline
		Gradient Training~\cite{sinha2018gradient} & Proactive & Training & No &  Yes \\ \hline
		Gradient Regularisation~\cite{lyu2015unified} & Proactive & Training & No & Yes \\ \hline
		Structured Regularisation~\cite{roth2018adversarially} & Proactive & Training & Yes & Yes \\ \hline
		Robust Training~\cite{shaham2015understanding} & Proactive & Robust Training & No & Yes \\ \hline
		Strong Adversary Training~\cite{huang2015learning} & Proactive & Robust Training & No & Yes \\ \hline
		CFOA Training~\cite{madry2017towards} & Proactive & Robust Training & Yes & Yes \\ \hline
		Ensemble Training~\cite{tramer2017ensemble} & Proactive & Robust Training & Yes & Yes \\ \hline
		Stochastic Pruning~\cite{dhillon2018stochastic} & Proactive & Robust Training & No &  Yes \\ \hline
		Distillation~\cite{hinton2015distilling} & Proactive & Architecture & No & Yes \\ \hline
		Parseval Networks~\cite{cisse2017parseval} & Proactive & Architecture & No &  - \\ \hline
		Deep Contractive Networks~\cite{gu2014towards} & Proactive & Architecture & No & Yes \\ \hline
		Biological Networks~\cite{nayebi2017biologically}& Proactive & Architecture & No & Yes \\ \hline
		DeepCloak~\cite{gao2017deepcloak} & Proactive & Architecture & No & Yes \\ \hline
		Fortified Networks~\cite{lamb2018fortified} & Proactive & Architecture & Yes & No \\ \hline
		Rotation-Equivariant Networks~\cite{dumont2018robustness} & Proactive & Architecture & No & Yes \\ \hline
		HyperNetworks~\cite{sun2017hypernetworks} & Proactive & Architecture & No & Yes \\ \hline
		Bidirectional Networks~\cite{pontes2018bidirectional} & Proactive & Architecture & No & Yes \\ \hline
		DAM~\cite{krotov2016dense} & Proactive & Architecture & No & - \\ \hline
		Certified Defences~\cite{raghunathan2018certified} & Proactive & Certified & - & - \\ \hline
		Formal Tools~\cite{katz2017reluplex, ehlers2017formal, huang2017safety, ruan2018reachability}  & Proactive & Certified & - & - \\ \hline
		Distributional Robustness~\cite{sinha2018certifying}  & Proactive & Certified &  - & - \\ \hline
		Convex Outer Polytope~\cite{kolter2017provable}  & Proactive & Certified & - & - \\ \hline
		Lischitz Margin~\cite{tsuzuku2018lipschitz}  & Proactive & Certified & - & - \\ \hline
		Defence Gan~\cite{samangouei2018defense}  & Proactive & Generative & Yes & Yes \\ \hline
		FB-GAN~\cite{bao2018featurized}  & Proactive & Genearative & No  & - \\
	\bottomrule
	\end{tabular}
	\caption{Catalog of defences against adversarial examples.}
	\label{tbl:defences}
\end{table}

	\subsection{Reactive Defences}
\label{subsec:reactive_defences}

Reactive defences do not change the model under attack.
These methods act early in the processing pipeline; before an input reaches a model.
Within this class, we distinguish between (1) defences that \emph{detect} adversarial examples and thus process them differently and (2) defences that apply \emph{input transformations} to all inputs, before they reach the classifier.
Since the latter do not distinguish between adversarial and normal inputs, a first requirement is for classifiers to maintain accuracy on normal, transformed, inputs.

\subsubsection{Detection of adversarial examples}
\label{subsubsec:detection}

The algorithms presented in this section assume that adversarial examples are sampled from a different distribution than benign data.
Thus, a classifier can be trained to distinguish between normal and adversarial inputs.
One can either add a new class, corresponding to adversarial examples, to the main model or train a separate detector; with a different architecture an features.
As we shall see, adversarial detection is strongly correlated with the attack types and the hyper-parameters.
Thus, an universal detector is hard to develop.

While most papers propose to discard adversarial examples, towards the end we introduce some attempts to recover the original input and correctly classify it.
This is an important step for safety critical applications where no input can be discarded.
Final remarks regarding the efficacy of adversarial examples are given in Section~\ref{subsec:overall_defences}

\subsubsection*{Statistical Detection of Adversarial Examples}
\label{subsubsec:statistical_detection}

Gross \etal~\cite{grosse2017statistical} use statistical testing to check if adversarial examples are outside the training distribution.
In particular, they use a statistical test designed for high dimensional data, formalised as the biased estimator of the maximum mean discrepancy~\cite{gretton2012kernel}.
The null hypothesis states that an adversarial and a normal input are drawn from the same distribution.

In order to reject the null hypothesis, the test needs at least $50-100$ examples from each attack and class.
This requirement limits the applicability of statistical testing. 
In order to overcome this limitation, the authors propose to add a new class to the model and train it to detect adversarial examples.
This augmentation mechanism allows the model to classify a sample as benign or adversarial and, in the former case, to return its true label.
Experimental results show the approach is feasible, however, the results are tested on models with very low capacity and a low number of classes.
It is not clear if this approach scales to bigger models or data sets.
Moreover, the procedure requires data about an attack before a detector can be trained.
Therefore, it remains vulnerable to new attacks.
\subsubsection*{Binary Adversarial Classification}
\label{subsubsec:adversarial_twins}

Gong, Wang and Ku~\cite{gong2017adversarial} suggest that training a binary classifier to detect between benign and adversarial examples is possible.
The authors train a separate neural network with adversarial examples developed using \ac{fgsm}, \ac{jsma} and \ac{illcm} (Section~\ref{subsec:sensitivity}).
While achieving almost maximum accuracy for one type of attack, their proposal does not generalise to different attacks or different hyper-parameters.
This suggests two directions: that each attack type generates a unique distribution of samples or their model heavily over-fits.
\subsubsection*{In Layer Detection}
\label{subsubsec:on_detecting_adv}

Metzen \etal~\cite{metzen2017detecting} train a separate neural network to detect adversarial examples.
However, instead of using the perturbed input for training, their approach trains a detector with the output of hidden layers.
In particular, they use a ResNet~\cite{he2016deep} architecture and 'plug-in' the detector at the output of various residual blocks.

The authors study two cases, corresponding to different attack models: (1) a static adversary where the attacker has access to the classification model and (2) a dynamic adversary where the attacker has access to both the classification model and the detector.
While the experimental results prove the feasibility of this approach, there are some peculiar properties which limit its applicability.
At first, the location for the detector varies with an attack type \ie~by plug-ing in the detector at different locations they obtain different results for different attacks.
Some detectors generalise well for a range of attacks, however, no detector can be considered universal.
Secondly, the precision of a detector is correlated with the attack's hyper-parameters.
These limitations suggest several detectors must be trained, accounting for different attacks or hyper-parameters.
\subsubsection*{Detecting Adversarial Examples from Artifacts}
\label{subsubsec:detecting_artifacts}

Feinman \etal~\cite{feinman2017detecting} train a linear adversarial detector using two features: (1) kernel density estimates in the subspace of the last layer and (2) bayesian uncertainty estimates extracted from the drop-out layers.
Both features measure if a point belongs to the probability distribution of a class.

The first one, kernel density estimates, is a distance-based metric that evaluates how far a point is from a manifold corresponding to a class.
The second feature, bayesian uncertainty, makes use of drop-out variance to estimate the uncertainty distribution of each input.

Experimental results show that a simple linear model using these two feature can be trained with good performance.
However, the defence is dependent on drop-out architectures, a requirement which drastically limits its impact.
Newer developments in batch normalisation~\cite{ioffe2015batch} achieve better regularisation results than drop-out.
\subsubsection*{SafetyNet}
\label{subsubsec:safetynet}

SafetyNet~\cite{lu2017safetynet} enforces an attacker to solve a discrete optimisation problem.
For a layer of ReLU, each activation is quantised in order to generate a discrete code, later used to train an RBF-SVM adversarial detector.
This enforces an attacker to solve an optimisation problem in order  to find optimal values for the ReLU thresholds.
The approach is interesting and scales well to models with high capacity and to larger data sets.


\subsubsection*{Detecting Perturbations with Saliency Data}
\label{subsubsec:detection_saliency}

Zhang \etal~\cite{zhang2018detecting} train a binary classifier with benign and saliency data.
Saliency data is used to identify significant pixels, later used to predict adversary examples.
The approach scales well on models with large capacity and on large data sets.
However, the model is tested only against attacks based on gradient methods, which by design exploit salience information.
\subsubsection*{Detection using Linear Transformations}
\label{subsubsec:detector_linear_transf}

Bhagoji \etal~\cite{bhagoji2018enhancing} use PCA~\cite{shlens2014tutorial} transformations to train adversarial detectors.
The intuition behind using PCA is its ability to identify the directions in which the input data has maximum variance.
By projecting the data data along these axes one can reduce the input dimensions, while preserving variance.

The authors train a separate detector using the same data set as the model under attack, transformed with PCA.
Experimental results prove the feasibility of this approach, however, fail to generalise to a multitude of attacks.
Moreover, the defence is tested for models with low capacity.
\subsubsection*{Detection with Key-based Networks}
\label{subsubsec:key_networks}

Zhao \etal~\cite{zhao2018detecting} propose a new detection mechanism that aims to hide the input label.
This will prevent an attacker from maximising an input given a label and, thus, from creating an adversarial example.
The authors define a ono-to-one encoding scheme from true labels to code vectors.

In order to detect adversarial examples, one can verify if the code vector computed from an input matches the signature of a class with certain precision.
If the output is negative, the input is treated as an adversarial example.
An interesting characteristic of this approach is that knowledge of adversarial examples is not necessary for the detector, thus enhancing generalisation across attacks.
Experimental results show good performance on iterative and adaptive attacks.
However, the approach is tested on networks with low capacity and small data sets.

\subsubsection*{Ensemble Detectors}
\label{subsubsec:ensemble_detection}

Ababsi and Gagne~\cite{abbasi2017robustness} developed an ensemble of detectors based on the confusion matrix of a classifier.
The underlying idea is that adversarial instances originating from a given class tend to fall into a small subset of incorrect classes.
Therefore, developing an ensemble of detectors, which can distinguish between confusion classes, can more easily spot adversarial examples.

While experimental results show an improvement in robustness for simple attacks, it is not clear if this approach will work against adaptive and iterative attacks.
\subsubsection*{Generative Adversarial Trainer}
\label{subsubsec:generative_trainer}

Lee \etal~\cite{lee2017generative} propose a generative training method (Section~\ref{subsec:gan_attacks}) in which two networks are trained alternatively.
The first network generates adversarial examples, while the second tries to correctly classify benign and adversarial examples.
This training procedure outperforms adversarial training with \ac{fgsm}, while maintaining a positive regularisation effect.

\subsubsection*{Detection using Convolution Filter Statistics}
\label{subsubsec:convolution_statistics}

Li and Li~\cite{li2017adversarial} develop an adversarial detector based on features extracted at every layer of a convolutional neural network.
They treat an image as a distribution of pixels that can be used to collect statistics and later used the statistics to train an adversarial detector.
Experimental results show it is not necessary to use statistics coming from all convolutional layers of a \ac{DNN}.
Selecting a number of initial layers is enough for good accuracy.

The selected features are non-differentiable in order to exclude adversaries that use gradient-based attacks.
In particular, the authors collect for each pixel the normalised PCA~\cite{shlens2014tutorial} coefficients, the minimal and maximal values and the 25-th, 50-th and 75-th percentile values of the distribution.
The rationale behind using PCA is that, at every layer, a linear transformation is applied before the non-linear one. 
Therefore, a significant part of the learning process lies within the linear transformation, for which PCA is a standard tool to analyse.
In order to include the features from all layers, the authors define a cascade classifier.

The detector shows good results when evaluated against L-BFGS attacks. 
The authors are also among the firsts to suggest that simple transformations applied to the first layer of a network can reduce the impact of adversarial examples and help recover the original input.
They show that, a simple $3x3$ average filter applied on adversarial examples improves the accuracy with over $73\%$.
\subsubsection*{Feature Squeezing}
\label{subsubsec:feature_squeezing}

Similar to the approach suggested in Section~\ref{subsubsec:convolution_statistics}, Xu and Qi~\cite{xu2017feature} apply small filters to the input image before processing.
This simple technique significantly alleviates the impact of adversarial examples.
In particular, the authors investigate reducing the colour depth and applying a local smoothing (blur) filter.

Using complex input representations can misguide \ac{DNN} into choosing irrelevant features.
Therefore, reducing the input space can restrict the adversarial examples space.
A simple technique for images is to reduce the number of bits used to indicate the colour of a pixel, thus decreasing quality and resolutions.
While human observers are sensitive to these changes, machine learning algorithms are not.
Another technique proposed by Xu and Qi is to reduce the amount of noise in an input.
In the image space, this technique is called spatial smoothing or blur.

The authors develop independent detectors for each feature and compare their output with a normal trained classifier.
Initial experimental results show these cheap techniques do reduce the adversarial space and deserve future investigations.
\subsubsection*{Pixel Defend}
\label{subsubsec:pixeldefend}

Song \etal~\cite{song2017pixeldefend} propose to not only detect adversarial examples, but try to \emph{purify} them and search for the true labels.
They leverage a generative adversarial network (Section~\ref{subsec:gan_attacks}), called PixelCNN~\cite{van2016conditional, salimans2017pixelcnn}, to learn the probability distribution of the training set and use statistical tests to detect if a new input belongs to this probability distribution.
If the input does not belong, it is probably an adversarial example.

However, instead of dis-regarding the input, the authors propose to \emph{purify} it and find its true class by searching for a training sample in the vicinity of the input.
Formally, given the input distribution $p(\vx)$ learned by PixelCNN and a new input $\vxadv$, the algorithm searches for a training sample $\vx$ that maximises $p(\vxadv)$ such that $\vx$ is within the $\epsilon_{\text{defend}}$-ball of $\vxadv$:
\begin{equation}
	\begin{aligned}
	\max_{\vx} p(\vx), \\
	\text{s.t.} \quad  \| \vx-\vxadv \|_\infty \le \epsilon_{\text{defend}}.
	\end{aligned}
\end{equation}
\noindent
Large values of $\epsilon_{\text{defend}}$ might change the meaning of $\vxadv$ while small values might be insufficient to return $\vxadv$ to the correct distribution.
In practice, $\epsilon_{\text{defend}}$ is chosen adaptively.

\noindent
Pixel Defend shows good performance when PixelCNN is used to estimate the probability distribution of a training set containing adversarial examples (as in adversarial training).
A similar approach for reconstruction using \ac{gan} was proposed by Santhanam and Grnarova~\cite{santhanam2018defending}.

\subsubsection*{MagNet}
\label{subsubsec:magnet}

MagNet~\cite{meng2017magnet} trains a model to detect how different a test example is from normal examples.
The detector approximates the distance between one example and the data manifold. 
If the data is bigger greater than a threshold, then the detector rejects the input sample.
The approach is model independent and can be applied to any neural network architecture.

When the adversary is close to the data manifold (bellow threshold), MagNet tries to recover it using auto-encoders (Section~\ref{subsec:gan_attacks}).
This approach shows high accuracy even when tested against iterative and adaptive attacks. 
A similar approach that uses sparse encoding and gaussian mixture models is presented in the next section. 
\subsubsection*{Gaussian Mixture Variational Autoencoders}
\label{subsubsec:gaussian_resisting}

The authors of~\cite{ghosh2018resisting} designed a generative model that finds a latent random variable such that the input and its label become conditionally independent given the latent variable.
The latent space is chosen as a mixture of Gaussians, such that each mixture component represents one of the classes in the data.
Inferring the label given the latent encoding is done by computing the contribution of the mixture components.
Adversarial samples are rejected based on thresholding the encoder and decoder outputs.
For example, if the distance between the sample encoding and the encoding of the predicted class in the latent space is bellow a threshold.

Similar to other defences using \ac{vae}, the authors propose a method for reconstruction and correct classification.
The experimental results show an increase on robustness for the COIL-100 data set.
However, it is not clear if this method can scale to ImageNet data set and can face adaptive attacks.

\subsubsection{Defences Based on Input Transformations}
\label{subsec:input_transform}


As suggested in Section~\ref{subsubsec:detection}, input transformation can help alleviate the impact of adversarial examples.
However, instead of using them to build detectors, the following publications propose the use of input transformations in a pre-processing step.
Although most publications focus on empirical proofs, some recent works~\cite{gopalakrishnan2018combating} introduced a theoretical view on how input sparsity can attenuate the distortion in the softmax layer.
%
A big disadvantage of input transformation techniques is their context dependence - each transformations is specific to a machine learning task.
Thus a method can not transfer from task to task (\eg~object to speech recognition).

\subsubsection*{Countering Adversarial Examples with Bit-depth Reduction and Compression}
\label{subsubsec:countering_transformations}

The authors of~\cite{guo2017countering} suggest the use of bit-depth reduction, JPEG compression, total variance minimisation and image quilting as a pre-processing step of a convolutional classifier.
The idea of using JPEG compression was also explored in~\cite{dziugaite2016study, das2017keeping, shaham2018defending}.
Variance minimisation and image quilting prove, in practice, more effective.
This result is emphasised when the models are trained with modified images - a technique similar to adversarial training.
The transformations scale to large data sets, however, they are not strong enough to withstand adaptive attacks.
\subsubsection*{Basis Functions Transformations}
\label{subsubsec:basis_transformations}

Shaham \etal~\cite{shaham2018defending} experiment with different input transformations: low-pass filters, PCA, JPEG compression, low resolution wavelet approximations and soft-thresholding.
However, neither of those techniques scales to large models and large data sets.
In some settings the model performs better without any employed defences.
\subsubsection*{Randomised transformation}
\label{subsubsec:defence_ransomisation}

Xie \etal~\cite{xie2017mitigating} propose to use two randomisation operations: (1) random resizing of input images and (2) random padding with zeros around the input images.
The approach performs and generalise surprisingly well even on large networks and data sets, without training the classifier with modified inputs.
However, the authors do not discuss accuracy-robustness trade-offs, making it unclear if the approach decreases the performance on clean examples or not.

\subsubsection*{Thermometer One Hot Encoding}
\label{subsubsec:thermometer}

Thermometer one hot encoding~\cite{buckman2018thermometer} breaks the linear extrapolation behaviour of \ac{DNN} by pre-processing the input with an extremely non-linear function.
However, instead of replacing a real number with its counterpart transformation, the authors replace each real number with a binary vector.
Multiplying the input vector with the network's weight enables different input values to use different parameters of the network.

The authors use pixel-wise one-hot encodings and thermometer encodings for discretising the inputs.
Experimental results show that thermometer encoding in combination with adversarial training is a powerful defence capable to withstand \ac{pgd} attacks in both white-box and black-box settings.
\subsubsection*{Blind Pre-Processing}
\label{subsubsec:blind_preprocessing}

Inspired by thermometer encoding~\cite{buckman2018thermometer}, Rakin \etal~\cite{rakin2018blind} propose to process the input data using an ensemble of methods which includes $tanh$ function, batch normalisation, thermometer encoding and one hot encoding.
The authors slightly change the attack model, giving the attacker access to all hyper-parameters except for input transformations.
The experiments, however, are only carried against the \ac{fgsm} and Carlini attacks, while thermometer encoding is tested against \ac{pgd}.
\subsubsection*{Data Specific Discretisation}
\label{subsubsec:data_discretisation}

Chen \etal~\cite{chen2018improving} propose a pre-processing technique that can successfully mask the gradients even for adaptive and iterative attackers.
Their proposal is based on encoding the whole input space using a small set of separable codewords and training a classifier on the encoded information.
The only requirement for finding the codewords is to have large pairwise distances under an appropriate metric.

In practice, the authors use density estimators in order to construct separable codes.
Experimental results demonstrate an increase in performance when tested on large data sets and against adaptive attacks.
However, the technique works only for small perturbations.
\subsubsection*{Adaptive Noise Reduction}
\label{subsubsec:adaptive_noise_detection}

Lian \etal~\cite{liang2017detecting} treat perturbations as noise and leverage noise reduction techniques to reduce their adversarial effects.
If the effect is downgraded properly, the de-noised adversarial example will be classified as a new class, distinct from the adversarial target.
The authors use two image processing techniques: scalar quantisation and smoothing spatial filter.
In order to improve the generality of this method, an adaptive noise reduction is used in the following way: for images with low entropy, an aggressive noise reduction strategy is adopted, while for images with high entropy light strategies are employed.
Upon receiving a sample to a classifier, the algorithm will first apply adaptive noise reduction in order to create a clean sample.
Later, both samples are sent to the original classifier and, if they get different labels, the initial input is treated as an adversarial example.

The experimental results show good performance of detecting \ac{fgsm} attacks.
However, a big requirement of this attack is that adversarial examples are generated with very small perturbations.
Thus, this approach is limited to use-cases where robustness against small requirements is needed.
Moreover, more testing against iterative attacks is needed.
	\subsection{Proactive Defences}
\label{subsec:proactive_defences}

This class of defences alter the model or its learning procedure in order to increase \ac{ML} robustness.
As opposed to reactive defences, where some degree of robustness is achieved  earlier in the processing pipeline, proactive defences aim to design better architectures or training procedures.
Thus, they eliminate any pre-processing steps.

Within this class of defences we distinguish between defences that use adversarial examples in the training set with (Section~\ref{subsec:adv_training}) or without normal inputs (Section~\ref{subsec:minmax}).
Moreover, we distinguish between defences that re-design the network's architecture (Section~\ref{subsec:defences_arch}) or just increase their non-linearity and capacity (Section~\ref{subsec:hyperparam}).
Lastly, we introduce a class of defences that aims to certify a degree of robustness (Section~\ref{subsubsec:provable}) and a class of defences that use generative models (Section~\ref{subsec:defences_gans}).

\subsubsection{Adversarial Training}
\label{subsec:adv_training}

Training with adversarial examples is a form of regularisation~\cite{girosi1995regularization} \ie~a strategy designed to reduce the test error, possibly affecting the training error. 
A common way to make a \ac{ML} algorithm better generalise on a task is to train it with more data. 
In practice, however, the amount of available data is limited. 
In order to overcome this barrier, one can create fake data (by applying noise, translations, rotations, \etc) and \emph{augment} the training set with the new examples.

Unlike common data augmentation schemes, generating adversarial examples is different because these inputs are not expected to occur naturally.
Adversarial examples expose flaws in the ways a model conceptualises its decision functions~\cite{goodfellow2014explaining} and adversarial training helps in overcoming them.

\subsubsection*{Adversarial Training with \emph{FGSM}}
\label{subsubsec:fgsm_training}

Many regularisation schemes limit the capacity of a model by adding a penalty term to the cost function. 
Goodfellow \etal~found that training with an adversarial error function based on the fast gradient sign method (Section~\ref{subsubsec:fgsm}) was an effective regulariser~\cite{goodfellow2014explaining}.
Formally, the adversarial error function is defined as:
\begin{equation}
	 \tilde{J}(\vtheta, \vx, y) = \alpha J(\vtheta, \vx, y) + (1-\alpha) J(\vtheta, \vx + \eps \sign \left( \nabla_\vx J(\vtheta, \vx, y) \right),	
\end{equation}
\noindent
where $\alpha$ is a hyper-parameter that weights the relative contribution of the adversarial penalty.
In practice, Goodfellow \etal~used a value equal to $\alpha = 0.5$.


An important take-away from experimenting with \ac{fgsm} adversarial training is that it is generally better to perturb the input, rather than a hidden layer of a neural network.
When applied to hidden units whose activations are unbounded, \ac{DNN} respond by making their hidden unit activations very large.

The results show that adversarial training is able to significantly improve the robustness of a model.
Moreover, adversarial training helps to better generalise and improve the original accuracy of a model.

\subsubsection*{Gradient Adversarial Training}
\label{subsubsec:gradient_adversarial}

Sinha \etal~\cite{sinha2018gradient} propose a new training framework which assumes that simultaneous gradient updates should be statistically indistinguishable from each other.
Thus the gradient can be regularised in order to remove salient information that can lead to adversarial examples.
Further on, the gradient tensor is processed in an auxiliary network and then passed to the main network via a gradient reversal procedure.
This training procedure adapts the cross-entropy loss function and adds weight to negative classes whose gradient tensors are similar to those of the primary class.
The weight weight is evaluated using the auxiliary network which indicates the gradient tensors similarities between the primary and negative classes.

Experimental results show better regularisation during training and an increase in robustness when tested against gradient adversarial attacks.
However, robustness is not always preserved when testing is done against iterative gradient methods (Section~\ref{subsubsec:illcm}).

\subsubsection*{Training with Gradient Regularisation}
\label{subsec:gradient_regularisation}

Lyu, Huang and Liang~\cite{lyu2015unified} propose a family of gradient based perturbations that can be used as a regularisation technique.
Their work can be seen as a generalisation of the \ac{fgsm} for all possible norms.
The regularisation follows from the Taylor expansion of the min-max problem (maximising the perturbation while minimising the loss - Section~\ref{subsec:minmax}) and yields:

 \begin{equation}
     \min_{\vtheta} J(\vtheta, \vx, y) + \epsilon \| \nabla_{x} J(\vtheta, \vx, y) \|_p.
 \end{equation}
\noindent
In the case of $p_{\infty}$ the perturbation is equivalent to \ac{fgsm} - Section~\ref{subsubsec:fgsm}.
However, the authors propose the use of $p_{2}$, which resembles marginalised Gaussian noise and propose an heuristic way to optimise the objective.
Using this regularisation technique increases performance against single step attacks, on small data sets.
However, the impact of iterative attacks is not empirically evaluated.


\subsubsection*{Training with Structured Gradient Regularisation}
\label{subsubsec:structured_regularisation}

Roth \etal~\cite{roth2018adversarially} propose to use a generative model (Section~\ref{subsec:gan_attacks}) in order to learn the distribution of adversarial corruptions from examples and use it as for regularisation.
The authors propose to define an objective function that includes a mixing distribution between un-corrupted and corrupted data.
Formally, the loss function can be defined as:
\begin{equation}
    L(\vtheta) = (1-\lambda) \mathbb{E}_p{\hat{P}} [L(\vtheta, \vx, y)] + \lambda \mathbb{E}_p{\hat{P}_Q} [L(\vtheta, \vx, y)],
\end{equation}
\noindent
where $\hat{P}$ denotes the empirical distribution of the input and $\hat{P}_Q$ the noise corrupted distribution.



\subsubsection{Learning in a Min-Max Setting}
\label{subsec:minmax}
Another way of training with adversarial examples is to exclude the clean examples from the training procedure.
This is equivalent to training on the 'worst' inputs possible, generated by adversarial attack methods.
The procedure draws from the field of \emph{robust optimisation}~\cite{ben2009robust}; an area of optimisation theory that aims to obtain solutions stable under some level of uncertainty.

Robust optimisation problems have a min-max formulation in which the objective function is being minimised with respect to maximum worst-case perturbations.
The assumption is that the perturbations can be drawn from specific sets called \emph{uncertainty sets} $\mathcal{U}$.
There is a number of cases that one can consider for the uncertainty sets.
One example is the $\mathcal{U}_i = B_p(\vx_i, r)$ norm ball centred at $\vx_i$, with radius $r$ and norm $p$.
Searching for a maximum perturbation, however, increases the training time considerably and might not always be a tractable problem.

In this section we introduce several attempts to use robust optimisation for training \ac{DNN}.
This procedure is also called \emph{strong} learning because we aim to find the worst adversarial example for each input in the training data set; as opposed to \emph{weak} learning which searches for an adversarial example within some bounds.

\subsubsection*{Robust Adversarial Training}
\label{subsubsec_understanding_adv}

Shaham, Yamada and Negahban~\cite{shaham2015understanding} propose a first training framework based on robust optimisation.
Formally, the minimisation-maximisation approach is stated as:
\begin{equation}
\label{eq:adv_robust_training}
\min_\vtheta \tilde{J}(\vtheta,\vx,y) = \min_\vtheta \sum_{i=1}^m \max_{\tilde{\vx}_i \in \mathcal{U}_i} J(\vtheta, \tilde{\vx}_i,y_i),
\end{equation}
\noindent
where $\mathcal{U}_i$ is the uncertainty set corresponding to input $\vx_i$.
The inner maximisation problem seems intractable for the authors (later, Madry \etal~\cite{madry2017towards} show this is in fact tractable).
Therefore, they propose to minimise a surrogate of \eqq~(\ref{eq:adv_robust_training}) in which each sub-procedure is reduced to a single ascent/descent step.
The formal expression for finding the worst-case perturbation for an input $\vx_i$ is based on the first-order Taylor expansion of the loss around the example, which yields:
\begin{equation}
  \label{eq:adv_robust_max}
  \eta_{i} = \arg \max_{\eta: \vx_i + \eta \in \mathcal{U}_i} J(\vtheta, \vx_i + \eta, y_i).
\end{equation}
\noindent
The outer minimisation problem is solved using a single descent step with respect to the perturbed data $\vx_i + \eta_i$.
The experimental results show that robust optimisation works best when using the $L_\infty$ norm and helps improving robustness of \ac{DNN} against adversarial examples on both MNIST an CIFAR-10 data sets.

\subsubsection*{Learning with a Strong Adversary}
\label{subsubsec:learning_strong_adversary}

Huang \etal~\cite{huang2015learning} also formulate the learning procedure as a min-max game.
However, the maximisation problem is solved using the approaches described in Section~\ref{subsubsec:strong_adversary} for upper-bounded uncertainty sets.
Formally, the objective function is stated as:
\begin{equation}
\label{eq:LWA}
	\min_\vtheta \sum_i \max_{\|\eta_{i}\|\leq \epsilon} J(\vtheta,  x_i + \eta_i, y_i).
\end{equation}
\noindent
Another interesting approach proposed in~\cite{huang2015learning} is to perturb only the representation learned by a neural network.
One can regard \ac{DNN} as consisting of two parts: (1) lower layers that can learn a representation of the input data (e.g. convolution layers) and (2) higher layers that learn a classification model on top of the representation (fully connected and softmax layers).
Instead of perturbing the raw input data, the authors propose to use the representation learned by the lower layers in order to train a robust neural network.
Formally, the proposed objective function is:
\begin{equation}
\label{eq:LWA2}
\min_{f{rep}, f{ cla}} \sum_i \max_{\|\eta_{i}\|\leq \epsilon} J \left(f_{cla}\left(f_{ rep}(\vx_i) + \eta_{i}\right), y_i\right),
\end{equation}
\noindent
where $f_{rep}$ are the lower layers in a neural network and $f_{cla}$ are the higher ones.
The perturbation is computed using one of the methods described in Section~\ref{subsubsec:strong_adversary} and
the experimental results show that training using the first method from Section~\ref{subsubsec:strong_adversary} achieves better results than adversarial training with \ac{fgsm} - Section~\ref{subsubsec:fgsm_training}.
\subsubsection*{Learning with a Complete First Order Adversary}
\label{subsubsec:madry_learning}

Madry \etal~\cite{madry2017towards} solve the inner maximisation problem using projected gradient descent (pgd).
The initial results presented in Section~\ref{subsubsec:madry} suggest that training a network with \ac{cfoa} will assure robustness against all first order methods.
Formally, the training procedure is defined as a min-max game:
\begin{equation}
	\label{eq:mandry}
	\min_{\vtheta} \rho(\vtheta), \quad \text{where} \quad \rho(\vtheta) = \mathbb{E}_{(\vx, y) \sim p_{\text{data}}} \left[\max_{\eta \in \mathcal{S}} J(\vtheta, \vx + \eta, y) \right],
\end{equation}
\noindent
The inner maximisation problem is tackled through projected gradient descent starting from a random perturbation around the natural example while the minimisation problem is tackled using stochastic gradient descent.
The results show state of the art performance at the moment of writing this paper.


\subsubsection*{\ac{eat}}
\label{subsubsec:ensemble}

\ac{eat}~\cite{tramer2017ensemble} is a technique that augments training data with adversarial examples crafted on other static, pre-trained, models.
The work is inspired by domain adaptation\footnote{In domain adaptation a model is trained on one or more input distributions and is evaluated on samples from a different, but related, distribution.}~\cite{zhang2012generalization} and tries to decouple the adversarial generation process from the parameters of a trained model.

Training with ensemble methods increases the diversity of perturbations seen during training.
This is equivalent to enhancing the uncertainty sets, $\mathcal{U}$, with examples crafted on other models.
Because adversarial examples are transferable between models, the intuition is that adversarial examples crafted for other models should help in cross-training robust models too.

Ensemble training uses the same procedure as learning with a complete first order adversary \eqq~(\ref{eq:mandry}), but applies a variety of attacks on pre-trained models in order to approximate the inner maximisation problem (e.g. \ac{fgsm}, \ac{rssa}). 
In a white-box setting, models trained through \ac{eat} are slightly less accurate when compared with standard adversarial training. 
However, \ac{eat} significantly boosts robustness to black-box attacks transferred from pre-trained models. 
This seems to be the expected behaviour, because the transfer models are similar to the tested ones (e.g. inceptionV3 vs InceptionV4 or IncRes v2 vs ResNet v2)


\subsubsection*{\ac{sap}}
\label{subsubsec:stochastic_pruning}

Dhillon \etal~\cite{dhillon2018stochastic} define a minimax zero-sum game between an adversary and \ac{DNN}.
The strategy for this game is to prune a random subset of activations (\eg~those with smaller magnitude) and scale-up the survivors in order to compensate.
This approach is similar to the dropout technique, where the activations with high absolute values have a higher chance of being sampled.

\ac{sap} can be applied to pre-trained networks and presents small improvements in accuracy when combined with adversarial training.


\subsubsection{Robust Deep Architectures}
\label{subsec:defences_arch}

This section introduces publications that alter the training method or the network's architecture in order to increase robustness to adversarial examples.
Some works draw inspiration from knowledge transfer, while others take inspiration from regularisations specific to de-noising auto-encoders - a technique used to recover original inputs from recover partially corrupted data.

\subsubsection*{Distillation Defence}
\label{subsubsection:distillation}

Distillation is a training procedure initially designed to train a \ac{DNN} using knowledge transferred from a different \ac{DNN} \cite{hinton2015distilling}.
The motivation behind distillation is to reduce the computational complexity of \ac{DNN} architectures by transferring knowledge from larger architectures to smaller ones.
A distilled network can be trained with \emph{soft} labels \ie~a probability distribution over all classes given by a softmax function, instead of \emph{hard} labels \ie~a discrete value.
The soft labels are the output of a large \ac{DNN} and used as knowledge transfer for smaller networks.
An important parameter is the \emph{temperature} of the softmax function because a higher temperature produces a smoother probability distribution over classes, thus influencing the knowledge transfer.

Papernot \etal~\cite{papernot2016distillation} define a new distillation mechanism that provides defence training \ie~instead of transferring knowledge between architectures, the knowledge extracted from a \ac{DNN} is used to improve its own resilience to adversarial examples.
This means that, as opposed to the original distillation mechanism, the same network architecture is used for training and distillation.

The knowledge extracted from defensive distillation is used to reduce the amplitude of network gradients.
In general, if the gradients are high, crafting adversarial examples becomes easier because small perturbations will induce high output variations.
The authors show that using distillation with a high temperature reduces the model's sensitivity to small variations of its inputs and increases the overall robustness.
However, Carlini \etal~\cite{carlini2017towards} introduced a series of attacks that break defence distillation. They are discussed in Section~\ref{subsubsec:carlini}.



\subsubsection*{Parseval Networks}
\label{subsubsection:parseval}

Parseval networks~\cite{cisse2017parseval} are a regularisation scheme that works by constraining the Lipschitz constant of each hidden layer of a \ac{DNN} to be smaller than a threshold.
Through this constraint any exponential growth of the Lipschitz constant is avoided.
In this setting a regularisation scheme (such as weight decay) at the last layer controls the overall Lipschitz constant of the network.

Szegedy \etal~\cite{szegedy2013intriguing} first suggested the idea that robustness of  \ac{DNN} can be proved inspecting the Lipschitz constant at each layer.
As mentioned in Section~\ref{sec:robustness_eval}, Weng \etal~\cite{weng2018evaluating} showed this bound might be loose.
In contrast, the experimental results in ~\cite{cisse2017parseval} show that a network's sensitivity to adversarial examples can be controlled using the Lipschitz constant.

\subsubsection*{Deep Contractive Networks}
\label{subsubsec:deep_contractive}


Gu and Rigazio~\cite{gu2014towards} propose a new network architecture, called deep contractive networks, which imposes a layer-wise penalty in a neural network.
This penalty minimises the network output variance \wrt~perturbations in the input s.t. a trained model can achieve robustness for perturbations around training data points.

The authors use contractive auto-encoders - a variation of auto-encoders with an additional penalty for minimising the squared norm of the Jacobian of the hidden representation with respect to input data~\cite{rifai2011higher}.
When applied to feed-forward networks, this penalty helps to explicitly learn invariant features at each layer.
Formally, the network optimises the following loss function:
\begin{equation}
	J_{DCN}(\vtheta)  = \sum_{i=1}^{m}  (L( \vx, \vtheta, y) + 
	\sum_{j=1}^{h+1} \lambda_j \| \frac{\partial y^{(i)}}{\partial(\vx)^{(i)}} \|_2 ),	
\end{equation}
\noindent
where h is a hidden layer and $\lambda$ is a scaling factor.

Experimental results show deep contractive networks help increase the average perturbation size needed to build an adversarial example.
However, the architecture trades accuracy on clean examples for these results.

\subsubsection*{Biologically Inspired Robust Neural Networks}
\label{subsubsec:biologically}

%

Nayebi and Ganguli~\cite{nayebi2017biologically} develop a new training scheme inspired by biophysical principles in neural circuits.
Following the idea suggested by Goodfellow \etal~\cite{goodfellow2014explaining} that adversarial examples are due to the linear summing of high dimensional input with small weights, the authors propose to force neural networks to operate in a non-linear, saturated, regime.

In order to enforce this constraint, the authors \emph{saturate} the network \ie~ensure that each element of the Jacobian matrix of the model is sufficiently small s.t. the model becomes insensitive to perturbations.
This approach is similar to deep contractive networks - Section~\ref{subsubsec:deep_contractive}.
However, the contractive penalty degrades test accuracy on clean examples and is difficult to compute for large networks.

In order to overcome these limitations, the authors propose the use of penalties specific to saturating auto-encoders~\cite{goroshin2013saturating} in order to explicitly encourage activations to be in the saturating regime of the nonlinearity.
Formally, for a given activation $h = W\vx + b$ and $\lambda \in \sR$, the saturation penalty is:
\begin{equation}
	\lambda \sum_{i=1}^{h} \phi_c(h_i),
\end{equation}
\noindent
where the complementary function is defined as:
\begin{equation}
	\phi_c{\vz} = \inf_{\vz' \in \sS} \| \vz - \vz' \|, \sS = \{\vz | \phi'(\vz) = 0\},
\end{equation}
\noindent
and reflects the distance of any individual activation to be the nearest saturation region.
\noindent
The experimental results show improvements on both first-order methods and adaptive, iterative, attacks.

\subsubsection*{DeepCloak Defence}
\label{subsubsec:deep_cloak}

DeepCloak Defence~\cite{gao2017deepcloak} propose to remove features not use in classification in order to increase robustness to adversarial examples.
To identify unnecessary features a pairs of adversarial samples are tested against the clean example.
In order to remove the features, a mask layer is introduced before the logit layer. 
The mask serves as a selector, keeping the necessary features and setting the unnecessary to 0.

DeepCloak is model independent and easy to implement.
However, the robustness increase is not substantial.
And, as in most architectures, the defence trades accuracy from the original model.

\subsubsection*{Fortified Networks}
\label{subsubsec:fortified}

Lamb \etal~\cite{lamb2018fortified} identify which hidden states are off the data manifold and map these states back to parts of the data manifold where the network performs better.
The fortification consists of inserting de-noising auto-encoders at crucial points between layers of the original network in order to clean up the transformed data points which may lie outside the data manifold.

The experimental results show good results even against the \ac{phd} attack (Section~\ref{subsubsec:madry}).
Similar techniques have been suggested in~\cite{xi2018manifold}.
However, both methods prove inefficient when faced with adaptive attacks~\cite{carlini2017magnet}.
\subsubsection*{Rotation-Equivariant Networks}
\label{subsubsec:rotation_equivariant}

Dumont, Maggio and Montalvo~\cite{dumont2018robustness} investigate the resistance to adversarial attacks of three rotation-equivariant network architectures~\cite{cohen2016group, worrall2017harmonic, zhou2017oriented}.
They discover that rotation equivariant networks are significantly more robust to attacks based on small translations and rotations, and marginally more robust to attacks based on local geometric distortions.
\subsubsection*{HyperNetworks}
\label{subsubsec:hypernetworks}

Sun \etal~\cite{sun2017hypernetworks} propose to use data dependent weights in the hidden layers of a neural network. 
The main idea is to adaptively filter convolution weights using HyperNetworks~\cite{ha2016hypernetworks} - a technqiue that uses a network to generate the weights for another network.

The experimental results show an increase in robustness, without trading off accuracy of the initial model. 
However, although HyperNetworks proved to be powerful, this architecture is not used in large scale vision tasks due to the high dimension of weights in the recent state-of-the-art convolutional neural network architecture.
\subsubsection*{Bidirectional Learning}
\label{subsubsec:bidirectional_learning}

The authors of \cite{pontes2018bidirectional} propose to use bidirectional learning in order to increase robustness to adversarial examples.
Bidirectional learning trains two models - the first on the input and the second on a reversed copy of the input.
However, in \cite{pontes2018bidirectional}, a single model is trained to behave both as a discriminative and generative model.
Therefore, the same model can be a classifier and a generator in the same time.
This behaviour is achieved using an undirected neural network that back-propagates the errors in both directions - each direction of the network has its own biases and the weights are shared.

The positive weights of the last layer of a generator become the first layers of a classifier.
The authors also introduce a hybrid method in which two models are trained (instead of an undirected network) and the weights are shared.
The experimental results show an increase in robustness, due to more robust weights.
Moreover, adversarial examples are harder to generate for bi-directional networks.
However, the results fail to scale to larger data sets.

\subsubsection*{\ac{dam} Models}
\label{subsubsec:dense_memory}

\ac{dam} models~\cite{krotov2016dense} store a set of memory vectors, corresponding to the learnt patterns.
At query time, the network is presented with an incomplete pattern resembling, but not identical to, one of the stored memories and the task is to recover the full memory.
For example, pixel intensities can be combined with the label of the image into one vector, which will serve as a memory for the associative memory.
One limitation of using associative memory models is the limited memory; the standard model of associative memory works fine in the limit when then umber of stored patterns is much smaller than the number of neurones, or equivalent the number of pixels in an image~\cite{krotov2016dense}.

Krotov and Hopfield suggest that \ac{dam} with higher order energy functions are closer to human visual perception that \ac{DNN} with RELUs.
Thus, \ac{dam} models are more robust to adversarial examples.
However, the statements are not empirically supported.

\subsubsection{Hyper-parameter Tuning}
\label{subsec:hyperparam}
\subsubsection*{Non-linearity}
\label{subsec:non_linearity}

Given the linear conjecture, introduced in Section~\ref{sec:causes}, which states that adversarial examples exploit the linear behaviour of \ac{DNN}, a normal attempt to increase robustness is to use more non-linear activation functions. 

Goodfellow \etal~\cite{goodfellow2014explaining} suggest that RBF networks - which behave in a non-linear fashion - are naturally immune to adversarial examples and have low confidence when they are fooled.
However, RBF units can not generalise very well and do not achieve the same performance as ReLU \ac{DNN}.
Krotov and Hopfield~\cite{krotov2016dense} tried to replace ReLU activations with higher rectified polynomials and observed an increase in robustness.
However, the increase is not sufficient to completely remove the adversarial phenomenon.
Moreover, the training procedure is slower.
Some architectural designs also exploit non-linearities and are presented in~\cite{nayebi2017biologically, krotov2016dense} - Section~\ref{subsubsec:biologically} and~\ref{subsubsec:dense_memory}.




\subsubsection*{Increased Capacity}
\label{subsec:capacity}

Several publications suggested that a high capacity of \ac{DNN} increases their robustness to adversarial examples\cite{madry2017towards, kurakin2016aadversarial, rozsa2016accuracy}.
However, no publication investigated the correlation in isolated experiments.
Increasing the capacity comes with additional costs in terms of processing power and training time.
It is interesting to follow if capacity and robustness are, indeed, correlated and which are the trade-offs between the two.
\subsubsection{Provable defences}
\label{subsubsec:provable}

In this section we introduce works that \emph{guarantee} a threshold for the size of a perturbation.
Following the discussion in Section~\ref{sec:robustness_eval}, the thresholds are chosen to guarantee either lower-bound or upper-bound robustness to adversarial examples.



\subsubsection*{Certified Defenses}
\label{subsubsec:certified}

The authors of~\cite{raghunathan2018certified} focus on computing and upper bound on the worst-case loss of linear classifiers and \ac{DNN} with two hidden layer.
This upper bound serves as a \emph{certificate of robustness} against all attacks for a given network and input.

They use integration to obtain an exact expression for upper bound loss using all gradients in a p-ball around an example and use a semidefinite relaxation in order to convert this to a convex, tractable, optimisation problem.
The loss is later used in training a model.

There are, however, some disadvantages with this method. 
At first, it is limited to linear models, therefore, has limited impact in the image recognition task.
Secondly, increasing the number of layers (from 2) also increases the complexity of the method and might result in intractable optimisation problems.


\subsubsection*{Formal Tools for DNN Robustness}
\label{subsubsec:reluplex}

Katz \etal~\cite{katz2017reluplex} propose the use of SMT solvers to prove upper bound robustness of \ac{DNN}.
This procedure encodes the neural network and the constraints regarding upper bound robustness (max $\epsilon$) as a set of linear equations.
The authors extend the simplex algorithm to handle ReLU activation functions - therefore the name Reluplex.
Reluplex uses SAT solving techniques to check if an adversarial example exists in the given constraint space.
If the constraint is not satisfied, the algorithm returns a counter-example which constitutes a valid adversarial example.
Some recent work extends Reluplex to \cite{carlini2018provably} to verify $L_1$ and $L_\infty$ norm, by encoding absolute values using ReLUs.

On the same path, Huang \etal~\cite{huang2017safety} use model checkers in order to guarantee robustness against \emph{known} adversarial perturbations,
Ehlers~\cite{ehlers2017formal} uses SMT solvers for neural networks with linear activation

An inherent disadvantage of using SMT solvers is processing time. 
However, as opposed to other provable defences, this approach is not constrained by network capacity.

In \cite{ruan2018reachability}, the authors design  and  implement  a  reachability  analysis tool  for  deep  neural  networks,  which  has  provable  guarantees and can be applied to \ac{DNN} with deep layers and nonlinear activation functions.  
Their work has the ability to work with larger networks lower  computational  complexity,  i.e.,  NP- completeness  with  respect  to  the  input  dimensions  to  be changed, instead of the number of hidden neurons.
\subsubsection*{Certifying Some Distributional Robustness}
\label{subsubsec:certified_distributional}

The authors of~\cite{sinha2018certifying} provide a training procedure that augments parameter update with worst-case perturbations of training data - similar to min-max games presented in Section~\ref{subsec:minmax}.
However, the uncertainty sets are chosen to be distributions at close Wasserstein-distance from the training distribution.

The main idea behind providing certified distributional robustness is to optimise a loss surrogate that allows adversarial perturbations within a certain range.
The adversarial range is regulated by controlling the distribution drift, through the Wasserstein-distance.
The experimental results show this method provides better robustness than most networks trained with adversarial examples (Section~\ref{subsec:adv_training}) and even for networks trained with \ac{pgd} (Section~\ref{subsec:minmax}).
\subsubsection*{Convex Outer Adversarial Polytope}
\label{subsubsec:convex_outer}

Kolter and Wong~\cite{kolter2017provable} present a method for training provably robust deep ReLU classifiers.
The approach provides robustness against any norm-bounded adversarial perturbations on the training set.
While no previously unseen example can bypass this defence, it might sometimes flag non-adversarial examples as adversarial.
The main idea is to construct an outer bound on the set of all final-layer activations that can be achieved by applying a norm-bounded perturbation to an input.
If one can guarantee that the class prediction of an example does not change within this outer bound, it is equivalent to a proof that the example can not be adversarial.

The algorithm shows that the feasible set of the dual problem can be expressed as a neural network.
Therefore, because finding a feasible dual solution provides a guaranteed lower bound on the solution of the primal, using a single backward pass through the second network can prove the lower bound for the primal network under analysis.
Some extensions that help with scalability are presented in~\cite{wong2018scaling}.

\subsubsection*{Lower bounds on the robustness to adversarial perturbations}
\label{subsubsec:lower_bounds}

Peck \etal~\cite{peck2017lower} derive lower-bounds on the magnitude of perturbations necessary to change the classification for various \ac{DNN} architectures.
The lower-bounds are expressed directly in terms of the neural network's parameters.
Although not used as a defence, this methodology allows comparisons between different classifiers, therefore, helping to evaluate defences.

\subsubsection*{Lipschitz-Margin Training}
\label{subsubsec:lipschitz_margin_training}
In~\cite{tsuzuku2018lipschitz}, the authors propose a method to enlarge the provable robust norm-ball during training.
At first, the robust norm ball is defined in terms of Lipschitz constant and the gap between the correct label and other labels, for each input.
This concept, called \emph{margin}~\cite{bartlett2017spectrally} is converted into a loss function in order to ensure non-trivial norm-balls during training. 
The main idea behind this process is to maximises the number of training data points that have guarded areas larger than a hyper-parameter, as long as the original training procedure maximises the number of inputs that are correctly classified.

The authors also introduce a fast method to calculate the upper bounds of the Lipschitz constant, making Lipschitz-margin training computational feasible.
Experimental results show an increase in the provable norm-ball for large networks and improved robustness against adaptive attacks.

\subsubsection{Defences based on \ac{gan}}
\label{subsec:defences_gans}

In this section we present defences based on generative models, first introduced in Section~\ref{subsec:gan_attacks}.

\subsubsection*{Defence-GAN}
\label{subsubsec:defence_gan}

Defence-GAN~\cite{samangouei2018defense} is trained to model the distribution of unperturbed images.
At inference time, it finds a close output to a given image, which does not contain the adversarial change.
This output is then fed to the classifier.
Since it does not alter the classifier's structure or training procedure, it can be used with any classifier architecture.
The experimental results, however, suggests Defence-GAN trades a lot of accuracy for robustness.
\subsubsection*{\ac{fbgan}}
\label{subsubsec:bidirectional_gan}

\ac{fbgan}~\cite{bao2018featurized} first captures the semantic information of any input, original or adversarial, and then retrieves the unperturbed input from this information.
This process is meant to remove the perturbation.
The authors use a bi-directional \ac{gan}~\cite{donahue2016adversarial, dumoulin2016adversarially} to learn the feature mapping and add mutual information to the latent space in order to reduce the dimension of latent codes and ensure they capture the semantical variation.

The defence shows good accuracy, however, it is only tested on small data sets.
It is unclear if the method will scale to large data sets and models with high capacity.

	\subsection{On the Overall Efficacy of Adversarial Defenses}
\label{subsec:overall_defences}

One limitation of most publications presented in the previous sections is the lack of rigorous evaluation.
With the exception of defenses that guarantee robustness within some bounds, other publications do not provide sufficient evidence for their effectiveness. 
Therefore, the efficacy of these defenses is often over-emphasized.
\if 0\mode  We illustrate this behavior in Appendix~\ref{sec:benchmarks}. \else\fi
Fortunately, some publications questioned their success and showed some defenses are, in fact, not effective.

\citeauthor{carlini2017adversarial}~\cite{carlini2017adversarial} evaluated the efficacy of adversarial detection methods and proved that most detectors perform poorly when faced with strong, iterative, attacks.
In particular, the authors use the Carlini and Wagner attack~\cite{carlini2017towards} to show that none of the detectors~--~\cite{bhagoji2018enhancing, feinman2017detecting, gong2017adversarial, grosse2017statistical, hendrycks2016early, metzen2017detecting, li2017adversarial}~--~efficiently detect adversarial examples and that their reported results do not reflect the reality.
Moreover, \citeauthor{athalye2017synthesizing}~\cite{athalye2017synthesizing} broke the \cite{lu2017standard} detector.
In another paper, \citeauthor{carlini2017magnet}~\cite{carlini2017magnet} show that MagNet~\cite{meng2017magnet} is also easily defeated by adaptive attacks.

Conversely, \citeauthor{athalye2018obfuscated}~\cite{athalye2018obfuscated} showed that 7 out of 9 defenses published at ICLR 2018 as non-certified defenses suffer from gradient masking (a phenomenon described in the next section) and are not efficient against the BPDA attack (Table~\ref{tbl:attacks}).
Although the other 2 defenses do not suffer from gradient masking, they have also been broken.
\citeauthor{he2017adversarial}~\cite{he2017adversarial} show that combining adversarial defenses in an ensemble defense is also not effective, while \citeauthor{sharma2017breaking}~\cite{sharma2017breaking} broke the adversary training method based on the Madry~attack~\cite{madry2017towards}.
Moreover, certified defenses can be bypassed using generative models~\cite{poursaeed2018generative} or with perturbations outside the pixel norm ball~\cite{liu2018beyond}.
\citeauthor{zhang2019limitations}~\cite{zhang2019limitations} showed that robust adversarial training and certified defenses only protect in regions close to the training data, but fail to protect against inputs away from these regions.

Given these results, the overall efficacy of all adversarial defenses is under scrutiny. A protocol for defense evaluation is presented in the following section.

\else 

	Similar to the attack evaluation introduced in Section~\ref{sec:attacks}, we  provide a qualitative and a quantitative evaluation of each defense.
The qualitative evaluation measures the performance of one defense against various attacks, its complexity and the testing setup, while the quantitative evaluation measures the literature impact.
Each dimensions of both the qualitative and the quantitative assessment is measured on a categorical scale, ranging from low (*), medium(**) and high(***).
The final score is computed by averaging over the attributes described below.

\begin{itemize}
	\item \emph{Defense strength.} The defense strength evaluates how powerful a defense is against different attacks (** or ***, depending on the attack's strength) or if a defense was broken (*).
	\item \emph{Defense complexity.} The defense complexity evaluates how easy it is to apply one defense (* if the defense is relatively easy to deploy, up to *** if the defense requires changing the model completely).
	\item \emph{Experimental setup.} The experimental setup measures how thoroughly the defense was tested; if a threat model is presented (* if no, ** if the threat model is incomplete and *** if a correct threat model is presented), against which attacks is the defense tested (*, ** or ***, depending on the attack strength), and which data sets were used during evaluation (same as for attacks).
\end{itemize}
\noindent
The quantitative evaluation is the same as for attacks: the ratio between the number of citations and the number of months since publication, as indicated by Google Scholar.
Although bibliometrics are not a direct indicator of quality, they can be a good proxy to influential papers for a novice reader.

The results are presented together with the defense strategies discussed in Section~\ref{sec:taxonomy_defense} in Table~\ref{tbl:defences_acm}.
We note that the strongest defenses are the certified ones, although they also involve higher complexity.
Whenever the perturbation space around a highly dimensional input can be well estimated, training with adversarial examples also leads  to strong defenses (as in the case of Madry).
Unfortunately the least complex defenses~--~\ie~guards~--~are also the least powerful.
More details follow in the next sections.
We conclude with a discussion the overall efficacy of the defenses (Section~\ref{subsec:overall_defences}) and with a note on the evaluation of future defenses (Section~\ref{subsec:gradient_ob}).

\subsection{Guards}
\label{subsec:reactive}

Guards add a pre-processing step to the classification pipeline, in which adversarial examples are either detected or their impact is diminished by transforming the input.
Because they do not alter the target model, guards have great potential.

Adversarial detectors are suited for tasks where refusing to process an input or discarding a classification is not mission critical.
For example, an object detector from a cloud storage provider can refuse to classify one input, but an autonomous vehicle might not be able do so (since the input can be an important traffic sign).
Moreover, based on the constructs used, adversarial detectors can also suffer from lack of robustness.
Input transformations show great potential because they do not require any training and are easy to deploy.
However, they can suffer from the same disadvantage as adversarial detectors because transforming an input might introduce other side effects.
For example, compressing a high resolution medical image can lead to loss of accuracy.

\subsubsection{Detection of Adversarial Examples.}
\label{subsubsec:detection}
Adversarial detectors use a variety of distinct features, as follows.
\citeauthor{grosse2017statistical}~\cite{grosse2017statistical} used a model agnostic statistical test to check if adversarial examples are outside the training data distribution.
They observe that adversarial examples generated with some attacks can be found in different regions of the output surface than normal inputs and can be detected using statistical testing.
\citeauthor{gong2017adversarial}~\cite{gong2017adversarial} trained a separate \ac{DNN} only with adversarial examples, able to successful detect adversarial examples with very small perturbations.
\citeauthor{metzen2017detecting}~\cite{metzen2017detecting} also trained a \ac{DNN} for adversarial examples but, this time, the authors use the output of the \ac{DNN}'s hidden layers.
This techniques shows good result against weak attacks, but the detector's performance diminishes agains iterative attacks.
Similarly, \citeauthor{li2017adversarial}\cite{li2017adversarial} developed an adversarial detector based on features extracted at every layer of a convolutional neural network.
The authors treat an image as a distribution of pixels that can be used to collect statistics and used statistics from the hidden layers to train an adversarial detector.

\citeauthor{feinman2017detecting}~\cite{feinman2017detecting} trained a linear  detector using two distinct features: (1) the kernel density estimates in the subspace of the last layer and (2) the bayesian uncertainty estimates extracted from the drop-out layers~\cite{gal2016dropout}.
They show that the bayesian uncertainty estimates are typically higher for adversarial examples, but not enough to successfully detect them.
SafetyNet~\cite{lu2017safetynet} enforces an attacker to solve a discrete optimization problem.
Each activation of a ReLU layer is quantized in order to generate a discrete code (assumed to be different for adversarial examples and normal inputs), later used to train an RBF-SVM adversarial detector with high accuracy.

\citeauthor{zhang2018detecting}~\cite{zhao2018detecting} proposed a new detection mechanism that hides the input labels from the adversary.
This will prevent an attacker from maximizing the loss, given an input and a label and, thus, from creating an adversarial example.
The authors define a one-to-one encoding scheme from true labels to code vectors.
In order to detect adversarial examples, one can verify if the code vector computed from an input matches the signature of a class with certain precision.
If the output is negative, the input is treated as an adversarial example.
This approach shows good results on small data sets, but was not tested on more complex problems.

\citeauthor{abbasi2017robustness}~\cite{abbasi2017robustness} developed an ensemble of detectors based on the confusion matrix of each classifier.
The underlying idea is that adversarial instances originating from a given class tend to fall into a small subset of incorrect classes.
Therefore, developing an ensemble of detectors which can distinguish between confusion classes can more easily spot adversarial examples.
However, this claim was only evaluated on toy networks and data sets.
\citeauthor{meng2017magnet}~\cite{meng2017magnet} trained a model which distinguishes between a test and training input using the distance between an input and the data manifold.
A thresholding function later decides if an input is normal or adversarial.
However, the method is based on the existence of a distance function between an input and the data manifold.

\citeauthor{lee2017generative}~\cite{lee2017generative} proposed a generative training method in which two \acp{DNN} are trained alternatively.
The first network generates adversarial examples, while the second tries to correctly classify benign and adversarial examples.
Similarly, \citeauthor{song2017pixeldefend}~\cite{song2017pixeldefend} used generative networks to \emph{purify} adversarial examples and search for their true labels.
The authors run, at first, a statistical test to detect if an input belongs to the data generation distribution and, if the test fails, recover the input using generative models.
\citeauthor{ghosh2018resisting}~\cite{ghosh2018resisting} designed a generative model that finds a latent random variable such that the input and its label become conditionally independent given the latent variable.
The latent space is chosen as a mixture of Gaussians, such that each mixture component represents one of the classes in the data.
Inferring the label given the latent encoding is done by computing the contribution of the mixture components.
Adversarial samples are rejected based on thresholding the encoder and decoder outputs.
For example, if the distance between the sample encoding and the encoding of the predicted class in the latent space is below a threshold.


Although a wide range of adversarial detectors have been developed, using very distinct features, many of them fail to detect strong, adaptive attacks.
Moreover, their ability to generalize to new attacks is limited~--~ as new training is needed for new attacks~--~thus reducing their applicability.

\subsubsection{Input Transformation.}
\label{subsubsec:input}
Input transformations are thought to restrict the space of adversarial examples, therefore diminishing their impact.
\citeauthor{guo2017countering}~\cite{guo2017countering} suggest the use of several transformations: "bit-depth reduction, JPEG compression, total variance minimization and image quilting"~\cite{guo2017countering} as a pre-processing step to a convolutional classifier.
The idea of using JPEG compression was also explored in~\cite{dziugaite2016study, das2017keeping, shaham2018defending}.
Variance minimization and image quilting prove, in practice, the most effective transformations.
\citeauthor{shaham2018defending}~\cite{shaham2018defending} experiment with other input transformations: low-pass filters, PCA, low resolution wavelet approximations and soft-thresholding.
They found all transformations provide some robustness against strong white and black box attacks.
\citeauthor{xie2017mitigating}~\cite{xie2017mitigating} propose to use two simple randomization operations: (1) random resizing of input images and (2) random padding with zeros around the input images.
However, none of these approaches provide robustness against strong attacks.

Thermometer one hot encoding~\cite{buckman2018thermometer} breaks the linear  behavior of \acp{DNN} (suggested with the linearity hypothesis) by pre-processing the input with an extremely non-linear function.
Instead of replacing a real number with its counterpart transformation, the authors replace each real number with a binary vector.
Multiplying the input vector with the network's weight enables different input values to use different parameters of the network.
Inspired by thermometer encoding~\cite{buckman2018thermometer}, \citeauthor{rakin2018blind}~\cite{rakin2018blind} proposed to process the input data using an ensemble of methods which includes the tanh function, batch normalization, thermometer encoding and one hot encoding.
When used in combination with adversarial training, such methods show an increase in robustness because the resulting \acp{DNN} are less linear than normal ones.
However, this increase is not powerful enough to protect against all adversarial examples.

\citeauthor{chen2018improving}~\cite{chen2018improving} proposed a pre-processing technique that can successfully mask the gradients, even for iterative attackers.
Therefore, without access to the gradients, an attacker can not mount certain types of attacks.
Their proposal is based on encoding the whole input space using a small set of separable codewords and training a classifier on the encoded information.
Similarly, \citeauthor{liang2017detecting}~\cite{liang2017detecting} treat perturbations as noise and use noise reduction methods in order to mitigate their threat.
These methods show improvements for small data sets such as MNIST or CIFAR-10, but no improvements on larger data sets such as ImageNet.

In general, input transformations lead to stochastic gradients, which make attacks based on sensitivity analysis harder to mount.
However, when tested against attacks using the optimization strategy, most  input transformation defenses fail.
We discuss this weakness in Section~\ref{subsec:gradient_ob}.

\subsection{Defense by Design}
\label{subsec:proactive}
This class of defenses alter the model, the data or the learning procedure in order to increase robustness.
As opposed to guards, where some degree of robustness is achieved  earlier in the processing pipeline, proactive defenses aim to design better architectures or training procedures.

\subsubsection{Adversarial Traning.}
\label{subsubsec:adv_training}
Including adversarial examples in the training data set	 is a form of regularization~\cite{girosi1995regularization} \ie~a strategy designed to reduce the test error, possibly affecting the training error.
%
\citeauthor{goodfellow2014explaining}~\cite{goodfellow2014explaining} found that training with adversarial examples generated with the \ac{fgsm} attack, is an effective regularizer.
However, because the attack is not strong, the training procedure does not increase robustness significantly.
\citeauthor{sinha2018gradient}~\cite{sinha2018gradient} proposed a new training framework which assumes that simultaneous gradient updates should be statistically indistinguishable from each other.
Thus, the gradient can be regularized in order to remove salient information, which can lead to adversarial examples.
However, this phenomenon leads to masking the gradient and only protects against sensitivity analysis attacks.
\citeauthor{lyu2015unified}~\cite{lyu2015unified} abstracted from these a family of gradient based perturbations that can be used as regularization techniques.
Their work is a generalization of the \ac{fgsm}-based training procedure for all possible norms.
\citeauthor{roth2018adversarially}~\cite{roth2018adversarially} proposed a similar method to regularize the gradients focused on the correlation structure of the perturbations.
However, this method is data dependent and can only be used with certain types of perturbations.

Another way of training with adversarial examples is to completely exclude clean examples from the training procedure.
This method is equivalent to training on the \emph{worst} inputs possible, as in \eqq~(\ref{eq:minmax}).
Solving the inner maximization problem increases the training time and might not always be tractable.
In practice, however, approximating the result works very well.

\citeauthor{shaham2015understanding}~\cite{shaham2015understanding} first investigated the min-max training procedure from \eqq~(\ref{eq:minmax}), but found the inner maximization problem intractable.
Therefore, they propose to minimize an alternative form, in which they only evaluate a single step of the gradient (ascent and descent, corresponding to the steps in \eqq~(\ref{eq:minmax})).
Trying to solve the same problem, \citeauthor{huang2015learning}~\cite{huang2015learning} use linear approximation to approximate the inner maximization problem.
\citeauthor{madry2017towards}~\cite{madry2017towards} use \ac{pgd} for the inner maximization problem and suggest the problem is, in fact, tractable.
In order to explore a large part of the loss landscape, \ac{pgd} is restarted from many points in the ball around an input.
Surprisingly, although there are many local maxima spread widely apart within this space, they tend to have very well concentrated loss values.
This suggests that an adversarial example found by this method is representative for \emph{all} adversarial examples generated with first order methods.

\citeauthor{tramer2017ensemble}~\cite{tramer2017ensemble} developed a technique that augments training data with adversarial examples crafted on other static, pre-trained, models.
Training with ensemble methods increases the diversity of perturbations seen during training.
This is equivalent to enhancing the uncertainty sets with examples crafted on other models.
\citeauthor{dhillon2018stochastic}~\cite{dhillon2018stochastic} define a min-max zero-sum game between an adversary and a \ac{DNN}.
The strategy for this game is to prune a random subset of activations (\eg~those with smaller magnitude) and scale-up the survivors in order to compensate.
This approach is similar to the dropout technique, where the activations with high absolute values have a higher chance of being sampled.

In practice, adversarial training with the worse case perturbation yields very good results.
The Madry~\cite{madry2017towards} training procedure, which approximates the inner maximization problem using \ac{pgd} is a state-of-the-art defense.
Finding new and faster approximation methods was not explored in depth and remains an open question.

\subsubsection{Architectural Defenses.}
\label{subsubsec:architectural_defenses}
Some defenses propose to change the model's architecture by either imposing layer-wise constraints or by altering the final layer.
One of the first proposals uses distillation~--~a transfer learning method in which smaller \acp{DNN} are trained with knowledge extracted from larger \acp{DNN}~\cite{hinton2015distilling}.
\citeauthor{papernot2016distillation}~\cite{papernot2016distillation} used distillation in order to increase the \acp{DNN} robustness.
However, instead of using multiple \acp{DNN} in the training process, distillation is used for a single \ac{DNN}.
As opposed to the original distillation mechanism, the same network architecture is used both for training and distillation.
\citeauthor{cisse2017parseval}~\cite{cisse2017parseval} introduced a regularization scheme that constrains the Lipschitz constant layer-wise, thus avoiding any exponential growth of the constant.
In this setting, a regularization scheme (such as weight decay) applied to the last layer controls the overall Lipschitz constant of the network.

Similarly, \citeauthor{gu2014towards}~\cite{gu2014towards} proposed a new network architecture, called deep contractive networks, which imposes a layer-wise constraint.
This constraint minimizes the network's output variance \wrt~perturbations in the input \stt~a trained model can achieve robustness for perturbations around training data points.
\citeauthor{nayebi2017biologically}~\cite{nayebi2017biologically} developed a new training scheme inspired by biophysical principles in neural circuits.
Following the idea suggested by \citeauthor{goodfellow2014explaining}~\cite{goodfellow2014explaining} that adversarial examples are due to the linear summing of high dimensional input with small weights, the authors propose to force neural networks to operate in a non-linear, saturated, regime.
In order to enforce this constraint, the authors ensure that each element of the Jacobian matrix of the model is sufficiently small \stt~the model becomes insensitive to perturbations.

DeepCloak~\cite{gao2017deepcloak} removes features not used in classification in order to increase robustness to adversarial examples.
To identify unnecessary features, adversarial samples are tested against the clean example.
In order to remove the features, a mask layer is introduced before the logits layer.
The mask serves as a selector, keeping the necessary features and setting the unnecessary to null.
\citeauthor{lamb2018fortified}~\cite{lamb2018fortified} identify which hidden states are off the data manifold and map these states back to parts of the data manifold where the network performs better.
This process consists of inserting de-noising auto-encoders at crucial points between layers of the original network in order to clean up the transformed data points which may lie outside the data manifold.
\citeauthor{dumont2018robustness}~\cite{dumont2018robustness} investigated the resistance to adversarial attacks of three rotation-equivariant network architectures~\cite{cohen2016group}.
They discover that rotation equivariant networks are significantly more robust to attacks based on small translations and rotations, but marginally robust against attacks based on local geometric distortions.

\citeauthor{sun2017hypernetworks}~\cite{sun2017hypernetworks} proposed to use data dependent weights for each hidden layers of \acp{DNN}.
The weights are generated using HyperNetworks~\cite{ha2016hypernetworks}~--~a training technique in which a \ac{DNN} generates the weights for another.
The weights create an inductive bias specific to the training set which alleviates the effect of adversarial examples for small perturbations.
\citeauthor{pontes2018bidirectional}~\cite{pontes2018bidirectional} proposed to use bidirectional learning in order to increase robustness to adversarial examples.
Bidirectional learning trains two models~--~the first on the inputs and the second on a reversed copy of the inputs.
However, in \cite{pontes2018bidirectional}, a single model is trained to behave both as a discriminative and generative model.
Therefore, the same model can be a classifier and a generator in the same time.
This behavior is achieved using an un-directed \ac{DNN} that back-propagates the errors in both directions~--~each direction of the network has its own biases and the weights are shared.

Dense associative models~\cite{krotov2016densemem} store a set of vectors in memory, corresponding to the learnt patterns.
For example, pixel characteristics (such as intensity) can be stored together with the corresponding label.
At query time, the model tries to recover the memory sequences from incomplete data (corresponding to variations of an input, or perturbations).
If strong perturbations are applied to an input, the model will fail to recover the original label and consider the input adversarial.

Given the linear conjecture, introduced in Section~\ref{sec:causes}, which states that adversarial examples exploit the linear behavior of \acp{DNN}, a normal attempt to increase robustness is to use more non-linear activation functions.
\citeauthor{goodfellow2014explaining}~\cite{goodfellow2014explaining} suggested that RBF networks~--~which behave in a non-linear manner~--~are naturally immune to adversarial examples and have low confidence when they are fooled.
However, RBF units can not generalize very well and do not achieve the same performance as ReLU based \acp{DNN}.
\citeauthor{krotov2016dense}~\cite{krotov2016dense} tried to replace ReLU activations with higher rectified polynomials and observed an increase in robustness.
However, the increase is not sufficient to completely remove the adversarial phenomenon.

Several publications suggested that \acp{DNN} capacity increases their robustness to adversarial examples\cite{madry2017towards, kurakin2016aadversarial, rozsa2016accuracy}.
However, this correlation was not further investigated .

\subsubsection{Certified Defenses}
\label{subsec:certified_defenses}
Certified defenses rely on formal verification techniques in order to \emph{guarantee} that robustness holds within the bounds defined in Section~\ref{sec:robustness_eval},for the \emph{training} data set.
Certified defenses can be broadly classified based on the guarantees they give in (1) exact, deterministic guarantees~--~ which give a proof of robustness, (2) one-sided guarantees~--~which find a lower or an upper bound for robustness, (3) converging lower or upper bound or (4) statistical guarantees,  which quantify the probability that a model is robust~\cite{huang2018safety}.

Proving \emph{deterministic guarantees} for robustness has been formulated as a constrained solving problem where solutions are searched using SMT or mixed integer solvers.
Two approaches, namely Reluplex~\cite{katz2017reluplex} and Planet~\cite{ehlers2017formal}, use SMT solvers to verify robustness constraints for \acp{DNN} with ReLU activations.
Reluplex adapts the Simplex algorithm with rules for non-convex optimization in order to handle ReLU function, while Planet uses linear approximation in order to over-approximate the network's behavior.
If a property is not satisfied, these algorithms return counterexamples, which constitutes valid adversarial examples.
\citeauthor{carlini2018provably}~\cite{carlini2018provably} extends Reluplex to  verify $p_1$ and $p_\infty$ norm adversarial examples by encoding absolute values using ReLUs.
Scaling SMT/SAT solvers for larger models is, however, difficult and remains an open issue.

Other approaches have focused on \emph{certifying lower or upper bounds} on the existence of adversarial examples.
\citeauthor{wong2017provable}~\cite{wong2017provable} consider a convex outer approximation of the set of activation values which can be reached using adversarial examples and use linear programming to minimize the loss in this region.
The dual of the linear program can be specified as a \ac{DNN}, making the process efficient.
\citeauthor{dvijotham2018dual}~\cite{dvijotham2018dual} propose a similar form in which the dual problem is formulated and solved using a Lagragian relaxation of the optimization problem, thus obtaining an upper bound on the robustness against adversarial examples.
A different method to certify lower or upper bounds is to encode the inputs in an abstract domain, which contains all perturbations (\eg~a zonotope), and train with it~\cite{mirman2018differentiable, gehr2018ai, yang2019analyzing}.
Similarly, \citeauthor{gowal2018effectiveness}~\cite{gowal2018effectiveness} use interval bound propagation for training verifiable robust models.
These approaches are very effective and can scale to larger models.

\citeauthor{huang2017safety}~\cite{huang2017safety} give \emph{proofs of convergence} using SMT solvers by exhaustively searching for perturbations in a given norm ball, at each layer of \acp{DNN}.
Later, \citeauthor{wicker2018feature}~\cite{wicker2018feature}~and~\citeauthor{wu2018game}~\cite{wu2018game} extend the search using Monte-Carlo tree search methods.
The publications discussed in Section~\ref{sec:robustness_eval}, however, give \emph{statistical guarantees} to robustness~\cite{bastani2016measuring, weng2018evaluating, ruan2018reachability}.

Through their ability to guarantee that perturbations within some bounds can not cause misclassifications, certified defenses are well suited for tasks where robustness is paramount.
However, some techniques (such as using SMT solvers) are very complex, require more training resources and often do not scale to deep models.
Therefore, their applicability is reduced.
Moreover, guarantees can be given only for the training data set. 
As discussed in the following section, there are attacks which can bypass these defenses.


	\subsection{Gradient Masking and Defense Evaluation}
\label{subsec:gradient_ob}

Many defenses alleviate the model's sensitivity to small changes in the input by minimizing the gradients during the learning phase or constructing models without useful gradients.
However, forbidding access to gradient information is not enough to limit an attacker from constructing adversarial examples~\cite{papernot2017practical}.
This phenomenon, called \emph{gradient masking}~\cite{Papernot2018, papernot2017practical} was identified to give a false sense of security and leads to an improper evaluation of adversarial defenses~\cite{athalye2018obfuscated, tramer2017ensemble, carlini2017adversarial}.
For example, \citeauthor{galloway2017attacking}~\cite{galloway2017attacking} found that binarized neural networks exhibit stronger gradient masking, which sometimes gives them an advantage over full precision models, however, these methods only provide apparent robustness.
Defenses that exploit gradient masking can be sometimes broken with stronger attacks~\cite{carlini2017adversarial, oneval}.

Moreover, applying a defense may lead to a false sense of security because current defenses imply an attacker does not know the defense is used.
The subject of correctly evaluating adversarial defenses is at the forefront of research in adversarial examples and has proven to be not trivial~\cite{oneval}.
\citeauthor{oneval}~\cite{oneval} provide a methodological foundation for evaluating adversarial defenses.
Their suggestions are based on skepticism of the results and evaluation against strong, adaptive, attacks: the attack models should include the possible defense mechanisms employed.
Therefore, the evaluation must include the budget an attacker needs to spend in order to break a defense she knows.
At this moment, their work suggest the best protocol for evaluating new defenses and should be considered whenever a new defense is designed.


\fi
    \section{Transferability}
\label{sec:transferability}


Besides the existence of adversarial examples, \citeauthor{szegedy2013intriguing}~\cite{szegedy2013intriguing} showed that adversarial examples can transfer between different \ac{ML} models.
This phenomenon was later explored by \citeauthor{papernot2016transferability}~\cite{papernot2016transferability}, 
who studied the ability to transfer adversarial examples not only between \acp{DNN}, but also between different \ac{ML} techniques.
The authors identified two types of transferable adversarial examples: (1) \emph{intra-technique} examples~--~which transfer between models trained with the same \ac{ML} method and (2) \emph{cross-technique} examples~--~which transfer between models using different  methods.
\citeauthor{papernot2016transferability} found that all \ac{ML} models are vulnerable to intra-technique adversarial examples.
The phenomenon shows stronger for differentiable models than for non-differentiable models.
In regard to cross-technique adversarial examples, linear models, \acp{SVM}, decision trees or ensembles of models are more vulnerable between themselves compared to \acp{DNN}, which maintain some resilience against such techniques.
Later, \citeauthor{demontis2019adversarial}~\cite{demontis2019adversarial} showed that transferability is dependent on the input gradient's size (for the target classifier) and on the loss variance for the base base classifier. Less complex or regularized models, which lead to smaller input gradients, tend to be more robust.

\citeauthor{liu2016delving}~\cite{liu2016delving} investigated the transferability phenomenon at larger scale, using the ImageNet data set and various \acp{DNN} architectures.
Moreover, the authors discuss transferability in close relation to the attacker's goals (Section~\ref{sec:taxonomy_attacks}): (1) to cause an untargeted or (2) a targeted misclassification.
The experimental results show untargeted attacks transfer easily intra-technique.
In contrast, targeted attacks do not maintain labels during transfer.
Moreover, when only the hyper-parameters vary (but the architecture is preserved), transferability is not consistent, revealing that transferability  depends on hyper-parameters.
The results are strengthened by \citeauthor{su2018robustness}~\cite{su2018robustness} who run a large scale study of transferability between different \ac{DNN} architectures and show that targeted adversarial examples do not transfer between architectures.
However, transferability can be used to reverse engineer models and find the base architecture.

\citeauthor{tramer2017space}~\cite{tramer2017space} show empirical evidence that different models draw similar decision boundaries and proposed a method to measure the space where adversarial examples can be found.
Following the hypothesis that adversarial examples inhabit large, continuous regions instead of small pockets of the manifold (Section~\ref{sec:causes}), the authors measure the number of orthogonal perturbations that can lead to misclassifications.
Although these vectors are not sufficient to represent a basis of the adversarial space, they are a good indicator of its size.
Experimental results show this number depends on the network architecture.
Nonetheless, even small adversarial spaces are sufficient to intersect across different models and give transferable examples.
%
%
This study also shows that transferability is inherent to models which learn feature spaces with non-robust properties in the input space; an idea further developed in~\cite{ilyas2019adversarial}.
If models were to learn (or designed to select) distinct features with robust properties in the input space, it would be impossible to transfer adversarial examples.
Therefore, transferability emerges as a consequence of algorithm design.

\if 0\mode \bigskip \else\fi
In summary, the literature brings empirical evidence to show that different \ac{ML} algorithms learn a close representation of the input space.
Therefore, evading a region corresponding to a correct class with sufficient distance from the boundaries would most certainly transfer between models.
However, while the representation of classes is similar, they are not distributed alike in the output space, making it harder to transfer an input in a desired target region.

\if 0\mode
Nevertheless, this is, at the moment, an assumption that deserves future investigations.
\else \fi
\if 0\mode
We argue that transferability should be an important topic (often neglected in current publications) when designing attacks or defences against adversarial examples, but less significant when trying to give robustness guarantees.
This statement also follows from the conclusions in~\cite{tramer2017space}, that transferability is not an inherent property of non-robust models, but a consequence of algorithm design.
It is, therefore, compulsory to evaluate and discuss the importance of transferability, together with the threat model, when designing new attacks or defences.
\else\fi

    \section{Distilled Knowledge}
\label{sec:distilled_knowledge}


We focused on several points related to the adversarial examples phenomenon.
In this section we summarise and comment on the most interesting directions, that can also shape future research.

At first, as discussed in Section~\ref{sec:causes}, an unanimously accepted conjecture on the existence of adversarial examples is still missing.
Following the two main classes of attacks, we distinguish between two different perspectives on the existence of adversarial examples: (1) the case in which adversarial examples are drawn from the same distribution as normal inputs and lie on the same data manifold and (2) the case in which adversarial examples are part of a different distribution and lie off the data manifold.
To this moment there is not enough empirical data to refute any of these conjectures - although the adherents to the second hypothesis do not try to reject the first one.

Carlini and Wagner~\cite{carlini2017adversarial} questioned the second conjecture by defeating all the defences built upon it.
However, this evidence is not sufficient to completely refute it.
Regarding the first conjecture, some interesting insights are presented in~\cite{gilmer2018adversarial}, on artificial tasks where the data manifold can be explored.
We argue that similar inquiries that can reveal information about models with higher capacity are needed.
Moreover, more fundamental research on the causes and effects of perturbations can lead to models with increased robustness.
For example, limiting (or saturating) the gradients~\cite{nayebi2017biologically} as a response to sensitive features is not enough to completely remove the effect of adversarial examples.
This reveals that sensitive features are not the only cause for adversarial examples and highlights the need for further investigations.

Secondly, although a consistent body of publications claims security consequences of adversarial examples, very few present concrete (and practical) threat models or scenarios.
Close to none take into consideration the economics of using adversarial examples as opposed to other attacks.
From this perspective most security claims of adversarial examples remain invalid.
\if 0\mode We argue, in agreement with~\cite{gilmer2018adversarial}, that whenever authors claim security implications, a clear threat model and use cases must be presented.\else\fi
This angle opens new research questions: when (and why) attacks using adversarial examples are better?
And which defences act better in these use-cases?

Thirdly, the lack of standardised evaluation methods makes it difficult to compare  adversarial defences.
New publications should evaluate and present their results on models with high capacity (\eg~\cite{he2016deep}) and trained on large data sets (\eg~ImageNet~\cite{ILSVRC15}).
Moreover, the authors should present the results obtained against iterative (adaptive) attacks~--~considered state of the art.
Large scale evaluations, like the one suggested, will lead to better defences and, ultimately, to more robust models.

Several definitions of robustness are presented in the literature (and outlined in Section~\ref{sec:robustness_eval}).
They focus on the area around an input where no adversarial examples can be found or on the maximum perturbation for which an adversarial examples, specific to an input, can not be found.
Both definitions gravitate around a specific input.
In these circumstances, even if a certified region around some inputs can be guaranteed to be adversarial-free, it is not clear if this is enough to declare a model robust.
Since the training data set is only an approximation of the data generation distribution, one can argue that a certainty obtained on the training data set is only an approximation of a certainty for the data generation distribution.
Judging the notion of certified robustness from this angle raises the question if certification around inputs are sufficient to certify classes of inputs.

Lastly, the phenomenon of adversarial examples sheds light on a more important topic, often neglected~--~deep learning models learn differently than human beings do~--~which opens a number of fundamental questions.
It is not clear (from the literature) why the sensitivity to adversarial examples must be removed.
Is this problem relevant when the objective is to design a computer vision system that resembles the human perceptual system and can be used in a complex system; able to achieve some form of intelligence?
Further on, what is the next step if the phenomenon of adversarial examples is intrinsic to \ac{DNN} models which have even the smallest error~--~as suggested in~\cite{gilmer2018adversarial}?
\ac{DNN} show impressive results on the object recognition task and the ability to distinguish patterns similar to the way humans do.
It is not clear how adversarial examples influence these patterns, however, when the perturbations are small the patterns are still differentiable.
Can \ac{DNN} make better use of such patterns?
Moreover, when is the problem of adversarial examples considered 'solved'?
When one has to change an image so much that it resembles another object?
All these questions are still to be answered in future literature.


%


    \small
    \bibliographystyle{abbrv}
    \bibliography{clean_bib}

    \normalsize
    \appendix
\section{Benchmarks}
\label{sec:benchmarks}
In this Appendix we give references to the tools, data sets and \ac{DNN} architectures used in the literature about adversarial examples.
Moreover, for each defence we present the data set and \ac{DNN} architecture used for evaluation.

\subsection{Tools and Libraries}
Open source tools used to generate attacks or defences against adversarial examples.

\begin{table}[h]
	\centering
	\begin{tabular}{|l|l|l|}
		\toprule
		Name & Link & Publication \\ 
		\midrule
		CleverHans & https://cleverhans.readthedocs.io & \cite{papernot2016cleverhans} \\ \hline
		Foolbox & https://foolbox.readthedocs.io & \cite{rauber2017foolbox} \\ \hline
		Adversarial Robustness Toolbox & https://github.com/IBM/adversarial-robustness-toolbox & \cite{nicolae2018adversarial} \\ 
		\bottomrule
	\end{tabular}
	\caption{List of libraries for adversarial examples.}
\end{table}

\subsection{data sets and \ac{DNN} architectures.}
Common data sets and \ac{DNN} architectures used in the literature of adversarial examples.

\begin{table}[h]
	\centering
	\begin{minipage}{0.4\linewidth}
		\begin{tabular}{|l|l|}
		\toprule
		data set & Reference \\ 
		\midrule
		MNIST & \cite{lecun1998mnist} \\ \hline
		F-MNIST & \cite{xiao2017fashion} \\ \hline
		DEBRIN & \cite{arp2014drebin} \\ \hline
		Micro-RNA & \cite{shimomura2016novel} \\ \hline
		CIFAR-10/100 & \cite{krizhevsky2009learning, krizhevsky2014cifar} \\ \hline
		ImageNet & \cite{krizhevsky2012imagenet}	 \\ \hline
		ImageNet-1000 & \cite{deng2009imagenet} \\ \hline
		SVHN & \cite{netzer2011reading} \\ \hline
		HAR & \cite{anguita2013public} \\ \hline
		COIL-100 & \cite{nayar1996columbia} \\ 
		\bottomrule
	\end{tabular}
	\caption{List of common data sets used \\ in the literature of adversarial examples.}
	\end{minipage}%
	\begin{minipage}{.4\linewidth}
	\begin{tabular}{|l|l|}
		\toprule
		Model & Reference \\ 
		\midrule
		LeNet & \cite{lecun1999object} \\ \hline
		Maxout & \cite{goodfellow2013maxout} \\ \hline
		MxNet & \cite{chen2015mxnet} \\ \hline
		AlexNet & \cite{krizhevsky2012imagenet} \\ \hline
		ResNet & \cite{he2016deep} \\ \hline
		Wide ResNet & \cite{zagoruyko2016wide} \\ \hline
		VGG19 & \cite{simonyan2014very} \\ \hline
		DenseNet & \cite{huang2017densely} \\ \hline
		MobileNet & \cite{howard2017mobilenets} \\ \hline
		InceptionResNet-v2 & \cite{szegedy2017inception} \\ \hline
		Inception-v3 & \cite{szegedy2016rethinking} \\ \hline
		Inception-v4 & \cite{szegedy2017inception} \\ \hline
		DAM & \cite{krotov2016dense} \\ \hline
		Defence-GAN & \cite{samangouei2018defense} \\ 
		\bottomrule
	\end{tabular}
	\caption{List of common \ac{ML} models \\ used for adversarial examples on the \\ object recognition task.}
	\end{minipage}
\end{table}

\subsection{Defences Benchmark}
\begin{table}[h]
	\centering
	\begin{tabular}{|l|l|l|}
		\toprule
		Defence &%
		data sets &%
		Models \\ 
		\midrule
		Statistical Detection~\cite{grosse2017statistical} & MNIST, DREBIN, MicroRNA  &  \ac{dt}, \ac{SVM}, 2 layers-CNN \\ \hline
		Binary Classification~\cite{gong2017adversarial} & MNIST, CIFAR-10, SVHN  &  AlexNet \\ \hline
		In-Layer Detection~\cite{metzen2017detecting} & CIFAR-10, 10-class ImageNet  &  ResNet  \\ \hline
		Detecting from Artifacts~\cite{feinman2017detecting} & MNIST, CIFAR-10, SVHN & LeNet, 12-layer CNN  \\ \hline
		SafetyNet~\cite{lu2017safetynet} & CIFAR-10, ImageNet-1000 & ResNet, VGG19  \\ \hline
		Saliency Data Detector~\cite{zhang2018detecting} & MNIST, CIFAR-10, ImageNet &  AlexNet, AlexNet, VGG19 \\ \hline
		Linear Transformations Detector~\cite{bhagoji2018enhancing} &  MNIST, HAR & SVM  \\ \hline
		Key-based Networks~\cite{zhao2018detecting} & MNIST & 2/3-layers CNN  \\ \hline
		Ensemble Detectors~\cite{abbasi2017robustness} & MNIST, CIFAR-10 &  3-layers CNN \\ \hline
		Generative  Detector~\cite{lee2017generative} & CIFAR-10, CIFAR-100 & 6-layers CNN  \\ \hline
		Convolutional Statistics Detector~\cite{li2017adversarial} & ImageNet & VGG-16  \\ \hline
		Feature Squeezing~\cite{xu2017feature} & MNIST, CIFAR-10, ImageNet & \specialcell{7-layers CNN, DenseNet \\ MobileNet}  \\ \hline
		PixelDefend~\cite{song2017pixeldefend}  & ImageNet & ResNet, VGG \\ \hline
		MagNet~\cite{meng2017magnet} & MNIST, CIFAR-10 & 4/9-layers CNN \\ \hline
		VAE Detector~\cite{ghosh2018resisting}  & MNIST, SVNH, COIL-100 &  - \\ \hline
		Bit-Depth~\cite{guo2017countering} & ImageNet  & ResNet, DenseNet, Inception-v4  \\ \hline
		Basis Transformations~\cite{shaham2018defending}  & ImageNet & Inception-v3, Inception-v4  \\ \hline
		Randomised Transformations~\cite{xie2017mitigating} & ImageNet & Inception-v3, ResNet \\ \hline
		Thermometer Encoding~\cite{buckman2018thermometer} & MNIST, CIFAR-10, CIFAR-100, SVHN & 30-layers CNN, Wide ResNet  \\ \hline
		Blind Pre-Processing~\cite{rakin2018blind}  & MNIST, CIFAR-10, SVHN & LeNet, ResNet-50, ResNet-18  \\ \hline
		Data Discretisation~\cite{chen2018improving} & MNIST, CIFAR-10, ImageNET & InceptionResnet-V2   \\ \hline
		Adaptive Noise~\cite{liang2017detecting} & MNIST, ImageNet &  -  \\ \hline
		FGSM Training~\cite{goodfellow2014explaining} & MNIST &   Maxout \\ \hline
		Gradient Training~\cite{sinha2018gradient} & CIFAR-10, SVHN & ResNet-18 \\ \hline
		Gradient Regularisation~\cite{lyu2015unified} & MNIST, CIFAR-10 & Maxout \\ \hline
		Structured Gradient Regularisation~\cite{roth2018adversarially} & MNIST, CIFAR-10 & 9-layers CNN \\ \hline
		Robust Training~\cite{shaham2015understanding} & MNIST, CIFAR-10 &  2-layers CNN, VGG \\ \hline
		Strong Adversary Training~\cite{huang2015learning} & MNIST, CIFAR-10 & MxNet  \\ \hline
		CFOA Training~\cite{madry2017towards} & MNIST, CIFAR-10 & 2/4/6-layers CNN, Wide ResNet  \\ \hline
		Ensemble Training~\cite{tramer2017ensemble} & ImageNet & ResNet, InceptionResNet-v2 \\ \hline
		Stochastic Pruning~\cite{dhillon2018stochastic} & CIFAR-10 & Resnet-20 \\ \hline
		Distillation~\cite{hinton2015distilling} & MNIST, CIFAR-10 & 4-layers CNN \\ \hline
		Parseval Networks~\cite{cisse2017parseval} & MNIST, CIFAR-10, CIFAR-100, SVHN & ResNet, Wide Resnet \\ \hline
		Deep Contractive Networks~\cite{gu2014towards} & MNIST & LeNet, AlexNet \\ \hline
		Biological Networks~\cite{nayebi2017biologically} & MNIST & 3-layers CNN \\ \hline
		DeepCloak~\cite{gao2017deepcloak} & CIFAR-10 & ResNet-164 \\  \hline
		Fortified Networks~\cite{lamb2018fortified} & MNIST & 2-layers CNN \\ \hline
		Rotation-Equivariant Networks~\cite{dumont2018robustness} & CIFAR-10, ImageNet & 	ResNet \\ \hline
		HyperNetworks~\cite{sun2017hypernetworks} & ImageNet & ResNet \\ \hline
		Bidirectional Networks~\cite{pontes2018bidirectional} & MNIST, CIFAR-10 & 3-layers CNN \\ \hline
		DAM~\cite{krotov2016dense} & MNIST & DAM \\ \hline
		Certified Defences~\cite{raghunathan2018certified} & MNIST & 2-layers FC \\ \hline
		Formal Tools~\cite{katz2017reluplex, ehlers2017formal, huang2017safety, ruan2018reachability}  & - & - \\ \hline
		Distributional Robustness~\cite{sinha2018certifying}  & MNIST  & 3-layers CNN \\ \hline
		Convex Outer Polytope~\cite{kolter2017provable}  & MNIST, F-MNIST & 2-layers CNN \\ \hline
		Lischitz Margin~\cite{tsuzuku2018lipschitz}  & SVHN & Wide ResNet  \\ \hline
		Defence Gan~\cite{samangouei2018defense}  & MNIST, F-MNIST  & Defene-GAN \\ \hline
		FB-GAN~\cite{bao2018featurized}  & MNIST, F-MNIST & 8-layers CNN \\ 

	\bottomrule
	\end{tabular}
	\caption{data sets and \ac{ML} models used to benchmark defences against adversarial examples.}
	\label{tbl:defences_benchmark}
\end{table}

    \end{document}

\else
    \documentclass[acmsmall]{acmart} 
    \citestyle{acmnumeric}
    \usepackage{booktabs} 
    \usepackage[ruled]{algorithm2e} 



    
    \let\subcaption\relax

    \acmJournal{CSUR}
    \acmVolume{53}
    \acmNumber{3}
    \acmArticle{66}
    \acmYear{2020}
    \acmMonth{6}


    \setcopyright{acmcopyright}

    \acmDOI{10.1145/3398394}

    \begin{document}
    \title{Adversarial Examples on Object Recognition: A~Comprehensive Survey
        }
        \author{Alex Serban}
        \author{Erik Poll}
        \affiliation{%
        \institution{Radboud University}
        \streetaddress{Toernooiveld 212}
        \city{Nijmegen}
        \postcode{6525 EC}
        \country{The Netherlands}}
        \author{Joost Visser}
            \affiliation{%
    	\institution{Leiden University}
    	\streetaddress{Niels Bohrweg 1}
    	\city{Nijmegen}
    	\postcode{2333 CA}
    	\country{The Netherlands}}
    \email{a.serban@cs.ru.nl}
    \renewcommand\shortauthors{A. Serban}

	\ccsdesc[300]{Computing methodologies~Neural networks}
	\keywords{adversarial examples, machine learning, security, robustness}

    \begin{abstract}
        
    \end{abstract}
    \maketitle

    \acrodef{ML}[ML]{Machine Learning}
\acrodef{DL}[DL]{Deep Learning}
\acrodef{DNN}[DNN]{Deep Neural Network}
\acrodef{SVM}[SVM]{Support Vector Machine}
\acrodef{RL}[RL]{Reinforcement learning}
\acrodef{PAC}[PAC]{Probably Approximately Correct}

\acrodef{MILP}[MILP]{mixed integer linear programming}

\acrodef{wrt}[\emph{w.r.t}]{with respect to}
\acrodef{st}[\emph{s.t.}]{such that}

\acrodef{fgsm}[FGS]{Fast Gradient Sign}
\acrodef{bim}[BI]{Basic Iterative}
\acrodef{illcm}[ILC]{Iterative Least-likely Class}
\acrodef{jsma}[JSMA]{Jacobian-based Saliency Map Attack}
\acrodef{uap}[UAP]{Universal Adversarial Perturbations}
\acrodef{opa}[OPA]{One Pixel Attack}
\acrodef{pgd}[PGD]{projected gradient descent}
\acrodef{rssa}[RSSA]{Randomised Single Step Attack}
\acrodef{eat}[EAT]{Ensemble Adversarial Training}
\acrodef{gaa}[GAA]{Generative Adversarial Attacks}
\acrodef{gan}[GAN]{Generative Adversarial Networks}
\acrodef{nae}[NAE]{Natural Adversarial Examples}
\acrodef{atn}[ATN]{Adversarial Transformation Networks}
\acrodef{vae}[VAE]{Variational Auto-Encoders}
\acrodef{cfoa}[CFOA]{Complete First Order Adversary}
\acrodef{iid}[i.i.d]{independent and identically distributed}
\acrodef{bpda}[BPDA]{Backward Pass Differentiable Approximation}
\acrodef{alp}[ALP]{Adversarial Logit Pairing}
\acrodef{fbgan}[FB-GAN]{Featurized Bidirectional Generative Adversarial Networks}
\acrodef{sap}[SAP]{Stochastic Activation Pruning}
\acrodef{mat}[MAT]{Multi-strength Adversarial Training}
\acrodef{dam}[DAM]{Dense Associative Memory}
\acrodef{zoo}[ZOO]{Zeroth Order optimisation}
\acrodef{sa}[STA]{Strong Adversary}
\acrodef{lm}[LM]{Linear Models}
\acrodef{dt}[DT]{Decision Trees}
\acrodef{knn}[KNN]{K-nearest Neighbour}


	\section{Discussion}
\label{sec:discussion}

The literature on adversarial examples makes different claims about their impact on security and safety, although in practice all publications use the definition of robustness from Section~\ref{sec:robustness_eval} as a proxy to either of these notions.
In this section we discuss the relevance of robustness to security and safety and the economics of building more robust models.
Moreover, since p-norm is the dominant similarity metric, we also comment on its relevance.
This section ends with a discussion on the representations learned by \acp{DNN} and how they impact adversarial examples.

\paragraph{On the relevance of robustness to security.}

\citeauthor{gilmer2018adversarial}~\cite{gilmer2018motivating} express some skepticism about
whether adversarial examples are always a serious security concern.
Here it is interesting to note again that much
of the early work on adversarial machine learning \cite{lowd2005adversarial, barreno2010security} concerned
applications of \ac{ML} for security tasks, such as detecting spam,
malware, or network intrusions. In such applications there is by
definition an attacker interested in causing misclassification, as the
whole point of the system is to defend against such an attacker, and
hence misclassifications - esp.\ false negatives -- clearly have a
security impact.  By contrast, most recent work on
adversarial learning focuses on computer vision. While adversarial
examples may seem worrying thinking of some applications of computer
vision, say automated driving, this does not imply that there is an
interesting way for attackers to exploit it.
For example, \citeauthor{eykholt2018robust}~\cite{eykholt2018robust} use perturbed stop signs to attack autonomous vehicles.
However, the perturbations are far from sensible and can be detected by human observers.
Simply obscuring or removing the stop sign or 
may be easier ways to achieve the same effect.

Since adversarial examples rely on small perturbations, which can not be distinguished by human observers, they seem to be appropriate for attacks on systems which interact with humans or when the content of the message should not be heavily modified.
However, until now there are not many scenarios in which the lack of robustness defined as in Section~\ref{sec:robustness_eval} has strong security consequences.
Searching for attacks which can only be mounted with adversarial
examples against systems that do not carry out security tasks is
important to assess their real impact on security.

\paragraph{On the relevance of robustness to safety.}
From an engineering perspective, safety is the ability of a system to protect its users from  harmful or non-desirable outcomes.
The distinction between security and safety is that security protects
a system against \emph{intentional}, malicious, attacks while safety
protects a system from \emph{unintended} mishaps in the operational
environment.
Some publications aim to improve or validate the safety of \acp{DNN}~--~\eg~\cite{huang2017safety, lu2017safetynet, gehr2018ai}.
However, safety is an inherent property of a system and not of an algorithm solely.
Moreover, safety becomes important when a system can produce physical or material damage to humans, assets or the environment.
Talking about safety for systems without such impact~--~\eg~an image based search engine using \ac{ML}~--~is futile.
In order to guarantee safety, one should make sure that possible errors are detected and contained inside the system without affecting its normal operation.
Take again the example of an autonomous vehicle.
If the outcome of its computer vision system is cross checked with information coming from maps, the effect of using adversarial examples on traffic signs can be detected and contained inside the system, reducing their impact on safety.

The discussion of \ac{ML} safety in relation to the robustness definition from Section~\ref{sec:robustness_eval} should take into account the operational environment of an algorithm because some perturbations (such as those needed to build adversarial examples) may never appear in some environments, but may be common in others.
Besides, it might also be interesting to benchmark an algorithm and increase its robustness to common corruptions and perturbations~\cite{hendrycks2019benchmarking}.
Following this direction may reveal spots where current algorithms fail although the operational settings are natural and can be more relevant for safety.
Moreover, searching for distance metrics to better reflect the uncertainties in the operational environment of a system may lead to new threats to safety.

\paragraph{On the relevance of p-norm.}
The dominant similarity metric in the literature is the p-norm distance defined in Section~\ref{sec:background}.
Choosing an adequate metric is still an open question.
However, since there are no solutions to robustness for the p-norm
distance, it is hard to believe that using other metrics will result in more robust models~\cite{oneval}.
Nonetheless, there is an increased interest to explore new distance
functions, \eg~the Wasserstein distance~\cite{wong2019wasserstein} or using physical parameters underlying the image formation process~\cite{liu2018beyond}.
This is an interesting direction to pursue, which may lead to new attack and defense strategies and new insights about \ac{ML} algorithms.
We argue that the p-norm remains relevant for experimental settings;
however, it must be paired with relevant threat models in order to
evaluate its impact in security and search for operational
environments where it impacts safety.
Exploring new metrics and scenarios in which lack of robustness poses threats to safety or security is an important direction for future research.

\paragraph{On the economics of defending against adversarial examples.}
Until now there seems to be a tradeoff between accuracy and robustness to adversarial examples, inherent to the algorithms and the training methods used.
Moreover, the results presented in Section~\ref{sec:causes} suggest that training models robust to adversarial examples requires more resources than non-robust ones.
This means robustness comes at a cost.
Whether these cost are acceptable, and how high they can be, will
depend on the application and the context.
Given that \ac{ML} systems are not widely deployed and, as mentioned
earlier, the real impact of adversarial examples on safety and
security is still to be determined, it remains to be seen which
defenses can be cost-effective in practice.

\paragraph{On the representations learned by \acp{DNN}.}
The sensitivity of \acp{DNN} to adversarial examples raises questions about their ability to learn high level abstractions from data.
Although it is believed that increasing the depth of a network helps increasing the level of abstraction and it was observed that early layers in convolutional networks learn filters that resemble contour extractors, while deeper layers learn more complex patterns, \acp{DNN} seem to learn superficial abstractions restricted to the space on which they operate.
In object recognition, the training objectives lie in pixel space, and not in a conceptual or relational space.
Pixel spaces are necessary for extracting first order information about the task, but seem to be insufficient for higher level abstractions needed to overcome complex perception systems.
Moreover, the capacity to create adversarial inputs which are not intelligible by humans (as in~\cite{nguyen2015deep}) shows that \acp{DNN} use different features than we wish for.
Research in adversarial examples strengthen the conclusions from \citeauthor{jo2017measuring}~\cite{jo2017measuring} which analyzed convolutional networks in different regimes and showed they exhibit a tendency to learn surface regularities rather than higher-level abstract concepts.
Therefore, adversarial examples might be intrinsic to the methods used to solve \ac{ML} related tasks or to the current training procedures.
In this context, it is interesting to search for models which learn a better representation of the world and which may solve the sensitivity to adversarial examples as a side effect.

	\section{Conclusions and Future Research}
\label{sec:conclusions}

We focused on several aspects of the adversarial examples phenomenon, in an attempt to provide a comprehensive and self contained survey of this field of research.
In particular, we focused on explaining the hypotheses on the existence of adversarial examples, position the phenomenon in the security and robustness context, describe and characterize the attacks and defenses proposed in the literature and present the ability of adversarial examples to transfer between different \ac{ML} techniques.
In the appendix we have included examples of software libraries and competitions for developing attacks and defenses.

To conclude, we note that adversarial examples are an intriguing phenomenon of \ac{ML} algorithms and their existence can raise both safety or security alarms.
However, their true impact on safety and securiity is hard to estimate at this moment in time.
A key  take away is that the phenomenon of adversarial examples has no generally accepted explanation or solution.
Moreover, until now all defenses (including the ones using formal verification)  have been broken.
Therefore, the field remains active and spans  several future research directions.


At first, since no universally accepted conjecture on the existence of adversarial examples exist, a more fundamental approach is required.
Such an inquiry can use methods from topology (to analyze the data manifold or the decision boundaries), from statistics (to extract information about distribution of adversarial examples) or from learning theory (to investigate in which theoretical settings the resources needed for training robust models can be decreased).
Nonetheless, such an inquiry must be accompanied by practical reasoning about the impact of adversarial examples on security by developing, for example, new threat models. Similarly for safety, by searching for operational scenarios in which the deployment of \ac{ML} algorithms is hindered by adversarial examples or searching for scenarios that require new ways to measure and certify robustness.

Secondly, the use of certified defenses suggests robustness within certain bounds can be achieved (or, at least, precisely measured).
It is interesting to search for new certified defenses with small computational complexity, which can better scale to large models.
When designing new certified defenses it is important to consider how new techniques can help increase the bounds for robustness and not only precisely measure it.
Moreover, in ideal settings, new defenses should alter the models as little as possible and should help already trained models too.
Besides, it is important to search for spaces where certify defenses fail or investigate attacks which can break them.
At the moment, this research is sparse, giving some indications that certified defenses can also be bypassed~\cite{poursaeed2018generative}.

Thirdly, although asymptotic results suggest solving the adversarial examples problem requires more resources (data or computational), searching for new approximate solutions is essential.
For other tasks, theoretical results suggest similar limits, although approximate solutions already achieve very good results.
Therefore, finding better methods to approximate the perturbation space and train robust models in low data regimes is an interesting research direction.

Lastly, one important take away from analyzing the phenomenon of adversarial examples is that \ac{ML} and, in particular, \acp{DNN}  operate in spaces less abstract than humans do.
Therefore, perturbations with no impact in the human space have a large impact in the algorithmic space.
We may compare adversarial examples with human optical illusions, where one is fooled by perceiving individual stimuli as a whole or illusions that create images different from the objects that make them (\eg~the rabbit-duck illusion).
These subjective stances humans experience, called qualia, have been long studied in philosophy and psychology and are common in the space of human cognition or emotion.
However, human beings are able to identify when such phenomenons occur (most humans can identify an optical illusion as an optical illusion, but can not classify its content correctly).
Moreover, the effect of illusions or subjective beliefs can be removed by humans through rational analysis.
This type of analysis has been identified as using the 'slower' part of our cognition system, which seems to be missing from the way \ac{ML} models are currently built.
If limited by time, humans can also be fooled by adversarial examples~\cite{elsayed2018adversarial}.
However, our capacity to engage in rational analysis and adapt to the environment removes such threats and remains to be implemented in machines.

    \bibliographystyle{ACM-Reference-Format}
	\bibliography{small_bib}


\begin{thebibliography}{186}


\ifx \showCODEN    \undefined \def \showCODEN     #1{\unskip}     \fi
\ifx \showDOI      \undefined \def \showDOI       #1{#1}\fi
\ifx \showISBNx    \undefined \def \showISBNx     #1{\unskip}     \fi
\ifx \showISBNxiii \undefined \def \showISBNxiii  #1{\unskip}     \fi
\ifx \showISSN     \undefined \def \showISSN      #1{\unskip}     \fi
\ifx \showLCCN     \undefined \def \showLCCN      #1{\unskip}     \fi
\ifx \shownote     \undefined \def \shownote      #1{#1}          \fi
\ifx \showarticletitle \undefined \def \showarticletitle #1{#1}   \fi
\ifx \showURL      \undefined \def \showURL       {\relax}        \fi
\providecommand\bibfield[2]{#2}
\providecommand\bibinfo[2]{#2}
\providecommand\natexlab[1]{#1}
\providecommand\showeprint[2][]{arXiv:#2}

\bibitem[\protect\citeauthoryear{Abbasi and Gagn{\'e}}{Abbasi and
  Gagn{\'e}}{2017}]%
        {abbasi2017robustness}
\bibfield{author}{\bibinfo{person}{Mahdieh Abbasi} {and}
  \bibinfo{person}{Christian Gagn{\'e}}.} \bibinfo{year}{2017}\natexlab{}.
\newblock \bibinfo{title}{Robustness to adversarial examples through an
  ensemble of specialists}.
\newblock \bibinfo{howpublished}{arXiv:1702.06856}.
\newblock


\bibitem[\protect\citeauthoryear{Akhtar and Mian}{Akhtar and Mian}{2018}]%
        {akhtar2018threat}
\bibfield{author}{\bibinfo{person}{Naveed Akhtar} {and} \bibinfo{person}{Ajmal
  Mian}.} \bibinfo{year}{2018}\natexlab{}.
\newblock \showarticletitle{Threat of adversarial attacks on deep learning in
  computer vision: A survey}.
\newblock \bibinfo{journal}{\emph{IEEE Access}} (\bibinfo{year}{2018}).
\newblock


\bibitem[\protect\citeauthoryear{Alzantot, Sharma, Chakraborty, and
  Srivastava}{Alzantot et~al\mbox{.}}{2018}]%
        {alzantot2018genattack}
\bibfield{author}{\bibinfo{person}{Moustafa Alzantot}, \bibinfo{person}{Yash
  Sharma}, \bibinfo{person}{Supriyo Chakraborty}, {and} \bibinfo{person}{Mani
  Srivastava}.} \bibinfo{year}{2018}\natexlab{}.
\newblock \bibinfo{title}{GenAttack: Practical Black-box Attacks with
  Gradient-Free Optimization}.
\newblock \bibinfo{howpublished}{arXiv:1805.11090}.
\newblock


\bibitem[\protect\citeauthoryear{Athalye, Carlini, and Wagner}{Athalye
  et~al\mbox{.}}{2018a}]%
        {athalye2018obfuscated}
\bibfield{author}{\bibinfo{person}{Anish Athalye}, \bibinfo{person}{Nicholas
  Carlini}, {and} \bibinfo{person}{David Wagner}.}
  \bibinfo{year}{2018}\natexlab{a}.
\newblock \showarticletitle{Obfuscated gradients give a false sense of
  security: Circumventing defenses to adversarial examples}.
\newblock \bibinfo{journal}{\emph{ICML}} (\bibinfo{year}{2018}).
\newblock


\bibitem[\protect\citeauthoryear{Athalye, Engstrom, Ilyas, and Kwok}{Athalye
  et~al\mbox{.}}{2018b}]%
        {athalye2017synthesizing}
\bibfield{author}{\bibinfo{person}{Anish Athalye}, \bibinfo{person}{Logan
  Engstrom}, \bibinfo{person}{Andrew Ilyas}, {and} \bibinfo{person}{Kevin
  Kwok}.} \bibinfo{year}{2018}\natexlab{b}.
\newblock \showarticletitle{Synthesizing robust adversarial examples}.
\newblock \bibinfo{journal}{\emph{ICML}} (\bibinfo{year}{2018}).
\newblock


\bibitem[\protect\citeauthoryear{Baluja and Fischer}{Baluja and
  Fischer}{2018}]%
        {baluja2017adversarial}
\bibfield{author}{\bibinfo{person}{Shumeet Baluja} {and} \bibinfo{person}{Ian
  Fischer}.} \bibinfo{year}{2018}\natexlab{}.
\newblock \showarticletitle{Adversarial transformation networks: Learning to
  generate adversarial examples}. In \bibinfo{booktitle}{\emph{AAAI}}.
\newblock


\bibitem[\protect\citeauthoryear{Barreno, Nelson, Joseph, and Tygar}{Barreno
  et~al\mbox{.}}{2010}]%
        {barreno2010security}
\bibfield{author}{\bibinfo{person}{Marco Barreno}, \bibinfo{person}{Blaine
  Nelson}, \bibinfo{person}{Anthony~D. Joseph}, {and} \bibinfo{person}{JD.
  Tygar}.} \bibinfo{year}{2010}\natexlab{}.
\newblock \showarticletitle{The security of machine learning}.
\newblock \bibinfo{journal}{\emph{Machine Learning}} (\bibinfo{year}{2010}).
\newblock


\bibitem[\protect\citeauthoryear{Barreno, Nelson, Sears, Joseph, and
  Tygar}{Barreno et~al\mbox{.}}{2006}]%
        {barreno2006can}
\bibfield{author}{\bibinfo{person}{Marco Barreno}, \bibinfo{person}{Blaine
  Nelson}, \bibinfo{person}{Russell Sears}, \bibinfo{person}{Anthony~D.
  Joseph}, {and} \bibinfo{person}{J.~Doug Tygar}.}
  \bibinfo{year}{2006}\natexlab{}.
\newblock \showarticletitle{Can machine learning be secure?}. In
  \bibinfo{booktitle}{\emph{ASIACCS}}. ACM.
\newblock


\bibitem[\protect\citeauthoryear{Bastani, Ioannou, Lampropoulos, Vytiniotis,
  Nori, and Criminisi}{Bastani et~al\mbox{.}}{2016}]%
        {bastani2016measuring}
\bibfield{author}{\bibinfo{person}{Osbert Bastani}, \bibinfo{person}{Yani
  Ioannou}, \bibinfo{person}{Leonidas Lampropoulos}, \bibinfo{person}{Dimitrios
  Vytiniotis}, \bibinfo{person}{Aditya Nori}, {and} \bibinfo{person}{Antonio
  Criminisi}.} \bibinfo{year}{2016}\natexlab{}.
\newblock \showarticletitle{Measuring neural net robustness with constraints}.
\newblock \bibinfo{journal}{\emph{NeurIPS}} (\bibinfo{year}{2016}).
\newblock


\bibitem[\protect\citeauthoryear{Behzadan and Munir}{Behzadan and
  Munir}{2017}]%
        {behzadan2017vulnerability}
\bibfield{author}{\bibinfo{person}{Vahid Behzadan} {and}
  \bibinfo{person}{Arslan Munir}.} \bibinfo{year}{2017}\natexlab{}.
\newblock \showarticletitle{Vulnerability of deep reinforcement learning to
  policy induction attacks}. In \bibinfo{booktitle}{\emph{Int. Conference on
  Machine Learning and Data Mining in Pattern Recognition}}. Springer.
\newblock


\bibitem[\protect\citeauthoryear{Ben-Tal, El~Ghaoui, and Nemirovski}{Ben-Tal
  et~al\mbox{.}}{2009}]%
        {ben2009robust}
\bibfield{author}{\bibinfo{person}{Aharon Ben-Tal}, \bibinfo{person}{Laurent
  El~Ghaoui}, {and} \bibinfo{person}{Arkadi Nemirovski}.}
  \bibinfo{year}{2009}\natexlab{}.
\newblock \bibinfo{booktitle}{\emph{Robust optimization}}.
\newblock \bibinfo{publisher}{Princeton University Press}.
\newblock


\bibitem[\protect\citeauthoryear{Bhagoji, Cullina, Sitawarin, and
  Mittal}{Bhagoji et~al\mbox{.}}{2018}]%
        {bhagoji2018enhancing}
\bibfield{author}{\bibinfo{person}{Arjun~Nitin Bhagoji},
  \bibinfo{person}{Daniel Cullina}, \bibinfo{person}{Chawin Sitawarin}, {and}
  \bibinfo{person}{Prateek Mittal}.} \bibinfo{year}{2018}\natexlab{}.
\newblock \showarticletitle{Enhancing robustness of machine learning systems
  via data transformations}. In \bibinfo{booktitle}{\emph{CISS}}. IEEE.
\newblock


\bibitem[\protect\citeauthoryear{Bhagoji, He, Li, and Song}{Bhagoji
  et~al\mbox{.}}{2017}]%
        {bhagoji2017exploring}
\bibfield{author}{\bibinfo{person}{Arjun~Nitin Bhagoji},
  \bibinfo{person}{Warren He}, \bibinfo{person}{Bo Li}, {and}
  \bibinfo{person}{Dawn Song}.} \bibinfo{year}{2017}\natexlab{}.
\newblock \bibinfo{title}{Exploring the Space of Black-box Attacks on Deep
  Neural Networks}.
\newblock \bibinfo{howpublished}{arXiv:1712.09491}.
\newblock


\bibitem[\protect\citeauthoryear{Biggio, Fumera, and Roli}{Biggio
  et~al\mbox{.}}{2014}]%
        {biggio2014security}
\bibfield{author}{\bibinfo{person}{Battista Biggio}, \bibinfo{person}{Giorgio
  Fumera}, {and} \bibinfo{person}{Fabio Roli}.}
  \bibinfo{year}{2014}\natexlab{}.
\newblock \showarticletitle{Security evaluation of pattern classifiers under
  attack}. In \bibinfo{booktitle}{\emph{TKDE}}. IEEE.
\newblock


\bibitem[\protect\citeauthoryear{Biggio, Nelson, and Laskov}{Biggio
  et~al\mbox{.}}{2012}]%
        {biggio2012poisoning}
\bibfield{author}{\bibinfo{person}{Battista Biggio}, \bibinfo{person}{Blaine
  Nelson}, {and} \bibinfo{person}{Pavel Laskov}.}
  \bibinfo{year}{2012}\natexlab{}.
\newblock \bibinfo{title}{Poisoning attacks against support vector machines}.
\newblock \bibinfo{howpublished}{arXiv:1206.6389}.
\newblock


\bibitem[\protect\citeauthoryear{Biggio and Roli}{Biggio and Roli}{2018}]%
        {Biggio2018}
\bibfield{author}{\bibinfo{person}{Battista Biggio} {and}
  \bibinfo{person}{Fabio Roli}.} \bibinfo{year}{2018}\natexlab{}.
\newblock \showarticletitle{Wild patterns: Ten years after the rise of
  adversarial machine learning}.
\newblock \bibinfo{journal}{\emph{Pattern Recognition}} (\bibinfo{year}{2018}).
\newblock


\bibitem[\protect\citeauthoryear{Boyd and Vandenberghe}{Boyd and
  Vandenberghe}{2004}]%
        {boyd2004convex}
\bibfield{author}{\bibinfo{person}{Stephen Boyd} {and} \bibinfo{person}{Lieven
  Vandenberghe}.} \bibinfo{year}{2004}\natexlab{}.
\newblock \bibinfo{booktitle}{\emph{Convex optimization}}.
\newblock \bibinfo{publisher}{Cambridge University Press}.
\newblock


\bibitem[\protect\citeauthoryear{Br{\"u}ckner, Kanzow, and
  Scheffer}{Br{\"u}ckner et~al\mbox{.}}{2012}]%
        {bruckner2012static}
\bibfield{author}{\bibinfo{person}{Michael Br{\"u}ckner},
  \bibinfo{person}{Christian Kanzow}, {and} \bibinfo{person}{Tobias Scheffer}.}
  \bibinfo{year}{2012}\natexlab{}.
\newblock \showarticletitle{Static prediction games for adversarial learning
  problems}.
\newblock \bibinfo{journal}{\emph{JMLR}} (\bibinfo{year}{2012}).
\newblock


\bibitem[\protect\citeauthoryear{Bubeck, Price, and Razenshteyn}{Bubeck
  et~al\mbox{.}}{2018}]%
        {bubeck2018adversarial}
\bibfield{author}{\bibinfo{person}{S{\'e}bastien Bubeck}, \bibinfo{person}{Eric
  Price}, {and} \bibinfo{person}{Ilya Razenshteyn}.}
  \bibinfo{year}{2018}\natexlab{}.
\newblock \bibinfo{title}{Adversarial examples from computational constraints}.
\newblock \bibinfo{howpublished}{arXiv:1805.10204}.
\newblock


\bibitem[\protect\citeauthoryear{Buckman, Roy, Raffel, and Goodfellow}{Buckman
  et~al\mbox{.}}{2018}]%
        {buckman2018thermometer}
\bibfield{author}{\bibinfo{person}{Jacob Buckman}, \bibinfo{person}{Aurko Roy},
  \bibinfo{person}{Colin Raffel}, {and} \bibinfo{person}{Ian Goodfellow}.}
  \bibinfo{year}{2018}\natexlab{}.
\newblock \showarticletitle{Thermometer encoding: One hot way to resist
  adversarial examples}.
\newblock \bibinfo{journal}{\emph{ICLR}} (\bibinfo{year}{2018}).
\newblock


\bibitem[\protect\citeauthoryear{Carlini, Athalye, Papernot, Brendel, Rauber,
  Tsipras, Goodfellow, Madry, and Kurakin}{Carlini et~al\mbox{.}}{2019}]%
        {oneval}
\bibfield{author}{\bibinfo{person}{Nicholas Carlini}, \bibinfo{person}{Anish
  Athalye}, \bibinfo{person}{Nicolas Papernot}, \bibinfo{person}{Wieland
  Brendel}, \bibinfo{person}{Jonas Rauber}, \bibinfo{person}{Dimitris Tsipras},
  \bibinfo{person}{Ian~J. Goodfellow}, \bibinfo{person}{Aleksander Madry},
  {and} \bibinfo{person}{Alexey Kurakin}.} \bibinfo{year}{2019}\natexlab{}.
\newblock \showarticletitle{On Evaluating Adversarial Robustness}.
\newblock \bibinfo{journal}{\emph{arXiv:1902.06705}} (\bibinfo{year}{2019}).
\newblock


\bibitem[\protect\citeauthoryear{Carlini, Katz, Berret, and Dill}{Carlini
  et~al\mbox{.}}{2018}]%
        {carlini2018provably}
\bibfield{author}{\bibinfo{person}{Nicholas Carlini}, \bibinfo{person}{Guy
  Katz}, \bibinfo{person}{Clark Berret}, {and} \bibinfo{person}{David Dill}.}
  \bibinfo{year}{2018}\natexlab{}.
\newblock \bibinfo{title}{Provably Minimally-Distorted Adversarial Examples}.
\newblock \bibinfo{howpublished}{arXiv:1711.00851}.
\newblock


\bibitem[\protect\citeauthoryear{Carlini, Mishra, Vaidya, Zhang, Sherr,
  Shields, Wagner, and Zhou}{Carlini et~al\mbox{.}}{2016}]%
        {carlini2016hidden}
\bibfield{author}{\bibinfo{person}{Nicholas Carlini}, \bibinfo{person}{Pratyush
  Mishra}, \bibinfo{person}{Tavish Vaidya}, \bibinfo{person}{Yuankai Zhang},
  \bibinfo{person}{Micah Sherr}, \bibinfo{person}{Clay Shields},
  \bibinfo{person}{David Wagner}, {and} \bibinfo{person}{Wenchao Zhou}.}
  \bibinfo{year}{2016}\natexlab{}.
\newblock \showarticletitle{Hidden Voice Commands}. In
  \bibinfo{booktitle}{\emph{USENIX Security}}.
\newblock


\bibitem[\protect\citeauthoryear{Carlini and Wagner}{Carlini and
  Wagner}{2017a}]%
        {carlini2017adversarial}
\bibfield{author}{\bibinfo{person}{Nicholas Carlini} {and}
  \bibinfo{person}{David Wagner}.} \bibinfo{year}{2017}\natexlab{a}.
\newblock \showarticletitle{Adversarial examples are not easily detected:
  Bypassing ten detection methods}. In \bibinfo{booktitle}{\emph{AISec}}. ACM.
\newblock


\bibitem[\protect\citeauthoryear{Carlini and Wagner}{Carlini and
  Wagner}{2017b}]%
        {carlini2017magnet}
\bibfield{author}{\bibinfo{person}{Nicholas Carlini} {and}
  \bibinfo{person}{David Wagner}.} \bibinfo{year}{2017}\natexlab{b}.
\newblock \bibinfo{title}{Magnet and {"Efficient defenses against adversarial
  attacks"} are not robust to adversarial examples}.
\newblock \bibinfo{howpublished}{arXiv:1711.08478}.
\newblock


\bibitem[\protect\citeauthoryear{Carlini and Wagner}{Carlini and
  Wagner}{2017c}]%
        {carlini2017towards}
\bibfield{author}{\bibinfo{person}{Nicholas Carlini} {and}
  \bibinfo{person}{David Wagner}.} \bibinfo{year}{2017}\natexlab{c}.
\newblock \showarticletitle{Towards evaluating the robustness of neural
  networks}. In \bibinfo{booktitle}{\emph{S\&P}}. IEEE.
\newblock


\bibitem[\protect\citeauthoryear{Carlini and Wagner}{Carlini and
  Wagner}{2018}]%
        {carlini2018audio}
\bibfield{author}{\bibinfo{person}{Nicholas Carlini} {and}
  \bibinfo{person}{David Wagner}.} \bibinfo{year}{2018}\natexlab{}.
\newblock \bibinfo{title}{Audio adversarial examples: Targeted attacks on
  speech-to-text}.
\newblock \bibinfo{howpublished}{arXiv:1801.01944}.
\newblock


\bibitem[\protect\citeauthoryear{Chen, Wu, Liang, and Jha}{Chen
  et~al\mbox{.}}{2018b}]%
        {chen2018improving}
\bibfield{author}{\bibinfo{person}{Jiefeng Chen}, \bibinfo{person}{Xi Wu},
  \bibinfo{person}{Yingyu Liang}, {and} \bibinfo{person}{Somesh Jha}.}
  \bibinfo{year}{2018}\natexlab{b}.
\newblock \bibinfo{title}{Improving Adversarial Robustness by Data-Specific
  Discretization}.
\newblock \bibinfo{howpublished}{arXiv:1805.07816}.
\newblock


\bibitem[\protect\citeauthoryear{Chen, Sharma, Zhang, Yi, and Hsieh}{Chen
  et~al\mbox{.}}{2018a}]%
        {chen2017ead}
\bibfield{author}{\bibinfo{person}{Pin-Yu Chen}, \bibinfo{person}{Yash Sharma},
  \bibinfo{person}{Huan Zhang}, \bibinfo{person}{Jinfeng Yi}, {and}
  \bibinfo{person}{Cho-Jui Hsieh}.} \bibinfo{year}{2018}\natexlab{a}.
\newblock \showarticletitle{EAD: elastic-net attacks to deep neural networks
  via adversarial examples}. In \bibinfo{booktitle}{\emph{AAAI}}.
\newblock


\bibitem[\protect\citeauthoryear{Chen, Zhang, Sharma, Yi, and Hsieh}{Chen
  et~al\mbox{.}}{2017}]%
        {chen2017zoo}
\bibfield{author}{\bibinfo{person}{Pin-Yu Chen}, \bibinfo{person}{Huan Zhang},
  \bibinfo{person}{Yash Sharma}, \bibinfo{person}{Jinfeng Yi}, {and}
  \bibinfo{person}{Cho-Jui Hsieh}.} \bibinfo{year}{2017}\natexlab{}.
\newblock \showarticletitle{{ZOO}: Zeroth order optimization based black-box
  attacks to deep neural networks without training substitute models}. In
  \bibinfo{booktitle}{\emph{AISec}}. ACM.
\newblock


\bibitem[\protect\citeauthoryear{Cisse, Bojanowski, Grave, Dauphin, and
  Usunier}{Cisse et~al\mbox{.}}{2017}]%
        {cisse2017parseval}
\bibfield{author}{\bibinfo{person}{Moustapha Cisse}, \bibinfo{person}{Piotr
  Bojanowski}, \bibinfo{person}{Edouard Grave}, \bibinfo{person}{Yann Dauphin},
  {and} \bibinfo{person}{Nicolas Usunier}.} \bibinfo{year}{2017}\natexlab{}.
\newblock \showarticletitle{Parseval networks: Improving robustness to
  adversarial examples}.
\newblock \bibinfo{journal}{\emph{ICML}} (\bibinfo{year}{2017}).
\newblock


\bibitem[\protect\citeauthoryear{Cohen and Welling}{Cohen and Welling}{2016}]%
        {cohen2016group}
\bibfield{author}{\bibinfo{person}{Taco Cohen} {and} \bibinfo{person}{Max
  Welling}.} \bibinfo{year}{2016}\natexlab{}.
\newblock \showarticletitle{Group equivariant convolutional networks}.
\newblock \bibinfo{journal}{\emph{ICML}} (\bibinfo{year}{2016}).
\newblock


\bibitem[\protect\citeauthoryear{Cullina, Bhagoji, and Mittal}{Cullina
  et~al\mbox{.}}{2018}]%
        {cullina2018pac}
\bibfield{author}{\bibinfo{person}{Daniel Cullina},
  \bibinfo{person}{Arjun~Nitin Bhagoji}, {and} \bibinfo{person}{Prateek
  Mittal}.} \bibinfo{year}{2018}\natexlab{}.
\newblock \bibinfo{title}{PAC-learning in the presence of evasion adversaries}.
\newblock \bibinfo{howpublished}{arXiv:1806.01471}.
\newblock


\bibitem[\protect\citeauthoryear{Dalvi, Domingos, Sanghai, Verma,
  et~al\mbox{.}}{Dalvi et~al\mbox{.}}{2004}]%
        {dalvi2004adversarial}
\bibfield{author}{\bibinfo{person}{Nilesh Dalvi}, \bibinfo{person}{Pedro
  Domingos}, \bibinfo{person}{Sumit Sanghai}, \bibinfo{person}{Deepak Verma},
  {et~al\mbox{.}}} \bibinfo{year}{2004}\natexlab{}.
\newblock \showarticletitle{Adversarial classification}. In
  \bibinfo{booktitle}{\emph{SIGKDD}}. ACM.
\newblock


\bibitem[\protect\citeauthoryear{Das, Shanbhogue, Chen, Hohman, Chen, Kounavis,
  and Chau}{Das et~al\mbox{.}}{2017}]%
        {das2017keeping}
\bibfield{author}{\bibinfo{person}{Nilaksh Das}, \bibinfo{person}{Madhuri
  Shanbhogue}, \bibinfo{person}{Shang-Tse Chen}, \bibinfo{person}{Fred Hohman},
  \bibinfo{person}{Li Chen}, \bibinfo{person}{Michael~E. Kounavis}, {and}
  \bibinfo{person}{Duen~Horng Chau}.} \bibinfo{year}{2017}\natexlab{}.
\newblock \bibinfo{title}{Keeping the bad guys out: Protecting and vaccinating
  deep learning with JPEG compression}.
\newblock \bibinfo{howpublished}{arXiv:1705.02900}.
\newblock


\bibitem[\protect\citeauthoryear{Demontis, Melis, Pintor, Jagielski, Biggio,
  Oprea, Nita-Rotaru, and Roli}{Demontis et~al\mbox{.}}{2019}]%
        {demontis2019adversarial}
\bibfield{author}{\bibinfo{person}{Ambra Demontis}, \bibinfo{person}{Marco
  Melis}, \bibinfo{person}{Maura Pintor}, \bibinfo{person}{Matthew Jagielski},
  \bibinfo{person}{Battista Biggio}, \bibinfo{person}{Alina Oprea},
  \bibinfo{person}{Cristina Nita-Rotaru}, {and} \bibinfo{person}{Fabio Roli}.}
  \bibinfo{year}{2019}\natexlab{}.
\newblock \showarticletitle{Why do adversarial attacks transfer? explaining
  transferability of evasion and poisoning attacks}. In
  \bibinfo{booktitle}{\emph{USENIX Security}}.
\newblock


\bibitem[\protect\citeauthoryear{Demontis, Russu, Biggio, Fumera, and
  Roli}{Demontis et~al\mbox{.}}{2016}]%
        {demontis2016security}
\bibfield{author}{\bibinfo{person}{Ambra Demontis}, \bibinfo{person}{Paolo
  Russu}, \bibinfo{person}{Battista Biggio}, \bibinfo{person}{Giorgio Fumera},
  {and} \bibinfo{person}{Fabio Roli}.} \bibinfo{year}{2016}\natexlab{}.
\newblock \showarticletitle{On security and sparsity of linear classifiers for
  adversarial settings}. In \bibinfo{booktitle}{\emph{Int. Workshops on SPR and
  SSPR}}. Springer.
\newblock


\bibitem[\protect\citeauthoryear{Dhillon, Azizzadenesheli, Lipton, Bernstein,
  Kossaifi, Khanna, and Anandkumar}{Dhillon et~al\mbox{.}}{2018}]%
        {dhillon2018stochastic}
\bibfield{author}{\bibinfo{person}{Guneet~S. Dhillon}, \bibinfo{person}{Kamyar
  Azizzadenesheli}, \bibinfo{person}{Zachary~C. Lipton},
  \bibinfo{person}{Jeremy Bernstein}, \bibinfo{person}{Jean Kossaifi},
  \bibinfo{person}{Aran Khanna}, {and} \bibinfo{person}{Anima Anandkumar}.}
  \bibinfo{year}{2018}\natexlab{}.
\newblock \bibinfo{title}{Stochastic activation pruning for robust adversarial
  defense}.
\newblock \bibinfo{howpublished}{arXiv:1803.01442}.
\newblock


\bibitem[\protect\citeauthoryear{Dong, Liao, Pang, Su, Zhu, Hu, and Li}{Dong
  et~al\mbox{.}}{2018}]%
        {dong2017boosting}
\bibfield{author}{\bibinfo{person}{Yinpeng Dong}, \bibinfo{person}{Fangzhou
  Liao}, \bibinfo{person}{Tianyu Pang}, \bibinfo{person}{Hang Su},
  \bibinfo{person}{Jun Zhu}, \bibinfo{person}{Xiaolin Hu}, {and}
  \bibinfo{person}{Jianguo Li}.} \bibinfo{year}{2018}\natexlab{}.
\newblock \showarticletitle{Boosting adversarial attacks with momentum}. In
  \bibinfo{booktitle}{\emph{CVPR}}. IEEE.
\newblock


\bibitem[\protect\citeauthoryear{Dumont, Maggio, and Montalvo}{Dumont
  et~al\mbox{.}}{2018}]%
        {dumont2018robustness}
\bibfield{author}{\bibinfo{person}{Beranger Dumont}, \bibinfo{person}{Simona
  Maggio}, {and} \bibinfo{person}{Pablo Montalvo}.}
  \bibinfo{year}{2018}\natexlab{}.
\newblock \bibinfo{title}{Robustness of Rotation-Equivariant Networks to
  Adversarial Perturbations}.
\newblock \bibinfo{howpublished}{arXiv:1802.06627}.
\newblock


\bibitem[\protect\citeauthoryear{Dvijotham, Stanforth, Gowal, Mann, and
  Kohli}{Dvijotham et~al\mbox{.}}{2018}]%
        {dvijotham2018dual}
\bibfield{author}{\bibinfo{person}{Krishnamurthy Dvijotham},
  \bibinfo{person}{Robert Stanforth}, \bibinfo{person}{Sven Gowal},
  \bibinfo{person}{Timothy Mann}, {and} \bibinfo{person}{Pushmeet Kohli}.}
  \bibinfo{year}{2018}\natexlab{}.
\newblock \showarticletitle{A dual approach to scalable verification of deep
  networks}.
\newblock \bibinfo{journal}{\emph{UAI}} (\bibinfo{year}{2018}).
\newblock


\bibitem[\protect\citeauthoryear{Dziugaite, Ghahramani, and Roy}{Dziugaite
  et~al\mbox{.}}{2016}]%
        {dziugaite2016study}
\bibfield{author}{\bibinfo{person}{Gintare~Karolina Dziugaite},
  \bibinfo{person}{Zoubin Ghahramani}, {and} \bibinfo{person}{Daniel~M. Roy}.}
  \bibinfo{year}{2016}\natexlab{}.
\newblock \bibinfo{title}{A study of the effect of jpg compression on
  adversarial images}.
\newblock \bibinfo{howpublished}{arXiv:1608.00853}.
\newblock


\bibitem[\protect\citeauthoryear{Ehlers}{Ehlers}{2017}]%
        {ehlers2017formal}
\bibfield{author}{\bibinfo{person}{Ruediger Ehlers}.}
  \bibinfo{year}{2017}\natexlab{}.
\newblock \showarticletitle{Formal verification of piece-wise linear
  feed-forward neural networks}.
\newblock \bibinfo{journal}{\emph{Int. Symposium on Automated Technology for
  Verification and Analysis}} (\bibinfo{year}{2017}).
\newblock


\bibitem[\protect\citeauthoryear{Elsayed, Shankar, Cheung, Papernot, Kurakin,
  Goodfellow, and Sohl-Dickstein}{Elsayed et~al\mbox{.}}{2018}]%
        {elsayed2018adversarial}
\bibfield{author}{\bibinfo{person}{Gamaleldin Elsayed}, \bibinfo{person}{Shreya
  Shankar}, \bibinfo{person}{Brian Cheung}, \bibinfo{person}{Nicolas Papernot},
  \bibinfo{person}{Alexey Kurakin}, \bibinfo{person}{Ian Goodfellow}, {and}
  \bibinfo{person}{Jascha Sohl-Dickstein}.} \bibinfo{year}{2018}\natexlab{}.
\newblock \showarticletitle{Adversarial examples that fool both computer vision
  and time-limited humans}.
\newblock \bibinfo{journal}{\emph{NeurIPS}} (\bibinfo{year}{2018}).
\newblock


\bibitem[\protect\citeauthoryear{Engstrom, Tsipras, Schmidt, and
  Madry}{Engstrom et~al\mbox{.}}{2017}]%
        {engstrom2017rotation}
\bibfield{author}{\bibinfo{person}{Logan Engstrom}, \bibinfo{person}{Dimitris
  Tsipras}, \bibinfo{person}{Ludwig Schmidt}, {and} \bibinfo{person}{Aleksander
  Madry}.} \bibinfo{year}{2017}\natexlab{}.
\newblock \bibinfo{title}{A rotation and a translation suffice: Fooling {CNNs}
  with simple transformations}.
\newblock \bibinfo{howpublished}{arXiv:1712.02779}.
\newblock


\bibitem[\protect\citeauthoryear{Evtimov, Eykholt, Fernandes, Kohno, Li,
  Prakash, Rahmati, and Song}{Evtimov et~al\mbox{.}}{2017}]%
        {evtimov2017robust}
\bibfield{author}{\bibinfo{person}{Ivan Evtimov}, \bibinfo{person}{Kevin
  Eykholt}, \bibinfo{person}{Earlence Fernandes}, \bibinfo{person}{Tadayoshi
  Kohno}, \bibinfo{person}{Bo Li}, \bibinfo{person}{Atul Prakash},
  \bibinfo{person}{Amir Rahmati}, {and} \bibinfo{person}{Dawn Song}.}
  \bibinfo{year}{2017}\natexlab{}.
\newblock \bibinfo{title}{Robust physical-world attacks on deep learning
  models}.
\newblock \bibinfo{howpublished}{arXiv:1707.08945}.
\newblock


\bibitem[\protect\citeauthoryear{Eykholt, Evtimov, Fernandes, Li, Rahmati,
  Xiao, Prakash, Kohno, and Song}{Eykholt et~al\mbox{.}}{2018}]%
        {eykholt2018robust}
\bibfield{author}{\bibinfo{person}{Kevin Eykholt}, \bibinfo{person}{Ivan
  Evtimov}, \bibinfo{person}{Earlence Fernandes}, \bibinfo{person}{Bo Li},
  \bibinfo{person}{Amir Rahmati}, \bibinfo{person}{Chaowei Xiao},
  \bibinfo{person}{Atul Prakash}, \bibinfo{person}{Tadayoshi Kohno}, {and}
  \bibinfo{person}{Dawn Song}.} \bibinfo{year}{2018}\natexlab{}.
\newblock \showarticletitle{Robust Physical-World Attacks on Deep Learning
  Visual Classification}. In \bibinfo{booktitle}{\emph{CVPR}}. IEEE.
\newblock


\bibitem[\protect\citeauthoryear{Fawzi, Fawzi, and Fawzi}{Fawzi
  et~al\mbox{.}}{2018a}]%
        {fawzi2018adversarial}
\bibfield{author}{\bibinfo{person}{Alhussein Fawzi}, \bibinfo{person}{Hamza
  Fawzi}, {and} \bibinfo{person}{Omar Fawzi}.}
  \bibinfo{year}{2018}\natexlab{a}.
\newblock \bibinfo{title}{Adversarial vulnerability for any classifier}.
\newblock \bibinfo{howpublished}{arXiv:1802.08686}.
\newblock


\bibitem[\protect\citeauthoryear{Fawzi, Fawzi, and Frossard}{Fawzi
  et~al\mbox{.}}{2015}]%
        {fawzi2015fundamental}
\bibfield{author}{\bibinfo{person}{Alhussein Fawzi}, \bibinfo{person}{Omar
  Fawzi}, {and} \bibinfo{person}{Pascal Frossard}.}
  \bibinfo{year}{2015}\natexlab{}.
\newblock \showarticletitle{Fundamental limits on adversarial robustness}. In
  \bibinfo{booktitle}{\emph{ICML Workshop}}.
\newblock


\bibitem[\protect\citeauthoryear{Fawzi, Fawzi, and Frossard}{Fawzi
  et~al\mbox{.}}{2018b}]%
        {fawzi2018analysis}
\bibfield{author}{\bibinfo{person}{Alhussein Fawzi}, \bibinfo{person}{Omar
  Fawzi}, {and} \bibinfo{person}{Pascal Frossard}.}
  \bibinfo{year}{2018}\natexlab{b}.
\newblock \showarticletitle{Analysis of classifiers' robustness to adversarial
  perturbations}.
\newblock \bibinfo{journal}{\emph{Machine Learning}} (\bibinfo{year}{2018}).
\newblock


\bibitem[\protect\citeauthoryear{Fawzi, Moosavi-Dezfooli, and Frossard}{Fawzi
  et~al\mbox{.}}{2016}]%
        {fawzi2016robustness}
\bibfield{author}{\bibinfo{person}{Alhussein Fawzi},
  \bibinfo{person}{Seyed-Mohsen Moosavi-Dezfooli}, {and}
  \bibinfo{person}{Pascal Frossard}.} \bibinfo{year}{2016}\natexlab{}.
\newblock \showarticletitle{Robustness of classifiers: from adversarial to
  random noise}.
\newblock \bibinfo{journal}{\emph{NeurIPS}} (\bibinfo{year}{2016}).
\newblock


\bibitem[\protect\citeauthoryear{Feinman, Curtin, Shintre, and Gardner}{Feinman
  et~al\mbox{.}}{2017}]%
        {feinman2017detecting}
\bibfield{author}{\bibinfo{person}{Reuben Feinman}, \bibinfo{person}{Ryan~R.
  Curtin}, \bibinfo{person}{Saurabh Shintre}, {and} \bibinfo{person}{Andrew~B.
  Gardner}.} \bibinfo{year}{2017}\natexlab{}.
\newblock \bibinfo{title}{Detecting adversarial samples from artifacts}.
\newblock \bibinfo{howpublished}{arXiv:1703.00410}.
\newblock


\bibitem[\protect\citeauthoryear{Gal and Ghahramani}{Gal and
  Ghahramani}{2016}]%
        {gal2016dropout}
\bibfield{author}{\bibinfo{person}{Yarin Gal} {and} \bibinfo{person}{Zoubin
  Ghahramani}.} \bibinfo{year}{2016}\natexlab{}.
\newblock \showarticletitle{Dropout as a bayesian approximation: Representing
  model uncertainty in deep learning}.
\newblock \bibinfo{journal}{\emph{ICML}} (\bibinfo{year}{2016}).
\newblock


\bibitem[\protect\citeauthoryear{Galloway, Taylor, and Moussa}{Galloway
  et~al\mbox{.}}{2017}]%
        {galloway2017attacking}
\bibfield{author}{\bibinfo{person}{Angus Galloway}, \bibinfo{person}{Graham~W.
  Taylor}, {and} \bibinfo{person}{Medhat Moussa}.}
  \bibinfo{year}{2017}\natexlab{}.
\newblock \bibinfo{title}{Attacking binarized neural networks}.
\newblock \bibinfo{howpublished}{arXiv:1711.00449}.
\newblock


\bibitem[\protect\citeauthoryear{Gao, Wang, Lin, Xu, and Qi}{Gao
  et~al\mbox{.}}{2017}]%
        {gao2017deepcloak}
\bibfield{author}{\bibinfo{person}{Ji Gao}, \bibinfo{person}{Beilun Wang},
  \bibinfo{person}{Zeming Lin}, \bibinfo{person}{Weilin Xu}, {and}
  \bibinfo{person}{Yanjun Qi}.} \bibinfo{year}{2017}\natexlab{}.
\newblock \showarticletitle{Deepcloak: Masking deep neural network models for
  robustness against adversarial samples}.
\newblock \bibinfo{journal}{\emph{ICLR}} (\bibinfo{year}{2017}).
\newblock


\bibitem[\protect\citeauthoryear{Gehr, Mirman, Drachsler-Cohen, Tsankov,
  Chaudhuri, and Vechev}{Gehr et~al\mbox{.}}{2018}]%
        {gehr2018ai}
\bibfield{author}{\bibinfo{person}{Timon Gehr}, \bibinfo{person}{Matthew
  Mirman}, \bibinfo{person}{Dana Drachsler-Cohen}, \bibinfo{person}{Petar
  Tsankov}, \bibinfo{person}{Swarat Chaudhuri}, {and} \bibinfo{person}{Martin
  Vechev}.} \bibinfo{year}{2018}\natexlab{}.
\newblock \showarticletitle{Ai 2: Safety and robustness certification of neural
  networks with abstract interpretation}. In \bibinfo{booktitle}{\emph{S\&P}}.
  IEEE.
\newblock


\bibitem[\protect\citeauthoryear{Ghosh, Losalka, and Black}{Ghosh
  et~al\mbox{.}}{2018}]%
        {ghosh2018resisting}
\bibfield{author}{\bibinfo{person}{Partha Ghosh}, \bibinfo{person}{Arpan
  Losalka}, {and} \bibinfo{person}{Michael~J. Black}.}
  \bibinfo{year}{2018}\natexlab{}.
\newblock \bibinfo{title}{Resisting Adversarial Attacks using Gaussian Mixture
  Variational Autoencoders}.
\newblock \bibinfo{howpublished}{arXiv:1806.00081}.
\newblock


\bibitem[\protect\citeauthoryear{Gilmer, Adams, Goodfellow, Andersen, and
  Dahl}{Gilmer et~al\mbox{.}}{2018a}]%
        {gilmer2018motivating}
\bibfield{author}{\bibinfo{person}{Justin Gilmer}, \bibinfo{person}{Ryan~P.
  Adams}, \bibinfo{person}{Ian Goodfellow}, \bibinfo{person}{David Andersen},
  {and} \bibinfo{person}{George~E. Dahl}.} \bibinfo{year}{2018}\natexlab{a}.
\newblock \bibinfo{title}{Motivating the Rules of the Game for Adversarial
  Example Research}.
\newblock \bibinfo{howpublished}{arXiv:1807.06732}.
\newblock


\bibitem[\protect\citeauthoryear{Gilmer, Metz, Faghri, Schoenholz, Raghu,
  Wattenberg, and Goodfellow}{Gilmer et~al\mbox{.}}{2018b}]%
        {gilmer2018adversarial}
\bibfield{author}{\bibinfo{person}{Justin Gilmer}, \bibinfo{person}{Luke Metz},
  \bibinfo{person}{Fartash Faghri}, \bibinfo{person}{Samuel~S. Schoenholz},
  \bibinfo{person}{Maithra Raghu}, \bibinfo{person}{Martin Wattenberg}, {and}
  \bibinfo{person}{Ian Goodfellow}.} \bibinfo{year}{2018}\natexlab{b}.
\newblock \bibinfo{title}{Adversarial spheres}.
\newblock \bibinfo{howpublished}{arXiv:1801.02774}.
\newblock


\bibitem[\protect\citeauthoryear{Girosi, Jones, and Poggio}{Girosi
  et~al\mbox{.}}{1995}]%
        {girosi1995regularization}
\bibfield{author}{\bibinfo{person}{Federico Girosi}, \bibinfo{person}{Michael
  Jones}, {and} \bibinfo{person}{Tomaso Poggio}.}
  \bibinfo{year}{1995}\natexlab{}.
\newblock \showarticletitle{Regularization theory and neural networks
  architectures}.
\newblock \bibinfo{journal}{\emph{Neural computation}} (\bibinfo{year}{1995}).
\newblock


\bibitem[\protect\citeauthoryear{Globerson and Roweis}{Globerson and
  Roweis}{2006}]%
        {globerson2006nightmare}
\bibfield{author}{\bibinfo{person}{Amir Globerson} {and} \bibinfo{person}{Sam
  Roweis}.} \bibinfo{year}{2006}\natexlab{}.
\newblock \showarticletitle{Nightmare at test time: robust learning by feature
  deletion}.
\newblock \bibinfo{journal}{\emph{ICML}} (\bibinfo{year}{2006}).
\newblock


\bibitem[\protect\citeauthoryear{Gong, Wang, and Ku}{Gong
  et~al\mbox{.}}{2017}]%
        {gong2017adversarial}
\bibfield{author}{\bibinfo{person}{Zhitao Gong}, \bibinfo{person}{Wenlu Wang},
  {and} \bibinfo{person}{Wei-Shinn Ku}.} \bibinfo{year}{2017}\natexlab{}.
\newblock \bibinfo{title}{Adversarial and clean data are not twins}.
\newblock \bibinfo{howpublished}{arXiv:1704.04960}.
\newblock


\bibitem[\protect\citeauthoryear{Goodfellow}{Goodfellow}{2018}]%
        {goodfellow2018gradient}
\bibfield{author}{\bibinfo{person}{Ian Goodfellow}.}
  \bibinfo{year}{2018}\natexlab{}.
\newblock \bibinfo{title}{Gradient Masking Causes CLEVER to Overestimate
  Adversarial Perturbation Size}.
\newblock \bibinfo{howpublished}{arXiv:1804.07870}.
\newblock


\bibitem[\protect\citeauthoryear{Goodfellow, Pouget-Abadie, Mirza, Xu,
  Warde-Farley, Ozair, Courville, and Bengio}{Goodfellow et~al\mbox{.}}{2014}]%
        {goodfellow2014generative}
\bibfield{author}{\bibinfo{person}{Ian Goodfellow}, \bibinfo{person}{Jean
  Pouget-Abadie}, \bibinfo{person}{Mehdi Mirza}, \bibinfo{person}{Bing Xu},
  \bibinfo{person}{David Warde-Farley}, \bibinfo{person}{Sherjil Ozair},
  \bibinfo{person}{Aaron Courville}, {and} \bibinfo{person}{Yoshua Bengio}.}
  \bibinfo{year}{2014}\natexlab{}.
\newblock \showarticletitle{Generative adversarial nets}.
\newblock \bibinfo{journal}{\emph{NeurIPS}} (\bibinfo{year}{2014}).
\newblock


\bibitem[\protect\citeauthoryear{Goodfellow, Shlens, and Szegedy}{Goodfellow
  et~al\mbox{.}}{2015}]%
        {goodfellow2014explaining}
\bibfield{author}{\bibinfo{person}{Ian Goodfellow}, \bibinfo{person}{Jonathon
  Shlens}, {and} \bibinfo{person}{Christian Szegedy}.}
  \bibinfo{year}{2015}\natexlab{}.
\newblock \showarticletitle{Explaining and Harnessing Adversarial Examples}.
\newblock \bibinfo{journal}{\emph{ICLR}} (\bibinfo{year}{2015}).
\newblock


\bibitem[\protect\citeauthoryear{Gowal, Dvijotham, Stanforth, Bunel, Qin,
  Uesato, Mann, and Kohli}{Gowal et~al\mbox{.}}{2018}]%
        {gowal2018effectiveness}
\bibfield{author}{\bibinfo{person}{Sven Gowal}, \bibinfo{person}{Krishnamurthy
  Dvijotham}, \bibinfo{person}{Robert Stanforth}, \bibinfo{person}{Rudy Bunel},
  \bibinfo{person}{Chongli Qin}, \bibinfo{person}{Jonathan Uesato},
  \bibinfo{person}{Timothy Mann}, {and} \bibinfo{person}{Pushmeet Kohli}.}
  \bibinfo{year}{2018}\natexlab{}.
\newblock \bibinfo{title}{On the effectiveness of interval bound propagation
  for training verifiably robust models}.
\newblock \bibinfo{howpublished}{arXiv:1810.12715}.
\newblock


\bibitem[\protect\citeauthoryear{Grosse, Manoharan, Papernot, Backes, and
  McDaniel}{Grosse et~al\mbox{.}}{2017}]%
        {grosse2017statistical}
\bibfield{author}{\bibinfo{person}{Kathrin Grosse}, \bibinfo{person}{Praveen
  Manoharan}, \bibinfo{person}{Nicolas Papernot}, \bibinfo{person}{Michael
  Backes}, {and} \bibinfo{person}{Patrick McDaniel}.}
  \bibinfo{year}{2017}\natexlab{}.
\newblock \bibinfo{title}{On the (statistical) detection of adversarial
  examples}.
\newblock \bibinfo{howpublished}{arXiv:1702.06280}.
\newblock


\bibitem[\protect\citeauthoryear{Grosse, Papernot, Manoharan, Backes, and
  McDaniel}{Grosse et~al\mbox{.}}{2016}]%
        {grosse2016adversarial}
\bibfield{author}{\bibinfo{person}{Kathrin Grosse}, \bibinfo{person}{Nicolas
  Papernot}, \bibinfo{person}{Praveen Manoharan}, \bibinfo{person}{Michael
  Backes}, {and} \bibinfo{person}{Patrick McDaniel}.}
  \bibinfo{year}{2016}\natexlab{}.
\newblock \bibinfo{title}{Adversarial perturbations against deep neural
  networks for malware classification}.
\newblock \bibinfo{howpublished}{arXiv:1606.04435}.
\newblock


\bibitem[\protect\citeauthoryear{Gu and Rigazio}{Gu and Rigazio}{2014}]%
        {gu2014towards}
\bibfield{author}{\bibinfo{person}{Shixiang Gu} {and} \bibinfo{person}{Luca
  Rigazio}.} \bibinfo{year}{2014}\natexlab{}.
\newblock \bibinfo{title}{Towards deep neural network architectures robust to
  adversarial examples}.
\newblock \bibinfo{howpublished}{arXiv:1412.5068}.
\newblock


\bibitem[\protect\citeauthoryear{Guo, Rana, Cisse, and van~der Maaten}{Guo
  et~al\mbox{.}}{2018}]%
        {guo2017countering}
\bibfield{author}{\bibinfo{person}{Chuan Guo}, \bibinfo{person}{Mayank Rana},
  \bibinfo{person}{Moustapha Cisse}, {and} \bibinfo{person}{Laurens van~der
  Maaten}.} \bibinfo{year}{2018}\natexlab{}.
\newblock \showarticletitle{Countering adversarial images using input
  transformations}.
\newblock \bibinfo{journal}{\emph{ICLR}} (\bibinfo{year}{2018}).
\newblock


\bibitem[\protect\citeauthoryear{Ha, Dai, and Le}{Ha et~al\mbox{.}}{2017}]%
        {ha2016hypernetworks}
\bibfield{author}{\bibinfo{person}{David Ha}, \bibinfo{person}{Andrew Dai},
  {and} \bibinfo{person}{Quoc~V Le}.} \bibinfo{year}{2017}\natexlab{}.
\newblock \showarticletitle{Hypernetworks}.
\newblock \bibinfo{journal}{\emph{ICLR}} (\bibinfo{year}{2017}).
\newblock


\bibitem[\protect\citeauthoryear{He, Zhang, Ren, and Sun}{He
  et~al\mbox{.}}{2016}]%
        {he2016deep}
\bibfield{author}{\bibinfo{person}{Kaiming He}, \bibinfo{person}{Xiangyu
  Zhang}, \bibinfo{person}{Shaoqing Ren}, {and} \bibinfo{person}{Jian Sun}.}
  \bibinfo{year}{2016}\natexlab{}.
\newblock \showarticletitle{Deep residual learning for image recognition}. In
  \bibinfo{booktitle}{\emph{CVPR}}. IEEE.
\newblock


\bibitem[\protect\citeauthoryear{He, Wei, Chen, Carlini, and Song}{He
  et~al\mbox{.}}{2017}]%
        {he2017adversarial}
\bibfield{author}{\bibinfo{person}{Warren He}, \bibinfo{person}{James Wei},
  \bibinfo{person}{Xinyun Chen}, \bibinfo{person}{Nicholas Carlini}, {and}
  \bibinfo{person}{Dawn Song}.} \bibinfo{year}{2017}\natexlab{}.
\newblock \showarticletitle{Adversarial example defense: Ensembles of weak
  defenses are not strong}. In \bibinfo{booktitle}{\emph{USENIX WOOT}}.
\newblock


\bibitem[\protect\citeauthoryear{Hein and Andriushchenko}{Hein and
  Andriushchenko}{2017}]%
        {hein2017formal}
\bibfield{author}{\bibinfo{person}{Matthias Hein} {and} \bibinfo{person}{Maksym
  Andriushchenko}.} \bibinfo{year}{2017}\natexlab{}.
\newblock \showarticletitle{Formal guarantees on the robustness of a classifier
  against adversarial manipulation}.
\newblock \bibinfo{journal}{\emph{NeurIPS}} (\bibinfo{year}{2017}).
\newblock


\bibitem[\protect\citeauthoryear{Hendrycks and Dietterich}{Hendrycks and
  Dietterich}{2019}]%
        {hendrycks2019benchmarking}
\bibfield{author}{\bibinfo{person}{Dan Hendrycks} {and} \bibinfo{person}{Thomas
  Dietterich}.} \bibinfo{year}{2019}\natexlab{}.
\newblock \showarticletitle{Benchmarking neural network robustness to common
  corruptions and perturbations}.
\newblock \bibinfo{journal}{\emph{ICLR}} (\bibinfo{year}{2019}).
\newblock


\bibitem[\protect\citeauthoryear{Hendrycks and Gimpel}{Hendrycks and
  Gimpel}{2016}]%
        {hendrycks2016early}
\bibfield{author}{\bibinfo{person}{Dan Hendrycks} {and} \bibinfo{person}{Kevin
  Gimpel}.} \bibinfo{year}{2016}\natexlab{}.
\newblock \showarticletitle{Early methods for detecting adversarial images}.
\newblock \bibinfo{journal}{\emph{ICLR Workshop}} (\bibinfo{year}{2016}).
\newblock


\bibitem[\protect\citeauthoryear{Hinton, Vinyals, and Dean}{Hinton
  et~al\mbox{.}}{2015}]%
        {hinton2015distilling}
\bibfield{author}{\bibinfo{person}{Geoffrey Hinton}, \bibinfo{person}{Oriol
  Vinyals}, {and} \bibinfo{person}{Jeff Dean}.}
  \bibinfo{year}{2015}\natexlab{}.
\newblock \bibinfo{title}{Distilling the knowledge in a neural network}.
\newblock \bibinfo{howpublished}{arXiv:1503.02531}.
\newblock


\bibitem[\protect\citeauthoryear{Hu and Tan}{Hu and Tan}{2017}]%
        {hu2017generating}
\bibfield{author}{\bibinfo{person}{Weiwei Hu} {and} \bibinfo{person}{Ying
  Tan}.} \bibinfo{year}{2017}\natexlab{}.
\newblock \bibinfo{title}{Generating adversarial malware examples for black-box
  attacks based on GAN}.
\newblock \bibinfo{howpublished}{arXiv:1702.05983}.
\newblock


\bibitem[\protect\citeauthoryear{Huang, Xu, Schuurmans, and
  Szepesv{\'a}ri}{Huang et~al\mbox{.}}{2015}]%
        {huang2015learning}
\bibfield{author}{\bibinfo{person}{Ruitong Huang}, \bibinfo{person}{Bing Xu},
  \bibinfo{person}{Dale Schuurmans}, {and} \bibinfo{person}{Csaba
  Szepesv{\'a}ri}.} \bibinfo{year}{2015}\natexlab{}.
\newblock \bibinfo{title}{Learning with a strong adversary}.
\newblock \bibinfo{howpublished}{arXiv:1511.03034}.
\newblock


\bibitem[\protect\citeauthoryear{Huang, Papernot, Goodfellow, Duan, and
  Abbeel}{Huang et~al\mbox{.}}{2017b}]%
        {huang2017adversarial}
\bibfield{author}{\bibinfo{person}{Sandy Huang}, \bibinfo{person}{Nicolas
  Papernot}, \bibinfo{person}{Ian Goodfellow}, \bibinfo{person}{Yan Duan},
  {and} \bibinfo{person}{Pieter Abbeel}.} \bibinfo{year}{2017}\natexlab{b}.
\newblock \showarticletitle{Adversarial attacks on neural network policies}.
\newblock \bibinfo{journal}{\emph{ICLR Workshop}} (\bibinfo{year}{2017}).
\newblock


\bibitem[\protect\citeauthoryear{Huang, Kroening, Kwiatkowska, Ruan, Sun,
  Thamo, Wu, and Yi}{Huang et~al\mbox{.}}{2018}]%
        {huang2018safety}
\bibfield{author}{\bibinfo{person}{Xiaowei Huang}, \bibinfo{person}{Daniel
  Kroening}, \bibinfo{person}{Marta Kwiatkowska}, \bibinfo{person}{Wenjie
  Ruan}, \bibinfo{person}{Youcheng Sun}, \bibinfo{person}{Emese Thamo},
  \bibinfo{person}{Min Wu}, {and} \bibinfo{person}{Xinping Yi}.}
  \bibinfo{year}{2018}\natexlab{}.
\newblock \bibinfo{title}{Safety and Trustworthiness of Deep Neural Networks: A
  Survey}.
\newblock \bibinfo{howpublished}{arXiv:1812.08342}.
\newblock


\bibitem[\protect\citeauthoryear{Huang, Kwiatkowska, Wang, and Wu}{Huang
  et~al\mbox{.}}{2017a}]%
        {huang2017safety}
\bibfield{author}{\bibinfo{person}{Xiaowei Huang}, \bibinfo{person}{Marta
  Kwiatkowska}, \bibinfo{person}{Sen Wang}, {and} \bibinfo{person}{Min Wu}.}
  \bibinfo{year}{2017}\natexlab{a}.
\newblock \showarticletitle{Safety verification of deep neural networks}. In
  \bibinfo{booktitle}{\emph{CAV}}. Springer.
\newblock


\bibitem[\protect\citeauthoryear{Huber}{Huber}{2011}]%
        {huber2011robust}
\bibfield{author}{\bibinfo{person}{Peter~J. Huber}.}
  \bibinfo{year}{2011}\natexlab{}.
\newblock \showarticletitle{Robust statistics}.
\newblock In \bibinfo{booktitle}{\emph{Int. Encyclopedia of Statistical
  Science}}. \bibinfo{publisher}{Springer}.
\newblock


\bibitem[\protect\citeauthoryear{Huster, Chiang, and Chadha}{Huster
  et~al\mbox{.}}{2018}]%
        {huster2018limitations}
\bibfield{author}{\bibinfo{person}{Todd Huster}, \bibinfo{person}{Cho-Yu~Jason
  Chiang}, {and} \bibinfo{person}{Ritu Chadha}.}
  \bibinfo{year}{2018}\natexlab{}.
\newblock \showarticletitle{Limitations of the Lipschitz constant as a defense
  against adversarial examples}. In \bibinfo{booktitle}{\emph{ECML PKDD}}.
  Springer.
\newblock


\bibitem[\protect\citeauthoryear{Ilyas, Engstrom, Athalye, and Lin}{Ilyas
  et~al\mbox{.}}{2018}]%
        {ilyas2018black}
\bibfield{author}{\bibinfo{person}{Andrew Ilyas}, \bibinfo{person}{Logan
  Engstrom}, \bibinfo{person}{Anish Athalye}, {and} \bibinfo{person}{Jessy
  Lin}.} \bibinfo{year}{2018}\natexlab{}.
\newblock \showarticletitle{Black-box Adversarial Attacks with Limited Queries
  and Information}.
\newblock \bibinfo{journal}{\emph{ICML}} (\bibinfo{year}{2018}).
\newblock


\bibitem[\protect\citeauthoryear{Ilyas, Santurkar, Tsipras, Engstrom, Tran, and
  Madry}{Ilyas et~al\mbox{.}}{2019}]%
        {ilyas2019adversarial}
\bibfield{author}{\bibinfo{person}{Andrew Ilyas}, \bibinfo{person}{Shibani
  Santurkar}, \bibinfo{person}{Dimitris Tsipras}, \bibinfo{person}{Logan
  Engstrom}, \bibinfo{person}{Brandon Tran}, {and} \bibinfo{person}{Aleksander
  Madry}.} \bibinfo{year}{2019}\natexlab{}.
\newblock \showarticletitle{Adversarial Examples Are Not Bugs, They Are
  Features}.
\newblock \bibinfo{journal}{\emph{arXiv:1905.02175}} (\bibinfo{year}{2019}).
\newblock


\bibitem[\protect\citeauthoryear{Izmailov, Sugrim, Chadha, McDaniel, and
  Swami}{Izmailov et~al\mbox{.}}{2018}]%
        {izmailov2018enablers}
\bibfield{author}{\bibinfo{person}{Rauf Izmailov}, \bibinfo{person}{Shridatt
  Sugrim}, \bibinfo{person}{Ritu Chadha}, \bibinfo{person}{Patrick McDaniel},
  {and} \bibinfo{person}{Ananthram Swami}.} \bibinfo{year}{2018}\natexlab{}.
\newblock \showarticletitle{Enablers of Adversarial Attacks in Machine
  Learning}. In \bibinfo{booktitle}{\emph{MILCOM}}. IEEE.
\newblock


\bibitem[\protect\citeauthoryear{Jo and Bengio}{Jo and Bengio}{2017}]%
        {jo2017measuring}
\bibfield{author}{\bibinfo{person}{Jason Jo} {and} \bibinfo{person}{Yoshua
  Bengio}.} \bibinfo{year}{2017}\natexlab{}.
\newblock \bibinfo{title}{Measuring the tendency of CNNs to learn surface
  statistical regularities}.
\newblock \bibinfo{howpublished}{arXiv:1711.11561}.
\newblock


\bibitem[\protect\citeauthoryear{Kanbak, Moosavi-Dezfooli, and Frossard}{Kanbak
  et~al\mbox{.}}{2018}]%
        {kanbak2017geometric}
\bibfield{author}{\bibinfo{person}{Can Kanbak}, \bibinfo{person}{Seyed-Mohsen
  Moosavi-Dezfooli}, {and} \bibinfo{person}{Pascal Frossard}.}
  \bibinfo{year}{2018}\natexlab{}.
\newblock \showarticletitle{Geometric robustness of deep networks: analysis and
  improvement}. In \bibinfo{booktitle}{\emph{CVPR}}. IEEE.
\newblock


\bibitem[\protect\citeauthoryear{Katz, Barrett, Dill, Julian, and
  Kochenderfer}{Katz et~al\mbox{.}}{2017}]%
        {katz2017reluplex}
\bibfield{author}{\bibinfo{person}{Guy Katz}, \bibinfo{person}{Clark Barrett},
  \bibinfo{person}{David~L. Dill}, \bibinfo{person}{Kyle Julian}, {and}
  \bibinfo{person}{Mykel~J. Kochenderfer}.} \bibinfo{year}{2017}\natexlab{}.
\newblock \showarticletitle{Reluplex: An efficient SMT solver for verifying
  deep neural networks}. In \bibinfo{booktitle}{\emph{CAV}}. Springer.
\newblock


\bibitem[\protect\citeauthoryear{Khrulkov and Oseledets}{Khrulkov and
  Oseledets}{2018}]%
        {khrulkov2017art}
\bibfield{author}{\bibinfo{person}{Valentin Khrulkov} {and}
  \bibinfo{person}{Ivan Oseledets}.} \bibinfo{year}{2018}\natexlab{}.
\newblock \showarticletitle{Art of singular vectors and universal adversarial
  perturbations}. In \bibinfo{booktitle}{\emph{CVPR}}. IEEE.
\newblock


\bibitem[\protect\citeauthoryear{Kingma and Welling}{Kingma and
  Welling}{2013}]%
        {kingma2013variational}
\bibfield{author}{\bibinfo{person}{Diederik~P. Kingma} {and}
  \bibinfo{person}{Max Welling}.} \bibinfo{year}{2013}\natexlab{}.
\newblock \bibinfo{title}{Auto-encoding variational Bayes.}
\newblock \bibinfo{howpublished}{arXiv:1312.6114}.
\newblock


\bibitem[\protect\citeauthoryear{Ko{\l}cz and Teo}{Ko{\l}cz and Teo}{2009}]%
        {kolcz2009feature}
\bibfield{author}{\bibinfo{person}{Aleksander Ko{\l}cz} {and}
  \bibinfo{person}{Choon~Hui Teo}.} \bibinfo{year}{2009}\natexlab{}.
\newblock \showarticletitle{Feature weighting for improved classifier
  robustness}. In \bibinfo{booktitle}{\emph{CEAS}}.
\newblock


\bibitem[\protect\citeauthoryear{Kreuk, Barak, Aviv-Reuven, Baruch, Pinkas, and
  Keshet}{Kreuk et~al\mbox{.}}{2018}]%
        {kreuk2018adversarial}
\bibfield{author}{\bibinfo{person}{Felix Kreuk}, \bibinfo{person}{Assi Barak},
  \bibinfo{person}{Shir Aviv-Reuven}, \bibinfo{person}{Moran Baruch},
  \bibinfo{person}{Benny Pinkas}, {and} \bibinfo{person}{Joseph Keshet}.}
  \bibinfo{year}{2018}\natexlab{}.
\newblock \bibinfo{title}{Adversarial Examples on Discrete Sequences for
  Beating Whole-Binary Malware Detection}.
\newblock \bibinfo{howpublished}{arXiv:1802.04528}.
\newblock


\bibitem[\protect\citeauthoryear{Krotov and Hopfield}{Krotov and
  Hopfield}{2016}]%
        {krotov2016densemem}
\bibfield{author}{\bibinfo{person}{Dmitry Krotov} {and}
  \bibinfo{person}{John~J. Hopfield}.} \bibinfo{year}{2016}\natexlab{}.
\newblock \showarticletitle{Dense associative memory for pattern recognition}.
\newblock \bibinfo{journal}{\emph{NeurIPS}} (\bibinfo{year}{2016}).
\newblock


\bibitem[\protect\citeauthoryear{Krotov and Hopfield}{Krotov and
  Hopfield}{2017}]%
        {krotov2016dense}
\bibfield{author}{\bibinfo{person}{Dmitry Krotov} {and}
  \bibinfo{person}{John~J. Hopfield}.} \bibinfo{year}{2017}\natexlab{}.
\newblock \showarticletitle{Dense associative memory is robust to adversarial
  inputs}.
\newblock \bibinfo{journal}{\emph{Neural Computation}} (\bibinfo{year}{2017}).
\newblock


\bibitem[\protect\citeauthoryear{Kurakin, Goodfellow, and Bengio}{Kurakin
  et~al\mbox{.}}{2016a}]%
        {kurakin2016adversarial}
\bibfield{author}{\bibinfo{person}{Alexey Kurakin}, \bibinfo{person}{Ian
  Goodfellow}, {and} \bibinfo{person}{Samy Bengio}.}
  \bibinfo{year}{2016}\natexlab{a}.
\newblock \bibinfo{title}{Adversarial examples in the physical world}.
\newblock \bibinfo{howpublished}{arXiv:1607.02533}.
\newblock


\bibitem[\protect\citeauthoryear{Kurakin, Goodfellow, and Bengio}{Kurakin
  et~al\mbox{.}}{2016b}]%
        {kurakin2016aadversarial}
\bibfield{author}{\bibinfo{person}{Alexey Kurakin}, \bibinfo{person}{Ian
  Goodfellow}, {and} \bibinfo{person}{Samy Bengio}.}
  \bibinfo{year}{2016}\natexlab{b}.
\newblock \bibinfo{title}{Adversarial machine learning at scale}.
\newblock \bibinfo{howpublished}{arXiv:1611.01236}.
\newblock


\bibitem[\protect\citeauthoryear{Lamb, Binas, Goyal, Serdyuk, Subramanian,
  Mitliagkas, and Bengio}{Lamb et~al\mbox{.}}{2018}]%
        {lamb2018fortified}
\bibfield{author}{\bibinfo{person}{Alex Lamb}, \bibinfo{person}{Jonathan
  Binas}, \bibinfo{person}{Anirudh Goyal}, \bibinfo{person}{Dmitriy Serdyuk},
  \bibinfo{person}{Sandeep Subramanian}, \bibinfo{person}{Ioannis Mitliagkas},
  {and} \bibinfo{person}{Yoshua Bengio}.} \bibinfo{year}{2018}\natexlab{}.
\newblock \bibinfo{title}{Fortified Networks: Improving the Robustness of Deep
  Networks by Modeling the Manifold of Hidden Representations}.
\newblock \bibinfo{howpublished}{arXiv:1804.02485}.
\newblock


\bibitem[\protect\citeauthoryear{Larochelle, Bengio, Louradour, and
  Lamblin}{Larochelle et~al\mbox{.}}{2009}]%
        {larochelle2009exploring}
\bibfield{author}{\bibinfo{person}{Hugo Larochelle}, \bibinfo{person}{Yoshua
  Bengio}, \bibinfo{person}{J{\'e}r{\^o}me Louradour}, {and}
  \bibinfo{person}{Pascal Lamblin}.} \bibinfo{year}{2009}\natexlab{}.
\newblock \showarticletitle{Exploring strategies for training deep neural
  networks}.
\newblock \bibinfo{journal}{\emph{JMLR}} \bibinfo{volume}{10},
  \bibinfo{number}{Jan} (\bibinfo{year}{2009}).
\newblock


\bibitem[\protect\citeauthoryear{Laskov et~al\mbox{.}}{Laskov
  et~al\mbox{.}}{2014}]%
        {laskov2014practical}
\bibfield{author}{\bibinfo{person}{Pavel Laskov} {et~al\mbox{.}}}
  \bibinfo{year}{2014}\natexlab{}.
\newblock \showarticletitle{Practical evasion of a learning-based classifier: A
  case study}. In \bibinfo{booktitle}{\emph{S\&P}}. IEEE.
\newblock


\bibitem[\protect\citeauthoryear{Lee, Han, and Lee}{Lee et~al\mbox{.}}{2017}]%
        {lee2017generative}
\bibfield{author}{\bibinfo{person}{Hyeungill Lee}, \bibinfo{person}{Sungyeob
  Han}, {and} \bibinfo{person}{Jungwoo Lee}.} \bibinfo{year}{2017}\natexlab{}.
\newblock \bibinfo{title}{Generative Adversarial Trainer: Defense to
  Adversarial Perturbations with GAN}.
\newblock \bibinfo{howpublished}{arXiv:1705.03387}.
\newblock


\bibitem[\protect\citeauthoryear{Li, Neupane, Paul, Song, Krishnamurthy,
  Chowdhury, and Swami}{Li et~al\mbox{.}}{2018}]%
        {li2018adversarial}
\bibfield{author}{\bibinfo{person}{Shasha Li}, \bibinfo{person}{Ajaya Neupane},
  \bibinfo{person}{Sujoy Paul}, \bibinfo{person}{Chengyu Song},
  \bibinfo{person}{Srikanth~V. Krishnamurthy}, \bibinfo{person}{Amit K.~Roy
  Chowdhury}, {and} \bibinfo{person}{Ananthram Swami}.}
  \bibinfo{year}{2018}\natexlab{}.
\newblock \bibinfo{title}{Adversarial perturbations against real-time video
  classification systems}.
\newblock \bibinfo{howpublished}{arXiv:1807.00458}.
\newblock


\bibitem[\protect\citeauthoryear{Li and Li}{Li and Li}{2017}]%
        {li2017adversarial}
\bibfield{author}{\bibinfo{person}{Xin Li} {and} \bibinfo{person}{Fuxin Li}.}
  \bibinfo{year}{2017}\natexlab{}.
\newblock \showarticletitle{Adversarial Examples Detection in Deep Networks
  with Convolutional Filter Statistics.}. In \bibinfo{booktitle}{\emph{ICCV}}.
  IEEE.
\newblock


\bibitem[\protect\citeauthoryear{Liang, Li, Su, Li, Shi, and Wang}{Liang
  et~al\mbox{.}}{2017}]%
        {liang2017detecting}
\bibfield{author}{\bibinfo{person}{Bin Liang}, \bibinfo{person}{Hongcheng Li},
  \bibinfo{person}{Miaoqiang Su}, \bibinfo{person}{Xirong Li},
  \bibinfo{person}{Wenchang Shi}, {and} \bibinfo{person}{Xiaofeng Wang}.}
  \bibinfo{year}{2017}\natexlab{}.
\newblock \bibinfo{title}{Detecting Adversarial Examples in Deep Networks with
  Adaptive Noise Reduction}.
\newblock \bibinfo{howpublished}{arXiv:1705.08378}.
\newblock


\bibitem[\protect\citeauthoryear{Lin, Hong, Liao, Shih, Liu, and Sun}{Lin
  et~al\mbox{.}}{2017}]%
        {lin2017tactics}
\bibfield{author}{\bibinfo{person}{Yen-Chen Lin}, \bibinfo{person}{Zhang-Wei
  Hong}, \bibinfo{person}{Yuan-Hong Liao}, \bibinfo{person}{Meng-Li Shih},
  \bibinfo{person}{Ming-Yu Liu}, {and} \bibinfo{person}{Min Sun}.}
  \bibinfo{year}{2017}\natexlab{}.
\newblock \showarticletitle{Tactics of adversarial attack on deep reinforcement
  learning agents}.
\newblock \bibinfo{journal}{\emph{IJCAI}} (\bibinfo{year}{2017}).
\newblock


\bibitem[\protect\citeauthoryear{Liu, Tao, Li, Nowrouzezahrai, and
  Jacobson}{Liu et~al\mbox{.}}{2018b}]%
        {liu2018beyond}
\bibfield{author}{\bibinfo{person}{Hsueh-Ti~Derek Liu},
  \bibinfo{person}{Michael Tao}, \bibinfo{person}{Chun-Liang Li},
  \bibinfo{person}{Derek Nowrouzezahrai}, {and} \bibinfo{person}{Alec
  Jacobson}.} \bibinfo{year}{2018}\natexlab{b}.
\newblock \bibinfo{title}{Beyond Pixel Norm-Balls: Parametric Adversaries using
  an Analytically Differentiable Renderer}.
\newblock
\newblock


\bibitem[\protect\citeauthoryear{Liu, Li, Zhao, Cai, Yu, and Leung}{Liu
  et~al\mbox{.}}{2018a}]%
        {liu2018survey}
\bibfield{author}{\bibinfo{person}{Qiang Liu}, \bibinfo{person}{Pan Li},
  \bibinfo{person}{Wentao Zhao}, \bibinfo{person}{Wei Cai},
  \bibinfo{person}{Shui Yu}, {and} \bibinfo{person}{Victor~CM Leung}.}
  \bibinfo{year}{2018}\natexlab{a}.
\newblock \showarticletitle{A Survey on Security Threats and Defensive
  Techniques of Machine Learning: A Data Driven View}.
\newblock \bibinfo{journal}{\emph{IEEE Access}} (\bibinfo{year}{2018}).
\newblock


\bibitem[\protect\citeauthoryear{Liu, Chen, Liu, and Song}{Liu
  et~al\mbox{.}}{2016}]%
        {liu2016delving}
\bibfield{author}{\bibinfo{person}{Yanpei Liu}, \bibinfo{person}{Xinyun Chen},
  \bibinfo{person}{Chang Liu}, {and} \bibinfo{person}{Dawn Song}.}
  \bibinfo{year}{2016}\natexlab{}.
\newblock \bibinfo{title}{Delving into transferable adversarial examples and
  black-box attacks}.
\newblock \bibinfo{howpublished}{arXiv:1611.02770}.
\newblock


\bibitem[\protect\citeauthoryear{Lowd and Meek}{Lowd and Meek}{2005}]%
        {lowd2005adversarial}
\bibfield{author}{\bibinfo{person}{Daniel Lowd} {and}
  \bibinfo{person}{Christopher Meek}.} \bibinfo{year}{2005}\natexlab{}.
\newblock \showarticletitle{Adversarial learning}. In
  \bibinfo{booktitle}{\emph{SIGKDD}}. ACM.
\newblock


\bibitem[\protect\citeauthoryear{Lu, Issaranon, and Forsyth}{Lu
  et~al\mbox{.}}{2017a}]%
        {lu2017safetynet}
\bibfield{author}{\bibinfo{person}{Jiajun Lu}, \bibinfo{person}{Theerasit
  Issaranon}, {and} \bibinfo{person}{David~A. Forsyth}.}
  \bibinfo{year}{2017}\natexlab{a}.
\newblock \showarticletitle{SafetyNet: Detecting and Rejecting Adversarial
  Examples Robustly}. In \bibinfo{booktitle}{\emph{ICCV}}. IEEE.
\newblock


\bibitem[\protect\citeauthoryear{Lu, Sibai, Fabry, and Forsyth}{Lu
  et~al\mbox{.}}{2017b}]%
        {lu2017standard}
\bibfield{author}{\bibinfo{person}{Jiajun Lu}, \bibinfo{person}{Hussein Sibai},
  \bibinfo{person}{Evan Fabry}, {and} \bibinfo{person}{David Forsyth}.}
  \bibinfo{year}{2017}\natexlab{b}.
\newblock \bibinfo{title}{Standard detectors aren't (currently) fooled by
  physical adversarial stop signs}.
\newblock \bibinfo{howpublished}{arXiv:1710.03337}.
\newblock


\bibitem[\protect\citeauthoryear{Luo, Boix, Roig, Poggio, and Zhao}{Luo
  et~al\mbox{.}}{2015}]%
        {luo2015foveation}
\bibfield{author}{\bibinfo{person}{Yan Luo}, \bibinfo{person}{Xavier Boix},
  \bibinfo{person}{Gemma Roig}, \bibinfo{person}{Tomaso Poggio}, {and}
  \bibinfo{person}{Qi Zhao}.} \bibinfo{year}{2015}\natexlab{}.
\newblock \bibinfo{title}{Foveation-based mechanisms alleviate adversarial
  examples}.
\newblock \bibinfo{howpublished}{arXiv:1511.06292}.
\newblock


\bibitem[\protect\citeauthoryear{Lyu, Huang, and Liang}{Lyu
  et~al\mbox{.}}{2015}]%
        {lyu2015unified}
\bibfield{author}{\bibinfo{person}{Chunchuan Lyu}, \bibinfo{person}{Kaizhu
  Huang}, {and} \bibinfo{person}{Hai-Ning Liang}.}
  \bibinfo{year}{2015}\natexlab{}.
\newblock \showarticletitle{A unified gradient regularization family for
  adversarial examples}. In \bibinfo{booktitle}{\emph{ICDM}}. IEEE.
\newblock


\bibitem[\protect\citeauthoryear{Madry, Makelov, Schmidt, Tsipras, and
  Vladu}{Madry et~al\mbox{.}}{2018}]%
        {madry2017towards}
\bibfield{author}{\bibinfo{person}{Aleksander Madry},
  \bibinfo{person}{Aleksandar Makelov}, \bibinfo{person}{Ludwig Schmidt},
  \bibinfo{person}{Dimitris Tsipras}, {and} \bibinfo{person}{Adrian Vladu}.}
  \bibinfo{year}{2018}\natexlab{}.
\newblock \showarticletitle{Towards deep learning models resistant to
  adversarial attacks}.
\newblock \bibinfo{journal}{\emph{ICLR}} (\bibinfo{year}{2018}).
\newblock


\bibitem[\protect\citeauthoryear{Meng and Chen}{Meng and Chen}{2017}]%
        {meng2017magnet}
\bibfield{author}{\bibinfo{person}{Dongyu Meng} {and} \bibinfo{person}{Hao
  Chen}.} \bibinfo{year}{2017}\natexlab{}.
\newblock \showarticletitle{Magnet: a two-pronged defense against adversarial
  examples}. In \bibinfo{booktitle}{\emph{ACM CCS}}. ACM.
\newblock


\bibitem[\protect\citeauthoryear{Metzen, Genewein, Fischer, and
  Bischoff}{Metzen et~al\mbox{.}}{2017}]%
        {metzen2017detecting}
\bibfield{author}{\bibinfo{person}{Jan~Hendrik Metzen}, \bibinfo{person}{Tim
  Genewein}, \bibinfo{person}{Volker Fischer}, {and} \bibinfo{person}{Bastian
  Bischoff}.} \bibinfo{year}{2017}\natexlab{}.
\newblock \showarticletitle{On detecting adversarial perturbations}.
\newblock \bibinfo{journal}{\emph{ICLR}} (\bibinfo{year}{2017}).
\newblock


\bibitem[\protect\citeauthoryear{Mirman, Gehr, and Vechev}{Mirman
  et~al\mbox{.}}{2018}]%
        {mirman2018differentiable}
\bibfield{author}{\bibinfo{person}{Matthew Mirman}, \bibinfo{person}{Timon
  Gehr}, {and} \bibinfo{person}{Martin Vechev}.}
  \bibinfo{year}{2018}\natexlab{}.
\newblock \showarticletitle{Differentiable abstract interpretation for provably
  robust neural networks}.
\newblock \bibinfo{journal}{\emph{ICML}} (\bibinfo{year}{2018}).
\newblock


\bibitem[\protect\citeauthoryear{Mitchell et~al\mbox{.}}{Mitchell
  et~al\mbox{.}}{1997}]%
        {mitchell1997machine}
\bibfield{author}{\bibinfo{person}{Tom~M. Mitchell} {et~al\mbox{.}}}
  \bibinfo{year}{1997}\natexlab{}.
\newblock \showarticletitle{Machine learning. 1997}.
\newblock \bibinfo{journal}{\emph{Burr Ridge, IL: McGraw Hill}}
  (\bibinfo{year}{1997}).
\newblock


\bibitem[\protect\citeauthoryear{Moosavi-Dezfooli, Fawzi, Fawzi, and
  Frossard}{Moosavi-Dezfooli et~al\mbox{.}}{2017a}]%
        {moosavi2017universal}
\bibfield{author}{\bibinfo{person}{Seyed-Mohsen Moosavi-Dezfooli},
  \bibinfo{person}{Alhussein Fawzi}, \bibinfo{person}{Omar Fawzi}, {and}
  \bibinfo{person}{Pascal Frossard}.} \bibinfo{year}{2017}\natexlab{a}.
\newblock \showarticletitle{Universal adversarial perturbations}. In
  \bibinfo{booktitle}{\emph{CVPR}}. IEEE.
\newblock


\bibitem[\protect\citeauthoryear{Moosavi-Dezfooli, Fawzi, Fawzi, Frossard, and
  Soatto}{Moosavi-Dezfooli et~al\mbox{.}}{2017b}]%
        {moosavi2017analysis}
\bibfield{author}{\bibinfo{person}{Seyed-Mohsen Moosavi-Dezfooli},
  \bibinfo{person}{Alhussein Fawzi}, \bibinfo{person}{Omar Fawzi},
  \bibinfo{person}{Pascal Frossard}, {and} \bibinfo{person}{Stefano Soatto}.}
  \bibinfo{year}{2017}\natexlab{b}.
\newblock \bibinfo{title}{Analysis of universal adversarial perturbations}.
\newblock \bibinfo{howpublished}{arXiv:1705.09554}.
\newblock


\bibitem[\protect\citeauthoryear{Moosavi-Dezfooli, Fawzi, Fawzi, Frossard, and
  Soatto}{Moosavi-Dezfooli et~al\mbox{.}}{2018}]%
        {moosavi-dezfooli2018robustness}
\bibfield{author}{\bibinfo{person}{Seyed-Mohsen Moosavi-Dezfooli},
  \bibinfo{person}{Alhussein Fawzi}, \bibinfo{person}{Omar Fawzi},
  \bibinfo{person}{Pascal Frossard}, {and} \bibinfo{person}{Stefano Soatto}.}
  \bibinfo{year}{2018}\natexlab{}.
\newblock \showarticletitle{Robustness of Classifiers to Universal
  Perturbations: A Geometric Perspective}.
\newblock \bibinfo{journal}{\emph{ICLR}} (\bibinfo{year}{2018}).
\newblock


\bibitem[\protect\citeauthoryear{Moosavi~Dezfooli, Fawzi, and
  Frossard}{Moosavi~Dezfooli et~al\mbox{.}}{2016}]%
        {moosavi2016deepfool}
\bibfield{author}{\bibinfo{person}{Seyed~Mohsen Moosavi~Dezfooli},
  \bibinfo{person}{Alhussein Fawzi}, {and} \bibinfo{person}{Pascal Frossard}.}
  \bibinfo{year}{2016}\natexlab{}.
\newblock \showarticletitle{Deepfool: a simple and accurate method to fool deep
  neural networks}. In \bibinfo{booktitle}{\emph{CVPR}}. IEEE.
\newblock


\bibitem[\protect\citeauthoryear{Narodytska and Kasiviswanathan}{Narodytska and
  Kasiviswanathan}{2017}]%
        {narodytska2017simple}
\bibfield{author}{\bibinfo{person}{Nina Narodytska} {and}
  \bibinfo{person}{Shiva~Prasad Kasiviswanathan}.}
  \bibinfo{year}{2017}\natexlab{}.
\newblock \bibinfo{title}{Simple Black-Box Adversarial Perturbations on Deep
  Neural Networks}.
\newblock \bibinfo{howpublished}{arXiv:1612.06299}.
\newblock


\bibitem[\protect\citeauthoryear{Nayebi and Ganguli}{Nayebi and
  Ganguli}{2017}]%
        {nayebi2017biologically}
\bibfield{author}{\bibinfo{person}{Aran Nayebi} {and} \bibinfo{person}{Surya
  Ganguli}.} \bibinfo{year}{2017}\natexlab{}.
\newblock \bibinfo{title}{Biologically inspired protection of deep networks
  from adversarial attacks}.
\newblock \bibinfo{howpublished}{arXiv:1703.09202}.
\newblock


\bibitem[\protect\citeauthoryear{Nguyen, Yosinski, and Clune}{Nguyen
  et~al\mbox{.}}{2015}]%
        {nguyen2015deep}
\bibfield{author}{\bibinfo{person}{Anh Nguyen}, \bibinfo{person}{Jason
  Yosinski}, {and} \bibinfo{person}{Jeff Clune}.}
  \bibinfo{year}{2015}\natexlab{}.
\newblock \showarticletitle{Deep neural networks are easily fooled: High
  confidence predictions for unrecognizable images}. In
  \bibinfo{booktitle}{\emph{CVPR}}. IEEE.
\newblock


\bibitem[\protect\citeauthoryear{Papernot}{Papernot}{2018}]%
        {papernot2018marauder}
\bibfield{author}{\bibinfo{person}{Nicolas Papernot}.}
  \bibinfo{year}{2018}\natexlab{}.
\newblock \bibinfo{title}{A Marauder's Map of Security and Privacy in Machine
  Learning}.
\newblock \bibinfo{howpublished}{arXiv:1811.01134}.
\newblock


\bibitem[\protect\citeauthoryear{Papernot, McDaniel, and Goodfellow}{Papernot
  et~al\mbox{.}}{2016a}]%
        {papernot2016transferability}
\bibfield{author}{\bibinfo{person}{Nicolas Papernot}, \bibinfo{person}{Patrick
  McDaniel}, {and} \bibinfo{person}{Ian Goodfellow}.}
  \bibinfo{year}{2016}\natexlab{a}.
\newblock \bibinfo{title}{Transferability in machine learning: from phenomena
  to black-box attacks using adversarial samples}.
\newblock \bibinfo{howpublished}{arXiv:1605.07277}.
\newblock


\bibitem[\protect\citeauthoryear{Papernot, McDaniel, Goodfellow, Jha, Celik,
  and Swami}{Papernot et~al\mbox{.}}{2017}]%
        {papernot2017practical}
\bibfield{author}{\bibinfo{person}{Nicolas Papernot}, \bibinfo{person}{Patrick
  McDaniel}, \bibinfo{person}{Ian Goodfellow}, \bibinfo{person}{Somesh Jha},
  \bibinfo{person}{Z~Berkay Celik}, {and} \bibinfo{person}{Ananthram Swami}.}
  \bibinfo{year}{2017}\natexlab{}.
\newblock \showarticletitle{Practical black-box attacks against machine
  learning}. In \bibinfo{booktitle}{\emph{ASIACCS}}. ACM.
\newblock


\bibitem[\protect\citeauthoryear{Papernot, McDaniel, Jha, Fredrikson, Celik,
  and Swami}{Papernot et~al\mbox{.}}{2016b}]%
        {papernot2016limitations}
\bibfield{author}{\bibinfo{person}{Nicolas Papernot}, \bibinfo{person}{Patrick
  McDaniel}, \bibinfo{person}{Somesh Jha}, \bibinfo{person}{Matt Fredrikson},
  \bibinfo{person}{Z.~Berkay Celik}, {and} \bibinfo{person}{Ananthram Swami}.}
  \bibinfo{year}{2016}\natexlab{b}.
\newblock \showarticletitle{The limitations of deep learning in adversarial
  settings}. In \bibinfo{booktitle}{\emph{EuroS\&P}}. IEEE.
\newblock


\bibitem[\protect\citeauthoryear{Papernot, McDaniel, Sinha, and
  Wellman}{Papernot et~al\mbox{.}}{2018}]%
        {Papernot2018}
\bibfield{author}{\bibinfo{person}{Nicolas Papernot}, \bibinfo{person}{Patrick
  McDaniel}, \bibinfo{person}{Arunesh Sinha}, {and} \bibinfo{person}{Michael~P.
  Wellman}.} \bibinfo{year}{2018}\natexlab{}.
\newblock \showarticletitle{SoK: Security and privacy in machine learning}. In
  \bibinfo{booktitle}{\emph{EuroS\&P}}. IEEE.
\newblock


\bibitem[\protect\citeauthoryear{Papernot, McDaniel, Wu, Jha, and
  Swami}{Papernot et~al\mbox{.}}{2016c}]%
        {papernot2016distillation}
\bibfield{author}{\bibinfo{person}{Nicolas Papernot}, \bibinfo{person}{Patrick
  McDaniel}, \bibinfo{person}{Xi Wu}, \bibinfo{person}{Somesh Jha}, {and}
  \bibinfo{person}{Ananthram Swami}.} \bibinfo{year}{2016}\natexlab{c}.
\newblock \showarticletitle{Distillation as a defense to adversarial
  perturbations against deep neural networks}. In
  \bibinfo{booktitle}{\emph{S\&P}}. IEEE.
\newblock


\bibitem[\protect\citeauthoryear{Pei, Cao, Yang, and Jana}{Pei
  et~al\mbox{.}}{2017}]%
        {pei2017towards}
\bibfield{author}{\bibinfo{person}{Kexin Pei}, \bibinfo{person}{Yinzhi Cao},
  \bibinfo{person}{Junfeng Yang}, {and} \bibinfo{person}{Suman Jana}.}
  \bibinfo{year}{2017}\natexlab{}.
\newblock \bibinfo{title}{Towards practical verification of machine learning:
  The case of computer vision systems}.
\newblock \bibinfo{howpublished}{arXiv:1712.01785}.
\newblock


\bibitem[\protect\citeauthoryear{Pontes-Filho and Liwicki}{Pontes-Filho and
  Liwicki}{2018}]%
        {pontes2018bidirectional}
\bibfield{author}{\bibinfo{person}{Sidney Pontes-Filho} {and}
  \bibinfo{person}{Marcus Liwicki}.} \bibinfo{year}{2018}\natexlab{}.
\newblock \bibinfo{title}{Bidirectional Learning for Robust Neural Networks}.
\newblock \bibinfo{howpublished}{arXiv:1805.08006}.
\newblock


\bibitem[\protect\citeauthoryear{Poursaeed, Katsman, Gao, and
  Belongie}{Poursaeed et~al\mbox{.}}{2018}]%
        {poursaeed2018generative}
\bibfield{author}{\bibinfo{person}{Omid Poursaeed}, \bibinfo{person}{Isay
  Katsman}, \bibinfo{person}{Bicheng Gao}, {and} \bibinfo{person}{Serge
  Belongie}.} \bibinfo{year}{2018}\natexlab{}.
\newblock \showarticletitle{Generative adversarial perturbations}. In
  \bibinfo{booktitle}{\emph{CVPR}}. IEEE.
\newblock


\bibitem[\protect\citeauthoryear{Rakin, He, Gong, and Fan}{Rakin
  et~al\mbox{.}}{2018}]%
        {rakin2018blind}
\bibfield{author}{\bibinfo{person}{Adnan~Siraj Rakin}, \bibinfo{person}{Zhezhi
  He}, \bibinfo{person}{Boqing Gong}, {and} \bibinfo{person}{Deliang Fan}.}
  \bibinfo{year}{2018}\natexlab{}.
\newblock \bibinfo{title}{Blind pre-processing: A robust defense method against
  adversarial examples}.
\newblock \bibinfo{howpublished}{arXiv:1802.01549}.
\newblock


\bibitem[\protect\citeauthoryear{Rezende and Wierstra}{Rezende and
  Wierstra}{2014}]%
        {danilo2014stochastich}
\bibfield{author}{\bibinfo{person}{Shakir~Mohamed Rezende, Danilo~Jimenez}
  {and} \bibinfo{person}{Daan Wierstra}.} \bibinfo{year}{2014}\natexlab{}.
\newblock \bibinfo{title}{Stochastic backpropagation and approximate inference
  in deep generative models.}
\newblock \bibinfo{howpublished}{arXiv:1401.4082}.
\newblock


\bibitem[\protect\citeauthoryear{Rosenberg, Shabtai, Rokach, and
  Elovici}{Rosenberg et~al\mbox{.}}{2017}]%
        {rosenberg2017generic}
\bibfield{author}{\bibinfo{person}{Ishai Rosenberg}, \bibinfo{person}{Asaf
  Shabtai}, \bibinfo{person}{Lior Rokach}, {and} \bibinfo{person}{Yuval
  Elovici}.} \bibinfo{year}{2017}\natexlab{}.
\newblock \bibinfo{title}{Generic Black-Box End-to-End Attack against RNNs and
  Other API Calls Based Malware Classifiers}.
\newblock \bibinfo{howpublished}{arXiv:1707.05970}.
\newblock


\bibitem[\protect\citeauthoryear{Roth, Lucchi, Nowozin, and Hofmann}{Roth
  et~al\mbox{.}}{2018}]%
        {roth2018adversarially}
\bibfield{author}{\bibinfo{person}{Kevin Roth}, \bibinfo{person}{Aurelien
  Lucchi}, \bibinfo{person}{Sebastian Nowozin}, {and} \bibinfo{person}{Thomas
  Hofmann}.} \bibinfo{year}{2018}\natexlab{}.
\newblock \bibinfo{title}{Adversarially Robust Training through Structured
  Gradient Regularization}.
\newblock \bibinfo{howpublished}{arXiv:1805.08736}.
\newblock


\bibitem[\protect\citeauthoryear{Rozsa, G{\"u}nther, and Boult}{Rozsa
  et~al\mbox{.}}{2016a}]%
        {rozsa2016accuracy}
\bibfield{author}{\bibinfo{person}{Andras Rozsa}, \bibinfo{person}{Manuel
  G{\"u}nther}, {and} \bibinfo{person}{Terrance~E. Boult}.}
  \bibinfo{year}{2016}\natexlab{a}.
\newblock \showarticletitle{Are accuracy and robustness correlated}. In
  \bibinfo{booktitle}{\emph{ICMLA}}. IEEE.
\newblock


\bibitem[\protect\citeauthoryear{Rozsa, Gunther, and Boult}{Rozsa
  et~al\mbox{.}}{2016b}]%
        {rozsa2016towards}
\bibfield{author}{\bibinfo{person}{Andras Rozsa}, \bibinfo{person}{Manuel
  Gunther}, {and} \bibinfo{person}{Terrance~E. Boult}.}
  \bibinfo{year}{2016}\natexlab{b}.
\newblock \bibinfo{title}{Towards robust deep neural networks with BANG}.
\newblock \bibinfo{howpublished}{arXiv:1612.00138}.
\newblock


\bibitem[\protect\citeauthoryear{Ruan, Huang, and Kwiatkowska}{Ruan
  et~al\mbox{.}}{2018}]%
        {ruan2018reachability}
\bibfield{author}{\bibinfo{person}{Wenjie Ruan}, \bibinfo{person}{Xiaowei
  Huang}, {and} \bibinfo{person}{Marta Kwiatkowska}.}
  \bibinfo{year}{2018}\natexlab{}.
\newblock \showarticletitle{Reachability analysis of deep neural networks with
  provable guarantees}.
\newblock \bibinfo{journal}{\emph{IJCAI}} (\bibinfo{year}{2018}).
\newblock


\bibitem[\protect\citeauthoryear{Rubinstein, Nelson, Huang, Joseph, Lau, Rao,
  Taft, and Tygar}{Rubinstein et~al\mbox{.}}{2009}]%
        {rubinstein2009antidote}
\bibfield{author}{\bibinfo{person}{Benjamin~IP. Rubinstein},
  \bibinfo{person}{Blaine Nelson}, \bibinfo{person}{Ling Huang},
  \bibinfo{person}{Anthony~D. Joseph}, \bibinfo{person}{Shing-hon Lau},
  \bibinfo{person}{Satish Rao}, \bibinfo{person}{Nina Taft}, {and}
  \bibinfo{person}{JD. Tygar}.} \bibinfo{year}{2009}\natexlab{}.
\newblock \showarticletitle{Antidote: understanding and defending against
  poisoning of anomaly detectors}. In \bibinfo{booktitle}{\emph{SIGCOMM}}. ACM.
\newblock


\bibitem[\protect\citeauthoryear{Russu, Demontis, Biggio, Fumera, and
  Roli}{Russu et~al\mbox{.}}{2016}]%
        {russu2016secure}
\bibfield{author}{\bibinfo{person}{Paolo Russu}, \bibinfo{person}{Ambra
  Demontis}, \bibinfo{person}{Battista Biggio}, \bibinfo{person}{Giorgio
  Fumera}, {and} \bibinfo{person}{Fabio Roli}.}
  \bibinfo{year}{2016}\natexlab{}.
\newblock \showarticletitle{Secure kernel machines against evasion attacks}. In
  \bibinfo{booktitle}{\emph{AISec}}. ACM.
\newblock


\bibitem[\protect\citeauthoryear{Sabour, Cao, Faghri, and Fleet}{Sabour
  et~al\mbox{.}}{2015}]%
        {sabour2015adversarial}
\bibfield{author}{\bibinfo{person}{Sara Sabour}, \bibinfo{person}{Yanshuai
  Cao}, \bibinfo{person}{Fartash Faghri}, {and} \bibinfo{person}{David~J.
  Fleet}.} \bibinfo{year}{2015}\natexlab{}.
\newblock \bibinfo{title}{Adversarial manipulation of deep representations}.
\newblock \bibinfo{howpublished}{arXiv:1511.05122}.
\newblock


\bibitem[\protect\citeauthoryear{Salman, Yang, Zhang, Hsieh, and Zhang}{Salman
  et~al\mbox{.}}{2019}]%
        {salman2019convex}
\bibfield{author}{\bibinfo{person}{Hadi Salman}, \bibinfo{person}{Greg Yang},
  \bibinfo{person}{Huan Zhang}, \bibinfo{person}{Cho-Jui Hsieh}, {and}
  \bibinfo{person}{Pengchuan Zhang}.} \bibinfo{year}{2019}\natexlab{}.
\newblock \showarticletitle{A Convex Relaxation Barrier to Tight Robust
  Verification of Neural Networks}.
\newblock \bibinfo{journal}{\emph{ICLR Workshop}} (\bibinfo{year}{2019}).
\newblock


\bibitem[\protect\citeauthoryear{Schmidt, Santurkar, Tsipras, Talwar, and
  Madry}{Schmidt et~al\mbox{.}}{2018}]%
        {schmidt2018adversarially}
\bibfield{author}{\bibinfo{person}{Ludwig Schmidt}, \bibinfo{person}{Shibani
  Santurkar}, \bibinfo{person}{Dimitris Tsipras}, \bibinfo{person}{Kunal
  Talwar}, {and} \bibinfo{person}{Aleksander Madry}.}
  \bibinfo{year}{2018}\natexlab{}.
\newblock \showarticletitle{Adversarially Robust Generalization Requires More
  Data}.
\newblock \bibinfo{journal}{\emph{NeurIPS}} (\bibinfo{year}{2018}).
\newblock


\bibitem[\protect\citeauthoryear{Shaham, Garritano, Yamada, Weinberger,
  Cloninger, Cheng, and Stanton}{Shaham et~al\mbox{.}}{2018}]%
        {shaham2018defending}
\bibfield{author}{\bibinfo{person}{Uri Shaham}, \bibinfo{person}{James
  Garritano}, \bibinfo{person}{Yutaro Yamada}, \bibinfo{person}{Ethan
  Weinberger}, \bibinfo{person}{Alex Cloninger}, \bibinfo{person}{Xiuyuan
  Cheng}, {and} \bibinfo{person}{Kelly Stanton}.}
  \bibinfo{year}{2018}\natexlab{}.
\newblock \bibinfo{title}{Defending against Adversarial Images using Basis
  Functions Transformations}.
\newblock \bibinfo{howpublished}{arXiv:1803.10840}.
\newblock


\bibitem[\protect\citeauthoryear{Shaham, Yamada, and Negahban}{Shaham
  et~al\mbox{.}}{2015}]%
        {shaham2015understanding}
\bibfield{author}{\bibinfo{person}{Uri Shaham}, \bibinfo{person}{Yutaro
  Yamada}, {and} \bibinfo{person}{Sahand Negahban}.}
  \bibinfo{year}{2015}\natexlab{}.
\newblock \bibinfo{title}{Understanding adversarial training: Increasing local
  stability of neural nets through robust optimization}.
\newblock \bibinfo{howpublished}{arXiv:1511.05432}.
\newblock


\bibitem[\protect\citeauthoryear{Sharif, Bhagavatula, Bauer, and Reiter}{Sharif
  et~al\mbox{.}}{2016}]%
        {sharif2016accessorize}
\bibfield{author}{\bibinfo{person}{Mahmood Sharif}, \bibinfo{person}{Sruti
  Bhagavatula}, \bibinfo{person}{Lujo Bauer}, {and} \bibinfo{person}{Michael~K.
  Reiter}.} \bibinfo{year}{2016}\natexlab{}.
\newblock \showarticletitle{Accessorize to a crime: Real and stealthy attacks
  on state-of-the-art face recognition}. In \bibinfo{booktitle}{\emph{SIGSAC}}.
  ACM.
\newblock


\bibitem[\protect\citeauthoryear{Sharma and Chen}{Sharma and Chen}{2017}]%
        {sharma2017breaking}
\bibfield{author}{\bibinfo{person}{Yash Sharma} {and} \bibinfo{person}{Pin-Yu
  Chen}.} \bibinfo{year}{2017}\natexlab{}.
\newblock \bibinfo{title}{Breaking the Madry Defense Model with L1-based
  Adversarial Examples}.
\newblock \bibinfo{howpublished}{arXiv:1710.10733}.
\newblock


\bibitem[\protect\citeauthoryear{Sinha, Chen, Badrinarayanan, and
  Rabinovich}{Sinha et~al\mbox{.}}{2018a}]%
        {sinha2018gradient}
\bibfield{author}{\bibinfo{person}{Ayan Sinha}, \bibinfo{person}{Zhao Chen},
  \bibinfo{person}{Vijay Badrinarayanan}, {and} \bibinfo{person}{Andrew
  Rabinovich}.} \bibinfo{year}{2018}\natexlab{a}.
\newblock \bibinfo{title}{Gradient Adversarial Training of Neural Networks}.
\newblock \bibinfo{howpublished}{arXiv:1806.08028}.
\newblock


\bibitem[\protect\citeauthoryear{Sinha, Namkoong, and Duchi}{Sinha
  et~al\mbox{.}}{2018b}]%
        {sinha2018certifying}
\bibfield{author}{\bibinfo{person}{Aman Sinha}, \bibinfo{person}{Hongseok
  Namkoong}, {and} \bibinfo{person}{John Duchi}.}
  \bibinfo{year}{2018}\natexlab{b}.
\newblock \showarticletitle{Certifying some distributional robustness with
  principled adversarial training}.
\newblock \bibinfo{journal}{\emph{ICLR}} (\bibinfo{year}{2018}).
\newblock


\bibitem[\protect\citeauthoryear{Song, Kim, Nowozin, Ermon, and Kushman}{Song
  et~al\mbox{.}}{2018a}]%
        {song2017pixeldefend}
\bibfield{author}{\bibinfo{person}{Yang Song}, \bibinfo{person}{Taesup Kim},
  \bibinfo{person}{Sebastian Nowozin}, \bibinfo{person}{Stefano Ermon}, {and}
  \bibinfo{person}{Nate Kushman}.} \bibinfo{year}{2018}\natexlab{a}.
\newblock \showarticletitle{{PixelDefend}: Leveraging generative models to
  understand and defend against adversarial examples}.
\newblock \bibinfo{journal}{\emph{ICLR}} (\bibinfo{year}{2018}).
\newblock


\bibitem[\protect\citeauthoryear{Song, Shu, Kushman, and Ermon}{Song
  et~al\mbox{.}}{2018b}]%
        {song2018constructing}
\bibfield{author}{\bibinfo{person}{Yang Song}, \bibinfo{person}{Rui Shu},
  \bibinfo{person}{Nate Kushman}, {and} \bibinfo{person}{Stefano Ermon}.}
  \bibinfo{year}{2018}\natexlab{b}.
\newblock \showarticletitle{Constructing unrestricted adversarial examples with
  generative models}.
\newblock \bibinfo{journal}{\emph{NeurIPS}} (\bibinfo{year}{2018}).
\newblock


\bibitem[\protect\citeauthoryear{Sra, Nowozin, and Wright}{Sra
  et~al\mbox{.}}{2012}]%
        {sra2012optimization}
\bibfield{author}{\bibinfo{person}{Suvrit Sra}, \bibinfo{person}{Sebastian
  Nowozin}, {and} \bibinfo{person}{Stephen~J. Wright}.}
  \bibinfo{year}{2012}\natexlab{}.
\newblock \bibinfo{booktitle}{\emph{Optimization for machine learning}}.
\newblock \bibinfo{publisher}{{MIT} Press}.
\newblock


\bibitem[\protect\citeauthoryear{Su, Zhang, Chen, Yi, Chen, and Gao}{Su
  et~al\mbox{.}}{2018}]%
        {su2018robustness}
\bibfield{author}{\bibinfo{person}{Dong Su}, \bibinfo{person}{Huan Zhang},
  \bibinfo{person}{Hongge Chen}, \bibinfo{person}{Jinfeng Yi},
  \bibinfo{person}{Pin-Yu Chen}, {and} \bibinfo{person}{Yupeng Gao}.}
  \bibinfo{year}{2018}\natexlab{}.
\newblock \showarticletitle{Is Robustness the Cost of Accuracy?--A
  Comprehensive Study on the Robustness of 18 Deep Image Classification
  Models}. In \bibinfo{booktitle}{\emph{ECCV}}.
\newblock


\bibitem[\protect\citeauthoryear{Su, Vargas, and Sakurai}{Su
  et~al\mbox{.}}{2019}]%
        {su2017one}
\bibfield{author}{\bibinfo{person}{Jiawei Su},
  \bibinfo{person}{Danilo~Vasconcellos Vargas}, {and} \bibinfo{person}{Kouichi
  Sakurai}.} \bibinfo{year}{2019}\natexlab{}.
\newblock \showarticletitle{One pixel attack for fooling deep neural networks}.
\newblock \bibinfo{journal}{\emph{IEEE Transactions on Evolutionary
  Computation}} (\bibinfo{year}{2019}).
\newblock


\bibitem[\protect\citeauthoryear{Sun, Ozay, and Okatani}{Sun
  et~al\mbox{.}}{2017}]%
        {sun2017hypernetworks}
\bibfield{author}{\bibinfo{person}{Zhun Sun}, \bibinfo{person}{Mete Ozay},
  {and} \bibinfo{person}{Takayuki Okatani}.} \bibinfo{year}{2017}\natexlab{}.
\newblock \bibinfo{title}{HyperNetworks with statistical filtering for
  defending adversarial examples}.
\newblock \bibinfo{howpublished}{arXiv:1711.01791}.
\newblock


\bibitem[\protect\citeauthoryear{Sutskever, Vinyals, and Le}{Sutskever
  et~al\mbox{.}}{2014}]%
        {sutskever2014sequence}
\bibfield{author}{\bibinfo{person}{Ilya Sutskever}, \bibinfo{person}{Oriol
  Vinyals}, {and} \bibinfo{person}{Quoc~V. Le}.}
  \bibinfo{year}{2014}\natexlab{}.
\newblock \showarticletitle{Sequence to sequence learning with neural
  networks}.
\newblock \bibinfo{journal}{\emph{NeurIPS}} (\bibinfo{year}{2014}).
\newblock


\bibitem[\protect\citeauthoryear{Szegedy, Zaremba, Sutskever, Bruna, Erhan,
  Goodfellow, and Fergus}{Szegedy et~al\mbox{.}}{2013}]%
        {szegedy2013intriguing}
\bibfield{author}{\bibinfo{person}{Christian Szegedy},
  \bibinfo{person}{Wojciech Zaremba}, \bibinfo{person}{Ilya Sutskever},
  \bibinfo{person}{Joan Bruna}, \bibinfo{person}{Dumitru Erhan},
  \bibinfo{person}{Ian Goodfellow}, {and} \bibinfo{person}{Rob Fergus}.}
  \bibinfo{year}{2013}\natexlab{}.
\newblock \bibinfo{title}{Intriguing properties of neural networks}.
\newblock \bibinfo{howpublished}{arXiv:1312.6199}.
\newblock


\bibitem[\protect\citeauthoryear{Tabacof and Valle}{Tabacof and Valle}{2016}]%
        {tabacof2015exploring}
\bibfield{author}{\bibinfo{person}{Pedro Tabacof} {and}
  \bibinfo{person}{Eduardo Valle}.} \bibinfo{year}{2016}\natexlab{}.
\newblock \showarticletitle{Exploring the space of adversarial images}.
\newblock \bibinfo{journal}{\emph{JCNN}} (\bibinfo{year}{2016}).
\newblock


\bibitem[\protect\citeauthoryear{Tanay and Griffin}{Tanay and Griffin}{2016}]%
        {tanay2016boundary}
\bibfield{author}{\bibinfo{person}{Thomas Tanay} {and} \bibinfo{person}{Lewis
  Griffin}.} \bibinfo{year}{2016}\natexlab{}.
\newblock \bibinfo{title}{A boundary tilting persepective on the phenomenon of
  adversarial examples}.
\newblock \bibinfo{howpublished}{arXiv:1608.07690}.
\newblock


\bibitem[\protect\citeauthoryear{Thys, Van~Ranst, and Goedem{\'e}}{Thys
  et~al\mbox{.}}{2019}]%
        {thys2019fooling}
\bibfield{author}{\bibinfo{person}{Simen Thys}, \bibinfo{person}{Wiebe
  Van~Ranst}, {and} \bibinfo{person}{Toon Goedem{\'e}}.}
  \bibinfo{year}{2019}\natexlab{}.
\newblock \bibinfo{title}{Fooling automated surveillance cameras: adversarial
  patches to attack person detection}.
\newblock \bibinfo{howpublished}{arXiv:1904.08653}.
\newblock


\bibitem[\protect\citeauthoryear{Tram{\`e}r, Kurakin, Papernot, Goodfellow,
  Boneh, and McDaniel}{Tram{\`e}r et~al\mbox{.}}{2017a}]%
        {tramer2017ensemble}
\bibfield{author}{\bibinfo{person}{Florian Tram{\`e}r}, \bibinfo{person}{Alexey
  Kurakin}, \bibinfo{person}{Nicolas Papernot}, \bibinfo{person}{Ian
  Goodfellow}, \bibinfo{person}{Dan Boneh}, {and} \bibinfo{person}{Patrick
  McDaniel}.} \bibinfo{year}{2017}\natexlab{a}.
\newblock \bibinfo{title}{Ensemble adversarial training: Attacks and defenses}.
\newblock \bibinfo{howpublished}{arXiv:1705.07204}.
\newblock


\bibitem[\protect\citeauthoryear{Tram{\`e}r, Papernot, Goodfellow, Boneh, and
  McDaniel}{Tram{\`e}r et~al\mbox{.}}{2017b}]%
        {tramer2017space}
\bibfield{author}{\bibinfo{person}{Florian Tram{\`e}r},
  \bibinfo{person}{Nicolas Papernot}, \bibinfo{person}{Ian Goodfellow},
  \bibinfo{person}{Dan Boneh}, {and} \bibinfo{person}{Patrick McDaniel}.}
  \bibinfo{year}{2017}\natexlab{b}.
\newblock \bibinfo{title}{The space of transferable adversarial examples}.
\newblock \bibinfo{howpublished}{arXiv:1704.03453}.
\newblock


\bibitem[\protect\citeauthoryear{Tsipras, Santurkar, Engstrom, Turner, and
  Madry}{Tsipras et~al\mbox{.}}{2019}]%
        {tsipras2018robustness}
\bibfield{author}{\bibinfo{person}{Dimitris Tsipras}, \bibinfo{person}{Shibani
  Santurkar}, \bibinfo{person}{Logan Engstrom}, \bibinfo{person}{Alexander
  Turner}, {and} \bibinfo{person}{Aleksander Madry}.}
  \bibinfo{year}{2019}\natexlab{}.
\newblock \showarticletitle{Robustness may be at odds with accuracy}.
\newblock \bibinfo{journal}{\emph{ICLR}} (\bibinfo{year}{2019}).
\newblock


\bibitem[\protect\citeauthoryear{Valiant}{Valiant}{1984}]%
        {valiant1984theory}
\bibfield{author}{\bibinfo{person}{Leslie~G. Valiant}.}
  \bibinfo{year}{1984}\natexlab{}.
\newblock \showarticletitle{A theory of the learnable}. In
  \bibinfo{booktitle}{\emph{ACM Symp. on Theory of Computing}}. ACM.
\newblock


\bibitem[\protect\citeauthoryear{Vaswani, Shazeer, Parmar, Uszkoreit, Jones,
  Gomez, Kaiser, and Polosukhin}{Vaswani et~al\mbox{.}}{2017}]%
        {vaswani2017attention}
\bibfield{author}{\bibinfo{person}{Ashish Vaswani}, \bibinfo{person}{Noam
  Shazeer}, \bibinfo{person}{Niki Parmar}, \bibinfo{person}{Jakob Uszkoreit},
  \bibinfo{person}{Llion Jones}, \bibinfo{person}{Aidan~N Gomez},
  \bibinfo{person}{{\L}ukasz Kaiser}, {and} \bibinfo{person}{Illia
  Polosukhin}.} \bibinfo{year}{2017}\natexlab{}.
\newblock \showarticletitle{Attention is all you need}.
\newblock \bibinfo{journal}{\emph{NeurIPS}} (\bibinfo{year}{2017}).
\newblock


\bibitem[\protect\citeauthoryear{Wei, Liang, Cao, and Zhu}{Wei
  et~al\mbox{.}}{2018}]%
        {wei2018transferable}
\bibfield{author}{\bibinfo{person}{Xingxing Wei}, \bibinfo{person}{Siyuan
  Liang}, \bibinfo{person}{Xiaochun Cao}, {and} \bibinfo{person}{Jun Zhu}.}
  \bibinfo{year}{2018}\natexlab{}.
\newblock \bibinfo{title}{Transferable Adversarial Attacks for Image and Video
  Object Detection}.
\newblock \bibinfo{howpublished}{arXiv:1811.12641}.
\newblock


\bibitem[\protect\citeauthoryear{Weng, Zhang, Chen, Yi, Su, Gao, Hsieh, and
  Daniel}{Weng et~al\mbox{.}}{2018}]%
        {weng2018evaluating}
\bibfield{author}{\bibinfo{person}{Tsui-Wei Weng}, \bibinfo{person}{Huan
  Zhang}, \bibinfo{person}{Pin-Yu Chen}, \bibinfo{person}{Jinfeng Yi},
  \bibinfo{person}{Dong Su}, \bibinfo{person}{Yupeng Gao},
  \bibinfo{person}{Cho-Jui Hsieh}, {and} \bibinfo{person}{Luca Daniel}.}
  \bibinfo{year}{2018}\natexlab{}.
\newblock \showarticletitle{Evaluating the Robustness of Neural Networks: An
  Extreme Value Theory Approach}.
\newblock \bibinfo{journal}{\emph{ICLR}} (\bibinfo{year}{2018}).
\newblock


\bibitem[\protect\citeauthoryear{Wicker, Huang, and Kwiatkowska}{Wicker
  et~al\mbox{.}}{2018}]%
        {wicker2018feature}
\bibfield{author}{\bibinfo{person}{Matthew Wicker}, \bibinfo{person}{Xiaowei
  Huang}, {and} \bibinfo{person}{Marta Kwiatkowska}.}
  \bibinfo{year}{2018}\natexlab{}.
\newblock \showarticletitle{Feature-guided black-box safety testing of deep
  neural networks}. In \bibinfo{booktitle}{\emph{TACAS 2018}}. Springer.
\newblock


\bibitem[\protect\citeauthoryear{Wong and Kolter}{Wong and Kolter}{2018}]%
        {wong2017provable}
\bibfield{author}{\bibinfo{person}{Eric Wong} {and} \bibinfo{person}{J.~Zico
  Kolter}.} \bibinfo{year}{2018}\natexlab{}.
\newblock \showarticletitle{Provable defenses against adversarial examples via
  the convex outer adversarial polytope}.
\newblock \bibinfo{journal}{\emph{ICML}} (\bibinfo{year}{2018}).
\newblock


\bibitem[\protect\citeauthoryear{Wong, Schmidt, and Kolter}{Wong
  et~al\mbox{.}}{2019}]%
        {wong2019wasserstein}
\bibfield{author}{\bibinfo{person}{Eric Wong}, \bibinfo{person}{Frank~R.
  Schmidt}, {and} \bibinfo{person}{J.~Zico Kolter}.}
  \bibinfo{year}{2019}\natexlab{}.
\newblock \bibinfo{title}{Wasserstein Adversarial Examples via Projected
  Sinkhorn Iterations}.
\newblock \bibinfo{howpublished}{arXiv:1902.07906}.
\newblock


\bibitem[\protect\citeauthoryear{Wu, Wicker, Ruan, Huang, and Kwiatkowska}{Wu
  et~al\mbox{.}}{2018}]%
        {wu2018game}
\bibfield{author}{\bibinfo{person}{Min Wu}, \bibinfo{person}{Matthew Wicker},
  \bibinfo{person}{Wenjie Ruan}, \bibinfo{person}{Xiaowei Huang}, {and}
  \bibinfo{person}{Marta Kwiatkowska}.} \bibinfo{year}{2018}\natexlab{}.
\newblock \bibinfo{title}{A Game-Based Approximate Verification of Deep Neural
  Networks with Provable Guarantees}.
\newblock \bibinfo{howpublished}{arXiv:1807.03571}.
\newblock


\bibitem[\protect\citeauthoryear{Xiao, Zhu, Li, He, Liu, and Song}{Xiao
  et~al\mbox{.}}{2018}]%
        {xiao2018spatially}
\bibfield{author}{\bibinfo{person}{Chaowei Xiao}, \bibinfo{person}{Jun-Yan
  Zhu}, \bibinfo{person}{Bo Li}, \bibinfo{person}{Warren He},
  \bibinfo{person}{Mingyan Liu}, {and} \bibinfo{person}{Dawn Song}.}
  \bibinfo{year}{2018}\natexlab{}.
\newblock \showarticletitle{Spatially transformed adversarial examples}.
\newblock \bibinfo{journal}{\emph{ICLR}} (\bibinfo{year}{2018}).
\newblock


\bibitem[\protect\citeauthoryear{Xie, Wang, Zhang, Ren, and Yuille}{Xie
  et~al\mbox{.}}{2018}]%
        {xie2017mitigating}
\bibfield{author}{\bibinfo{person}{Cihang Xie}, \bibinfo{person}{Jianyu Wang},
  \bibinfo{person}{Zhishuai Zhang}, \bibinfo{person}{Zhou Ren}, {and}
  \bibinfo{person}{Alan Yuille}.} \bibinfo{year}{2018}\natexlab{}.
\newblock \showarticletitle{Mitigating Adversarial Effects through
  Randomization}.
\newblock \bibinfo{journal}{\emph{ICLR}} (\bibinfo{year}{2018}).
\newblock


\bibitem[\protect\citeauthoryear{Xie, Wang, Zhang, Zhou, Xie, and Yuille}{Xie
  et~al\mbox{.}}{2017}]%
        {xie2017adversarial}
\bibfield{author}{\bibinfo{person}{Cihang Xie}, \bibinfo{person}{Jianyu Wang},
  \bibinfo{person}{Zhishuai Zhang}, \bibinfo{person}{Yuyin Zhou},
  \bibinfo{person}{Lingxi Xie}, {and} \bibinfo{person}{Alan Yuille}.}
  \bibinfo{year}{2017}\natexlab{}.
\newblock \showarticletitle{Adversarial examples for semantic segmentation and
  object detection}. In \bibinfo{booktitle}{\emph{CV}}. IEEE.
\newblock


\bibitem[\protect\citeauthoryear{Xu, Qi, and Evans}{Xu et~al\mbox{.}}{2016}]%
        {xu2016automatically}
\bibfield{author}{\bibinfo{person}{Weilin Xu}, \bibinfo{person}{Yanjun Qi},
  {and} \bibinfo{person}{David Evans}.} \bibinfo{year}{2016}\natexlab{}.
\newblock \showarticletitle{Automatically Evading Classifiers}. In
  \bibinfo{booktitle}{\emph{Network and Distributed Systems Symposium}}.
\newblock


\bibitem[\protect\citeauthoryear{Yakura and Sakuma}{Yakura and Sakuma}{2018}]%
        {yakura2018robust}
\bibfield{author}{\bibinfo{person}{Hiromu Yakura} {and} \bibinfo{person}{Jun
  Sakuma}.} \bibinfo{year}{2018}\natexlab{}.
\newblock \bibinfo{title}{Robust audio adversarial example for a physical
  attack}.
\newblock \bibinfo{howpublished}{arXiv:1810.11793}.
\newblock


\bibitem[\protect\citeauthoryear{Yang, Liu, Li, Chen, and Huang}{Yang
  et~al\mbox{.}}{2019}]%
        {yang2019analyzing}
\bibfield{author}{\bibinfo{person}{Pengfei Yang}, \bibinfo{person}{Jiangchao
  Liu}, \bibinfo{person}{Jianlin Li}, \bibinfo{person}{Liqian Chen}, {and}
  \bibinfo{person}{Xiaowei Huang}.} \bibinfo{year}{2019}\natexlab{}.
\newblock \bibinfo{title}{Analyzing Deep Neural Networks with Symbolic
  Propagation: Towards Higher Precision and Faster Verification}.
\newblock \bibinfo{howpublished}{arXiv:1902.09866}.
\newblock


\bibitem[\protect\citeauthoryear{Yeung, Cloete, Shi, and Y~Ng}{Yeung
  et~al\mbox{.}}{2010}]%
        {yeung2010sensitivity}
\bibfield{author}{\bibinfo{person}{Daniel~S. Yeung}, \bibinfo{person}{Ian
  Cloete}, \bibinfo{person}{Daming Shi}, {and} \bibinfo{person}{Wing Y~Ng}.}
  \bibinfo{year}{2010}\natexlab{}.
\newblock \bibinfo{booktitle}{\emph{Sensitivity analysis for neural networks}}.
\newblock \bibinfo{publisher}{Springer}.
\newblock


\bibitem[\protect\citeauthoryear{Zhang, Yang, and Ye}{Zhang
  et~al\mbox{.}}{2018}]%
        {zhang2018detecting}
\bibfield{author}{\bibinfo{person}{Chiliang Zhang}, \bibinfo{person}{Zhimou
  Yang}, {and} \bibinfo{person}{Zuochang Ye}.} \bibinfo{year}{2018}\natexlab{}.
\newblock \bibinfo{title}{Detecting Adversarial Perturbations with Saliency}.
\newblock \bibinfo{howpublished}{arXiv:1803.08773}.
\newblock


\bibitem[\protect\citeauthoryear{Zhang, Chen, Song, Boning, Dhillon, and
  Hsieh}{Zhang et~al\mbox{.}}{2019}]%
        {zhang2019limitations}
\bibfield{author}{\bibinfo{person}{Huan Zhang}, \bibinfo{person}{Hongge Chen},
  \bibinfo{person}{Zhao Song}, \bibinfo{person}{Duane Boning},
  \bibinfo{person}{Inderjit~S Dhillon}, {and} \bibinfo{person}{Cho-Jui Hsieh}.}
  \bibinfo{year}{2019}\natexlab{}.
\newblock \showarticletitle{The limitations of adversarial training and the
  blind-spot attack}.
\newblock \bibinfo{journal}{\emph{ICLR}} (\bibinfo{year}{2019}).
\newblock


\bibitem[\protect\citeauthoryear{Zhao, Fu, Hu, Wang, et~al\mbox{.}}{Zhao
  et~al\mbox{.}}{2018b}]%
        {zhao2018detecting}
\bibfield{author}{\bibinfo{person}{Pinlong Zhao}, \bibinfo{person}{Zhouyu Fu},
  \bibinfo{person}{Qinghua Hu}, \bibinfo{person}{Jun Wang}, {et~al\mbox{.}}}
  \bibinfo{year}{2018}\natexlab{b}.
\newblock \bibinfo{title}{Detecting Adversarial Examples via Key-based
  Network}.
\newblock \bibinfo{howpublished}{arXiv:1806.00580}.
\newblock


\bibitem[\protect\citeauthoryear{Zhao, Dua, and Singh}{Zhao
  et~al\mbox{.}}{2018a}]%
        {zhao2017adversarial}
\bibfield{author}{\bibinfo{person}{Zhengli Zhao}, \bibinfo{person}{Dheeru Dua},
  {and} \bibinfo{person}{Sameer Singh}.} \bibinfo{year}{2018}\natexlab{a}.
\newblock \showarticletitle{Generating Natural Adversarial Examples}.
\newblock \bibinfo{journal}{\emph{ICLR}}.
\newblock


\end{thebibliography}


	\clearpage
\appendix

\section{Tools and Competitions}
The libraries from Table~\ref{tbl:tools} implement some of the attacks and defenses presented in this paper.
Moreover, there is an increasing number of competitions for developing new attacks and defense, some of which are mentioned in Table~\ref{tbl:comp}.

\begin{table}[h]
\parbox{.45\linewidth}{
\centering
\scalebox{0.605}{
\begin{tabular}{|l|l|}
		\toprule
		Name & Link  \\ 
		\midrule
		CleverHans & https://cleverhans.readthedocs.io  \\ \hline
		Foolbox & https://foolbox.readthedocs.io  \\ \hline
		Adversarial Toolbox & https://adversarial-robustness-toolbox.readthedocs.io/ \\  \hline
		Advertorch &
		 https://github.com/BorealisAI/advertorch \\ 
		\bottomrule
	\end{tabular}
}
	\caption{List of libraries for attacks and defenses.}
  \label{tbl:tools}
}
\hfill
\parbox{.45\linewidth}{
\centering
	\scalebox{0.7}{
\begin{tabular}{|l|l|}
		\toprule
		Name & Link  \\ 
		\midrule
		Unrestricted Adv. Examples & https://tinyurl.com/y3xrgoy2  \\ \hline
		Adv. Vision Challenge & https://tinyurl.com/y5hhwomk \\  \hline
		CAAD & https://tinyurl.com/rrk7e78 \\ 
		\bottomrule
	\end{tabular}
}
	\caption{List of  adversarial competitions.}
  \label{tbl:comp}
}
\end{table}

    \end{document}

\fi